# Convolutional Conditional Neural Processes

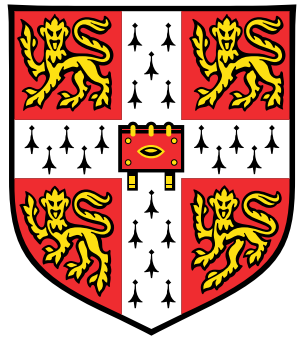

**Wessel Pieter Bruinsma**

Department of Engineering

University of Cambridge

This dissertation is submitted for the degree of

*Doctor of Philosophy*

Christ's College                                                                       15 July 2022

# Declaration

I hereby declare that, except where specific reference is made to the work of others, the contents of this dissertation are original and have not been submitted in whole or in part for consideration for any other degree or qualification in this or any other university. This dissertation is my own work and contains nothing which is the outcome of work done in collaboration with others, except as specified in the text. This dissertation contains fewer than 65,000 words including appendices, bibliography, footnotes, tables, and equations and has fewer than 150 figures.

<div style="text-align: right;">
Wessel Pieter Bruinsma<br>
15 July 2022
</div>

# Convolutional Conditional Neural Processes

Neural processes are a family of models which use neural networks to directly parametrise a map from data sets to predictions. Directly parametrising this map enables the use of expressive neural networks in small-data problems where neural networks would traditionally overfit. Neural processes can produce well-calibrated uncertainties, effectively deal with missing data, and are simple to train. These properties make this family of models appealing for a breadth of applications areas, such as healthcare or environmental sciences.

This thesis advances neural processes in three ways.

First, we propose *convolutional neural processes* (ConvNPs). ConvNPs improve data efficiency of neural processes by building in a symmetry called *translation equivariance*. ConvNPs rely on convolutional neural networks rather than multi-layer perceptrons.

Second, we propose *Gaussian neural processes* (GNPs). GNPs directly parametrise dependencies in the predictions of a neural process. Current approaches to modelling dependencies in the predictions depend on a latent variable, which consequently requires approximate inference, undermining the simplicity of the approach.

Third, we propose *autoregressive conditional neural processes* (AR CNPs). AR CNPs train a neural process without any modifications to the model or training procedure and, at test time, roll out the model in an autoregressive fashion. AR CNPs equip the neural process framework with a new knob where modelling complexity and computational expense at training time can be traded for computational expense at test time.

In addition to methodological advancements, this thesis also proposes a software abstraction that enables a compositional approach to implementing neural processes. This approach allows the user to rapidly explore the space of neural process models by putting together elementary building blocks in different ways.

<div style="text-align: right">Wessel Pieter Bruinsma</div>

# Acknowledgements

First and foremost, I must thank my supervisor Rich. Rich, you are one of the sharpest, most able, and most knowledgeable persons I know. It has been a privilege to pursue a PhD under your supervision.

I have only reached the end smiling because great friends make Cambridge feel like home. Eric and Coz, thank you for your friendship and innumerable great times. Our nights in '90s will be remembered. Will and Joel, I really value hanging out together. Together with Phil, the legacy of 308 lives on.

My time in Cambridge has been incredibly inspiring. Andrew, David, James, Stratis, Jiří, Jonathan, Anna, and Tom, I thoroughly enjoyed working together. Thank you for many insightful discussions. I have learned a great amount from every one of you. I thank Invenia for the support throughout the years, and all Invenians for interesting and fruitful discussions.

Back home, I also enjoyed the support of great friends. Pep en Oscar, thank you for countless many good nights. May there be many more to come. *Adt fundum!* Bram, Yann, and Mike, thank you for your long-standing friendships. Peter en Manouk, our new buurtjes, I look forward to spending more time together.

Coming home after having been away has always been a source of comfort. Pap, mam, and Merel, thank you for everything. Opa, you would have loved to see this thesis. I hope this thesis makes you proud.

Finally, Liesje, my dearest, thank you for your love and endless support.

# Contents

# List of Figures

# List of Tables

# List of Mathematical Statements

# List of Models

# Notation

$x \coloneqq y$         $x$ is defined as $y$

$x \eqqcolon y$         $y$ is defined as $x$

## Scalars, Vectors, and Matrices

$x$         Scalar

$x \wedge y$         $\min(x, y)$

$x \vee y$         $\max(x, y)$

$\mathbf{x}$         Vector

$\mathbf{x}_{i:j}$         Subvector of $\mathbf{x}$ consisting of elements $i$ through $j$

$\mathbf{x}_A$         Subvector of $\mathbf{x}$: $(x_i)_{i \in A}$

$\mathbf{x} \oplus \mathbf{y}$         Concatenation of $\mathbf{x}$ and $\mathbf{y}$

$|\mathbf{x}|$         Dimensionality of vector

$\mathbf{0}$         Vector of zeros

$\mathbf{X}$         Matrix

$\mathbf{I}$         Identity matrix

## Sets and Functions

$A \subsetneq B$         $A \subseteq B$, but $A \neq B$

$A/\sim$         All equivalence classes for an equivalence relation $\sim$ on $A$

$x \mapsto \ldots$         Function of $x$ without name; an anonymous function

$f = x \mapsto g(x)$         $f$ is equal to the function $g$

| | |
|---|---|
| $B^A$ | All functions $A \to B$ |
| $C(\mathcal{X}, \mathcal{Y})$ | All continuous functions $\mathcal{X} \to \mathcal{Y}$ |
| $C_b(\mathcal{X}, \mathcal{Y})$ | All continuous and bounded functions $\mathcal{X} \to \mathcal{Y}$ |
| $f\vert_A$ | Restriction of $f$ to the domain $A$ |
| $f \circ g$ | Composition of $f$ and $g$ |
| $f(\mathbf{x})$ | If $f\colon \mathbb{R} \to \mathbb{R}$, shorthand for $(f(x_1), \ldots, f(x_n))$ |
| $k(\mathbf{x}, \mathbf{y})$ | If $k\colon \mathbb{R} \times \mathbb{R} \to \mathbb{R}$, shorthand for |

$$\begin{bmatrix} k(x_1, y_1) & \cdots & k(x_1, y_m) \\ \vdots & \ddots & \vdots \\ k(x_n, y_1) & \cdots & k(x_n, y_m) \end{bmatrix}$$

## Topology and Analysis

| | |
|---|---|
| $\to$ | Convergence |
| $\rightharpoonup$ | Convergence in the weak topology |
| $d$ | Metric |
| $d_X$ | Metric on the space $X$ |
| $\|\cdot\|$ | Norm |
| $\|\cdot\|_p$ | $p$-norm; for example, $\|\cdot\|_2$ is the Euclidean norm |
| $\|\cdot\|_{\mathbb{H}}$ | Norm on the space $\mathbb{H}$ |
| $\langle \cdot, \cdot \rangle$ | Inner product |
| $\langle \cdot, \cdot \rangle_{\mathbb{H}}$ | Inner product on the space $\mathbb{H}$ |

## Probability

| | |
|---|---|
| $p$ | Probability density |
| $\mathbb{P}$ | Probability measure |
| $\mathbb{E}$ | Expectation |
| $\mathbb{E}_X$ | Expectation with respect to the random variable $X$ |
| $\mathbb{E}_p$ | Expectation with respect to the density $p$ |
| $\mathbb{E}_\mu$ | Expectation with respect to the probability measure $\mu$ |
| $\mathrm{cov}(X, Y)$ | Covariance between random variables $X$ and $Y$ |
| $\mathrm{var}(X)$ | Variance of random variable $X$ |
| $\|\cdot\|_{L^p}$ | $L^p$ norm: $\mathbb{E}[(\,\cdot\,)^p]^{\frac{1}{p}}$ |
| $\|\cdot\|_{L^p(\mu)}$ | $L^p$ norm with respect to $\mu$: $\mathbb{E}_\mu[(\,\cdot\,)^p]^{\frac{1}{p}}$ |
| $\mathrm{KL}(\mu, \nu)$ | Kullback–Leibler divergence of $\mu$ with respect to $\nu$ |
| $\mathbb{H}(\mu)$ | Differential entropy of $\mu$ with respect to the Lebesgue measure |
| $\mathcal{N}(\boldsymbol{\mu}, \boldsymbol{\Sigma})$ | Gaussian distribution with mean vector $\boldsymbol{\mu}$ and covariance matrix $\boldsymbol{\Sigma}$ |
| $\mathrm{Ber}(p)$ | Bernoulli distribution with probability $p$ |
| $o_\mathbb{P}(1)$ | A random variable that converges to zero in probability |

## Measure Theory

| | |
|---|---|
| $B$ | A typical Borel set |
| $\mathcal{B}(X)$ | Borel $\sigma$-algebra on $X$ |
| $\sigma(\mathcal{G})$ | $\sigma$-algebra generated by $\mathcal{G}$ |
| $f \in \mathcal{F}$ | $f$ is measurable with respect to $\mathcal{F}$ |
| $T(\mu)$ | Pushforward measure: $T(\mu)(B) = \mu(T^{-1}(B))$ |

## Coordinate Projections

| | |
|---|---|
| $P_{\mathbf{x}}$ | Projection onto coordinates $\mathbf{x}$: $f \mapsto (f(x_1), \ldots, f(x_n))$ |
| $P_{\mathbf{x}}\mu$ | Law of $(f(x_1), \ldots, f(x_n))$, where $f \sim \mu$ |
| $P_{\mathbf{x}}^\sigma \mu$ | Law of $(f(x_1)+\varepsilon_1, \ldots, f(x_n)+\varepsilon_n)$, where $f \sim \mu$ and $(\varepsilon_i)_{i=1}^n \overset{\text{i.i.d.}}{\sim} \mathcal{N}(0, \sigma^2)$ |

## Miscellaneous

| | |
|---|---|
| $\mathbb{S}^n$ | All permutations of size $n$ |
| $\sigma$ | A typical permutation |
| $\mathsf{T}_\tau$ | Translation by $\tau$ (Definition 2.3) |

## Meta-Learning and Neural Processes

| | |
|---|---|
| $\mathbf{x}^{(c)}$ | Context inputs |
| $\mathbf{y}^{(c)}$ | Context outputs |
| $D^{(c)}$ | Context set |
| $\mathbf{x}^{(t)}$ | Target inputs |
| $\mathbf{y}^{(t)}$ | Target outputs |
| $D^{(t)}$ | Target set |
| $q$ | Density of a prediction by a neural process |
| $\mathcal{Q}$ | Variational family |
| $\mathcal{L}_M$ | Empirical neural process objective (Definition 2.2) |
| $\mathcal{L}_{\mathrm{NP}}$ | Infinite-sample neural process objective (Definition 3.10) |
| enc | Encoder |
| dec | Decoder |

## Inputs, Outputs, and Data Sets

| | |
|---|---|
| $\mathcal{X}$ | Space of inputs |
| $\mathcal{Y}$ | Space of outputs |
| $I_N$ | Collection of all $N$ inputs: $\mathcal{X}^N$ |
| $I$ | Collection of all finite collections of inputs: $\bigcup_{n \geq 0} I_N$ |
| $\mathcal{D}_N$ | All data sets of size $N$: $(\mathcal{X} \times \mathcal{Y})^N$ |
| $\mathcal{D}_{\leq N}$ | All data sets up to size $N$: $\bigcup_{n=0}^{N} \mathcal{D}_n$ |
| $\mathcal{D}$ | All data sets: $\bigcup_{N \geq 0} \mathcal{D}_N$ |
| $D$ | A typical data set |
| $[D]$ | Equivalence class of $D$ (Section 4.2) |

## Stochastic Processes and Prediction Maps

| | |
|---|---|
| $\mathcal{P}$ | All stochastic processes |
| $\pi$ | A typical prediction map $\mathcal{D} \to \mathcal{P}$ |
| $\pi_f$ | Posterior prediction map (Definitions 3.1 and 3.6) |
| $m_\pi$ | Mean map of a prediction map $\pi$ (Definition 3.18) |
| $k_\pi$ | Kernel map of a prediction map $\pi$ (Definition 3.19) |
| $v_\pi$ | Variance map of a prediction map $\pi$ (Definition 3.20) |
| $\widetilde{\mathcal{D}}$ | Data sets of interest (Section 3.2) |
| $\widetilde{I}$ | Target inputs of interest (Section 3.2) |

# Abbreviations

TE            Translation equivariance (Definition 2.4)

DTE          Diagonal translation equivariance (Definitions 4.10 and 5.10)

CNPA        Conditional neural process approximation (Definition 3.24)

GNPA        Gaussian neural process approximation (Definition 3.25)

AR            Autoregressive

## Classes of Neural Process Models

CNPs         Conditional neural processes

LNPs         Latent-variable neural processes

GNPs         Gaussian neural processes (Section 5.5)

ConvNPs     Convolutional neural processes (Chapter 5)

## Neural Process Models

CNP          Conditional Neural Process (Model 5.1)

NP            Neural Process

GNP          Gaussian Neural Process (Model 5.9)

ACNP         Attentive Conditional Neural Process

ANP          Attentive Neural Process

AGNP         Attentive Gaussian Neural Process

ConvCNP    Convolutional Conditional Neural Process (Model 5.4)

ConvNP      Convolutional Neural Process

ConvGNP         Convolutional Gaussian Neural Process (Model 5.12)

FullConvGNP    Fully Convolutional Gaussian Neural Process (Model 5.13)

# 1 | Introduction

Before embarking on our journey through the land of neural processes, in Sections 1.1 and 1.2, we first gently introduce meta-learning. Afterwards, in Section 1.3, we introduce neural processes and explain the main contribution of this thesis. Section 1.4 then positions this thesis in historical context, and Section 1.5 provides an outline of the following chapters. Lastly, Section 1.6 lists publications and software published by the author during the PhD.

## 1.1 From Supervised Learning to Meta-Learning

What is the species of this animal? Does this MRI scan show anything of concern? What will the number of daily new coronavirus cases be tomorrow? Without having seen numerous examples of various species of animals; without having analysed many MRI scans; and without having seen extensive historical numbers of daily coronavirus cases; accurately and confidently answering these questions is nigh impossible.

These questions are all examples of *supervised learning* problems, which follow the same abstract structure. Given observed data, what is the output for a new, yet unobserved input? For example, given many observed photos of different species of animals, what is the species of the animal on this new, unobserved photo? Many statistical estimation techniques have been developed towards answering supervised learning problems. These techniques algorithmically process observed data to make predictions for new inputs. However, for complex problems, like predicting the species of an animal, these algorithms may require large amounts of data.

Unfortunately, in practice, data can be scarce. Suppose, as a running example, that we happen to spot a beautiful new species of *agapornis*, colloquially called lovebirds. No one has ever seen this species before, but we manage to take five photographs of it. To share our discovery with the world, we would like to distribute an algorithm that can process a photograph and say "Yes, this is the new species!" or "No, this is not it." Recognising the species of a bird is not an easy task, so we will require a complex algorithm, like a neural network (McCulloch et al., 1943; Rosenblatt, 1958; Ivakhnenko et al., 1965; Fukushima et al., 1982; Werbos, 1982; LeCun et al., 1989). If you have not come across neural networks before, just think of it as an incredibly flexible and versatile classification algorithm. As you might have guessed, five photographs is *far* too few to train a neural network classifier. Perhaps one hundred or a thousand example photographs would suffice, but five certainly will not.

In some sense, that a neural network requires a large amount of data to accurately determine the species of bird is reasonable. After all, the algorithm has never seen a bird before, so it must first learn what a bird even is before it can start to learn what tells apart different bird species, and that simply requires a lot of data. But this is very wasteful, because it is certainly not the first time that we encounter a bird! Perhaps, as a very young kid, having only seen a few birds in our life, we would have trouble recognising further occurrences of unusual new birds. However, as we grow older, we learn the general anatomy of a bird and what generally distinguishes different species. We learn to be able to quickly tell the general features of a species, even if we have never seen it before.

A key feature is that this learning process does not consist of just one attempt to recognise just one new species, but, throughout life, consists of many repeated attempts to recognise many new species. Hence, the life-long learning process can be seen as a stream of many small recognition problems which are all related. Crucially, attempting to solve one of these problems improves your ability to solve future ones. *Meta-learning* refers to scenarios like this where learning happens at two scales: at the scale of every small problem, attempting to solve that problem (*learning*); and across these problems, slowly improving the ability to solve any one problem (*learning to learn*). Think of it as *fast learning* and *slow learning*.

This idea of *learning to learn* lies at the heart of the meta-learning paradigm in machine learning (Schmidhuber, 1987; Thrun et al., 1998). The meta-learning paradigm attempts to build algorithms that have the ability to improve their learning mechanism. Indeed, it has been argued that this ability is necessary to build algorithms which learn like people (Lake et al., 2015; Lake et al., 2017). This two-tiered learning structure can sometimes explicitly be recognised in modern meta-learning algorithms as an *inner loop* and *outer loop* (Andrychowicz et al., 2016; Finn et al., 2017; Grant et al., 2018). However, it need not stop at two tiers (Schmidhuber, 1987).

Learning to recognise new species is a natural example of a meta-learning problem: a collection of many supervised learning problems, often small, with shared statistical structure. In meta-learning problems, solving any one of the small supervised learning problems is often too hard; but, by considering many simultaneously, an algorithm can pick up on the shared statistical structure, enabling it to perform better in any one of them. In this sense, meta-learning is closely related to *transfer learning*, where the idea is to better solve a problem by making use of different, but related problems. Whereas transfer learning is about *what* is learned from (related data), meta-learning is additionally about *how* the model learns (learning to learn).

This thesis focusses on advancing meta-learning algorithms for spatial, temporal, or spatio–temporal meta-learning problems. We now give two representative examples of such problems. For the first example, we consider electroencephalograms (EEG) of a multitude of patients (Zhang et al., 1995). For many of these EEGs, we wish to estimate a derived signal, like the patient's mood or, more simply, the signal for an electrode that we did not measure. We will develop an algorithm which *learns to estimate* the signal for a missing electrode by learning from EEGs of other patients. The second example is a problem from climate science called statistical downscaling (Maraun et al., 2018). In climate science, a large effort is spent on building simulators which can make predictions for the past and for the future (Dee et al., 2011). The output of these simulators, however, is sometimes very coarse grained, such as a single predicted temperature for every $100\,\mathrm{km}$. Downscaling methods attempt to refine these coarse-grained outputs into more fine-grained predictions by using auxiliary information, such as local topological features like elevation. We will develop an algorithm which *learns to downscale* predictions of climate simulators on a future day by learning from predictions on past days, where we recorded the true weather.

We have seen three examples of meta-learning problems: learning to recognise new bird species, learning to impute missing electrodes in EEGs (or other derived signals), and learning to downscale the output of a climate simulator. The variety of these examples attests to the flexibility of the meta-learning paradigm. It almost looks like we can just learn-to-learn anything! In the next section, we explore in more detail how an algorithm can learn to learn.

## 1.2 Learning to Learn

Before explaining how a meta-learning algorithm can improve its own learning mechanism, we take a step back to supervised learning and establish slightly more precise notation. In a supervised learning problem, we are given observed data, and the goal is to make a prediction for a new input. Denote the observed data by

$$D = \{(x_1, y_1), \ldots, (x_N, y_N)\}, \tag{1.1}$$

and denote the new input by $x_*$. Let $y_*$ be the unobserved output corresponding to $x_*$. For example, $x_n$ might be the $n^{\text{th}}$ photo of a bird, $y_n$ the species of the bird on that photo, and $x_*$ a photo of a bird for which we do not know the species $y_*$. To predict the species $y_*$ for the new photo of a bird $x_*$, supervised learning algorithms typically parametrise a function $f_\theta$ which takes in $x_*$ and gives back a prediction for $y_*$:

$$f_\theta(x_*) \quad \text{predicts} \quad y_*. \tag{1.2}$$

This function $f_\theta$ depends on some parameters $\theta$, which are learned from the observed data $D$. A typical way to learn these parameters $\theta$ from the observed data $D$ is by minimising some loss function $\ell$ which measures how well a prediction matches the data:

$$\hat\theta \in \arg\min_\theta \frac{1}{N} \sum_{n=1}^{N} \ell(f_\theta(x_n), y_n). \tag{1.3}$$

We denote a supervised learning problem by the triple $(D, x_*, y_*)$.

In a meta-learning problem, we are given not one but many supervised learning problems, often small, with shared statistical structure. Henceforth, we will refer to the many supervised learning problems as *tasks*:

$$D_{\text{related}} = \{\underbrace{(D^{(1)}, x_*^{(1)}, y_*^{(1)})}_{\text{task 1}}, \ldots, \underbrace{(D^{(M)}, x_*^{(M)}, y_*^{(M)})}_{\text{task } M}\}. \tag{1.4}$$

For example, $x_*^{(1)}$ could be a photo of some *diomedea* (a genus of albatrosses) and $D^{(1)}$ the few occurrences of *diomedea* that you have seen in the past; $x_*^{(2)}$ could be a photo of some *chrysocolaptes* (a genus of woodpeckers) and $D^{(2)}$ the few occurrences of *chrysocolaptes* that you have seen in the past; *et cetera*.

In this meta-learning problem, we are also given a new, unobserved data set $D^{(*)}$ with a new, unobserved input $x_*^{(*)}$. The goal is to predict the output $y_*^{(*)}$ corresponding to $x_*^{(*)}$. To clarify, in supervised learning, there is only one new thing: a new input $x_*$. In meta-learning, there are two new things: a new data set $D^{(*)}$ with a new input $x_*^{(*)}$. Table 1.1 compares the settings of meta-learning and supervised learning.

Whereas supervised learning requires sufficiently many data points, in meta-learning, the number of data points in any one data set can be small. This perfectly suits our quest to share the new species of *agapornis* with the world! Namely, $D^{(*)}$ could be the five photographs of the new species of *agapornis* that we managed to take. If $x_*^{(*)}$ is then anyone's photo of some *agapornis*, the meta-learning algorithm would tell whether the photo $x_*^{(*)}$ is of the new species. Note that the tasks $D_{\text{related}}$ are related—all recognising species of genea of birds—but perhaps not directly relevant for recognising *agapornes*. Meta-learning algorithms attempt to leverage both $D_{\text{related}}$ and $D^{(*)}$ to estimate the output $y_*^{(*)}$ corresponding to $x_*^{(*)}$. This is not always straightforward, because the observed tasks in $D_{\text{related}}$ may vary in how much they inform the new data set $D^{(*)}$ at hand.

Observing the outcome for the tasks in $D_{\text{related}}$ will allow the meta-learning algorithm to improve its own learning mechanism, to *learn to learn*. The algorithm will attempt to predict $y_*^{(1)}$ for $x_*^{(1)}$ by learning from $D^{(1)}$, and then slightly adjust itself based on how well it did (*learning to learn*). With the improved settings, it will attempt to predict $y_*^{(2)}$ for $x_*^{(2)}$ by learning from $D^{(2)}$, and again adjust itself based on how well it did (*learning to learn*). *Et cetera.* After the meta-algorithm has improved itself sufficiently many times, it is finally ready to predict $y_*^{(*)}$ for $x_*^{(*)}$ by learning from $D^{(*)}$. To explain how a meta-learning algorithm achieves this self-improving behaviour, we first consider what a supervised learning algorithm would do.

Table 1.1: Comparison of the settings of supervised learning and meta-learning. In supervised learning, there is one data set $D$, and one wishes to predict the output $y_*$ for a new input $x_*$. In meta-learning, there is a collection of data sets $D_{\text{related}}$. For a new data set $D^{(*)}$, one wishes to predict the output $y_*^{(*)}$ for a new input $x_*^{(*)}$.

| Setting | Observed | New | To predict |
|---|---|---|---|
| Supervised learning | $\{(x_1, y_1), \ldots, (x_N, y_N)\}$ | $x_*$ | $y_*$ |
| Meta-learning | $\{(D^{(1)}, x_*^{(1)}, y_*^{(1)}), \ldots, (D^{(M)}, x_*^{(M)}, y_*^{(M)})\}$ | $(D^{(*)}, x_*^{(*)})$ | $y_*^{(*)}$ |

For task $m$, to predict $y_*^{(m)}$ for $x_*^{(m)}$ by learning from $D^{(m)}$, a supervised learning algorithm might parametrise a function $f_\theta$ and learn its parameters from the data $D^{(m)}$:

$$D^{(m)} \rightarrow \hat{\theta} \in \arg\min_\theta \frac{1}{N} \sum_{n=1}^{N} \ell(f_\theta(x_n^{(m)}), y_n^{(m)}) \rightarrow f_{\hat{\theta}}(x_*^{(m)}) \quad \uparrow x_*^{(m)} \tag{1.5}$$

where we recall that $f_{\hat{\theta}}(x_*^{(m)})$ is the prediction for $y_*^{(m)}$. Equation (1.5) can be interpreted as a *learning pipeline*. This pipeline takes in $D^{(m)}$ and $x_*^{(m)}$ and produces a prediction for $y_*^{(m)}$. The goal of a meta-learning algorithm is to automatically improve this learning pipeline, to *learn to learn*. To enable this, meta-learning algorithms propose something radical. They propose to *replace the optimisation procedure in* (1.5) *with a learnable function* $\pi_\psi$ depending on some new parameters $\psi$:

$$D^{(m)} \rightarrow \pi_\psi(x_*^{(m)}, D^{(m)}) \quad \uparrow x_*^{(m)} \tag{1.6}$$

where $\pi_\psi(x_*^{(m)}, D^{(m)})$ now directly predicts $y_*^{(m)}$. We call $\pi_\psi$ the meta-learning algorithm.

As an example, the supervised learning function $f_\theta$ could be a neural network which takes in an image $x_*^{(m)}$ and which produces a probability for every class that $x_*^{(m)}$ could belong to.

The weights of the neural network, $\theta$, are learned by comparing the predictions of the neural network for examples in $D^{(m)}$ to the true classes. In comparison, a meta-learning approach would parametrise a neural network $\pi_\psi$ which takes in $x_*^{(m)}$ *and the observed examples* $D^{(m)}$. That is, $D^{(m)}$ is *not* used to learn the weights of the neural network; rather, just like the image $x_*^{(m)}$, the observed examples $D^{(m)}$ *are fed directly to the network as an input.*

To learn the parameters of $\psi$ of the meta-learning algorithm $\pi_\psi$, we optimise its performance on the observed tasks in $D_{\text{related}}$:

$$\hat\psi \in \underbrace{\arg\min_\psi \frac{1}{M} \sum_{m=1}^M \ell(\overbrace{\pi_\psi(x_*^{(m)}, D^{(m)})}^{\text{learning for task } m}, y_*^{(m)})}_{\text{learning to learn}} \qquad (1.7)$$

This optimisation illustrates that meta-learning algorithms learn at two scales: at the scale of every task, attempting to solve that task (*learning*); and across tasks, optimising $\psi$ to become better at solving any one task (*learning to learn*).

In summary, a meta-learning algorithm directly parametrises the learning pipeline with a function $\pi_\psi$. This function $\pi_\psi$ takes in the new input $x_*^{(m)}$ *and* the observed data $D^{(m)}$ and gives back a prediction for $y_*^{(m)}$. This is unlike in supervised learning, where only $x_*^{(m)}$ is an input to $f_\theta$. By adjusting the parameters $\psi$, the meta-algorithm can improve its own learning mechanism.

## 1.3 Main Contribution

Practical meta-learning algorithms implement $\pi_\psi$ in various ways. Neural processes (NPs; Garnelo et al., 2018a; Garnelo et al., 2018b), albeit a fairly recently introduced technique, take the seemingly simple approach of directly parametrising $\pi_\psi$ with a neural network. But this is not so straightforward, for how do we parametrise a neural network that can take a data set $D$ as an input? Namely, the number of data points in a data set is variable, so the neural network needs to accept an input of variable dimensionality. Moreover, the order in which we observe data points should not matter, so the neural network needs to not depend on the ordering of $D$. Neural processes use an existing tool from the literature to address these issues (Zaheer et al., 2017; Edwards et al., 2017).

The bigger challenge, however, that neural processes face is that learning the mapping from a data set $D$ and a new input $x_*$ to a prediction for $y_*$ can be very data intensive, simply because there are *so many* possible phenomena to predict. To appreciate this, consider the meta-learning problem where we learn to predict the trajectory of a thrown ball. Let us build an imaginary collection of tasks by throwing a ball under some random angle and with some random momentum at various points of time in the day, like at 10:00, 11:00, and 14:00. For every throw, we capture the position of the ball at two different times. For example, if we throw the ball at time $t = 10{:}00$, we could record the position $\mathbf{x}_t$ of the ball at times $t = 10{:}01$ and $t = 10{:}02$. This makes a mini data set $D = \{(10{:}01, \mathbf{x}_{10:01}), (10{:}02, \mathbf{x}_{10:02})\}$, which would be one of the tasks. The neural process will be trained by observing a few full trajectories, and must then use $D$ to predict the trajectory of a new throw.

We could solve this prediction problem with physics, though it would be hard to take into account all external circumstances that might affect the trajectory, such as the weather. Nevertheless, we expect that physics would give a reasonable prediction. The hope is that, after observing a few full trajectories, the neural process would learn to also apply the laws of physics. The problem is that the neural process does not know that we live in a universe with fixed laws. It does not know that a ball thrown at 10:00 is subject to the same laws as a ball thrown at 11:00! For what if the laws of the universe had changed between 10:00 and 11:00? As a consequence, the neural process will need to establish the laws of physics over and over again, for every point in time. Clearly, this is not very data efficient.

This thesis attempts, in a limited way, to build fundamental laws of physics, like conservation of energy and conservation of momentum, into the neural process. Building in these laws will greatly improve data efficiency and consequently improve predictions. To build in conservation laws, we will use a deep result called Noether's theorem (Noether, 1918). Noether's theorem says that conservation of energy is a consequence of *time-translation equivariance* and that conservation of moment is a consequence of *space-translation equivariance*. Time-translation equivariance says that it does not matter whether you throw the ball at 10:00 or at 11:00: if the angle, momentum, and all weather conditions are the same, then the resulting trajectory should be the same. Phrased differently, if you delay the throw by an hour, then you delay the trajectory by an hour. Similarly, space-translation equivariance says that if you move forward 10 meters, then the trajectory should be repositioned by 10 meters as well. In essence, these symmetries say that the laws of physics do not change with time or space. The main contribution of this thesis is to build a general notion of translation equivariance into a neural process.

## 1.4  Historical Context and Positioning

In 1987, Schmidhuber first put forward the idea of meta-learning by proposing algorithms which can improve their own learning strategy (Schmidhuber, 1987; Schmidhuber, 1992; Schmidhuber, 1993). The idea of giving a system the ability to learn its learning mechanism then took hold and was further developed. Y. Bengio et al. (1990) and S. Bengio et al. (1995) proposed the idea of directly parametrising a learning rule, and later Hochreiter et al.

(2001) and Younger et al. (2001) proposed the presently ubiquitous approach of performing meta-learning with gradient descent, with a reference to this idea dating back to 1991 by Schmidhuber. In 1998, Thrun et al.'s treatise on *learning to learn* put the field of meta-learning on solid grounding.

Meta-learning is a varied and diverse field, especially nowadays, and a plethora of modern methods have been developed (Andrychowicz et al., 2016; Vinyals et al., 2016; Ravi et al., 2017; Edwards et al., 2017; Snell et al., 2017; Finn et al., 2017; Garnelo et al., 2018a; Gordon et al., 2019; Requeima et al., 2019a). In this thesis, we shall only be concerned with neural processes, and we shall exclusively consider the setting of spatial, temporal, or spatio–temporal regression. Whereas other existing meta-learning techniques could be adjusted to compete with neural processes in the applications that we will consider, this would require additional research, so we limit our scope here. The most relevant coexisting line of work are gradient-based methods (Finn et al., 2017; Finn et al., 2018b), which come with universality guarantees like neural processes do (Finn et al., 2018a). See Section 7.2.1 by Gordon (2020) for a more detailed comparison of neural processes and gradient-based meta-learning.

## 1.5   Outline of Thesis

This thesis is the result of wonderful collaborations with Jonathan Gordon, Andrew Y. K. Foong, James Requeima, Stratis Markou, Anna Vaughan, Yann Dubois, and my supervisor Richard E. Turner (Gordon et al., 2020; Foong et al., 2020; Bruinsma et al., 2021c; Markou et al., 2022). The results in the chapters are often in a changed and extended form and, compared to these publications, are not presented in the order that they originally appeared.

In Chapter 2, we will introduce meta-learning and neural processes on a more technical level. This introduction will be centred around the concept of a *prediction map*: a mapping from data sets to stochastic processes. The concept of a prediction map was first introduced by Foong et al. (2020).

Chapter 3 develops a theoretical framework to rigorously analyse neural processes. Amongst other definitions, we propose the concept of a *neural process approximation* (NPA; Definition 3.15), which defines the object that a neural process converges to in the limit of infinite data. In Proposition 3.26, we characterise the NPA for the class of conditional neural processes (CNPs; Garnelo et al., 2018a); and in Proposition 3.27, we characterise the NPA for the class of Gaussian neural processes (GNPs; Section 5.5). On a theoretical level, these characterisations enable us to generally reason about the behaviour of CNPs and GNPs. On the practical side, these characterisations allow us to assess convergence of CNPs and GNPs by describing what these classes should converge to. In Section 3.5, we briefly touch upon consistency of neural processes, proving preliminary consistency results for CNPs (Proposition 3.34) and for GNPs (Proposition 3.35). The analysis in this chapter is not published, but builds on the analysis by Bruinsma et al. (2021c).

In Chapter 4, we develop two new representation theorems. In the context of neural processes, representation theorems generally characterise functions on data sets. We extend deep sets (Theorem 4.5; Zaheer et al., 2017; Edwards et al., 2017; Wagstaff et al., 2019) to *convolutional deep sets*, Theorems 4.8 and 4.9. Theorem 4.8 characterises functions on data sets which are *translation equivariant* (Definition 2.4), and Theorem 4.9 characterises functions on data sets which are *diagonally translation equivariant* (Definition 4.10). Theorem 4.8 was first presented by Gordon (2020) and Theorem 4.9 by Bruinsma et al. (2021c).

In Chapter 5, we construct new neural process models by applying representation theorems to the prediction map. First, we propose the class of *convolutional neural processes* (ConvNPs; Section 5.3). ConvNPs exploit stationarity of the data by parametrising a translation-equivariant prediction map (Proposition 5.2) with convolutional deep sets. In particular, we propose the *Convolutional Conditional Neural Process* (ConvCNP; Model 5.4), a translation-equivariant extension of the original Conditional Neural Process (CNP; Garnelo et al., 2018a). Second, we propose the class of *Gaussian neural processes* (GNPs; Section 5.5). GNPs model dependencies between target outputs by directly parametrising the covariance between target points. Within this class, we propose the *Convolutional Gaussian Neural Process* (ConvGNP; Model 5.12), an extension of the ConvCNP. We also propose the *Fully Convolutional Gaussian Neural Process* (FullConvGNP; Model 5.13). The FullConvGNP fixes representational capacity issues of the ConvGNP at the cost of increased computational expense. Finally, we propose the class of *autoregressive conditional neural processes* (AR CNPs; Section 5.6). AR CNPs deploy CNPs in an autoregressive fashion (Procedure 5.14) to obtain dependent and non-Gaussian predictions. The ConvCNP was first presented by Gordon et al. (2020), the ConvGNP by Markou et al. (2022), and the FullConvGNP by Bruinsma et al. (2021c). AR CNPs have not been published.

In Chapter 6, we put these new models to the test. We first perform a large-scale bake-off to establish the general weaknesses and strengths of five existing and eight newly proposed neural process models. Afterwards, we perform three experiments involving real-world data, demonstrating that neural processes can be deployed in a variety of settings. In the last of these experiments, we use the ConvCNP, ConvGNP, and AR ConvCNP for statistical downscaling (Maraun et al., 2018), extending the setup by Vaughan et al. (2022) to models that can produce coherent samples. These experimental results have not been published, but build on experiment setups by Gordon et al. (2020), Foong et al. (2020), Bruinsma et al. (2021c), and Markou et al. (2022).

Finally, Chapter 7 presents a simple software abstraction that enables a compositional approach to implementing neural processes. This approach allows the user to rapidly explore neural processes models by putting together elementary building blocks in different ways. The software abstraction forms the basis of a Python package `neuralprocesses` (Bruinsma et al., 2022a) available at https://github.com/wesselb/neuralprocesses. `neuralprocesses` was used to implement all models in this thesis and perform all experiments in Chapter 6. The software abstraction was conceived in `NeuralProcesses.jl` in collaboration with Jonathan Gordon (Bruinsma et al., 2022b). `neuralprocesses` is primarily developed by the author, but features contributions from Tom Andersson, Stratis Markou, and James Requeima.

## 1.6  List of Publications and Software

This section lists publications and software the author published during the course of the PhD. The authors of software are the contributors of the repository ordered by additions at the listed time.

**Peer-Reviewed Publications**

W. P. Bruinsma, M. Tegnér, and R. E. Turner (2022f). "Modelling Non-Smooth Signals With Complex Spectral Structure". In *Proceedings of the 25th International Conference on Artificial Intelligence and Statistics*. Proceedings of Machine Learning Research. PMLR. Electronic print: https://arxiv.org/abs/2203.06997.

B. Coker, W. P. Bruinsma, D. R. Burt, W. Pan, and F. Doshi-Velez (2022). "Wide Mean-Field Bayesian Neural Networks Ignore the Data". In *Proceedings of the 25th International Conference on Artificial Intelligence and Statistics*. Proceedings of Machine Learning Research. PMLR. Electronic print: https://arxiv.org/abs/2202.11670.

S. Markou, J. Requeima, W. P. Bruinsma, A. Vaughan, and R. E. Turner (2022). "Practical Conditional Neural Processes Via Tractable Dependent Predictions". In *Proceedings of the 10th International Conference on Learning Representations*. Electronic print: https://arxiv.org/abs/2203.08775.

A. Y. K. Foong, W. P. Bruinsma, D. R. Burt, and R. E. Turner (2021). "How Tight Can PAC-Bayes be in the Small Data Regime?" In *Advances in Neural Information Processing Systems 34*. Curran Associates, Inc. Electronic print: https://arxiv.org/abs/2106.03542.

A. Y. K. Foong, W. P. Bruinsma, J. Gordon, Y. Dubois, J. Requeima, and R. E. Turner (2020). "Meta-Learning Stationary Stochastic Process Prediction With Convolutional Neural

**Lightly Peer-Reviewed Workshop Submissions**

**Other Software**

# 2 | Neural Processes

**Abstract.** This chapter is a technical introduction to neural processes. It also introduces the idea of translation equivariance, which the remainder of this thesis builds upon.

**Outline.** In Section 2.1, we coin the concept of *prediction maps* and introduce meta-learning from this prespective. Then, in Section 2.2, we introduce neural processes; and in Section 2.3, we investigate consistency requirements for neural processes. In Section 2.4, we open up a neural process, explaining the general anatomy of neural process architectures. Finally, in Section 2.5, we introduce the notion of a translation-equivariant neural process.

**Attributions and relationship to prior work.** Prediction maps and translation equivariance of prediction maps were first considered by Foong, Bruinsma, Gordon, Dubois, Requeima, and Turner (2020) and later further analysed by Bruinsma, Requeima, Foong, Gordon, and Turner (2021c).

## 2.1 Prediction Maps

Neural processes are a meta-learning algorithm with the distinguishing feature that they produce internally consistent predictions. In this section, we will make this precise with the notion of a *consistent* meta-learning algorithm, and we will use this notion to introduce *prediction maps*. The concept of prediction maps captures the essence of neural processes. Chapter 3 will use prediction maps to engage in a rigorous theoretical analysis of neural processes, and Chapter 5 will parametrise prediction maps to construct practical neural process architectures. To begin with, we generally present the meta-learning setting that neural processes operate in.

For simplicity, assume that data points have one-dimensional inputs and one-dimensional outputs. Let $\mathcal{X} \subseteq \mathbb{R}$ be a compact input space and let $\mathcal{Y} = \mathbb{R}$ be the output space. Let $\mathcal{D}_N = (\mathcal{X} \times \mathcal{Y})^N$ be the collection of all $N$ input–output pairs, and let $\mathcal{D} = \bigcup_{N=0}^{\infty} \mathcal{D}_N$ be the collection of all finite numbers of input–output pairs, which includes the empty set $\varnothing$. We call elements of $\mathcal{D}$ *data sets*.[1] For a data set $D \in \mathcal{D}^N$, denote $D = (\mathbf{x}, \mathbf{y})$ where $\mathbf{x} \in \mathcal{X}^N$ is the concatenation of the inputs and $\mathbf{y} \in \mathcal{Y}^N$ the concatenation of the outputs. For a vector $\mathbf{z}$, let $|\mathbf{z}|$ denote its number of elements.

---

[1] A data set should not depend on the order of the data points. It would therefore be apt to identify data sets whose input–output pairs agree up to a permutation. We will return to this issue in Chapter 4.

In the meta-learning setting, we are given a collection of data sets $(D_m)_{m=1}^M$ (Vinyals et al., 2016; Ravi et al., 2017). This collection of data sets is called a *meta–data set*, and the individual data sets $D_m$ are called *tasks*. Every task $D_m$ is split up $D_m = D_m^{(c)} \cup D_m^{(t)}$ into a *context set* $D_m^{(c)} = (\mathbf{x}_m^{(c)}, \mathbf{y}_m^{(c)})$ and a *target set* $D_m^{(t)} = (\mathbf{x}_m^{(t)}, \mathbf{y}_m^{(t)})$. Here $\mathbf{x}_m^{(c)}$ are called the *context inputs*, $\mathbf{y}_m^{(c)}$ the *context outputs*, $\mathbf{x}_m^{(t)}$ the *target inputs*, and $\mathbf{y}_m^{(t)}$ the *target outputs*. The goal of meta-learning is to devise an algorithm which takes in a context set $D_m^{(c)}$ and which produces the best possible prediction for the corresponding target set $D_m^{(t)}$, by which we mean a prediction for the target outputs $\mathbf{y}_m^{(t)}$ given the target inputs $\mathbf{x}_m^{(t)}$. This means that what goes into meta-learning algorithm is a context set $D_m^{(c)}$ and some target inputs $\mathbf{x}_m^{(t)}$ and that what comes out is a prediction for $\mathbf{y}_m^{(t)}$. If the inputs $\mathbf{x}$ are images, the outputs $\mathbf{y}$ are categories, and the number of context data is small, then this setting is called *few-shot image classification* (Fei-Fei et al., 2006; Lake et al., 2015). We, however, shall be concerned with low-dimensional inputs and real-valued outputs.

This thesis focusses on *probabilistic* meta-learning algorithms. Such algorithms take in a context set $D_m^{(c)}$ and the target inputs $\mathbf{x}_m^{(t)}$ and produce a *probability distribution* for $\mathbf{y}_m^{(t)}$. This, however, immediately raises concerns. As we show now, probabilistic meta-learning algorithms can produce the same prediction for $\mathbf{y}_m^{(t)}$ in multiple ways, and there is no guarantee that these predictions are consistent with each other. For notational convenience, let us denote the context set by $D$, the target inputs by $\mathbf{x}$, and the target outputs by $\mathbf{y}$. Suppose that we have just two target inputs and two target outputs: $\mathbf{x} = (x_1, x_2)$ and $\mathbf{y} = (y_1, y_2)$. By running the meta-learning algorithm on $(D, (x_1, x_2))$, we obtain a prediction for $(y_1, y_2)$. What we can now do is to *discard* the prediction for $y_2$, which means that we are left with just a prediction for $y_1$. However, we could also have obtained a prediction for $y_1$ by running the meta-learning algorithm on just $(D, x_1)$, and there no guarantee that these predictions will be the same! Additionally, suppose that we run the meta-learning algorithm on $(D, (x_2, x_1))$, swapping around $x_2$ and $x_1$. We then obtain a prediction for $(y_2, y_1)$. By then swapping around the dimensions of the prediction, we obtain a prediction for $(y_1, y_2)$. Again, we could also have obtained a prediction for $(y_1, y_2)$ by running the meta-learning on just $(D, (x_1, x_2))$, and again there is no guarantee that these predictions would line up. We call a probabilistic meta-learning algorithm *consistent* if it always produces the same prediction, regardless of whether you *discard* (*marginalise*) or *permute* target inputs and outputs.

The discussion of consistency does not end here. By feeding the output of a meta-learning algorithm back into itself, it is possible to come up with more consistency requirements. We will come back to this in Section 2.3. For now, it suffices to define consistency in the above sense: predictions do not depend discarding or permuting target inputs and outputs.

Consider a consistent probabilistic meta-learning algorithm. It turns out that consistency is very naturally satisfied. For a context set $D$ and target inputs $\mathbf{x}$, denote the prediction for $\mathbf{y}$ by $\pi_{\mathbf{x}}(D)$, which is a probability distribution. (The reason for denoting this distribution as a function of $D$ will become clear shortly.) Then consistency of the meta-learning algorithm implies that the collection of $\{\pi_{\mathbf{x}}(D) : \mathbf{x} \in \mathbb{R}^N,\ N \in \mathbb{N}\}$ is *consistent under marginalisation* and *consistent under permutations*. Consistency under marginalisation means that, for all $N_1, N_2 \in \mathbb{N}$, inputs $\mathbf{x}_1 \in \mathcal{X}^{N_1}$, inputs $\mathbf{x}_2 \in \mathcal{X}^{N_2}$, and Borel sets $B_1 \in \mathcal{B}(\mathcal{Y}^{N_1})$,

$$\pi_{\mathbf{x}_1 \oplus \mathbf{x}_2}(D)(B_1 \times \mathcal{Y}^{N_2}) = \pi_{\mathbf{x}_1}(D)(B_1). \tag{2.1}$$

where $\mathbf{x}_1 \oplus \mathbf{x}_2$ concatenates $\mathbf{x}_1$ and $\mathbf{x}_2$. Consistency under permutations means that, for all $N \in \mathbb{N}$, inputs $\mathbf{x} \in \mathcal{X}^N$, Borel sets $B_1, \ldots, B_N \in \mathcal{B}(\mathcal{Y}^N)$, and permutations $\sigma \in \mathbb{S}^N$,

$$\pi_{x_{\sigma(1)},\ldots,x_{\sigma(N)}}(D)(B_{\sigma(1)} \times \cdots \times B_{\sigma(N)}) = \pi_{x_1,\ldots,x_N}(D)(B_1 \times \cdots \times B_N). \tag{2.2}$$

If consistency under marginalisation and consistency under permutations are satisfied, then Kolmogorov's extension theorem (Theorem 12.1.2; Dudley, 2002) implies that there exists a $\mathcal{Y}$-valued stochastic process on $\mathcal{X}$ such that every finite-dimensional distribution (f.d.d.) at $\mathbf{x}$ is equal to $\pi_{\mathbf{x}}(D)$. Denote this stochastic process by $\pi(D)$ (no subscript).

Let $\mathcal{P}$ be the collection of all $\mathcal{Y}$-valued stochastic processes on $\mathcal{X}$. We have seen that, for every context set $D \in \mathcal{D}$, a consistent probabilistic meta-learning algorithm produces a stochastic process $\pi(D) \in \mathcal{P}$. This means that every consistent probabilistic meta-learning algorithm is in correspondence with a map $\pi \colon \mathcal{D} \to \mathcal{P}$. The map $\pi \colon \mathcal{D} \to \mathcal{P}$ may seem like a complicated object, but it captures a simple and important idea. For every data set $D \in \mathcal{D}$, the map $\pi \colon \mathcal{D} \to \mathcal{P}$ produces a stochastic process $\pi(D) \in \mathcal{P}$. We argue that this stochastic process can be interpreted very simply as *a prediction*. "A prediction for what?" you may ask. Well, the stochastic process $\pi(D)$ has a finite-dimensional distribution for all possible target inputs $\mathbf{x}$, so $\pi(D)$ implies a prediction for any choice of target inputs $\mathbf{x}$. In particular, the stochastic process $\pi(D)$ does not depend on any choice of target inputs $\mathbf{x}$; instead, $\pi(D)$ predicts *everywhere*. The idea of letting a prediction be represented by a stochastic process takes the target inputs out of the equation and therefore simplifies the setup. Under this interpretation, consistent probabilistic meta-learning algorithms can be interpreted, very simply, as maps from data sets $\mathcal{D}$ to predictions $\mathcal{P}$. We capture this idea in the following definition.

**Definition 2.1** (Prediction map). *A prediction map $\pi$ is a map $\pi \colon \mathcal{D} \to \mathcal{P}$.*

In summary, a probabilistic meta-learning algorithm is a map from context sets $D$ and target inputs $\mathbf{x}$ to probability distributions for the target outputs $\mathbf{y}$. We call a probabilistic meta-learning algorithm *consistent* if predictions do not depend on the way they are produced. Consistency is a desirable property and turns out to be naturally satisfied: every consistent probabilistic meta-learning algorithm is in correspondence with a *prediction map*.

In this section, we have seen what a desirable probabilistic meta-learning algorithm *is*, namely, a prediction map. In the next section, we will introduce neural processes as an approach that is fundamentally based on prediction maps.

## 2.2 Neural Processes

Neural processes (Garnelo et al., 2018a; Garnelo et al., 2018b) approach a meta-learning problem by directly parametrising a prediction map using neural networks (McCulloch et al., 1943; Rosenblatt, 1958; Ivakhnenko et al., 1965; Fukushima et al., 1982; Werbos, 1982; LeCun et al., 1989). We shall not yet be concerned with how this parametrisation works. What is important is that the parametrisation is direct and simple.

Although we will not yet discuss precisely how neural processes parametrise prediction maps, to appreciate the elegance of the approach, it is helpful to keep the following example in mind:

$$\pi_\theta(D) = x \mapsto \mathrm{dec}_\theta(x, \mathbf{z}) \qquad \text{where} \qquad \mathbf{z} = \sum_{(x,y) \in D} \phi_\theta(x, y) \tag{2.3}$$

where $\phi_\theta \colon \mathcal{X} \times \mathcal{Y} \to \mathbb{R}^K$ is a neural network operating on every context data point $(x, y) \in D$ and $\mathrm{dec}_\theta \colon \mathcal{X} \times \mathbb{R}^K \to \mathcal{Y}$ is another neural network mapping a target input $x$ and the vector $\mathbf{z}$ to the predicted value. The parameters $\theta$ are the weights of the two neural networks. The notation in (2.3) means that $\pi_\theta(D)$ returns the deterministic function $x \mapsto \mathrm{dec}_\theta(x, \mathbf{z})$ as the prediction. Note that this example does not fully exploit the prediction map formalism, because $\pi_\theta(D)$ may return a random function, that is, a stochastic process. We will return to the general form of neural process architectures in Section 2.4.

The parametrisation of the prediction map by a neural process is often not able to produce every possible stochastic process. For example, a large class of neural processes is restricted to producing only Gaussian processes. We therefore define a neural process as a map $\pi_\theta \colon \mathcal{D} \to \mathcal{Q}$, where $\mathcal{Q} \subseteq \mathcal{P}$ is called the *variational family* and $\theta \in \Theta$ are the parameters of the neural process. The variational family $\mathcal{Q}$ determines the approximation properties of the neural process and can be interpreted just like the variational family in variational inference (Wainwright et al., 2008). By ranging over $\theta \in \Theta$, an equivalent definition of a neural process is that it is a collection of prediction maps $\{\pi_\theta : \theta \in \Theta\}$. This is very similar to the definition of a statistical model, a collection of probability distributions $\{\mathbb{P}_\theta : \theta \in \Theta\}$. By considering $\pi_\theta$ for every $\theta \in \Theta$ to be an hypothesis, one could also call $\{\pi_\theta : \theta \in \Theta\}$ the *hypothesis class*, which is more standard in learning theory.

Following more conventional notation, for all data sets $D \in \mathcal{D}$ and inputs $\mathbf{x} \in \mathbb{R}^N$, we will denote the density of the finite-dimensional distribution of $\pi_\theta(D)$ at $\mathbf{x}$ with respect to the Lebesgue measure by $q_\theta(\,\cdot\,|\,\mathbf{x}, D)$, assuming that this density exists.

To learn the parameters $\theta$, neural processes propose an objective which maximises the likelihood of the target sets under the predictions given the context sets:

$$\hat{\theta} \in \arg\max_{\theta \in \Theta} \frac{1}{M} \sum_{m=1}^{M} \log q_\theta(\mathbf{y}_m^{(\text{t})} \,|\, \mathbf{x}_m^{(\text{t})}, D_m^{(\text{c})}). \tag{2.4}$$

This objective was originally proposed by Garnelo et al. (2018a) and also considered by Gordon et al. (2019). Throughout this thesis, we will call it the *empirical neural process objective* or just the *neural process objective* when "empirical" is clear from the context.

**Definition 2.2** (Empirical neural process objective). *The* empirical neural process objective *is given by*

$$\mathcal{L}_M(\pi_\theta) = -\frac{1}{M} \sum_{m=1}^{M} \log q_\theta(\mathbf{y}_m^{(\text{t})} \,|\, \mathbf{x}_m^{(\text{t})}, D_m^{(\text{c})}) \tag{2.5}$$

*where $q_\theta(\,\cdot\,|\,\mathbf{x}, D)$ is the density of the finite-dimensional distribution of $\pi_\theta(D)$ at $\mathbf{x}$ with respect to the Lebesgue measure, assuming that this density exists.*

For a general $\mathcal{Q}$, note that the density $q_\theta(\,\cdot\,|\,\mathbf{x}_m^{(\text{t})}, D_m^{(\text{c})})$ might not be tractable, which means that the empirical neural process objective cannot always be evaluated exactly.

The class of neural processes proposed originally by Garnelo et al. (2018a) is the class of *conditional neural processes* (CNPs). CNPs choose $\mathcal{Q}$ to be the collection of all Gaussian processes which *do not model dependencies between target outputs*. This means that the prediction of a CNP is independent at any two different target inputs. For CNPs, the density $q_\theta(\,\cdot\,|\,\mathbf{x}_m^{(\text{t})}, D_m^{(\text{c})})$ is Gaussian, so the neural process objective can be evaluated exactly. Examples of CNPs are the original Conditional Neural Process (CNP; Garnelo et al., 2018a), the Attentive Conditional Neural Process (ACNP; Kim et al., 2019), the Convolutional Conditional Neural Process (ConvCNP; Section 5.3; Gordon et al., 2020), the Group-Equivariant Conditional Neural Process (EquivCNP; Kawano et al., 2021), and the Steerable Convolutional Conditional Neural Processes (SteerCNP; Holderrieth et al., 2021). The ConvCNP is introduced in this thesis in Section 5.3, and the EquivCNP and SteerCNP are both models based on the ConvCNP.

**A word of caution.** There are many neural processes and many classes of neural processes which all have similar names. For example, there is the class of conditional neural processes (CNPs), and there is the Conditional Neural Process (CNP). Even though these names differ only by two characters, they are not the same! The Conditional Neural Process is a specific model in the class of conditional neural processes. We admit that this naming might be confusing, but it is what is used in the literature, so we will stick to it. To help the reader, classes of neural processes will always be lower case (*e.g.*, "conditional neural processes") and the abbreviation will always end with an "s" (*e.g.*, "CNPs"). Specific neural process models, on the other hand, will always be capitalised (*e.g.*, "the Conditional Neural Process") and the abbreviation will never end with an "s" (*e.g.*, "the CNP").

In addition to CNPs, another commonly encountered class is the class of *latent-variable neural processes* (LNPs; Garnelo et al., 2018b). LNPs use a latent variable to induce a variational family $\mathcal{Q}$ of *non-Gaussian processes*. Importantly, unlike CNPs, LNPs *do* model dependencies between target outputs. For LNPs, to optimise the neural process objective, approximations are necessary. Examples of LNPs are the Neural Process (NP; Garnelo et al., 2018b), the Attentive Neural Process (ANP; Kim et al., 2019), the Functional Neural Process (FNP; Louizos et al., 2019), the Sequential Neural Process (SNP; G. Singh et al., 2019), and the Convolutional Neural Process (CovnNP; Foong et al., 2020). The ConvNP is derived from the ConvCNP, but not introduced in this thesis.

We have discussed CNPs, which choose $\mathcal{Q}$ to be the collection of *Gaussian* processes which *do not* model dependencies between target outputs; and LNPs, which choose $\mathcal{Q}$ to be a collection of *non-Gaussian* processes which *do* model dependencies between target outputs. One might wonder about the variational family $\mathcal{Q}$ of *Gaussian* processes which *do* model dependencies between target outputs. This class of neural processes is introduced in Section 5.5 and called the class of *Gaussian neural processes* (GNPs). GNPs have the unique ability to model dependencies between target outputs without requiring approximations to evaluate the neural process objective. Like for CNPs, for GNPs, the neural process objective can be evaluated exactly. Examples of GNPs are the Gaussian Neural Process (GNP; Chapter 6; Markou et al., 2022), the Attentive Gaussian Neural Process (AGNP; Chapter 6; Markou et al., 2022), the Convolutional Gaussian Neural Process (ConvGNP; Section 5.5; Markou et al., 2022), and the Fully Convolutional Gaussian Neural Process (FullConvGNP; Section 5.5; Bruinsma et al., 2021c). All these models are introduced in this thesis.

A helpful hierarchical organisation of neural process models, which extends the division of neural processes into CNPs and LNPs, is the Neural Process Family (Gordon, 2020; Dubois et al., 2020). The class of GNPs forms a new subfamily of the Neural Process Family.

## 2.3 More on Consistency

In Section 2.1, we defined a meta-learning algorithm to be *consistent* if predictions do not depend on whether you *discard* (*marginalise*; see (2.1)) or *permute* (see (2.2)) target inputs and outputs. Let us call consistency in this sense *consistency with respect to the target set*. As we then alluded to, by feeding the output of the meta-learning algorithm back into itself, it is possible to come up with more consistency requirements.

In this section, we define an additional consistency requirement called *consistency with respect to the context set*. If a meta-learning algorithm is consistent with respect to both the target set and the context set, then we will see that the algorithm essentially performs Bayesian inference in an underlying probabilistic model. Since exact Bayesian inference is generally computationally intractable, giving up consistency with respect to the context set can therefore be interpreted as a way of circumventing these computational challenges. Consistency with respect to the context set will be an important point of discussion when we introduce the class of *autoregressive conditional neural processes* (AR CNPs) in Section 5.6.

Like in Section 2.1, consider just two target inputs and outputs: $\mathbf{x} = (x_1, x_2)$ and $\mathbf{y} = (y_1, y_2)$. To make a prediction for $(y_1, y_2)$, it is simplest to just run the meta-learning algorithm on $(D, (x_1, x_2))$. Alternatively, we could attempt a two-stage procedure: we could first run the meta-learning algorithm on $(D, x_1)$ to obtain a prediction for $y_1$, and then feed this prediction back into the algorithm to obtain a prediction for $y_2$. Let $\tilde{y}_1$ be a sample from the prediction for $y_1$, and let $D \cup (x_1, \tilde{y}_1)$ denote the context set with the data point $(x_1, \tilde{y}_1)$ appended; recall that $D$ is just a collection of input–output pairs, so it is perfectly fine to append another input–output pair. In the two-stage procedure, after obtaining a prediction for $y_1$, we run the meta-learning algorithm on $(D \cup (x_1, \tilde{y}_1), x_2)$ to obtain a prediction for $y_2$. If the prediction for $(y_1, y_2)$ produced by the two-stage procedure agrees with just running the meta-learning algorithm on $(D, (x_1, x_2))$, then we say that the algorithm is *consistent with respect to the context set*.

More precisely, in terms of the family $\{\pi_{\mathbf{x}}(D) : D \in \mathcal{D}, \mathbf{x} \in \mathbb{R}^N, N \in \mathbb{N}\}$, the meta-learning algorithm is consistent with respect to the context set if, for all $D \in \mathcal{D}$, $N_1, N_2 \in \mathbb{N}$, inputs $\mathbf{x}_1 \in \mathcal{X}^{N_1}$ and $\mathbf{x}_2 \in \mathcal{X}^{N_2}$, and Borel sets $B_1 \in \mathcal{B}(\mathcal{Y}^{N_1})$ and $B_2 \in \mathcal{B}(\mathcal{Y}^{N_2})$,

$$\int_{B_1} \pi_{\mathbf{x}_2}(D \cup (\mathbf{x}_1, \mathbf{y}_1))(B_2) \, \mathrm{d}(\pi_{\mathbf{x}_1}(D))(\mathbf{y}_1) = \pi_{\mathbf{x}_1 \oplus \mathbf{x}_2}(D)(B_1 \times B_2). \tag{2.6}$$

Note that $\pi_{\mathbf{x}_1}(D)$ is a measure, so the left-hand side of (2.6) is an integral with respect to the measure $\pi_{\mathbf{x}_1}(D)$, *i.e.* an expectation over the law $\pi_{\mathbf{x}_1}(D)$ of the random variable $\mathbf{y}_1$. The left-hand side can therefore also be written as $\mathbb{E}_{\mathbf{y}_1 \sim \pi_{\mathbf{x}_1}(D)}[\mathbb{1}_{B_1}(\mathbf{y}_1) \pi_{\mathbf{x}_2}(D \cup (\mathbf{x}_1, \mathbf{y}_2))(B_2)]$.

Consistency with respect to the context set, when combined with consistency with respect to the target set, is a very strong requirement. Consider a probabilistic meta-learning algorithm which is consistent with respect to both the target set and the context set. For every data set $D \in \mathcal{D}$, we then argued in Section 2.1 that there exists a $\mathcal{Y}$-valued stochastic process on $\mathcal{X}$ such that every f.d.d. at $\mathbf{x}$ is equal to $\pi_{\mathbf{x}}(D)$. Let us more informally denote the density of $\pi_{\mathbf{x}}(D)$ by $p_D(f(\mathbf{x}))$. Then (2.6) says that, for all context sets $D \in \mathcal{D}$, additional context inputs $\mathbf{x}^{(c)}$, and target inputs $\mathbf{x}^{(t)}$,

$$p_{D \cup (\mathbf{x}^{(c)}, \mathbf{y}^{(c)})}(f(\mathbf{x}^{(t)})) = p_D(f(\mathbf{x}^{(t)}) \mid f(\mathbf{x}^{(c)}) = \mathbf{y}^{(c)}), \qquad (2.7)$$

where the right-hand side is a conditional distribution of $p_D(f(\mathbf{x}^{(t)}), f(\mathbf{x}^{(c)}))$. Specifically, suppose that the context set were empty: $D = \varnothing$. Then, for all additional context inputs $\mathbf{x}^{(c)}$ and target inputs $\mathbf{x}^{(t)}$,

$$\text{density of } \pi_{\mathbf{x}^{(t)}}(\underbrace{(\mathbf{x}^{(c)}, \mathbf{y}^{(c)})}_{D^{(c)}}) = p_{\underbrace{(\mathbf{x}^{(c)}, \mathbf{y}^{(c)})}_{D^{(c)}}}(f(\mathbf{x}^{(t)})) = p_\varnothing(f(\mathbf{x}^{(t)}) \mid \underbrace{f(\mathbf{x}^{(c)}) = \mathbf{y}^{(c)}}_{D^{(c)}}). \qquad (2.8)$$

Since this equality holds for all target inputs $\mathbf{x}^{(t)}$, relabelling $D^{(c)}$ to $D$, we find that *there exists some underlying stochastic process $f \sim p_\varnothing(f)$ such that, for all data sets $D \in \mathcal{D}$, $\pi(D)$ is the posterior of $f$ conditioned on $D$*. This observation is surprising! A prediction map $\pi \colon \mathcal{D} \to \mathcal{P}$ is simply a map from data sets to stochastic processes without any further structure, so it is not at all necessary that $\pi$ computes posteriors of some underlying stochastic process.

To conclude, if a probabilistic meta-learning algorithm is consistent with respect to both the target set and context set, then it computes posteriors of some underlying stochastic process. In practice, designing a neural network architecture that satisfies consistency with respect to the context set is extremely difficult, so nearly all neural process approaches—if not all—are consistent only with respect to the target set. A notable exception is BRUNO (Korshunova et al., 2018; Korshunova et al., 2020), an approach which shares strong similarities with neural processes. BRUNO achieves consistency with respect to the context set by embedding a simple probabilistic model into the approach. That nearly all neural processes are inconsistent respect to the context set poses theoretical and practical challenges for the class of AR CNPs that we will introduce in Section 5.6.

Posteriors of stochastic processes are generally computationally intractable, especially if neural networks are involved. We have therefore uncovered an interesting perspective on the computational benefits of neural processes: by giving up consistency with respect to the context set, neural processes are able to circumvent the computational challenges that come with computing Bayes' rule.

## 2.4 The Anatomy of a Neural Process

We have seen that neural processes approach meta-learning problems by directly parametrising a prediction map $\pi_\theta \colon \mathcal{D} \to \mathcal{Q}$ using neural networks. Section 2.2 presented (2.3) as an example of a neural process, illustrating what a neural process architecture looks like:

$$\pi_\theta(D) = x \mapsto \mathsf{dec}_\theta(x, \mathbf{z}) \qquad \text{where} \qquad \mathbf{z} = \sum_{(x,y) \in D} \phi_\theta(x, y). \tag{2.3}$$

In this section, we will explain the general form of neural process architectures. This general form will be motivated from theoretical principles in Chapter 4.

To parametrise a prediction map, a neural process architectures needs to overcome two challenges:

1. Data sets $D \in \mathcal{D}$ have a variable number of elements, which means that the neural network must process inputs of varying dimensionality.

2. Two data sets $D_1 \in \mathcal{D}$ and $D_2 \in \mathcal{D}$ with the same data points but a different ordering of the data points should be considered the same. In other words, $\pi_\theta(D)$ should not depend on the ordering of the elements of $D$. We more formally say that $\pi_\theta$ should be *permutation invariant*.

Neural processes approach both challenges by parametrising the prediction map with a so-called *encoder–decoder architecture*. This means that $\pi_\theta$ is the composition of an *encoder* $\mathsf{enc}_\theta$ and a *decoder* $\mathsf{dec}_\theta$:

$$\pi_\theta = \mathsf{dec}_\theta \circ \mathsf{enc}_\theta. \tag{2.9}$$

When we compute $\pi_\theta(D)$, the encoder first computes an *encoding* $\mathsf{enc}_\theta(D)$, which we temporarily denote by $\mathbf{z}$. This encoding $\mathbf{z}$ is then given to the decoder, which finally computes the output, $\mathsf{dec}_\theta(\mathbf{z})$:

$$\pi_\theta(D) = \mathsf{dec}_\theta(\mathbf{z}) \qquad \text{where} \qquad \mathbf{z} = \mathsf{enc}_\theta(D). \tag{2.10}$$

The encoder should be thought of as a lightweight component which is specifically designed to address the above two challenges. The decoder, on the other hand, will be more heavyweight, containing most of the representational capacity of the neural process. To address the above two challenges, the encoder will adhere to two design requirements:

1. The encoding $\mathbf{z}$ is of a *fixed format*, regardless of the dimensionality of $D$.

2. The encoding $\mathbf{z}$ does *not depend on the ordering* of the data points in $D$.

By adhering to these two requirements, the encoder takes the problems of varying dimensionality and permutation invariance out of the way for the decoder. Therefore, in a sense, the decoder can forget that varying dimensionality and permutation invariance were a problem in the first place. Since the encoding **z** is of a fixed format, *e.g.* a fixed-dimensional vector, the decoder can process the encoding using conventional approaches, *e.g.* using a feed-forward neural network. In the example (2.3), the encoder satisfies the first requirement by mapping every data point $(x, y)$ to a fixed-dimensional vector $\phi_\theta(x, y)$ and satisfies the second requirement by summing over the data points: $\mathbf{z} = \sum_{(x,y) \in D} \phi_\theta(x, y)$.

Although many neural processes follow an encoder–decoder architecture, this is not a strict and universal rule: neural process architectures may sometimes deviate from the encoder–decoder structure. In addition, for neural processes that do follow an encoder–decoder architecture, precisely where the encoder ends and the decoder starts might sometimes be ambiguous. Nevertheless, encoder–decoder architectures are a useful way to understand the design of many neural processes, which is why we propose it as the basic mental model.

To design a neural process, one must design an encoder and a decoder. It is important that these designs appropriately balance *flexibility* and *parameter efficiency*. Namely, one can design very flexible encoders and decoders which can learn nearly every possible prediction map, but these designs will involve an inordinate number of parameters which cannot feasibly be learned from finite data. Conversely, one can design encoders and decoders depending only on a few parameters, but these designs might be too constrained to satisfactorily solve the meta-learning problem at hand. In Chapter 4, we will consider current approaches to designing encoders and decoders. We will propose new approaches which, for spatial, temporal, and spatio–temporal meta-learning problems, can make a better trade-off between flexibility and parameter efficiency. These new approaches will be based on the idea of *translation equivariance*, which we discuss next.

## 2.5 Translation Equivariance

The main contribution of this thesis is to improve parameter efficiency of neural processes by building in a symmetry called *translation equivariance*. Although translation equivariance is not appropriate for every application, when it is appropriate, it can substantially improve in-distribution and generalisation performance. In spatial, temporal, and spatio–temporal meta-learning problems, translation equivariance is often helpful. In this section, we will define translation equivariance without motivation and demonstrate, in a real example, that it can be helpful. Chapter 5 will show that translation equivariance is related to *stationarity* of the underlying stochastic process.

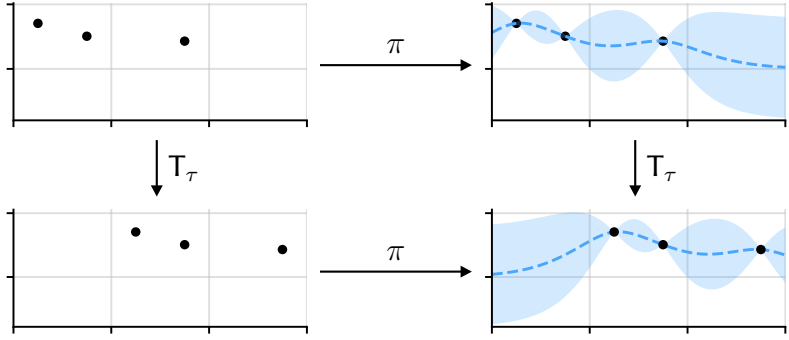

Figure 2.1: Commutative diagram illustrating translation equivariance of a neural process $\pi\colon \mathcal{D} \to \mathcal{P}$. The predictions of the neural process $\pi$ are shown in dashed blue.

**Definition 2.3** (Translation). *For $\tau \in \mathcal{X}$, let $\mathsf{T}_\tau$ denote a translation by $\tau$. Translations act on real numbers, data sets, and functions in the following way.*

- *For an input $x \in \mathcal{X}$, $\mathsf{T}_\tau x$ produces another input in $\mathcal{X}$:*

$$\mathsf{T}_\tau x = x + \tau. \tag{2.11}$$

- *For a data set $D = (\mathbf{x}, \mathbf{y}) \in \mathcal{D}$, $\mathsf{T}_\tau D$ produces another data set in $\mathcal{D}$:*

$$\mathsf{T}_\tau D = ((\mathsf{T}_\tau x_1, \ldots, \mathsf{T}_\tau x_n), \mathbf{y}). \tag{2.12}$$

- *For a function $f\colon \mathcal{X} \to Z$, $\mathsf{T}_\tau f$ produces another function $\mathcal{X} \to Z$:*

$$\mathsf{T}_\tau f(x) = f(x - \tau). \tag{2.13}$$

- *For a stochastic process $\mu \in \mathcal{P}$, $\mathsf{T}_\tau \mu$ produces another stochastic process in $\mathcal{P}$:*

$$\mathsf{T}_\tau \mu(B) = \mu(\mathsf{T}_\tau^{-1}(B)) \quad \text{for all cylinder sets } B. \tag{2.14}$$

**Definition 2.4** (Translation equivariance; TE). *Consider a map $\pi\colon A \to B$ where the elements in $A$ and $B$ can be translated. Then $\pi$ is* translation equivariant *(TE) if*

$$\pi \circ \mathsf{T}_\tau = \mathsf{T}_\tau \circ \pi \quad \text{for all } \tau \in \mathcal{X}. \tag{2.15}$$

Intuitively, if a map is translation equivariant, then, whenever the input is translated, the output is translated similarly. In the case of a neural process, translation equivariance is illustrated in Figure 2.1. In this case, the map $\pi_\theta\colon \mathcal{D} \to \mathcal{Q}$ is from data sets $\mathcal{D}$ to stochastic processes $\mathcal{Q}$. Translation equivariance then means that, whenever a data set $D$ is shifted by some amount, $\mathsf{T}_\tau D$, the corresponding prediction $\pi(\mathsf{T}_\tau D)$ is equal to the original prediction, $\pi(D)$, shifted by the same amount, $\mathsf{T}_\tau \pi(D)$. Technically, one says that application of the neural process and translation commute. Another important example of a translation equivariant map is a convolutional neural network (Fukushima et al., 1982; LeCun et al., 1989). If we pass a translated version of an input image to a CNN, then the CNN produces the original output translated by the same amount.

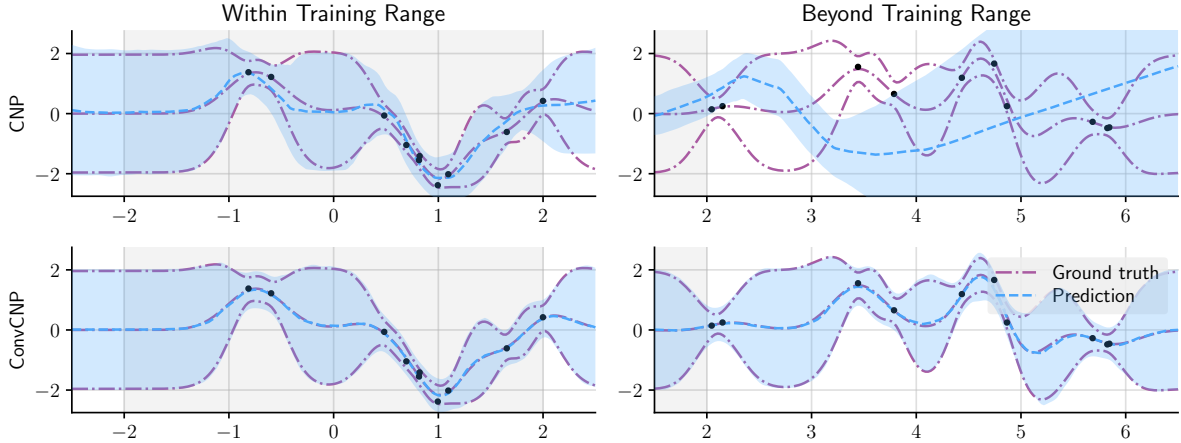

Figure 2.2: Comparison of the prediction of a trained CNP, a non-convolutional neural process, and ConvCNP, a convolutional conditional neural process. Shows predictions by the models in dashed blue and predictions by the ground truth in dot-dashed purple. The models were trained by observing data on $[-2, 2]$. We call $[-2, 2]$ the *training range*. In the plots, the training range is shaded. The left column shows predictions for observations in the training range, and the right column shows predictions for observations beyond the training range. Filled regions are central 95%-credible regions. The CNP and ConvCNP are taken from the experiment in Section 6.2.

Section 5.3 will propose general parametrisations of prediction maps $\pi_\theta \colon \mathcal{D} \to \mathcal{Q}$ that are translation equivariant. We call this class of neural processes *convolutional neural processes* (ConvNPs). Whereas non-convolutional neural processes are implemented with multi-layer perceptrons, convolutional neural processes are driven by convolutional neural networks. Examples of ConvNPs are the Convolutional Conditional Neural Process (ConvCNP; Section 5.3; Gordon et al., 2020), the Convolutional Gaussian Neural Process (ConvGNP; Section 5.5; Markou et al., 2022), and the Fully Convolutional Gaussian Neural Process (FullConvGNP; Section 5.5; Bruinsma et al., 2021c).

Figure 2.2 compares a trained CNP, a non-convolutional neural process, to a trained ConvCNP, a convolutional neural process. In the training range (see legend of Figure 2.2), the CNP shows a reasonable fit, perhaps slightly underfitting in some places. However, when it is evaluated outside the training range, the model completely breaks down. On the other hand, the ConvCNP shows a tight fit, closely recovering the ground truth, and seamlessly generalises to observations beyond the training range. This demonstrates, in a real example, that translation equivariance can improve in-distribution and generalisation performance.

## 2.6 Summary and Outlook

Section 1.5 provided a brief outline of this thesis. Whilst we summarise this chapter, we again connect to future chapters to elucidate the overarching structure in more detail.

A desirable probabilistic meta-learning algorithm is one that is *consistent (with respect to the target set)* (Section 2.1), and such a probabilistic meta-learning algorithm can be identified with a *prediction map* (Definition 2.1). This forms the basis for *neural processes*. A neural process approaches a meta-learning problem in a natural way by directly parametrising a prediction map using neural networks (Section 2.2). Compared to stochastic process models such as Gaussian processes, neural processes the circumvent computational challenges associated with computing Bayes' rule by giving up *consistency with respect to the context set* (Section 2.3). In Chapter 3, we will build on the idea of prediction maps to engage in a rigorous theoretical analysis of neural processes.

To parametrise a prediction map, neural processes often use *encoder–decoder architectures* (Section 2.4). Although using encoder–decoder architectures is not a necessity, they are a helpful mental model. In Chapter 4, we will study *representation theorems*. In the context of neural processes, representation theorems are general characterisations of functions on data sets. Representation theorems can be used to theoretically motivate encoder–decoder architectures.

One of the main contributions of this thesis is to build *translation equivariance* (Definition 2.4) into a neural process (Section 2.5). For spatial, temporal, and spatio–temporal meta-learning problems, translation-equivariant neural processes can make a better trade-off between flexibility and parameter efficiency. In Chapter 5, we will use the representation theorems from Chapter 4 to construct translation-equivariant neural processes. We call this class *convolutional neural processes* (ConvNPs). In addition to ConvNPs, Chapter 5 will also introduce the classes of *Gaussian neural processes* (GNPs) and *autoregressive conditional neural processes* (AR CNPs). GNPs have the unique ability to model dependencies between target outputs without requiring approximations for the neural process objective. AR CNPs, which we did not discuss in this chapter, trade the desirable property of consistency for better performance. Hence, AR CNPs are no longer consistent probabilistic meta-learning algorithms, but they may offer improved predictions. In Chapter 6, we will put these new models and existing approaches to the test, establishing general weaknesses and strengths.

Over the course of this chapter, one might have noticed that there are *many* flavours of neural processes. To help with the implementation of all of these flavours, Chapter 7 will present a software abstraction that enables the user to rapidly explore neural processes models by putting together elementary building blocks in different ways.

# 3 | Prediction Map Approximation

**Abstract.** This chapter presents a theoretical framework called *prediction map approximation* to rigorously analyse neural processes. The starting point is the definition of the so-called *posterior prediction map*. The posterior prediction map is the map from a data set to the posterior distribution of a stochastic process given that data set. It is the main object of interest. The premise of the framework is that neural processes form various classes of approximations of the posterior prediction map.

**Outline.** In Section 3.1, we motivate the key elements of the framework. Afterwards, in Section 3.2, we introduce the main technical concepts, setting us up for more formal analysis. In Section 3.3, we formally introduce and analyse the *neural processes objective*. As the name suggests, this is the objective that neural processes aim to optimise. Then, in Section 3.4, we define *neural process approximations*. A neural process approximation precisely defines what a class of neural processes targets. Finally, in Section 3.5, we analyse how a neural process approximation is estimated in practice. We engage in a discussion about convergence.

**Attributions and relationship to prior work.** Prediction maps and translation equivariance of prediction were first considered by Foong, Bruinsma, Gordon, Dubois, Requeima, and Turner (2020) and later further analysed by Bruinsma, Requeima, Foong, Gordon, and Turner (2021c). Although the analysis of this chapter has not been published in its current form, the analysis borrows considerably from Bruinsma et al. (2021c). The analysis by Bruinsma et al. (2021c) was primarily conducted by the author and James Requeima and checked by Jonathan Gordon and Andrew Y. K. Foong. All work was supervised by Richard E. Turner.

## 3.1 Introduction

In this chapter, we develop a theoretical framework to rigorously analyse neural processes. Building on the idea of a *prediction map* (Definition 2.1), this framework is called *prediction map approximation*. The primary goal of this chapter is an attempt to reduce the gap between the theory and practice of neural processes: to formally present neural processes to the more theoretically minded audience, and to bring subtle but important theoretical issues to the attention of the more practically oriented.

In the current literature, theoretical analyses of meta-learning use a variety of techniques to study generalisation of meta-learning algorithms: Baxter (1998) and Baxter (2000) use classical learning-theoretic techniques; Maurer (2005) investigates algorithmic stability; Pentina et al. (2014), Alquier et al. (2017), Amit et al. (2018), Yin et al. (2020), T. Liu et al. (2021), and Rothfuss et al. (2021) use PAC-Bayes bounds; Farid et al. (2021) combine algorithmic stability with PAC-Bayes bounds; Jose et al. (2021), Chen et al. (2021), and Rezazadeh et al. (2021) use bounds based on mutual information, which are similar to PAC-Bayes bounds; and Khodak et al. (2019) and Denevi et al. (2019) apply tools from convex analysis. Although the theoretical definitions that we propose in this chapter could be used to engage in similar analyses for neural processes, we will not get that far. We will spend our effort on establishing a solid theoretical grounding for neural processes.

Henceforth, assume the setting of a meta-learning problem as presented in Section 2.1. To motivate the theoretical framework, the starting point is a reasonable generative model for the meta–data set $(D_m)_{m=1}^M$. Specifically, we will assume that the tasks $(D_m)_{m=1}^M$ are independently and identically distributedly (i.i.d.) generated from some noise-corrupted ground-truth $\mathcal{Y}$-valued stochastic process $f$ on $\mathcal{X}$ with law $p(f)$. That is, for $m = 1, \ldots, M$, independently and identically sample

$$f_m \sim p(f), \tag{3.1}$$

$$\mathbf{y}_m^{(c)} \mid f_m, \mathbf{x}_m^{(c)}, \boldsymbol{\varepsilon}_m^{(c)} = f_m(\mathbf{x}_m^{(c)}) + \boldsymbol{\varepsilon}_m^{(c)} \quad \text{with} \quad \boldsymbol{\varepsilon}_m^{(c)} \sim \mathcal{N}(\mathbf{0}, \sigma_f^2 \mathbf{I}), \tag{3.2}$$

$$\mathbf{y}_m^{(t)} \mid f_m, \mathbf{x}_m^{(t)}, \boldsymbol{\varepsilon}_m^{(t)} = f_m(\mathbf{x}_m^{(t)}) + \boldsymbol{\varepsilon}_m^{(t)} \quad \text{with} \quad \boldsymbol{\varepsilon}_m^{(t)} \sim \mathcal{N}(\mathbf{0}, \sigma_f^2 \mathbf{I}), \tag{3.3}$$

where $\sigma_f > 0$ is the standard deviation of the observation noise. Importantly, in this setup, the realisation of the ground-truth process is *different for every task*. In fact, the realisations $(f_m)_{m=1}^M$ are independently and identically drawn from $p(f)$. That all $f_m$ have the same distribution may seem restrictive, because in practice tasks can be very different, like recognising different bird species. This can be modelled by letting $p(f)$ be a mixture distribution. For example, roll a biased $K$-faced die, and then draw $f$ from one of $K$ different stochastic processes. We hence see that the assumption of identical distributions is not necessarily so restrictive, which leaves us with the assumption that all $(f_m)_{m=1}^M$ are drawn independently. This independence assumption might or might not be applicable to a particular application, but it is an assumption that we make to simplify the analysis.

Assume that the context inputs $(\mathbf{x}_m^{(c)})_{m=1}^M$ are sampled i.i.d. from some distribution $p(\mathbf{x}^{(c)})$ and that the target inputs $(\mathbf{x}_m^{(t)})_{m=1}^M$ are sampled i.i.d. from possibly a different distribution $p(\mathbf{x}^{(t)})$. The distributions $p(\mathbf{y}^{(c)} \mid \mathbf{x}^{(c)})$ and $p(\mathbf{x}^{(c)})$ define a distribution over context sets $p(D^{(c)})$. Henceforth, to simplify the notation, we will drop the scripts $\cdot_m$, $\cdot^{(c)}$, and $\cdot^{(t)}$ and denote a context set by $D \sim p(D)$, the ground-truth stochastic process by $f \sim p(f)$, the target inputs by $\mathbf{x} \sim p(\mathbf{x})$, and the target outputs by $\mathbf{y} \sim p(\mathbf{y} \mid \mathbf{x})$.

Meta-learning algorithms aim to make the best possible prediction for a target set given a context set. For a meta–data set sampled from the generative model, the best possible prediction is given by the posterior over a target set given the context set. More specifically, for a context set $D$ and target inputs $\mathbf{x}$, the desired prediction for the target outputs $\mathbf{y}$ is given by $p(f(\mathbf{x}) + \boldsymbol{\varepsilon} \mid \mathbf{x}, D)$. We more concisely define this solution with the *posterior prediction map*.

**Definition 3.1** (Posterior prediction map). *The* posterior prediction map *is defined by*

$$\pi_f \colon \mathcal{D} \to \mathcal{P}, \quad \pi_f(D) = p(f \mid D). \tag{3.4}$$

In words, the posterior prediction map maps a data set to the prediction for that data set by the ground-truth stochastic process. For a $\mathcal{Y}$-valued stochastic process $g$ on $\mathcal{X}$ with law $\mu$ and target inputs $\mathbf{x}$, denote the law of $g(\mathbf{x})$ by $P_\mathbf{x}\mu$. In addition, for noise standard deviation $\sigma > 0$, the law of $g(\mathbf{x})$ plus additive Gaussian noise with variance $\sigma^2$ denote by $P_\mathbf{x}^\sigma \mu$:

$$g(\mathbf{x}) + \boldsymbol{\varepsilon} \sim P_\mathbf{x}^\sigma \mu \quad \text{with} \quad \boldsymbol{\varepsilon} \sim \mathcal{N}(\mathbf{0}, \sigma^2 \mathbf{I}). \tag{3.5}$$

Note that $P_\mathbf{x}^\sigma \mu$ always has a density with respect to the Lebesgue measure given by $\mathbf{x} \mapsto \mathbb{E}_g[\mathcal{N}(\mathbf{y} \mid g(\mathbf{x}), \sigma^2 \mathbf{I})]$. With this notation, to solve the meta-learning problem, for a context set $D$ and target inputs $\mathbf{x}$, the desired prediction for the target outputs $\mathbf{y}$ is given by $P_\mathbf{x}^{\sigma_f} \pi_f(D)$.

The central premise of the prediction map framework is that a neural process aims to approximate the posterior prediction map $\pi_f$. This means that, for every data set $D \in \mathcal{D}$, the neural process attempts to approximate $\pi_f(D)$. Let us elaborate on this simple, but important observation.

What a neural process is *not* doing is finding an approximation of the ground-truth stochastic process $f$, and then using this approximate prior to approximate any posterior $p(f \mid D)$. Instead, a neural process aims to *directly approximate* the posteriors $p(f \mid D)$ without approximating the prior as an intermediate step. This distinction is best understood by comparing the empirical neural process objective (Definition 2.2) proposed by Garnelo et al. (2018a) to the usual maximum likelihood objective. Let $D^{(c)}$ denote some context set, let $D^{(t)}$ denote some target set, and let $\theta$ be the parameters of some model. Whereas the usual application of maximum-likelihood estimation would maximise $\log p_\theta(D^{(c)}, D^{(t)})$, neural processes maximise only $\log p_\theta(D^{(t)} \mid D^{(c)})$. As the following application of the product rule illustrates, this means that neural processes give up modelling the context data $D^{(c)}$:

$$\underbrace{\log p_\theta(D^{(c)}, D^{(t)})}_{\substack{\text{usual maximum-likelihood}\\\text{objective}}} = \underbrace{\log p_\theta(D^{(t)} \mid D^{(c)})}_{\substack{\text{neural process}\\\text{objective}}} + \underbrace{\log p_\theta(D^{(c)})}_{\substack{\text{what neural processes}\\\text{give up}}} \quad (3.6)$$

To see why giving up modelling the context data $D^{(c)}$ can be helpful, consider the ground-truth stochastic process given by $f = h_0$ with probability $\frac{1}{2}$ and $f = h_1$ otherwise, where $h_0, h_1 \colon \mathcal{X} \to \mathcal{Y}$ are fixed deterministic functions. By construction, this prior $p(f)$ is bimodal, allowing only one of two possible realisations. A posterior, on the other hand, can be much simpler: if the observed data $D$ is able to roughly pin down whether $f$ is $h_0$ or $h_1$, then $p(f \mid D)$ will be a roughly unimodal distribution centred around $h_0$ or $h_1$. We therefore see that directly approximating a posterior $p(f \mid D)$ can be a simpler problem than trying to model the prior $p(f)$. But we can make an even stronger case by considering what happens if we do approximate $p(f)$ and then use this approximate prior to approximate posteriors $p(f \mid D)$. Suppose that we approximate $p(f)$ with a Gaussian process. One reasonable approximation is given by the Gaussian process with mean function $x \mapsto \mathbb{E}_{p(f)}[f(x)]$ and covariance function $(x,y) \mapsto \mathrm{cov}_{p(f)}(f(x), f(y))$[1]: $f \approx (\frac{1}{2} + \frac{1}{2}Z)h_1 + (\frac{1}{2} - \frac{1}{2}Z)h_2$ with $Z \sim N(0,1)$[2]. This Gaussian process will cover not just $h_0$ and $h_1$, but it will cover all affine combinations of $h_0$ and $h_1$. In words, this approximate prior will not just consider $h_0$ and $h_1$ as possible outcomes, but it will also consider all affine combinations of $h_0$ and $h_1$ as possible outcomes! Since this range of affine combinations is a much bigger collection than just $\{h_0, h_1\}$, approximations of posteriors formed by using this approximate prior will have strongly inflated uncertainty and might give rise to realisations like $0.7h_0 + 0.3h_1$, which can be totally unlike $h_0$ or $h_1$. By giving up modelling the prior using the context set $D^{(c)}$ and directly approximating the posterior without approximating the prior as an intermediate step, neural processes attempt to circumvent this issue.

To approximate the posterior prediction map $\pi_f$, the framework will define a loss function

---

[1] Specifically, this approximation is the *Gaussian neural process approximation* (GNPA) defined in Definition 3.25 and characterised in Proposition 3.27. Intuitively, the GNPA constructs the *moment-matched Gaussian process* by taking the mean function and covariance function of the non-Gaussian process. See Section 3.4 for a more detailed discussion.

[2] Write $f = Bh_1 + (1-B)h_2$ with $B \sim \mathrm{Ber}(\frac{1}{2})$. Then $f - \mathbb{E}[f] = (B - \frac{1}{2})h_1 - (B - \frac{1}{2})h_2$, so

$$\mathbb{E}[(f(x) - \mathbb{E}[f(x)])(f(y) - \mathbb{E}[f(y)])]$$
$$= \mathbb{E}[((B - \tfrac{1}{2})h_1(x) - (B - \tfrac{1}{2})h_2(x))((B - \tfrac{1}{2})h_1(y) - (B - \tfrac{1}{2})h_2(y))] \quad (3.7)$$
$$= \tfrac{1}{4}(h_1(x) - h_2(x))(h_1(y) - h_2(y)), \quad (3.8)$$

noting that $\mathbb{E}[(B - \tfrac{1}{2})^2] = \tfrac{1}{2} \cdot \tfrac{1}{4} + \tfrac{1}{2} \cdot \tfrac{1}{4} = \tfrac{1}{4} = \tfrac{1}{2} \cdot \tfrac{1}{2}$. The covariance function in (3.8) is the covariance function of $\tfrac{1}{2}Z(h_1 - h_2)$ with $Z \sim \mathcal{N}(0,1)$, so $f \approx \mathbb{E}[f] + \tfrac{1}{2}Z(h_1 - h_2) = (\tfrac{1}{2} + \tfrac{1}{2}Z)h_1 + (\tfrac{1}{2} - \tfrac{1}{2}Z)h_2$.

$\mathcal{L}(\pi, \sigma)$ which measures how well a candidate approximation $\pi$ and associated noise $\sigma > 0$ approximate the posterior prediction map $\pi_f$ and true noise $\sigma_f$. The particular loss function that we choose is called the *neural process objective* $\mathcal{L}_{\mathrm{NP}}$:

$$\mathcal{L}_{\mathrm{NP}}(\pi, \sigma) = \mathbb{E}_{p(D)p(\mathbf{x})}[\mathrm{KL}(P_{\mathbf{x}}^{\sigma_f}\pi_f(D), P_{\mathbf{x}}^{\sigma}\pi(D))]. \tag{3.9}$$

This loss function is called the neural process objective because a Monte Carlo approximation over the tasks $(D_m)_{m=1}^M$ recovers the *empirical* neural process objective (Definition 2.2) up to a constant that does not depend on $(\pi, \sigma)$. Consistent with the notation previously introduced for $q$, let $q_\theta(\,\cdot\mid \mathbf{x}, D)$ denote the density of $P_{\mathbf{x}}^{\sigma}\pi(D)$ with respect to the Lebesgue measure. Then

$$\mathcal{L}_{\mathrm{NP}}(\pi, \sigma) = -\mathbb{E}_{p(D)p(\mathbf{x})p(\mathbf{y})}[\log q_\theta(\mathbf{y}\mid \mathbf{x}, D)] - \mathbb{E}_{p(D)p(\mathbf{x})}[\mathbb{H}(P_{\mathbf{x}}^{\sigma_f}\pi_f(D))] \tag{3.10}$$

$$\approx -\frac{1}{M}\sum_{m=1}^M \log q_\theta(\mathbf{y}_m^{(t)}\mid \mathbf{x}_m^{(t)}, D_m^{(c)}) + \text{constant indep. of } (\pi, \sigma) \tag{3.11}$$

which agrees with Definition 2.2. In (3.10), $\mathbb{H}(P_{\mathbf{x}}^{\sigma_f}\pi_f(D))$ denotes the differential entropy of the probability measure $P_{\mathbf{x}}^{\sigma_f}\pi_f(D)$. The neural process objective raises three important points of concern, which we address in turn in the next sections:

1. The neural process objective $\mathcal{L}_{\mathrm{NP}}$ is defined by taking an expectation of the function $(D, \mathbf{x}) \mapsto \mathrm{KL}(P_{\mathbf{x}}^{\sigma_f}\pi_f(D), P_{\mathbf{x}}^{\sigma}\pi(D))$ over $p(D)p(\mathbf{x})$. However, it is not at all clear whether this function is even measurable. For example, measurability of this function depends on the regularity of $D \mapsto \pi_f(D)$ and $D \mapsto \pi(D)$, which thus far remain unaddressed. In Section 3.3, we address these issues and carefully define the neural process objective $\mathcal{L}_{\mathrm{NP}}$.

2. The neural process objective $\mathcal{L}_{\mathrm{NP}}$ depends on $p(D)$ and $p(\mathbf{x})$. Therefore, if we minimise $\mathcal{L}_{\mathrm{NP}}$ over a potentially restricted collection of prediction maps, then the minimiser depends on $p(D)$ and $p(\mathbf{x})$. In Section 3.4, we define two classes of minimisers of $\mathcal{L}_{\mathrm{NP}}$. We characterise these minimisers to determine how they depend on $p(D)$ and $p(\mathbf{x})$.

3. Even though (3.11) is a Monte Carlo approximation of $\mathcal{L}_{\mathrm{NP}}$, a minimiser of (3.11) need not necessarily converge to a minimiser of $\mathcal{L}_{\mathrm{NP}}$. In Section 3.5, we will determine conditions under which a minimiser of (3.11) converges to a minimiser of $\mathcal{L}_{\mathrm{NP}}$.

Having motivated the setup of the framework, we proceed to definitions that will form the technical foundation.

## 3.2 Technical Preliminaries

In Section 2.1, we defined $\mathcal{D}_N$ as the collection of all data sets of size $N$ and $\mathcal{D}$ as the collection of all data sets of finite size. It will be convenient to establish similar notation for inputs. Let $I_N = \mathcal{X}^N$ be the collection of all $N$ inputs, and let $I = \bigcup_{N=1}^{\infty} I_N$ be the collection of all finite collections of inputs. For a vector $\mathbf{x}$, recall that $|\mathbf{x}|$ denotes the dimensionality of $\mathbf{x}$. Endow $\mathcal{D}$ with the metric $d_\mathcal{D}(D_1, D_2) = \|\mathbf{x}_1 - \mathbf{x}_2\|_2 + \|\mathbf{y}_1 - \mathbf{y}_2\|_2$ if $|\mathbf{x}_1| = |\mathbf{x}_2|$ and $\infty$ otherwise. Similarly, endow $I$ with the metric $d_I(\mathbf{x}_1, \mathbf{x}_2) = \|\mathbf{x}_1 - \mathbf{x}_2\|_2$ if $|\mathbf{x}_1| = |\mathbf{x}_2|$ and $\infty$ otherwise. Allowing $d_\mathcal{D}$ and $d_I$ to attain the value infinity for arguments of different dimensionality is nothing to worry about. It simply means that the metric spaces associated to $d_\mathcal{D}$ and $d_I$ naturally break down into disjoint unions of metric spaces of fixed dimensionalities. Alternatively, we could separately define metrics for each of these fixed-dimensional spaces, but allowing $d_\mathcal{D}$ and $d_I$ to attain infinity is a more concise and arguably more elegant construction.

Whereas in the motivating section the distribution $p(D)$ was induced by randomly sampling context inputs and passing these to a sample of the ground-truth stochastic process $f$, the prediction map approximation framework more simply assumes that we are just given some distribution $p(D)$. In some sense, this means that $p(D)$ is *decoupled* from the ground-truth stochastic process $f$. Similarly, the framework assumes also that we are just given some distribution over $p(\mathbf{x})$. We now define these distributions.

Let $\widetilde{\mathcal{D}} \subseteq \mathcal{D}$ be a collection of *data sets of interest*, and let $\widetilde{I} \subseteq I$ be a collection of *target inputs of interest*. The choices $\widetilde{\mathcal{D}} = \mathcal{D}$ and $\widetilde{I} = I$ are allowed. The reason for considering subsets of $\mathcal{D}$ is practical: we most likely do not have context sets $(D^{(c)})_{m=1}^M$ that span all possible data sets. Rather, $(D^{(c)})_{m=1}^M$ might visit only a particular subspace of $\mathcal{D}$, and this can now be modelled by constraining $\widetilde{\mathcal{D}}$. Similarly, we most likely will not sample target inputs of arbitrary size, which can be modelled by constraining $\widetilde{I}$. Although $\widetilde{I}$ and $\widetilde{\mathcal{D}}$ may be chosen arbitrarily, the approximation properties of neural processes will depend on $\widetilde{I}$ and $\widetilde{\mathcal{D}}$.

For a choice of $\widetilde{\mathcal{D}}$ and $\widetilde{I}$, let $p(D)$ be any distribution on the Borel space of $\widetilde{\mathcal{D}}$ and let $p(\mathbf{x})$ be any distribution on the Borel space of $\widetilde{I}$. This means that $D \sim p(D)$ is a random element of $\widetilde{\mathcal{D}}$ and $\mathbf{x} \sim p(\mathbf{x})$ is a random element of $\widetilde{I}$. We will require two technical conditions on $p(\mathbf{x})$ and $p(D)$.

**Assumption 3.2.** *Assume that $p(\mathbf{x})$ and $p(D)$ assign positive probability to every open set. Moreover, assume that $p(\mathbf{x})$ and $p(D)$ are complete, possibly by completing the probability spaces.*

A technical subtlety that the framework will have to deal with is that conditional neural processes do *not* define continuous processes. Namely, conditional neural processes define only marginal statistics and do not specify any dependencies between different function values. An example of such a prediction is the stochastic process $f$ such that $f(\mathbf{x}) \sim \mathcal{N}(\mathbf{0}, \mathbf{I})$ for all $\mathbf{x} \in I$. By Kolmogorov's extension theorem (Theorem 12.1.2; Dudley, 2002), this construction defines a perfectly valid stochastic process. The problem, however, is that the lack of regularity requires the process to be defined on the sample space $\mathcal{Y}^\mathcal{X}$ with the associated cylindrical $\sigma$-algebra, and this $\sigma$-algebra contains too few measurable sets to be useful. Namely, a set is in the cylindrical $\sigma$-algebra if and only if it depends on only countably many values of $f$; that is, if it depends on $f(x)$ with $x \in T \subseteq \mathcal{X}$ where $T$ is countable. But the space $\mathcal{Y}^\mathcal{X}$ lacks any kind of regularity, so, even for basic properties like continuity, we must put a condition on $f(x)$ for *all* $x \in \mathcal{X}$. As a consequence, the resulting probability space is not able to express, for example, whether $f$ is continuous or not.

Even though many conditional neural processes are defined on inexpressive probability spaces, we still require some kind of regularity. For inputs $\mathbf{x}$, let $\mu_\mathbf{x}$ denote the distribution of $f(\mathbf{x})$. Instead of assuming continuity of $f$, we will assume that $\mathbf{x} \mapsto \mu_\mathbf{x}$ is weakly continuous: if $\mathbf{x}_i \to \mathbf{x}$, then $\mu_{\mathbf{x}_i} \rightharpoonup \mu_\mathbf{x}$ where $\rightharpoonup$ denotes convergence in the weak topology. Note that this condition is automatically satisfied if $f$ is a continuous process, which means that the condition is weaker than pathwise continuity. Unfortunately, this assumption is still too strong, which we illustrate with an example. Consider the previous example of $f$ such that $f(\mathbf{x}) \sim \mathcal{N}(\mathbf{0}, \mathbf{I})$ for all $\mathbf{x} \in I$; that is, like conditional neural processes, $f$ does not model dependencies. Since $\operatorname{cov}(f(1/n), f(0)) \not\to \operatorname{cov}(f(0), f(0))$, $(x, y) \mapsto \operatorname{cov}(f(x), f(y))$ is not continuous at $(x, y) = (0, 0)$. Consequently, $\mathbf{x} \mapsto \mu_\mathbf{x}$ is not weakly continuous at inputs $\mathbf{x}$ with repeated elements. Fortunately, this observation is inconsequential, because the collection of all $\mathbf{x}$ with repeated elements is a $p(\mathbf{x})$-null set whenever, *e.g.*, $p(\mathbf{x})$ has a density with respect to the Lebesgue measure. This motivates the following two definitions.

**Definition 3.3** (Regular stochastic process). *Call a $\mathcal{Y}$-valued stochastic processes $\mu$ on $\mathcal{X}$ $p(\mathbf{x})$-regular if $\mathbf{x} \mapsto \mu_\mathbf{x}$ is weakly continuous $p(\mathbf{x})$-almost everywhere. That is, $\mu$ is $p(\mathbf{x})$-regular if there exists a measurable $A \subseteq \widetilde{I}$ such that $\int_A p(\mathbf{x}) \, \mathrm{d}\mathbf{x} = 1$ and $\mathbf{x} \mapsto \mu_\mathbf{x}$ is weakly continuous at all $\mathbf{x} \in A$. Denote the collection of all $p(\mathbf{x})$-regular processes by $\mathcal{P}$.*

**Definition 3.4** (Continuous stochastic process). *Let $\mathcal{P}_\mathrm{c}$ be the collection of all $\mathcal{Y}$-valued stochastic processes on $\mathcal{X}$ which are continuous. Note that $\mathcal{P}_\mathrm{c} \subseteq \mathcal{P}$.*

Note that the definition of $\mathcal{P}$ depends on our choice of $p(\mathbf{x})$, but $\mathcal{P}_\mathrm{c}$ does not. Whereas we will assume $p(\mathbf{x})$-regularity for predictions of neural processes, for the ground-truth stochastic process $f$ we will require a slightly stronger assumption. In particular, we will assume that $f$ is Hölder continuous with respect to the $L^p$-norm.

**Assumption 3.5** (Regularity of ground-truth stochastic process). *There exist an exponent $p \geq 2$, Hölder exponent $\beta \in (\frac{1}{p}, 1]$, constant $c > 0$, and radius $r > 0$ such that*

$$\|f(x) - f(y)\|_{L^p} \leq c|x - y|^\beta \quad \text{whenever} \quad |x - y| < r. \tag{3.12}$$

*In addition, assume that $f(x) \in L^p$ for all $x \in \mathcal{X}$.*

By the triangle inequality, it holds that $\|f(x)\|_{L^p} \leq \|f(x) - f(y)\|_{L^p} + \|f(y)\|_{L^p}$. Therefore, by (3.12), the additional assumption that $f(x) \in L^p$ for *all* $x \in \mathcal{X}$ is true if $f(y) \in L^p$ is true for *some* $y \in \mathcal{X}$.

Assumption 3.5 is a fairly standard assumption, except that we require the exponent $p$ at least be two. For example, (3.12) is the assumption in Kolmogorov's continuity criterion, which says that $L^p$-Hölder continuity implies pathwise Hölder continuity by giving up $\frac{1}{p}$ in the exponent (Section 4.2, Theorem 1.4.2; Norris, 2018). In particular, it implies that $f$ is a continuous process, so $f \in \mathcal{P}_c$. For example, if $f$ is a zero-mean stationary Gaussian process with covariance function $k\colon \mathcal{X} \to \mathbb{R}$, then Assumption 3.5 is satisfied if

$$\lim_{r \to 0} \frac{k(0) - k(r)}{r^{2/p - \varepsilon}} = 0 \tag{3.13}$$

for some $\varepsilon > 0$. Choosing $p = 4$ and $\varepsilon = \frac{1}{4}$ works for many kernels. More generally, if $k$ is Hölder continuous with any Hölder exponent, then we can always find large enough $p \geq 2$ and small enough $\varepsilon > 0$ such that (3.13) is satisfied.

The prediction map framework is centred around approximating the posterior prediction map $\pi_f$ with another simpler, more tractable prediction map $\pi$. Let $\mu_f$ denote the law of the ground-truth stochastic process $f$.

**Definition 3.6** (Posterior prediction map, formal). *Let $\sigma_f > 0$ be some observation noise. Then define the posterior prediction map $\pi_f\colon \mathcal{D} \to \mathcal{P}$ by the following Radon–Nikodym derivatives: for all $D \in \mathcal{D}$,*

$$\frac{\mathrm{d}\pi_f(D)}{\mathrm{d}\mu_f}(f) = \frac{\mathcal{N}(\mathbf{y} \mid f(\mathbf{x}), \sigma_f^2 \mathbf{I})}{\mathbb{E}_f[\mathcal{N}(\mathbf{y} \mid f(\mathbf{x}), \sigma_f^2 \mathbf{I})]} = \frac{r(\mathbf{y} - f(\mathbf{x}))}{Z(\mathbf{x}, \mathbf{y})} \tag{3.14}$$

*where $D = (\mathbf{x}, \mathbf{y})$, $r(\mathbf{x}) = \exp(-\frac{1}{2\sigma_f^2}\|\mathbf{x}\|_2^2)$, and $Z(\mathbf{x}, \mathbf{y}) = \mathbb{E}_f[r(\mathbf{y} - f(\mathbf{x}))]$.*

Note that $Z(\mathbf{x}, \mathbf{y}) > 0$, for otherwise $|f(x)| = \infty$ with positive probability for some $x \in \mathcal{X}$, contradicting Assumption 3.5. To approximate $\pi_f$, we must be concerned with regularity of $\pi_f$. The following definition captures the basic form of regularity of a prediction map: continuity. Recall that $\rightharpoonup$ denotes convergence in the weak topology.

**Definition 3.7** (Continuous prediction map). *Call a prediction map $\pi\colon \mathcal{D} \to \mathcal{P}$ continuous if $D_i \to D$ implies that $P_\mathbf{x}\pi(D_i) \rightharpoonup P_\mathbf{x}\pi(D)$ for all inputs $\mathbf{x} \in I$. Denote the collection of all prediction maps $\mathcal{D} \to \mathcal{P}$ which are continuous by $\mathcal{M}$, and denote the collection of all prediction maps $\mathcal{D} \to \mathcal{P}_\mathrm{c}$ which are continuous by $\mathcal{M}_\mathrm{c}$. Note that $\mathcal{M}_\mathrm{c} \subseteq \mathcal{M}$.*

The following proposition says that any posterior $\pi_f(D)$ satisfies the same $L^p$-Hölder and integrability condition as $f$ (Assumption 3.5) and that $\pi_f$ is indeed continuous.

**Proposition 3.8** (Regularity of posterior prediction map, part one).

(1) *For all data sets $D \in \mathcal{D}$, there exists a constant $c_D > 0$ such that*

$$\|f(x) - f(y)\|_{L^p(\pi_f(D))} \leq c_D |x - y|^\beta \quad \text{whenever} \quad |x - y| < r. \tag{3.15}$$

*In addition, $\|f(x)\|_{L^p(\pi_f(D))} < \infty$ for all $x \in \mathcal{X}$.*

(2) *$\pi_f$ is continuous.*

*Proof.* See Appendix A.1. □

We have setup the basic definitions of the theoretical framework. We will now proceed with our formal analysis of neural processes. Our first matter of concern is the neural process objective.

## 3.3 The Neural Process Objective

In this section, we formally define the neural process objective $\mathcal{L}_{\mathrm{NP}}$ (Definition 3.10) and the space over which $\mathcal{L}_{\mathrm{NP}}$ will be optimised (Definition 3.9). Intuitively, the neural process objective $\mathcal{L}_{\mathrm{NP}}$ is the infinite-sample limit of the *empirical* neural process objective $\mathcal{L}_M$ (Definition 2.2). We will show that $\mathcal{L}_{\mathrm{NP}}$ is lower semi-continuous (Proposition 3.12) and that the posterior prediction map $\pi_f$ is the unique minimiser of $\mathcal{L}_{\mathrm{NP}}$ (Proposition 3.13). This presents the neural process objective $\mathcal{L}_{\mathrm{NP}}$ as an appropriate objective that we can use to approximate the posterior prediction map $\pi_f$. In this section and the following sections, all results assume Assumptions 3.2 and 3.5.

The goal of prediction map approximation is to approximate the posterior prediction map $\pi_f$ with a simpler, more tractable prediction map $\pi$. The posterior prediction map $\pi_f$ is associated to some observation noise $\sigma_f$. We will therefore also associate our approximation with some observation noise $\sigma > 0$. This means that uncertainty in predictions by $\pi_f$ and $\pi$ are a sum of two components: uncertainty about the ground-truth stochastic process $f$, represented by the variances of $\pi_f$ and $\pi$, and observation noise, represented by $\sigma_f$ and $\sigma$. These two uncertainties are fundamentally different, because only the former can be reduced by observing more data. We therefore call the former *epistemic uncertainty* and the latter *aleatoric uncertainty*. One potential issue is that an approximation $\pi$ of $\pi_f$ might conflate epistemic and aleatoric uncertainty. We will return to this issue in Section 3.4.

Since our approximation $\pi$ is associated to some observation noise $\sigma$, we call the tuple $(\pi, \sigma)$ a *noisy prediction map*. We now formalise the space of noisy prediction maps.

**Definition 3.9** (Noisy prediction maps). *Call $\overline{\mathcal{M}} = \mathcal{M} \times (0, \infty)$ the collection of noisy prediction maps. Similarly define $\overline{\mathcal{M}}_c = \mathcal{M}_c \times (0, \infty)$. Note that $\overline{\mathcal{M}}_c \subseteq \overline{\mathcal{M}}$. Endow $\overline{\mathcal{M}}$ and $\overline{\mathcal{M}}_c$ with the following topology of pointwise convergence: $(\pi_i, \sigma_i)_{i \geq 1}$ converges to $(\pi, \sigma)$ whenever (1) $P_\mathbf{x} \pi_i(D) \rightharpoonup P_\mathbf{x} \pi(D)$ for all $\mathbf{x} \in I$ and $D \in \mathcal{D}$ and (2) $\sigma_i \to \sigma$.*

Fox example, under Definition 3.9, a function $g \colon \overline{\mathcal{M}} \to \mathbb{R}$ on the space of noisy prediction maps is continuous if $P_\mathbf{x} \pi_i(D) \rightharpoonup P_\mathbf{x} \pi(D)$ for all $\mathbf{x} \in I$ and $D \in \mathcal{D}$ and $\sigma_i \to \sigma$ imply that $g(\pi_i, \sigma_i) \to g(\pi, \sigma)$. Having defined the space over which we will optimise the neural process objective, we are now in a position to formally define $\mathcal{L}_{\text{NP}}$.

**Definition 3.10** (The neural process objective). *Call the function*

$$\mathcal{L}_{\text{NP}} \colon \overline{\mathcal{M}} \to [0, \infty], \quad \mathcal{L}_{\text{NP}}(\pi, \sigma) = \mathbb{E}_{p(D)p(\mathbf{x})}[\text{KL}(P_\mathbf{x}^{\sigma_f} \pi_f(D), P_\mathbf{x}^\sigma \pi(D))] \quad (3.16)$$

*the* neural process objective.

It is not at all clear that the expectations in Definition 3.10 are well defined. The following proposition addresses this. Proposition 3.11 looks innocent, but its proof unfortunately is considerably technical.

**Proposition 3.11.** *The neural process objective $\mathcal{L}_{\text{NP}}$ is well defined.*

*Proof.* See Appendix A.2. □

We next show that $\mathcal{L}_{\text{NP}}$ is lower semi-continuous. It is also true that $\mathcal{L}_{\text{NP}}$ is convex, which follows directly from convexity of the Kullback–Leibler divergence (Lemma 7.2; Gray, 2011) and linearity of expectation, but we will not need this property.

**Proposition 3.12.** *The neural process objective $\mathcal{L}_{\text{NP}}$ is lower semi-continuous.*

*Proof.* See Appendix A.2. □

We finally arrive at Proposition 3.13. Modulo a few caveats, Proposition 3.13 states that $(\pi_f, \sigma_f)$ is the unique minimiser of $\mathcal{L}_{\text{NP}}$ over all noisy prediction maps. Unfortunately, minimising $\mathcal{L}_{\text{NP}}$ over all possible noisy prediction maps is not practically feasible. Instead, we could hope to approximate $(\pi_f, \sigma_f)$ by minimising $\mathcal{L}_{\text{NP}}$ over a tractable, large-enough class of noisy prediction maps. This is the approach that we will consider in the next section.

**Proposition 3.13.** *Assume that $\widetilde{I} \subseteq I$ is dense. Then $(\pi, \sigma) \in \arg\min_{\overline{\mathcal{M}}_c} \mathcal{L}_{\text{NP}}$ if and only if*

$$\pi_f|_{\widetilde{\mathcal{D}}} = \pi|_{\widetilde{\mathcal{D}}} \quad \text{and} \quad \sigma = \sigma_f. \tag{3.17}$$

*Proof.* See Appendix A.2. □

Note that Proposition 3.13 minimises over $\overline{\mathcal{M}}_c$ rather than $\overline{\mathcal{M}}$. This restriction is necessary, because otherwise some of the aleatoric noise $\sigma_f$ can be absorbed in the epistemic uncertainty, which means that the minimiser would not be unique. Also note that Proposition 3.13 requires $\widetilde{I}$ to be dense in $I$. In particular, this means that the target inputs of interest $\widetilde{I}$ must consist of inputs of arbitrarily large size, that is, $\mathbf{x} \in \mathcal{X}^N$ for arbitrarily large $N$. Finally, if $\widetilde{\mathcal{D}} = \mathcal{D}$, then note that $(\pi_f, \sigma_f)$ is the unique minimiser of $\mathcal{L}_{\text{NP}}$. On the other hand, if $\widetilde{\mathcal{D}} \subsetneq \mathcal{D}$, then a minimiser of $\mathcal{L}_{\text{NP}}$ recovers the posterior prediction map $\pi_f$ only on $\widetilde{\mathcal{D}}$. This means that the collection of data sets of interest $\widetilde{\mathcal{D}}$ determines the approximation properties of minimisers of $\mathcal{L}_{\text{NP}}$.

## 3.4  Neural Process Approximations

Having formally defined the neural process objective $\mathcal{L}_{\text{NP}}$ (Definition 3.10), we turn our attention to the problem of approximating the posterior prediction map $(\pi_f, \sigma_f)$ with a simpler, more tractable prediction map $(\pi, \sigma)$. We will form such an approximations by first defining a *variational family* $\mathcal{Q} \subseteq \overline{\mathcal{M}}$ (Definitions 3.14, 3.22 and 3.23). A variational family $\mathcal{Q}$ is intended to consist of simpler, more tractable candidate approximations. An approximation of $(\pi_f, \sigma_f)$ can then be obtained by minimising the neural process objective $\mathcal{L}_{\text{NP}}$ over the variational family $\mathcal{Q}$ (Definitions 3.15, 3.24 and 3.25). Different variational families define different *classes of neural processes* and we will call the associated minimisers *neural process approximations*. This means that neural process approximations formalise what different classes of neural processes precisely target. Importantly, not all neural processes target the same approximation (Propositions 3.26 and 3.27). Analysing the properties of neural process approximations is a way to understand the behaviour of neural processes in practice.

**Definition 3.14** (Variational family, neural process). *A variational family $\mathcal{Q} \subseteq \overline{\mathcal{M}}$ is a collection of noisy prediction maps. Variational families are also called* neural processes.

**Definition 3.15** (Neural process approximation). *A neural process approximation is the collection of minimisers of the neural process objective $\mathcal{L}_{\text{NP}}$ over a variational family.*

Figure 3.1 illustrates how all objects we have defined thus far connect.

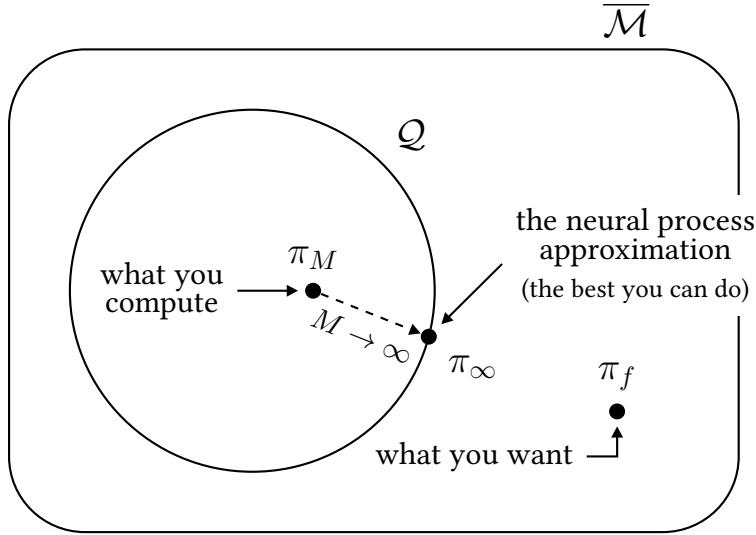

Figure 3.1: Connection between the posterior prediction map $\pi_f$ (Definition 3.1), a neural process approximation $\pi_\infty$ (Definition 3.15), and what you compute in practice $\pi_M$. Shows the collection of all noisy prediction maps $\overline{\mathcal{M}}$ (Definition 3.9) and a variational family $\mathcal{Q}$ of a neural process (Definition 3.14). For this variational family, shows a minimiser $\pi_M \in \arg\min_\mathcal{Q} \mathcal{L}_M$ of the empirical neural process objective $\mathcal{L}_M$ (Definition 2.2) and a minimiser $\pi_\infty \in \arg\min_\mathcal{Q} \mathcal{L}_{\mathrm{NP}}$ of the neural process objective $\mathcal{L}_{\mathrm{NP}}$ (Definition 3.10). In the figure, distance is as measured by the neural process objective $\mathcal{L}_{\mathrm{NP}}$. The idea is that what you compute in practice $\pi_M$ converges to the neural process approximation $\pi_\infty$ in the limit of infinite data $M \to \infty$.

The class of neural processes originally presented by Garnelo et al. (2018a) will be represented by the variational family $\mathcal{Q}_{\mathrm{G,MF}}$ called the collection of *conditional neural processes* (CNPs; Definition 3.22). CNPs are prediction maps which map to Gaussian processes that *do not* model dependencies between different function values. This section also presents a new class of neural processes $\mathcal{Q}_\mathrm{G}$ called the collection of *Gaussian neural processes* (GNPs; Definition 3.23). Contrary to CNPs, GNPs are prediction maps which map to Gaussian processes that *do* model dependencies between different function values. The following definitions set up the notation that we will use to define CNPs and GNPs.

**Definition 3.16** (Gaussian process). *Let $\mathcal{P}_\mathrm{G}$ be the collection of processes in $\mathcal{P}$ which are Gaussian.*

**Definition 3.17** (Gaussian prediction map). *Call a prediction map $\pi\colon \mathcal{D} \to \mathcal{P}_\mathrm{G}$ Gaussian if it maps to $\mathcal{P}_\mathrm{G}$.*

**Definition 3.18** (Mean map). *For any prediction map $\pi\colon \mathcal{D} \to \mathcal{P}$, not necessarily Gaussian, define the* mean map $m_\pi$ *by*

$$m_\pi \colon \mathcal{D} \to \mathcal{Y}^\mathcal{X}, \qquad m_\pi(D) = x \mapsto \mathbb{E}_{\pi(D)}[f(x)]. \tag{3.18}$$

**Definition 3.19** (Kernel map). *For any prediction map $\pi \colon \mathcal{D} \to \mathcal{P}$, not necessarily Gaussian, define the* kernel map $k_\pi$ *by*

$$k_\pi \colon \mathcal{D} \to \mathbb{R}^{\mathcal{X} \times \mathcal{X}}, \qquad k_\pi(D) = (x, y) \mapsto \mathrm{cov}_{\pi(D)}(f(x), f(y)). \tag{3.19}$$

**Definition 3.20** (Variance map). *For any prediction map $\pi \colon \mathcal{D} \to \mathcal{P}$, not necessarily Gaussian, define the* variance map $v_\pi$ *by*

$$v_\pi \colon \mathcal{D} \to [0, \infty)^{\mathcal{X} \times \mathcal{X}}, \qquad v_\pi(D) = x \mapsto \mathrm{var}_{\pi(D)}(f(x)). \tag{3.20}$$

If a prediction map $\pi_i$ is subscripted, for example by $i$, we more simply denote $m_i = m_{\pi_i}$, $k_i = k_{\pi_i}$, and $v_i = v_{\pi_i}$. Similarly, for the ground-truth stochastic process $f$, we more simply denote $m_f = m_{\pi_f}$, $k_f = k_{\pi_f}$, and $v_f = v_{\pi_f}$.

We will now define the classes of conditional neural processes and Gaussian neural processes. CNPs require the following technical condition, which we already alluded to in Section 3.2.

**Assumption 3.21.** *Assume that the collection of inputs $\mathbf{x} \in I$ with repeated elements is a $p(\mathbf{x})$-null set.*

**Definition 3.22** (Conditional neural process; CNP). *Let the collection of* conditional neural processes *(CNPs) be*

$$\mathcal{Q}_{\mathrm{G,MF}} = \left\{ (\pi, \sigma) \in \overline{\mathcal{M}} \;\middle|\; \begin{array}{c} \pi \text{ is Gaussian,} \\ k_\pi(D)(x, y) = 0 \text{ for all } x \neq y \text{ and } D \in \mathcal{D}, \\ m_\pi(D) \text{ and } v_\pi(D) \text{ are continuous for all } D \in \mathcal{D} \end{array} \right\}. \tag{3.21}$$

Note that $\mathcal{Q}_{\mathrm{G,MF}}$ is a subset of $\overline{\mathcal{M}}$ rather than $\overline{\mathcal{M}}_{\mathrm{c}}$, meaning that all prediction maps in $\mathcal{Q}_{\mathrm{G,MF}}$ map to $p(\mathbf{x})$-regular processes rather than continuous processes. This generality is necessary, because assuming that $k_\pi(D)(x, y) = 0$ for all $x \neq y$ but allowing $k_\pi(D)(x, x) > 0$ can create discontinuities of $\mathbf{x} \mapsto P_\mathbf{x} \pi(D)$ at inputs $\mathbf{x}$ with repeated elements. Fortunately, this is allowed, because we assumed that the collection of all inputs $\mathbf{x}$ with repeated elements is a $p(\mathbf{x})$-null set (Assumption 3.21). It also means, however, that the mean maps $m_\pi$ and variance maps $v_\pi$ map to functions continuous only almost everywhere. We strengthen this by assuming, in the definition of $\mathcal{Q}_{\mathrm{G,MF}}$, that $m_\pi$ and $v_\pi$ map to functions continuous everywhere.

The collection of conditional neural processes should be interpreted as the collection of prediction maps which map to Gaussian processes that *do not* model dependencies between different function values. Definition 3.22 makes this precise by the condition on $k_\pi(D)$. The "MF" in the subscript of $\mathcal{Q}_{\text{G,MF}}$ stands for "mean field" to remind the reader of this. A conditional neural process $(\pi, \sigma)$ is characterised by its mean map $m_\pi$, variance map $v_\pi$, and noise variance $\sigma$.

**Definition 3.23** (Gaussian neural process; GNP). *Let the collection of* Gaussian neural processes *(GNPs) be*
$$\mathcal{Q}_{\text{G}} = \{(\pi, \sigma) \in \overline{\mathcal{M}}_{\text{c}} : \pi \text{ is Gaussian}\}. \tag{3.22}$$

Note that $\mathcal{Q}_{\text{G}}$ is a subset of $\overline{\mathcal{M}}_{\text{c}}$ rather than $\overline{\mathcal{M}}$, meaning that all prediction maps in $\mathcal{Q}_{\text{G}}$ map to continuous Gaussian processes. Consequently, for all $\pi \in \mathcal{Q}_{\text{G}}$, the mean maps $m_\pi$ and kernel maps $k_\pi$ map to continuous functions.

The collection of Gaussian neural processes should be interpreted as the collection of prediction maps which map to Gaussian processes that *do* model dependencies between different function values. Definition 3.23 makes this precise by allowing any continuous kernel map, which includes ones such that $k_\pi(D)(x, y) \neq 0$ for $x \neq y$; compare this to Definition 3.22, where $k_\pi(D)(x, y) = 0$ for all $x \neq y$. A Gaussian neural process $(\pi, \sigma)$ is characterised by its mean map $m_\pi$, kernel map $k_\pi$, and noise variance $\sigma$.

Having defined two variational families $\mathcal{Q}_{\text{G,MF}}$ and $\mathcal{Q}_{\text{G}}$, we next define two classes of *neural processes approximations* by minimising the neural process objective $\mathcal{L}_{\text{NP}}$ over these variational families.

**Definition 3.24** (Conditional neural process approximation; CNPA). *Let the collection of* conditional neural process approximations *(CNPAs) be the minimisers of $\mathcal{L}_{\text{NP}}$ over $\mathcal{Q}_{\text{G,MF}}$.*

**Definition 3.25** (Gaussian neural process approximation; GNPA). *Let the collection of* Gaussian neural process approximations *(GNPAs) be the minimisers of $\mathcal{L}_{\text{NP}}$ over $\mathcal{Q}_{\text{G}}$.*

The following propositions characterise CNPAs and GNPAs. Whereas Proposition 3.13 requires the assumption that $\widetilde{I}$ is dense in $I$, GNPAs and CNPAs admit weaker assumptions on $\widetilde{I}$.

**Proposition 3.26** (Characterisation of CNPA). *Assume that $\inf_{\mathcal{Q}_{\text{G,MF}}} \mathcal{L}_{\text{NP}} < \infty$. Also assume that $\widetilde{I}$ is dense in $I_N$ for some $N \geq 1$. Then a noisy prediction map $(\pi, \sigma) \in \mathcal{Q}_{\text{G,MF}}$ is a CNPA if and only if*
$$m_\pi|_{\widetilde{\mathcal{D}}} = m_f|_{\widetilde{\mathcal{D}}} \quad \text{and} \quad v_\pi|_{\widetilde{\mathcal{D}}} + \sigma^2 = v_f|_{\widetilde{\mathcal{D}}} + \sigma_f^2, \tag{3.23}$$

*Proof.* See Appendix A.3. □

If $\widetilde{\mathcal{D}} = \mathcal{D}$, then $(\pi, \sigma) \in \mathcal{Q}_{\text{G,MF}}$ is a CNPA if and only if $m_\pi = m_f$ and $v_\pi + \sigma^2 = v_f^2 + \sigma_f^2$. This means that a CNPA is *not* unique: for every $\sigma \in (0, \sigma_f]$, choosing $v_\pi = v_f^2 + \sigma_f^2 - \sigma^2$ gives a CNPA. On the other hand, if $\widetilde{\mathcal{D}} \subsetneq \mathcal{D}$, then a CNPA recovers $m_f$ and $v_f + \sigma^2$ only on $\widetilde{\mathcal{D}}$. We therefore see that $\widetilde{\mathcal{D}}$ determines the approximation properties of CNPAs.

Proposition 3.26 requires that $\widetilde{I}$ is dense in $I_N$ for some $N \geq 1$. This means that a CNPA can be found even if the target set size is only one.

**Proposition 3.27** (Characterisation of GNPA). *Assume that $\inf_{\mathcal{Q}_{\text{G}}} \mathcal{L}_{\text{NP}} < \infty$. Also assume that $\widetilde{I}$ is dense in $I_N$ for some $N \geq 2$. Then a noisy prediction map $(\pi, \sigma) \in \mathcal{Q}_{\text{G}}$ is a GNPA if and only if*

$$m_\pi|_{\widetilde{\mathcal{D}}} = m_f|_{\widetilde{\mathcal{D}}}, \quad k_\pi|_{\widetilde{\mathcal{D}}} = k_f|_{\widetilde{\mathcal{D}}}, \quad \text{and} \quad \sigma = \sigma_f. \tag{3.24}$$

*Proof.* See Appendix A.3. □

If $\widetilde{\mathcal{D}} = \mathcal{D}$, then $(\pi, \sigma) \in \mathcal{Q}_{\text{G}}$ is then a GNPA if and only if $m_\pi = m_f$, $k_\pi = k_f$, and $\sigma = \sigma_f$. In particular, a GNPA is unique, so we may speak of *the* GNPA. On the other hand, if $\widetilde{\mathcal{D}} \subsetneq \mathcal{D}$, then a GNPA recovers $m_f$ and $k_f$ only on $\widetilde{\mathcal{D}}$. We therefore see that $\widetilde{\mathcal{D}}$ also determines the approximation properties of GNPAs.

Proposition 3.27 requires that $\widetilde{I}$ is dense in $I_N$ for some $N \geq 2$. Hence, to find a GNPA, one must consider target set sizes of at least two. This is a stronger requirement than for the CNPA, where one can consider target set sizes of just one.

If we interpret the mean map as the "first moment" of a prediction map and the kernel map as the "second moment", then the Gaussian neural process approximation *moment matches the posterior prediction map*. Such behaviour has previously been noted by Ma et al. (2018). One might wonder whether moment matching the posterior prediction map yields a prediction map which maps to continuous processes. As the following proposition shows, the GNPA inherits $L^p$-Hölder regularity from the posterior prediction map (Assumption 3.5), but with $p = 2$. In particular, this means that the GNPA indeed maps to continuous processes.

**Proposition 3.28** (Regularity of GNPA). *Let $(\pi, \sigma) \in \mathcal{Q}_{\text{G}}$ be a Gaussian neural process approximation. Then, for all data sets $D \in \widetilde{\mathcal{D}}$,*

$$\|f(x) - f(y)\|_{L^2(\pi(D))} \leq c_D |x - y|^\beta \quad \text{whenever} \quad |x - y| < r, \tag{3.25}$$

*where $c_D > 0$ is the constant from Proposition 3.8.(1) and the Hölder exponent $\beta \in (\frac{1}{p}, 1]$ and radius $r > 0$ are from Assumption 3.5.*

*Proof.* See Appendix A.3. □

We have seen that the GNPA is unique, but CNPAs are not: CNPAs define a range of solutions by absorbing part of the observation noise $\sigma_f$ in the variance map. This highlights a fundamental distinction between Gaussian neural processes and conditional neural processes. Since Gaussian neural processes model dependencies between function values, they are able to correctly separate epistemic and aleatoric uncertainty. Conditional neural processes, on the contrary, do not model these dependencies, and consequently they are unable to separate epistemic and aleatoric uncertainty; they can only identify the sum of the two. We will further compare GNPs and CNPs in Section 5.5.

## 3.5 Consistency

In the previous section, we defined two approximations of the posterior prediction map $\pi_f$ by minimising the neural process objective $\mathcal{L}_{\text{NP}}$ over two different variational families $\mathcal{Q}$ (Definitions 3.22 and 3.23). Unfortunately, in practice, we cannot compute either of these approximations, because we do not have access to $\mathcal{L}_{\text{NP}}$. Although we cannot compute $\mathcal{L}_{\text{NP}}$ exactly, we can compute a Monte Carlo approximation of it:

$$\mathcal{L}_M(\pi, \sigma) = -\frac{1}{M} \sum_{m=1}^{M} \log q_\theta(\mathbf{y}_m^{(t)} \mid \mathbf{x}_m^{(t)}, D_m^{(c)}) + \text{constant}. \tag{3.26}$$

We earlier defined this Monte Carlo approximation as the *empirical* neural process objective (Definition 2.2), which is the objective originally proposed by Garnelo et al. (2018a). To approach either of the approximations, it seems reasonable that we could minimise $\mathcal{L}_M$ over the variational family $\mathcal{Q}$, and hope that the minimiser converges to the minimiser of $\mathcal{L}_{\text{NP}}$ as $M \to \infty$, a property called *consistency*. See also the illustration in Figure 3.1.

Sadly, for the variational families $\mathcal{Q}_{\text{G,MF}}$ and $\mathcal{Q}_{\text{G}}$ we introduced in Definitions 3.22 and 3.23, consistency does not hold. To see this, simply note that (3.26) can be made arbitrarily small by setting $m_\pi(D_m^{(c)})(\mathbf{x}_m^{(t)}) = \mathbf{y}_m^{(t)}$ and making the variance sufficiently small. We can always pick such $m_\pi$, because the variational families require $m_\pi$ to only be continuous and map to continuous functions, which means that we can arbitrarily choose the value of $m_\pi$ at any finite number of points. What is essentially happening is that (3.26) is optimised over the collection of all continuous functions, which is too large, because a continuous function can always fit a data set exactly! Consequently, our estimator will *overfit*.

To combat overfitting, we must constrain our variational family $\mathcal{Q}$ to a *smaller* family $\widetilde{\mathcal{Q}} \subseteq \mathcal{Q}$ such that estimation with $\widetilde{\mathcal{Q}}$ is consistent and, crucially, minimising $\mathcal{L}_M$ over $\widetilde{\mathcal{Q}}$ still gives the desired neural process approximation. As we will discuss shortly, a sufficient condition for $\widetilde{\mathcal{Q}}$ to be appropriately small is that $\widetilde{\mathcal{Q}}$ be *compact* in an appropriate topology.

In this section, we will show that there exist reasonable assumptions and a compact $\widetilde{\mathcal{Q}}$ such that, for any ground-truth stochastic process satisfying these assumptions, the CNPA and GNPA are contained in $\widetilde{\mathcal{Q}}$. For the moment, let us very roughly compare $\mathcal{Q}$ to $\mathbb{R}^n$. Then the sought-after neural process approximation would be some $z \in \mathbb{R}^n$. An appropriate choice for $\widetilde{\mathcal{Q}}$ is then apparent: just take $\widetilde{\mathcal{Q}}$ to be the equivalent of $[-M, M]^n$ for large enough $M > 0$. Namely, $[-M, M]^n$ is compact and eventually captures all of $\mathbb{R}^n$ as $M \to \infty$. While this idea is promising, the analogy is flawed: $\mathcal{Q}$ is *infinite dimensional*, so, unlike $\mathbb{R}^n$, a sequence of compact sets which eventually capture all of $\mathcal{Q}$ does not exist. In more technical terms, every *infinite-dimensional* topological vector space is *not locally compact* (Theorem 1.22; Rudin, 1991). Therefore, it is not at all clear that we can find an appropriate $\widetilde{\mathcal{Q}}$. The main result of this section, however, is that we can (Proposition 3.33), leading us to consistency results for neural processes (Propositions 3.34 and 3.35).

To establish consistency of an estimator of a neural process approximation, we will follow Wald's (1949) proof of consistency of maximum-likelihood estimators. At the heart of Wald's approach lies *compactness of the parameter space*. As we revealed in the previous paragraph, in our setting, this means that we require *compactness of the variational family* $\mathcal{Q}$. In particular, this means that we require a topology on $\mathcal{Q}$, and this topology must be carefully selected. For example, consider all mean maps associated to noisy prediction maps in $\mathcal{Q}_G$. Then a candidate topology would be the initial topology generated by $\{m_\pi \mapsto m_\pi(D)(\mathbf{x}) : D \in \mathcal{D}, \mathbf{x} \in I\}$. This topology is problematic, because it admits "too large" compact sets. As a result, optimising (3.26) over a set compact in this topology does not constrain the optimiser enough to avoid the problem of exactly fitting the data. For example, by Tychonoff's theorem (Theorem 37.3; Munkres, 2000), the collection of mean functions defined by all possible functions $\mathcal{D} \to [0,1]^{\mathcal{X}}$ would be compact in this topology, but clearly we cannot optimise over this collection without running the risk of overfitting!

We therefore choose a topology on $\mathcal{Q}_{G,MF}$ and $\mathcal{Q}_G$ which does not admit "too large" compact sets. Henceforth, in this section, we view mean maps more simply as functions $\mathcal{D} \times \mathcal{X} \to \mathcal{Y}$ rather than functions $\mathcal{D} \to \mathcal{Y}^{\mathcal{X}}$. Similarly, we view kernel maps more simply as functions $\mathcal{D} \times \mathcal{X} \times \mathcal{X} \to \mathbb{R}$ and variance maps more simply as functions $\mathcal{D} \times \mathcal{X} \to [0, \infty)$. For mean maps, define the supremum norm relative to $\widetilde{\mathcal{D}}$ as follows:

$$\|m_1 - m_2\|_{\widetilde{\infty}} = \sup_{D \in \widetilde{\mathcal{D}}, x \in \mathcal{X}} |m_1(D, x) - m_2(D, x)|. \qquad (3.27)$$

Similarly define this norm for kernel maps and variance maps. For the class of conditional neural processes $\mathcal{Q}_{G,MF}$, define the metric

$$d_{G,MF}((\pi_1, \sigma_1), (\pi_2, \sigma_2)) = \|m_1 - m_2\|_{\widetilde{\infty}} + \|v_1 - v_2\|_{\widetilde{\infty}} + |\sigma_1 - \sigma_2|. \qquad (3.28)$$

For the class of Gaussian neural processes $\mathcal{Q}_\mathrm{G}$, define the metric

$$d_\mathrm{G}((\pi_1, \sigma_1), (\pi_2, \sigma_2)) = \|m_1 - m_2\|_{\widetilde{\infty}} + \|k_1 - k_2\|_{\widetilde{\infty}} + |\sigma_1 - \sigma_2|. \tag{3.29}$$

We will consider the topologies on $\mathcal{Q}_\mathrm{G,MF}$ and $\mathcal{Q}_\mathrm{G}$ induced by these metrics. Under these metrics, a characterisation of compactness is given by the Arzelà–Ascoli theorem (Theorem 7.25; Rudin, 1976): a subset of $\mathcal{Q}$ of $\mathcal{Q}_\mathrm{G,MF}$ (respectively, $\mathcal{Q}_\mathrm{G}$) is compact if and only if it is closed and the associated mean maps and variance maps (respectively, kernel maps) are equicontinuous and uniformly bounded. Equicontinuity roughly means the following: the collection of mean maps over which we optimise the finite-sample objective may only *vary limitedly quickly*. It is precisely this property that will disallow arbitrarily choosing the value at any finite number of points to exactly fit the data.

The remainder of this section proceeds as follows. We first establish a number of reasonable conditions. Based on these conditions, we define a compact subset $\widetilde{\mathcal{Q}}$ of $\mathcal{Q}_\mathrm{G,MF}$ (respectively, $\mathcal{Q}_\mathrm{G}$). We then prove that, for any ground-truth stochastic process satisfying these conditions, the CNPA (respectively, GNPA) is contained in $\widetilde{\mathcal{Q}}$. To establish these conditions, we will concoct pathological sequences which attempt to break equicontinuity. In particular, we want to avoid that the assumptions permit a sequence of data sets $(D_i)_{i \geq 1}$ and ground-truth stochastic processes $(f_i)_{i \geq 1}$ such that $m_{f_i}(D_i)$ becomes arbitrarily steeply sloped at any input. We will take this as a guiding criterion.

To begin with, for $n \geq 1$, consider $D_n = \{(-1, -n), (1, n)\}$. Then $m_f(D_n)$ will approximately interpolate $(-1, -n)$ and $(1, n)$, so the slope of $m_f(D)$ at $x = 0$ is approximately $2n$. Hence, as $n \to \infty$, the slope of $m_f(D_n)$ at $x = 0$ becomes arbitrarily large, contradicting equicontinuity. To prevent this from happening, we must assume that the magnitude of the observations is bounded.

Similarly, consider $D_n = \{(-2^0, -1), \ldots, (-2^{-n}, -1), (2^{-n}, 1), \ldots, (2^0, 1)\}$. Then $m_f(D)$ will approximately interpolate $(-2^{-n}, -1)$ and $(2^{-n}, 1)$. As $n \to \infty$, the sheer number of observations will overwhelm the prior, so $m_f(D)$ will interpolate $(-2^{-n}, -1)$ and $(2^{-n}, 1)$ more and more closely. Hence, as $n \to \infty$, the slope of $m_f(D)$ at $x = 0$ will again become arbitrarily large. We must therefore also assume that the number of observations is bounded.

**Assumption 3.29** (Boundedness of context sets). *The collection of data sets of interest $\widetilde{\mathcal{D}}$ satisfies*

$$B_{\widetilde{\mathcal{D}}} := \sup \{|\mathbf{x}| \vee \|\mathbf{y}\|_2 : (\mathbf{x}, \mathbf{y}) \in \widetilde{\mathcal{D}}\} < \infty. \tag{3.30}$$

Instead of assuming that the magnitude of the observations is bounded, we could have assumed that $f$ is a bounded process. Although this alternative assumption works and is arguably simpler, it is also more restrictive. For example, it would exclude the common choice $f \sim \mathcal{GP}(0, e^{-\frac{1}{2}(\cdot-\cdot)^2})$. By assuming that only the context sets are bounded, whenever $D \in \widetilde{\mathcal{D}}$, we can still approximate $p(f \mid D)$ for unbounded $f$.

Finally, consider $D_n = \{(-2^{-n}, -1), (2^{-n}, 1)\}$ and a sequence of ground-truth stochastic processes $f_n$ with observation noises $\sigma_{f_n} = 2^{-n}$. Then, as $n \to \infty$, the observation noise becomes smaller and smaller, so $m_{f_n}(D_n)$ will interpolate $(-2^{-n}, -1)$ and $(2^{-n}, 1)$ more and more closely, meaning that the slope of $m_{f_n}(D_n)$ at $x = 0$ will again become arbitrarily large. We therefore require a universal lower bound on the allowed values for $\sigma_f$.

**Assumption 3.30** (Boundedness of noise). *The observation noise $\sigma_f$ is larger than some universal lower bound $\underline{\sigma} > 0$.*

In addition to a universal lower bound $\underline{\sigma}$, we will also assume that $\sigma_f \leq \overline{\sigma} < \infty$ for some universal upper bound $\overline{\sigma}$. Then $\sigma_f$ will be contained in the interval $[\underline{\sigma}, \overline{\sigma}]$, which indeed is compact.[3]

Our final assumption is a technical one, stating that we require a little more than the second moment to be uniformly bounded. Note that, if $p > 2$, then this condition is implied by Assumption 3.5.

**Assumption 3.31** (Boundedness of ground-truth stochastic process). *There exists a universal $\gamma > 0$ such that*
$$B_f := \sup_{x \in \mathcal{X}} \|f(x)\|_{L^{2+\gamma}} < \infty. \tag{3.31}$$

We collect the parameters of all assumptions thus far in a tuple $\mathsf{U}$ called the *universal parameters*.

**Definition 3.32** (Universal parameters). *The universal parameters $\mathsf{U}$ are the parameters defined in Assumptions 3.5 and 3.29 to 3.31:*
$$\mathsf{U} = (p, \beta, c, r, \gamma, B_f, B_{\widetilde{\mathcal{D}}}, \underline{\sigma}). \tag{3.32}$$

Henceforth, when we say that an object exists universally, we mean that it depends only on the universal parameters and nothing else. We now come to the first result of this section, which says that the mean map and the kernel map of the posterior prediction map satisfy a

---

[3] This upper bound can likely be made redundant by considering the compactification $[\underline{\sigma}, \infty]$. See Example 5.16 by van der Vaart (1998) for an illustration of how Wald's proof can be applied to a non-compact parameter space by "adding $-\infty$ and $\infty$". The upper bound is therefore mainly for technical convenience.

Hölder condition. Crucially, the Hölder constants and exponents are universal, meaning that they only depend on the universal parameters.

**Proposition 3.33** (Regularity of posterior prediction map, part two).

(1) *There exist universal constants $c_m > 0$ and $c_k > 0$ such that, for any two $D_1, D_2 \in \widetilde{\mathcal{D}}$, whenever $d_{\mathcal{D}}(D_1, D_2) < 1$, then*

$$\sup_{x \in \mathcal{X}} |\mathbb{E}_{\pi_f(D_1)}[f(x)] - \mathbb{E}_{\pi_f(D_2)}[f(x)]| \leq c_m \, d_{\mathcal{D}}(D_1, D_2)^{\frac{2+2\gamma}{3+2\gamma}(\beta - \frac{1}{p})}, \quad (3.33)$$

$$\sup_{x,y \in \mathcal{X}} |\mathbb{E}_{\pi_f(D_1)}[f(x)f(y)] - \mathbb{E}_{\pi_f(D_2)}[f(x)f(y)]| \leq c_k \, d_{\mathcal{D}}(D_1, D_2)^{\frac{\gamma}{1+\gamma}(\beta - \frac{1}{p})}. \quad (3.34)$$

(2) *Consequently, for any $D_1, D_2 \in \widetilde{\mathcal{D}}$ and $x, y \in \mathcal{X}$, whenever $d_{\mathcal{D}}(D_1, D_2) < 1$, $|x_1 - x_2| < r$, and $|y_1 - y_2| < r$, then*

$$|\mathbb{E}_{\pi_f(D_1)}[f(x_1)] - \mathbb{E}_{\pi_f(D_2)}[f(x_2)]| \leq c_m \, d_{\mathcal{D}}(D_1, D_2)^{\frac{2+2\gamma}{3+2\gamma}(\beta - \frac{1}{p})} + \frac{c}{c_\ell}|x_1 - x_2|^\beta \quad (3.35)$$

*and*

$$|\mathbb{E}_{\pi_f(D_1)}[f(x_1)f(y_1)] - \mathbb{E}_{\pi_f(D_2)}[f(x_2)f(y_2)]|$$
$$\leq c_k \, d_{\mathcal{D}}(D_1, D_2)^{\frac{\gamma}{1+\gamma}(\beta - \frac{1}{p})} + B_f \frac{c}{c_\ell^2}|x_1 - x_2|^\beta + B_f \frac{c}{c_\ell^2}|y_1 - y_2|^\beta, \quad (3.36)$$

*where $c_\ell > 0$ is a universal constant from [Lemma A.4](#).*

*Proof.* See [Appendix A.4](#). □

[Proposition 3.33.(2)](#) states that $m_f$, $k_f$, and $v_f$ satisfy particular universal Hölder conditions. Let $\mathsf{H}_m$, $\mathsf{H}_k$, and $\mathsf{H}_v$ be the collections of respectively all mean maps, kernel maps, and variance maps satisfying these universal Hölder conditions. We now use these collections to define our desired compact variational family $\widetilde{\mathcal{Q}}$. Define the subfamilies $\widetilde{\mathcal{Q}}_{\text{G,MF}} \subseteq \mathcal{Q}_{\text{G,MF}}$ and $\widetilde{\mathcal{Q}}_{\text{G}} \subseteq \mathcal{Q}_{\text{G}}$ as follows:

$$\widetilde{\mathcal{Q}}_{\text{G,MF}} = \{(m_\pi, v_\pi, \sigma) \in \mathcal{Q}_{\text{G}} : m_\pi \in \mathsf{H}_m, \, v_\pi \in \mathsf{H}_v, \, \sigma \in [\underline{\sigma}, \overline{\sigma}]\}, \quad (3.37)$$

$$\widetilde{\mathcal{Q}}_{\text{G}} = \{(m_\pi, k_\pi, \sigma) \in \mathcal{Q}_{\text{G}} : m_\pi \in \mathsf{H}_m, \, k_\pi \in \mathsf{H}_k, \, \sigma \in [\underline{\sigma}, \overline{\sigma}]\}. \quad (3.38)$$

By construction, $(m_f, v_f, \sigma_f) \in \widetilde{\mathcal{Q}}_{\text{G,MF}}$ and $(m_f, k_f, \sigma_f) \in \widetilde{\mathcal{Q}}_{\text{G}}$. Crucially, by the Hölder conditions, the variational families $\widetilde{\mathcal{Q}}_{\text{G,MF}}$ and $\widetilde{\mathcal{Q}}_{\text{G}}$ are closed, equicontinuous, and uniformly bounded, which means that they are compact! We have therefore arrived at our desired consistency result.

Recall the metric $d_{\text{G,MF}}$ on $\widetilde{\mathcal{Q}}_{\text{G,MF}}$ defined in (3.28) and the metric $d_{\text{G}}$ on $\widetilde{\mathcal{Q}}_{\text{G}}$ defined in (3.29). By $o_{\mathbb{P}}(1)$, we denote a random variable that converges to zero in probability as $M \to \infty$. In the following statements, $o_{\mathbb{P}}(1)$ is models noise deriving stochastic optimisation procedures.

**Proposition 3.34** (Consistency of CNPA). *Assume that $\widetilde{I}$ is dense in $I_N$ for some $N \geq 1$, and assume that $\sup_{\mathbf{x} \in \widetilde{I}} |\mathbf{x}| < \infty$. Let $(\pi_M, \sigma_M) \in \widetilde{\mathcal{Q}}_{\text{G,MF}}$ be such that*

$$\mathcal{L}_M(\pi_M, \sigma_M) \leq \inf_{\widetilde{\mathcal{Q}}_{\text{G,MF}}} \mathcal{L}_M + o_{\mathbb{P}}(1). \tag{3.39}$$

*Then, as $M \to \infty$, the distance of $(\pi_M, \sigma_M)$ to the closest CNPA converges to zero in probability.*

*Proof.* See Appendix A.4. □

**Proposition 3.35** (Consistency of GNPA). *Assume that $\widetilde{I}$ is dense in $I_N$ for some $N \geq 2$, and assume that $\sup_{\mathbf{x} \in \widetilde{I}} |\mathbf{x}| < \infty$. Let $(\pi_M, \sigma_M) \in \widetilde{\mathcal{Q}}_{\text{G}}$ be such that*

$$\mathcal{L}_M(\pi_M, \sigma_M) \leq \inf_{\widetilde{\mathcal{Q}}_{\text{G}}} \mathcal{L}_M + o_{\mathbb{P}}(1). \tag{3.40}$$

*Then, as $M \to \infty$, the distance of $(\pi_M, \sigma_M)$ to the GNPA converges to zero in probability.*

*Proof.* See Appendix A.4. □

We end our discussion of consistency of neural processes with a remark on the mode of convergence for Gaussian neural processes. Consider a GNPA $\pi \in \widetilde{\mathcal{Q}}_{\text{G}}$ and let $D \in \widetilde{\mathcal{D}}$ be some data set. Proposition 3.35 gives convergence in $d_{\text{G}}$, which in turn implies that $P_{\mathbf{x}} \pi_M(D) \rightharpoonup P_{\mathbf{x}} \pi(D)$ for all $\mathbf{x} \in I$. For estimating expectations of functions which depend on only finitely many function values, e.g. $\mathbb{E}_{\pi(D)}[L(f(\mathbf{x}))]$ where $L$ is continuous and bounded, this mode of convergence is appropriate. However, we might want to estimate expectations of functions which depend on the entire realisation, like the position of the maximum: $\mathbb{E}_{\pi(D)}[\arg\max_{x \in \mathcal{X}} f(x)]$. To estimate such expectations, we require a stronger mode of convergence, namely $\pi_M(D) \rightharpoonup \pi(D)$. Conveniently, again owing to the universal Hölder condition, Proposition A.8 in Appendix A.4 combined with Theorem 7.5 by Billingsley (1999) show that $P_{\mathbf{x}} \pi_i(D) \rightharpoonup P_{\mathbf{x}} \pi(D)$ for all $\mathbf{x} \in I$ in fact implies that $\pi_i(D) \rightharpoonup \pi(D)$.

## 3.6 Conclusion

This chapter has presented a number of definitions and results collectively called *prediction map approximation*. This theoretical framework allowed us to formally define and analyse neural processes. We now summarise the main results and state the main takeaways.

Neural processes are various approximations of the *posterior prediction map* (Definition 3.1). Although all neural processes attempt to approach the posterior prediction map, some classes of neural processes come closer than others. In particular, in the limit of infinite data, different classes of neural processes approach different objects. We call the object that a class of neural processes approaches a *neural process approximation* (Definition 3.15).

The class of neural processes first proposed by Garnelo et al. (2018a) is the class of *conditional neural processes* (Definition 3.22). Conditional neural processes have Gaussian predictions which *do not* model dependencies. In the limit of infinite data, conditional neural processes approach a *conditional neural process approximation* (Definition 3.24). A conditional neural process approximation recovers the mean map (Definition 3.18) of the posterior prediction map (Proposition 3.26). It also recovers the sum of the variance map (Definition 3.20) and the observation noise. A conditional neural process approximation, however, cannot separate the two. We therefore say that conditional neural processes fails to separate epistemic and aleatoric uncertainty.

In addition to conditional neural processes, we introduced the class of *Gaussian neural processes* (Definition 3.23). Gaussian neural processes have Gaussian predictions which *do* model dependencies. In the limit of infinite data, Gaussian neural processes approach the *Gaussian neural process approximation* (Definition 3.25). The Gaussian neural process approximation recovers the mean map, kernel map (Definition 3.19), and observation noise of the posterior prediction map (Proposition 3.27). In particular, Gaussian neural processes are able to separate epistemic and aleatoric uncertainty. Because the Gaussian neural process approximation recovers the mean map and kernel map, we also say that the Gaussian neural process approximation *moment matches* the posterior prediction map.

In practice, neural processes are learned by maximising the probability of the target sets under the predictions given the context sets (Garnelo et al., 2018b). We can consider this objective in the limit of infinite data. We call this infinite-data objective the *neural process objective* (Definition 3.10). Neural process approximations, the infinite-data limits of neural processes, are minimisers of the neural process objective (Definitions 3.15, 3.24 and 3.25). To successfully deploy a neural process, we want what we compute in practice to converge to the minimiser of the neural process objective. We call this property *consistency*.

First, to achieve consistency, in the general case the target inputs must be of arbitrarily large size (Proposition 3.13). This is not practical. Conditional and Gaussian neural processes relax this requirement. To approximate the Gaussian neural process approximation, it suffices to consider target inputs of size just $N$, as long as $N \geq 2$ (Proposition 3.27); and to approximate a conditional neural processes approximation, one may even consider target inputs of size only one (Proposition 3.26).

Second, to achieve consistency, we must prevent overfitting. Overfitting happens if the neural process is arbitrarily flexible. To prevent overfitting, the appropriate technical notion is that the variational family of the neural process must be *compact* (Section 3.5). By making reasonable assumptions about the data and the ground-truth stochastic process (Assumptions 3.29 to 3.31), for conditional and Gaussian neural processes, compactness can reasonably be satisfied. This, in turn, guarantees consistency of conditional and Gaussian neural processes (Propositions 3.34 and 3.35).

Our investigation shed light on some of the more subtle issues concerning neural processes. However, many important questions still remain unsolved. To begin with, although we established a consistency result for conditional and Gaussian neural processes, the more important question is how much data we need, quantitatively, to come sufficiently close to the neural process approximations. Additionally, the aleatoric noise has been Gaussian and homogeneous. It would be more realistic to generalise this to non-Gaussian and heterogeneous noise. As long as the noise distribution satisfies reasonable conditions and the parameters vary continuously with the input, this generalisation should work out. Finally, in practice, we do not optimise over the compact family proposed by Section 3.5, but use neural networks. Although we will not hope to achieve finite-sample guarantees for neural networks, perhaps the proposed compact family can be better connected with practice.

# 4 | Representation Theorems

**Abstract.** In Section 2.4, we saw that neural processes approach a meta-learning problem by parametrising a prediction map $\pi_\theta \colon \mathcal{D} \to \mathcal{Q}$ using an *encoder–decoder architecture*. In this chapter, we will see how encoder–decoder architectures can be motivated theoretically with *representation theorems*. In the context of neural processes, representation theorems say that prediction maps may be decomposed as encoder–decoder architectures without losing representational power.

**Outline.** In Section 4.1, we introduce representation theorems; and in Section 4.2, we introduce functions on data sets. Then, in Section 4.3, we review a basic but important representation theorem called *deep sets* (Zaheer et al., 2017; Edwards et al., 2017; Wagstaff et al., 2019). In Section 4.4, we present an extension of deep sets called *convolutional deep sets*. Convolutional deep sets incorporate *translation equivariance*. Finally, in Section 4.5, we generalise translation equivariance to *diagonal translation equivariance* and present a generalisation of convolutional deep sets to this symmetry.

**Attributions and relationship to prior work.** Theorem 4.8 was first presented as Theorem 1 by Gordon, Bruinsma, Foong, Requeima, Dubois, and Turner (2020). The proof for Theorem 1 was originally developed by the author and Jonathan Gordon and later improved by Andrew Y. K. Foong. We furthermore acknowledge David Burt and Mark Rowland for reviewing the proof. Theorem 4.9 was first presented as Theorem E.1 by Bruinsma, Requeima, Foong, Gordon, and Turner (2021c). The proof of Theorem E.1 was originally developed by the author and James Requeima and later checked by Jonathan Gordon and Andrew Y. K. Foong. All work was supervised by Richard E. Turner.

## 4.1 Introduction

As we discussed in Section 2.4, designing a neural process involves parametrising a prediction map $\pi_\theta \colon \mathcal{D} \to \mathcal{P}$. In parametrising a prediction map, we argued that there are two main challenges. First, data sets vary in number of data points, which means that $\pi_\theta$ must process inputs of varying dimensionality. Second, data sets with different orderings of the data points are fundamentally the same, which means that $\pi_\theta(D)$ should not depend on an ordering of the elements in $D$. We then argued that these challenges can be addressed by parametrising $\pi_\theta$ with an *encoder–decoder architecture*. Encoder–decoder architectures decompose $\pi_\theta = \text{dec}_\theta \circ \text{enc}_\theta$ into an *encoder* and *decoder*. The encoder does not depend on the order of the elements and produces an *encoding* that always has the same format. In Section 4.3, we will review deep sets, which is a *representation theorem* proved by Zaheer et al. (2017) that motivates the encoder–decoder architecture on a theoretical level.

In the context of neural processes, the study of representation theorems more generally means the study of functions on data sets $\mathcal{D} \to Z$, where $Z$ is some codomain. The supposition of this study is that we would like to efficiently represent such functions on a computer. The main problem, as have seen, is that it is not at all clear how to do that, due to the structure of $\mathcal{D}$. Representation theorems provide general decompositions of functions $\mathcal{D} \to Z$ without losing representational power. Importantly, if the components of this decomposition can be implemented on a computer, then the theorems provide ways to generally implement functions $\mathcal{D} \to Z$ on a computer. In particular, by applying representation theorems to prediction maps, we find ways to concretely parametrise and implement prediction maps on a computer. The application of representation theorems to prediction maps will be what we focus on in the next chapter, Chapter 5. In this chapter, we take a step back from the neural process setup and more generally consider the study of functions on data sets. We will, however, remain to use the language of "encoder", "encoding" and "decoder" to smoothly transition back in the next chapter.

In practice, the components of a representation theorem will be implemented with neural networks. For the neural network to be able to approximate these components, the components must be *continuous* (Cybenko, 1989). For this reason, representation theorems always guarantee that their components are continuous. Without the requirement of continuity, more powerful statements are possible, but these statements are of little practical utility (Section 3; Wagstaff et al., 2019). Before proceeding to our study of representation theorems, we first more carefully define functions on data sets.

## 4.2   Functions on Data Sets

For simplicity, like in the previous chapter, we will assume that the inputs and outputs are one-dimensional: $\mathcal{X} \subseteq \mathbb{R}$ and $\mathcal{Y} \subseteq \mathbb{R}$. All results, however, may readily be extended to multidimensional inputs and outputs.

In our developments up to this point, a data set $D \in \mathcal{D}_N$ has been an element of $(\mathcal{X} \times \mathcal{Y})^N$. In particular, $D$ is associated to an ordering of the input–output pairs. However, two data sets with different orderings of the data points are fundamentally the same. To this end, we will consider data sets equal if their input–output pairs are equal up to a permutation, which we now formalise. For a permutation $\sigma \in \mathbb{S}^N$ and vector $\mathbf{z} \in \mathbb{R}^N$, denote $\sigma \mathbf{z} = (z_{\sigma(1)}, \ldots, z_{\sigma(N)})$. For a data set $D = (\mathbf{x}, \mathbf{y}) \in \mathcal{D}^N$, denote $\sigma D = (\sigma \mathbf{x}, \sigma \mathbf{y})$. Then define the equivalence relation $\sim$ by $D_1 \sim D_2$ whenever there exists a permutation $\sigma \in \mathbb{S}^N$ such that $\sigma D_1 = D_2$. Denote the equivalence class of $D$ by $[D]$, and denote the collection of all equivalence classes of $\mathcal{D}_N$ by $[\mathcal{D}_N] = \mathcal{D}_N/\sim$. An element $[D] \in [\mathcal{D}_N]$ represents a data set that does not depend on the ordering of the data points. Like before, let $[\mathcal{D}] = \bigcup_{N \geq 0} [\mathcal{D}_N]$ be the collection of all data sets of finite size, which includes the empty set $\varnothing$. In this chapter, we additionally let $[\mathcal{D}_{\leq N}] = \bigcup_{n=0}^{N} [\mathcal{D}_n]$ be the collection of all data sets of size at most $N$, again including the empty set $\varnothing$. More generally, for a subset $A \subseteq \mathcal{D}$, denote $[A] = \{[D] : D \in A\}$. We then have the following definition.

**Definition 4.1** (Function on data sets). *A function on data sets $f$ is a function $f\colon [\mathcal{D}] \to Z$, for some codomain $Z$.*

An example of a function on data sets is the function which computes the average observed value:

$$f\colon [\mathcal{D}] \to \mathbb{R}, \quad f([D]) = \sum_{n=1}^{|\mathbf{x}|} y_n. \tag{4.1}$$

We will be concerned with *continuity* of functions on data sets, which means that we must endow $[\mathcal{D}]$ with a topology. We previously endowed $\mathcal{D}$ with the metric

$$d_{\mathcal{D}}(D_1, D_2) = \begin{cases} \|\mathbf{x}_1 - \mathbf{x}_2\|_2 + \|\mathbf{y}_1 - \mathbf{y}_2\|_2 & \text{if } |\mathbf{x}_1| = |\mathbf{x}_2|, \\ \infty & \text{otherwise.} \end{cases} \tag{4.2}$$

For a data set $D$, let $|D|$ denote the number of data points. Endow $[\mathcal{D}]$ with the following metric:

$$d_{[\mathcal{D}]}([D_1], [D_2]) = \inf_{\sigma_2 \in \mathbb{S}^{|D_2|}} d_{\mathcal{D}}(D_1, \sigma_2 D_2). \tag{4.3}$$

We collect some basic results in the following proposition. Call a set $A \subseteq \mathcal{D}$ *permutation invariant* if $D \in A$ implies that $\sigma D \in A$ for all permutations $\sigma \in \mathbb{S}^{|D|}$.

**Proposition 4.2.**

(1) *The function $d_{[\mathcal{D}]}$ is a metric on $[\mathcal{D}]$.*

(2) *If $A \subseteq \mathcal{D}$ is open, then $[A]$ is open in the topology of $d_{[\mathcal{D}]}$.*

(3) *If $A \subseteq \mathcal{D}$ is closed and permutation invariant, then $[A]$ is closed in the topology of $d_{[\mathcal{D}]}$.*

(4) *The topology on $[\mathcal{D}]$ induced by $d_{[\mathcal{D}]}$ coincides with the quotient topology.*

*Proof.* See Appendix B.1. □

For Propositions 4.2.(2) and 4.2.(3), since the topology on $[\mathcal{D}]$ induced by $d_{[\mathcal{D}]}$ coincides with the quotient topology, $[A]$ is also respectively open and closed in the quotient topology.

Call a function $f\colon \mathcal{D} \to Z$ permutation invariant if $f(\sigma D) = f(D)$ for all $\sigma \in \mathbb{S}^{|D|}$. We then have the following result.

**Proposition 4.3.** *Suppose that $f\colon \mathcal{D} \to Z$ is continuous and permutation invariant. Then $[f]\colon [\mathcal{D}] \to Z$ defined by $[f]([D]) = f(D')$ for any $D'$ such that $D' = [D]$ is well defined and continuous. Conversely, suppose that $[f]\colon [\mathcal{D}] \to Z$ is continuous. Then $f\colon \mathcal{D} \to Z$ defined by $f(D) = [f]([D])$ is continuous and permutation invariant.*

*Proof.* See Appendix B.1. □

Proposition 4.3 says that it does not matter whether we study continuous functions on $\mathcal{D}$ which are permutation invariant or continuous functions on $[\mathcal{D}]$; they are essentially the same. Therefore, in this chapter, we will confine ourselves to continuous functions on $[\mathcal{D}]$ and will not be concerned with the notion of permutation invariance.

Qi et al. (2017) investigate functions on sets rather than on data sets. In this more general setting, Qi et al. use the Hausdorff distance to measure distance between sets. In our case, the collection of all data sets $[\mathcal{D}]$ has more structure, because it is a quotient by a group of isometries. This structure allows us to employ the infimum-based metric in (4.3); in general, such a definition might violate the triangle inequality. In this light, Proposition 4.2.(4) is a specialisation of Theorem 2.1.(ii) by Cagliari et al. (2015) to $[\mathcal{D}]$.

## 4.3 Deep Sets

The starting point of our study of functions on data sets is the following theorem by Zaheer et al. (2017), which characterises functions on $[\mathcal{D}_N]$.

**Theorem 4.4** (Deep set, preliminary; Zaheer et al., 2017)**.** *Assume that $\mathcal{X} \subseteq \mathbb{R}$ and $\mathcal{Y} \subseteq \mathbb{R}$ are compact. A map $\pi\colon [\mathcal{D}_N] \to Z$ is continuous if and only if it is of the form*

$$\pi = \mathsf{dec} \circ \mathsf{enc} \qquad \textit{where} \qquad \mathsf{enc}(D) = \sum_{(x,y)\in D} \phi(x,y) \qquad (4.4)$$

*with $\mathsf{enc}\colon [\mathcal{D}_N] \to \mathbb{R}^{2(N+1)}$, $\phi\colon \mathcal{X} \times \mathcal{Y} \to \mathbb{R}^{2(N+1)}$, and $\mathsf{dec}\colon \mathbb{R}^{2(N+1)} \to Z$ continuous. Moreover, the choice of $\mathsf{enc}$ and $\phi$ do not depend on $\pi$.*

*Proof.* See the proof of Theorem 2 by Zaheer et al. (2017). □

Thoughout this thesis, we will call functions of the form of (4.4) *deep sets*. In terms of encoders, note that deep sets encode on two levels. For every data point $(x, y) \in D$, first $\phi(x, y)$ produces an encoding of that data point. These data-point-specific encodings are then summed over all data points in the data set, finally producing an encoding of the whole data set $D$. Although Theorem 4.4 was originally presented by Zaheer et al. (2017), in concurrent work Edwards et al. (2017) employed a similar approach for a similar problem. Theorem 4.4 is also comparable to Theorem 1 by Qi et al. (2017). Theorem 1 by Qi et al. (2017) uses a maximum instead of a sum for enc and characterises function continuous with respect to a different topology.

To prove Theorem 4.4, Zaheer et al. proceed by defining

$$\phi(x, y) = (x^0, x^1, \ldots, x^{N+1}, y^0, y^1, \ldots, y^{N+1}). \qquad (4.5)$$

Indeed, $\phi(x, y)$ is of dimensionality $2(N+1)$. It is clear that enc is continuous. The key step is to show that $\text{enc} \colon [\mathcal{D}_N] \to \mathbb{R}^{2(N+1)}$ is also *injective*, which means that it always maps two different $[D_1] \in [\mathcal{D}_N]$ and $[D_2] \in [\mathcal{D}_N]$ to distinct encodings. Zaheer et al. show this in their Lemma 4. Having shown that $\phi$ is a continuous injection, it follows that it is a continuous bijection from onto its image $\text{enc}([\mathcal{D}_N])$. The fact that we now use is that all continuous bijections $A \to B$ with $A$ compact and $B$ Hausdorff have a continuous inverse (Theorem 26.6; Munkres, 2000). Since $[\mathcal{D}_N]$ is compact, we may conclude that $\phi \colon [\mathcal{D}_N] \to \phi([\mathcal{D}_N])$ has a continuous inverse $\phi^{-1}$. Theorem 4.4 then follows by setting $\text{dec} = \pi \circ \phi^{-1}$.

Theorem 4.4 has an important limitation: it only applies to data sets of a single fixed size $N$. Wagstaff et al. later extended Theorem 4.4, addressing this restriction:

**Theorem 4.5** (Deep set; Zaheer et al., 2017; Wagstaff et al., 2019). *Assume that $\mathcal{X} \subseteq \mathbb{R}$ and $\mathcal{Y} \subseteq \mathbb{R}$ are compact. A map $\pi \colon [\mathcal{D}_{\leq N}] \to Z$ is continuous if and only if it is of the form*

$$\pi = \text{dec} \circ \text{enc} \qquad \text{where} \qquad \text{enc}(D) = \sum_{(x,y) \in D} \phi(x, y) \qquad (4.6)$$

*with $\text{enc} \colon [\mathcal{D}_N] \to \mathbb{R}^{2N}$, $\phi \colon \mathcal{X} \times \mathcal{Y} \to \mathbb{R}^{2N}$, and $\text{dec} \colon \mathbb{R}^{2(N+1)} \to Z$ continuous. Moreover, the choice of enc and $\phi$ do not depend on $\pi$.*

*Proof.* See the proof of Theorem 4.4 by Wagstaff et al. (2019). □

Compared to Theorem 4.4, there are two differences. First, in Theorem 4.5, $\phi$ maps into $\mathbb{R}^{2N}$ rather than $\mathbb{R}^{2(N+1)}$. Wagstaff et al. make this improvement by noting that the elements $x^0$ and $y^0$ in (4.5) are always one and can therefore be removed. Second, in Theorem 4.5, the theorem applies to any function on $[\mathcal{D}_{\leq N}]$ rather than $[\mathcal{D}_N]$. This means that it applies to all data set of size at most $N$, addressing the aforementioned limitation of Theorem 4.4. Wagstaff et al. make this improvement by designating a special element in $\mathbb{R} \setminus \mathcal{X}$. Let us denote this special element by missing $\in \mathbb{R} \setminus \mathcal{X}$. Using missing, any data set $D \in \mathcal{D}_{N'}$ with $N' < N$ can be made a data set of $N$ data points by appending the appropriate number of missings. In this way, $\mathcal{D}_{\leq N}$ can be embedded into the collection of data sets with $N$ data points with inputs in $\mathcal{X} \cup \{\text{missing}\}$ and outputs in $\mathcal{Y} \cup \{\text{missing}\}$. Since $\mathcal{X} \cup \{\text{missing}\}$ and $\mathcal{Y} \cup \{\text{missing}\}$ are still compact, Theorem 4.4 can be applied.

Theorem 4.5 is an important result, because it provides a way to concretely parametrise a map $\pi \colon [\mathcal{D}_{\leq N}] \to Z$ on a computer, supposing that $Z$ is some Euclidean space. Namely, in (4.6), the components $\phi$ and dec are just continuous functions between Euclidean spaces, which can therefore be approximated with multi-layer perceptrons (Cybenko, 1989).

## 4.4 Convolutional Deep Sets

Although Theorem 4.5 provides a way to generally implement a map $\pi \colon [\mathcal{D}_{\leq N}] \to Z$, this implementation requires two multi-layer perceptrons: one for $\phi$, and one for dec. Although multi-layer perceptrons (MLPs) can approximate any continuous function (Cybenko, 1989), in some domains, MLPs are so parameter inefficient that they cannot feasibly be used. One such domain is images. Consider a $100 \times 100$ image, consisting of three colour channels. Then a one-layer MLP mapping to an image of the same size would consist of nearly a billion parameters, which is on the order of the biggest models that can nowadays be trained (Brown et al., 2020). Specifically for the domain of images, convolutional neural networks (CNNs) were developed (Fukushima et al., 1982), which are MLPs with certain weights set to zero. In comparison, a one-layer CNN mapping a $100 \times 100$ image with three colour channels to an image of the same size may only consist of only 81 parameters. CNNs achieve this parameter efficiency by incorporating a symmetry called *translation equivariance*. Recall that, in Definitions 2.3 and 2.4 in Section 2.5, we defined translations and translation equivariance. We also explained the intuition behind translation equivariance. In this section, we present a translation-equivariant version of deep sets (Theorem 4.5) called *convolutional deep sets* (Theorem 4.8). As the name suggests, convolutional deep sets are to deep sets what CNNs are to MLPs. In particular, convolutional deep sets inherit the parameter efficiency of CNNs.

In this section, for simplicity, we still assume one-dimensional inputs and outputs, and we emphasise again that all results readily generalise to higher-dimensional inputs and outputs. For the notion of translation equivariance to be well defined, we require that $\mathcal{X} = \mathbb{R}$ entirely. If $\mathcal{X} \subsetneq \mathbb{R}$, then $\mathsf{T}_\tau x$ could shift an input outside $\mathcal{X}$. As an example, if $\mathcal{X} = [0, 1]$, then $\mathsf{T}_{\frac{1}{2}} 1 = \frac{3}{2} \notin \mathcal{X}$.

Every CNN is a translation-equivariant MLP. It turns out that every MLP which is transla-tion equivariant is also a CNN (Kondor et al., 2018). This means that, amongst all MLPs, translation equivariance *characterises* CNNs. In a similar spirit, the main result of this section is a characterisation of functions on data sets which are translation equivariant. Let us for a moment contemplate such a result. Suppose that

$$\pi = \mathsf{enc} \circ \mathsf{dec} \quad \text{where} \quad \mathsf{enc}(D) = \sum_{(x,y) \in D} \phi(x,y) \tag{4.7}$$

were translation equivariant: $\pi \circ \mathsf{T}_\tau = \mathsf{T}_\tau \circ \pi$ for all $\tau \in \mathcal{X}$. To establish translation equivariance of $\pi$, one reasonable construction would be to ensure that enc and dec are translation equivariant. If that were true, then, by associativity of function composition,

$$\mathsf{dec} \circ \mathsf{enc} \circ \mathsf{T}_\tau \stackrel{\text{(enc is TE)}}{=} \mathsf{dec} \circ \mathsf{T}_\tau \circ \mathsf{enc} \stackrel{\text{(dec is TE)}}{=} \mathsf{T}_\tau \circ \mathsf{dec} \circ \mathsf{enc} \tag{4.8}$$

which would mean that $\pi$ is indeed translation equivariant. But there is a big problem with this construction: enc is a function mapping into $\mathbb{R}^{2N}$ (Theorem 4.5), and, crucially, vectors in $\mathbb{R}^{2N}$ cannot naturally be translated by translations in $\mathcal{X} = \mathbb{R}$! One way to translate a vector in $\mathbb{R}^{2N}$ is to shift along its elements, wrapping around the outermost elements:

$$\mathsf{T}_1(z_1, z_2, \ldots, z_{2N-1}, z_{2N}) = (z_{2N}, z_1, \ldots, z_{2N-2}, z_{2N-1}). \tag{4.9}$$

This, however, only works for integer-valued translations. Another proposal might be to simply add the translation to all elements of the vector:

$$\mathsf{T}_\tau(z_1, \ldots, z_{2N}) = (z_1 + \tau, \ldots, z_{2N} + \tau). \tag{4.10}$$

But then what would you do for two-dimensional inputs? In general, there is no natural way to translate vectors in $\mathbb{R}^{2N}$ by $\mathcal{X}$-valued translations. We therefore see that the Euclidean space $\mathbb{R}^{2N}$ in which the encoding lives is an obstacle for deriving a translation-equivariant version of deep sets. Instead, we would like the encoding to live in a space on which translations can naturally act. And the most natural such space is a space of *functions*, which motivates the following idea.

**Definition 4.6** (Functional encoding). *Call an encoding* functional *if the output of* enc *and hence the input of* dec *is a function on* $\mathcal{X}$:

$$\mathsf{enc} \colon [\mathcal{D}] \to Z^\mathcal{X}, \quad \mathsf{dec} \colon Z^\mathcal{X} \to C \tag{4.11}$$

*where $Z$ is the codomain of the functional encoding and $C$ the codomain of the decoder.*

If the encoding were functional, then translation equivariance is a perfectly well-defined notion for the output of enc and the input of dec. Therefore, in that case, we could pursue the construction of building translation equivariance into $\pi$ by building it into enc and dec.

Before we state the main result, we need a few definitions. Call a space $A$ a *translation space* if elements in $A$ can be translated. Call a translation space $A$ *topological* if $(\tau, a) \mapsto \mathsf{T}_\tau a$ is continuous. Call a subset $A' \subset A$ of a translation space $A$ *closed under translations* if $a \in A'$ implies that $\mathsf{T}_\tau a \in A'$ for all $\tau \in \mathcal{X}$. Call a function $k\colon \mathcal{X} \times \mathcal{X} \to \mathbb{R}$ *(strictly) positive definite* if, for all $N \in \mathbb{N}$ and $\mathbf{x} \in \mathcal{X}^N$, the matrix

$$\begin{bmatrix} k(x_1, x_1) & \cdots & k(x_1, x_n) \\ \vdots & \ddots & \vdots \\ k(x_n, x_1) & \cdots & k(x_n, x_n) \end{bmatrix} \tag{4.12}$$

is (strictly) positive definite. Call a function $k\colon \mathcal{X} \to \mathbb{R}$ (strictly) positive definite if $(x, y) \mapsto k(x - y)$ is (strictly) positive definite. For a Hilbert space $(\mathbb{H}, \langle \,\cdot\,, \,\cdot\, \rangle_{\mathbb{H}})$ of functions on $\mathcal{X}$, call a positive-definite function $k$ a *reproducing kernel* if $k(\,\cdot\,, x) \in \mathbb{H}$ for all $x \in \mathcal{X}$ and $\langle f, k(\,\cdot\,, x) \rangle_{\mathbb{H}} = f(x)$ for all $f \in \mathbb{H}$ and $x \in \mathcal{X}$. For every positive-definite function $k$, there exists one and only one Hilbert space of functions on $\mathcal{X}$ for which $k$ is a reproducing kernel (Moore–Aronszajn theorem; Section 2; Aronszajn, 1950). We call this Hilbert space the *reproducing kernel Hilbert space* (RKHS) of $k$. Finally, we define the *multiplicity* of a data set $[D] \in [\mathcal{D}]$, which counts how many times an input is repeated exactly.

**Definition 4.7** (Multiplicity of data set). *For a data set $[D] \in [\mathcal{D}]$, let the* multiplicity *of $[D]$ be the maximum number of times an input is repeated exactly. Denote the multiplicity of $[D]$ by* $\mathrm{mult}\,[D]$. *Then*

$$\mathrm{mult}\,[D] = \sup\,\{|\{i \in [N] : x_i = x\}| : x \in \{x_1, \ldots, x_N\}\} \;\; \text{where}\; D = (\mathbf{x}, \mathbf{y}). \tag{4.13}$$

*For a subset $[\mathcal{D}'] \subseteq [\mathcal{D}]$, define* $\mathrm{mult}\,[\mathcal{D}'] = \sup_{[D] \in [\mathcal{D}']} \mathrm{mult}\,[D]$.

**Theorem 4.8** (Convolutional deep set). *Let $\mathcal{Y} \subseteq \mathbb{R}$ be compact. Suppose that $[\mathcal{D}'] \subseteq [\mathcal{D}]$ is closed, is closed under translations, has multiplicity $K$, and has maximum data set size $N < \infty$. Let $k\colon \mathcal{X} \to \mathbb{R}$ be a continuous strictly-positive-definite function such that (1) $k(0) = \sigma^2 > 0$ (2) $k \geq 0$, and (3) $k(\tau) \to 0$ as $|\tau| \to \infty$. Denote the reproducing kernel Hilbert space of $k$ by $\mathbb{H}$. Let $Z$ be a translation space. Then a function $\pi\colon [\mathcal{D}'] \to Z$ is continuous and translation equivariant if and only if it is of the form*

$$\pi = \mathsf{dec} \circ \mathsf{enc} \quad \text{where} \quad \mathsf{enc}(D) = \sum_{(x,y) \in D} \phi(y) k(\,\cdot\, - x) \tag{4.14}$$

*with* $\mathsf{enc}\colon [\mathcal{D}'] \to \mathbb{H}'$ *continuous and translation equivariant,* $\mathsf{dec}\colon \mathbb{H}' \to Z$ *continuous and* *translation equivariant, and $\phi(y) = (0, y^1, \ldots, y^K)$. Here $\mathbb{H}' = \mathsf{enc}([\mathcal{D}'])$ is a subspace of $\mathbb{H}^{K+1}$ which is closed and closed under translations.*

*Proof.* The proof proceeds like the proof for Theorem 4.5. We show that enc is injective, therefore bijective onto its image. In this case, however, the image of enc is not compact. Instead, we use the structure of the RKHS $\mathbb{H}$ to prove continuity of the inverse of enc. See Appendix B.2. □

Theorem 4.8 is a characterisation of functions $\pi \colon [\mathcal{D}'] \to Z$ on data sets which are translation equivariant. In the following paragraphs, we will carefully analyse various aspects of the theorem.

A restrictive assumption of Theorem 4.8 is that the domain $[\mathcal{D}']$ must be simultaneously closed and have multiplicity $K$. For example, suppose that we choose $[\mathcal{D}']$ to be the data sets in $[\mathcal{D}_2]$ with distinct inputs. Then $\mathrm{mult}\,[\mathcal{D}'] = 1$. However, $[\mathcal{D}']$ is not closed: $[((1/n, 0), (0, 0))] \in [\mathcal{D}']$ for all $n \geq 1$, but

$$\lim_{n \to \infty} [((1/n, 0), (0, 0))] = [((0, 0), (0, 0))], \tag{4.15}$$

and $[((0,0), (0,0))]$ has multiplicity two, so it is not in $[\mathcal{D}']$. A reasonable way to construct subsets of $[\mathcal{D}']$ which are simultaneously closed and have multiplicity $K$ is to fix some $\varepsilon > 0$, and to consider the data sets in $[\mathcal{D}']$ with at most $K$ inputs exactly repeated and all other inputs $\varepsilon$ or more apart.

Like Theorem 4.5, Theorem 4.8 also lets $\phi$ be a power expansion (see (4.5)). However, the highest power of $\phi$ in Theorem 4.5 is dictated by the maximum data set size, whereas the highest power of $\phi$ in Theorem 4.8 is set by the multiplicity of $[\mathcal{D}']$. In particular, if the multiplicity is low, then the highest power of $\phi$ in Theorem 4.8 is low, regardless of the maximum data sets size.

As we alluded to before, Theorem 4.8 constructs a translation-equivariant encoder and decoder by employing a functional encoding (Definition 4.6). The particular function space that the encoding lives is the $(K+1)$-fold product of the reproducing kernel Hilbert space $\mathbb{H}$ of a continuous positive-definite kernel $k$. We are free to choose $k$. In Chapter 5, we will choose $k$ to be a simple Gaussian kernel: $k(\tau) = \exp(-\frac{1}{2\ell^2}\tau^2)$ for some length scale $\ell > 0$.

The decoder is defined only on a subspace $\mathbb{H}' \subseteq \mathbb{H}^{K+1}$ which is closed and closed under translations. It would be desirable to continuously extend the decoder to all of $\mathbb{H}^{K+1}$ whilst preserving translation equivariance. Continuously extending functions from a closed subspace to the entire space is classically done with the Tietze extension theorem (Theorem 35.1; Munkres, 2000). The Tietze extension theorem, however, assumes that $Z$ is a finite-dimensional Euclidean space and does not preserve translation equivariance. In 1950, Gleason extended Tietze's theorem to preserve symmetries with respect to compact groups by averaging over the Haar measure. The group of translations, unfortunately, is not compact, though it is locally compact. In 1951, Dugundji extended Tietze's theorem to any $Z$ which is a locally convex vector space. Later Jaworowski (1981) and more recently Feragen (2006) proposed extensions of Tietze's theorem which generalise in both directions. We leave it to future work to determine whether an equivariant extension theorem can be applied to extend the decoder from $\mathbb{H}'$ to all of $\mathbb{H}^{K+1}$.

Finally, we compare the form of convolution deep sets (Theorem 4.8) to the form deep sets (Theorem 4.5). There are two important differences. First, for deep sets, the encoder is implemented with an MLP, and the decoder is also implemented with an MLP. For convolutional deep sets, the encoder only depends on a choice for $k$, like the Gaussian kernel. It can therefore trivially be implemented and, in particular, it *does not depend on any neural network*. Additionally, the decoder is now a translation-equivariant map between translation spaces. Crucially, such maps are exactly what CNNs approximate (Theorem 3.1; Yarotsky, 2022)! Therefore, for convolutional deep sets, the decoder can be implemented with a CNN. The second difference pertains to the choice of maximum data set size $N$. For deep sets, the dimensionality of the encoding depends on $N$. For convolutional deep sets, on the other hand, the encoding always lives in $\mathbb{H}^{K+1}$, regardless of the maximum data set size. We could therefore imagine applying Theorem 4.8 to a sequence of $N$s going to infinity to obtain a version of convolution deep sets that holds for data sets of all sizes. We leave such a construction for future work.

After introducing Theorem 4.8 (Gordon et al., 2020), Xu et al. (2020) also introduced functional representations in the context of meta-learning for the purpose of improving representational capacity. Whereas Theorem 4.8 uses a functional encoding, Xu et al. argue that the encoder in Theorem 4.8 is limited. We, however, emphasise that the simplicity of the encoder is without loss of generality, because Theorem 4.8 applies to *any* continuous function $[\mathcal{D}'] \to Z$ which is translation equivariant. The simplicity of the encoder can therefore be seen as a feature rather than a downside.

## 4.5 Diagonal Translation Equivariance

Convolutional deep sets (Theorem 4.8) characterise translation-equivariant functions on data sets. In this section, we investigate a slightly more general class: functions on data sets which are *equivariant with respect to diagonal translations*. Before stating the definition of *diagonal translation equivariance* (DTE), we first explain where DTE comes from.

In Section 3.4, we defined *mean maps* (Definition 3.18), functions of the form $m\colon [\mathcal{D}] \to \mathcal{Y}^{\mathcal{X}}$, and *kernel maps* (Definition 3.19), functions of the form $k\colon [\mathcal{D}] \to \mathbb{R}^{\mathcal{X} \times \mathcal{X}}$. In Section 5.5, we will encounter mean maps $m$ which are translation equivariant:

$$m \circ \mathsf{T}_\tau = \mathsf{T}_\tau \circ m \qquad \text{for all } \tau \in \mathcal{X}. \tag{4.16}$$

Theorem 4.8 can be used to characterise such mean maps $m$. We will also encounter kernel maps $k$ which are translation equivariant. Kernel maps, however, turn out to be translation equivariant in a slightly different sense:

$$k \circ \mathsf{T}_\tau = \mathsf{T}_{(\tau,\tau)} \circ k \qquad \text{for all } \tau \in \mathcal{X}. \tag{4.17}$$

Compared to (4.16), (4.17) has $\mathsf{T}_{(\tau,\tau)}$ on the right-hand side instead of $\mathsf{T}_\tau$, where $\mathsf{T}_{(\tau,\tau)}$ is a translation on $\mathcal{X} \times \mathcal{X}$. Recall that the codomain of $k$ consists of functions on $\mathcal{X} \times \mathcal{X}$, not on $\mathcal{X}$. Because (4.17) has different symbols on both sides, we cannot directly apply Theorem 4.8. We will see that the symmetry in (4.17) is closely related to the notion of *diagonal translation equivariance* (Definition 4.10), which we later define. The main result of this section is a characterisation of functions which are *diagonally translation equivariant* in the sense of (4.17).

**Theorem 4.9** (Convolutional deep set for DTE)**.** *Let $\mathcal{Y} \subseteq \mathbb{R}$ be compact. Suppose that $[\mathcal{D}'] \subseteq [\mathcal{D}]$ is closed, is closed under translations, has multiplicity $K$, and has maximum data set size $N < \infty$. Let $k\colon \mathcal{X} \times \mathcal{X} \to \mathbb{R}$ be a continuous strictly-positive-definite function such that (1) $k(\mathbf{0}) = \sigma^2 > 0$, (2) $k \geq 0$, and (3) $k(\boldsymbol{\tau}) \to 0$ as $\|\boldsymbol{\tau}\|_2 \to \infty$. Denote the reproducing kernel Hilbert space associated to $k$ by $\mathbb{H}$. Let $Z$ be a topological $(\mathcal{X} \times \mathcal{X})$-translation space. Let $C$ be another topological $(\mathcal{X} \times \mathcal{X})$-translation space, let $c \in C$ be diagonally translation invariant and anti-diagonal discriminating, and denote $C' = \{\mathsf{T}_{\boldsymbol{\tau}} c : \boldsymbol{\tau} \in \mathcal{X} \times \mathcal{X}\}$. Then a function $\pi\colon [\mathcal{D}'] \to Z$ is continuous and diagonally translation equivariant in the sense of*

$$\pi \circ \mathsf{T}_\tau = \mathsf{T}_{(\tau,\tau)} \circ \pi \quad \text{for all } \tau \in \mathcal{X} \tag{4.18}$$

*if and only if it is of the form*

$$\pi = \mathrm{dec} \circ \mathrm{enc} \quad \text{where} \quad \mathrm{enc}(D) = \begin{bmatrix} \sum_{(x,y) \in C} \phi(y) k(\,\cdot\, - (x,x)) \\ c \end{bmatrix} \tag{4.19}$$

*with $\mathrm{enc}\colon [\mathcal{D}'] \to \mathbb{H}' \times C'$ continuous and translation equivariant, $\mathrm{dec}\colon \mathbb{H}^{K+1} \to Z$ continuous and translation equivariant, and $\phi(y) = (0, y^1, \ldots, y^K)$. Here $\mathbb{H}' = \mathrm{enc}([\mathcal{D}'])$ is a subspace of $\mathbb{H}^{K+1}$ which is closed and closed under translations.*

If $C$ is the space of functions $\mathcal{X} \times \mathcal{X} \to \mathbb{R}$ and $\mathcal{X} = \mathbb{R}$, then a simple and appropriate choice for $c$ is the Gaussian $c(\mathbf{x}) = \exp(-\frac{1}{2\ell^2}\langle \mathbf{x}, \mathbf{e}_\perp \rangle^2)$ for some length scale $\ell > 0$. Intuitively, this choice for $c$ measures the distance to the diagonal and decays based on that distance.

*Proof.* To prove Theorem 4.9, our approach will be to first rewrite (4.18) to have $\mathsf{T}_{(\tau,\tau)}$ on both sides. Once (4.18) is in this form, we will apply Theorem 4.8 with $\mathcal{X} \times \mathcal{X}$ as the space of inputs. Recall that Theorem 4.8 generalises to higher-dimensional inputs.

To rewrite (4.17), we use a simple trick: *duplicating the inputs*. Let $\mathcal{D}_{\text{dbl}}$ be the collection of data sets with inputs in $\mathcal{X}_{\text{dbl}} := \mathcal{X} \times \mathcal{X}$ and outputs in $\mathcal{Y}$. The "dbl" in the subscript of $\mathcal{D}_{\text{dbl}}$ stands for "<u>d</u>ou<u>bl</u>e" and serves to remind the reader that there are double the number of inputs. Let $\text{dup}\colon \mathcal{D} \to \mathcal{D}_{\text{dbl}}$ be the function that maps a data set into $\mathcal{D}_{\text{dbl}}$ by duplicating the inputs. More precisely, for $D = (\mathbf{x}, \mathbf{y}) \in \mathcal{D}_N$,

$$\text{dup}(D) = (((x_1, x_1), y_1), \ldots ((x_N, x_N), y_N)). \tag{4.20}$$

In (4.20), every input on the right-hand side has duplicated elements, so we say that such data sets have duplicated inputs. Let $\mathcal{D}_{\text{dup}} \subsetneq \mathcal{D}_{\text{dbl}}$ be the subset of $\mathcal{D}_{\text{dbl}}$ with duplicated inputs. Note that $\mathcal{D}_{\text{dbl}}$ contains data sets with and without duplicated inputs. Let $\text{dup}^{-1}\colon \mathcal{D}_{\text{dup}} \to \mathcal{D}$ be right inverse of $\text{dup}$: $\text{dup}^{-1}$ takes in a data set with duplicated inputs and recovers the original by *deduplicating the inputs*.

With these definitions, we can relate $\mathsf{T}_\tau \colon \mathcal{D} \to \mathcal{D}$ to $\mathsf{T}_{(\tau,\tau)}\colon \mathcal{D}_{\text{dbl}} \to \mathcal{D}_{\text{dbl}}$:

$$\mathsf{T}_\tau = \text{dup}^{-1} \circ \mathsf{T}_{(\tau,\tau)} \circ \text{dup}. \tag{4.21}$$

Define $\pi_{\text{dup}} := \pi \circ \text{dup}^{-1}$, and note that $\pi_{\text{dup}}\colon \mathcal{D}_{\text{dup}} \to Z$ operates on data sets with duplicated inputs. Then, by substituting (4.21) in (4.18), we find that[1]

$$\pi_{\text{dup}} \circ \mathsf{T}_{(\tau,\tau)} = \mathsf{T}_{(\tau,\tau)} \circ \pi_{\text{dup}} \qquad \text{for all } \tau \in \mathcal{X}. \tag{4.31}$$

---

[1] To carefully argue this, note that

$$\begin{aligned}
\mathsf{T}_{(\tau,\tau)} \circ \pi &= \pi \circ \mathsf{T}_\tau & \text{(apply (4.18))} & \quad (4.22) \\
&= \pi \circ (\text{dup}^{-1} \circ \mathsf{T}_{(\tau,\tau)} \circ \text{dup}) & \text{(definitions of dup and dup}^{-1}) & \quad (4.23) \\
&= (\pi \circ \text{dup}^{-1}) \circ \mathsf{T}_{(\tau,\tau)} \circ \text{dup}, & & \quad (4.24) \\
\mathsf{T}_{(\tau,\tau)} \circ \pi_{\text{dup}} &= \mathsf{T}_{(\tau,\tau)} \circ (\pi \circ \text{dup}^{-1}) & \text{(definition of } \pi_{\text{dup}}) & \quad (4.25) \\
&= (\mathsf{T}_{(\tau,\tau)} \circ \pi) \circ \text{dup}^{-1} & & \quad (4.26) \\
&= ((\pi \circ \text{dup}^{-1}) \circ \mathsf{T}_{(\tau,\tau)} \circ \text{dup}) \circ \text{dup}^{-1} & \text{(apply (4.24))} & \quad (4.27) \\
&= (\pi_{\text{dup}} \circ \mathsf{T}_{(\tau,\tau)} \circ \text{dup}) \circ \text{dup}^{-1} & \text{(definition of } \pi_{\text{dup}}) & \quad (4.28) \\
&= \pi_{\text{dup}} \circ \mathsf{T}_{(\tau,\tau)} \circ (\text{dup} \circ \text{dup}^{-1}) & & \quad (4.29) \\
&= \pi_{\text{dup}} \circ \mathsf{T}_{(\tau,\tau)}. & (\text{dup}^{-1} \text{ is right inverse of dup}) & \quad (4.30)
\end{aligned}$$

Crucially, (4.31) has $\mathsf{T}_{(\tau,\tau)}$ on both sides, so we may try to characterise $\pi_{\text{dup}}$ by applying Theorem 4.8 with $\mathcal{X} \times \mathcal{X}$ as the space of inputs! Having characterised $\pi_{\text{dup}}$, we also find a characterisation of $\pi = \pi_{\text{dup}} \circ \text{dup}$, which is what we are ultimately after. To apply Theorem 4.8 with $\mathcal{X} \times \mathcal{X}$ as the space of inputs, however, we require $\pi_{\text{dup}}$ to be $(\mathcal{X} \times \mathcal{X})$-translation equivariant, that is, equivariant with respect to *all* translations on $\mathcal{X} \times \mathcal{X}$:

$$\pi_{\text{dup}} \circ \mathsf{T}_{(\tau_1,\tau_2)} \stackrel{?}{=} \mathsf{T}_{(\tau_1,\tau_2)} \circ \pi_{\text{dup}} \qquad \text{for all } \tau_1, \tau_2 \in \mathcal{X}. \tag{4.32}$$

Unfortunately, (4.31) says that $\pi_{\text{dup}}$ is only equivariant with respect to translations along the *diagonal* of the space $\mathcal{X} \times \mathcal{X}$: translations of the form $\mathsf{T}_{(\tau_1,\tau_2)}$ where $\tau_1 = \tau_2$. This motivates the general definition of *diagonal translation equivariance*.

**Definition 4.10** (Diagonal translation equivariance; DTE). *Consider a map $\pi \colon A \to B$ where $A$ and $B$ are $(\mathcal{X} \times \mathcal{X})$-translation spaces. Then $\pi$ is* equivariant with respect to diagonal translations *or* diagonally translation equivariant *(DTE) if*

$$\pi \circ \mathsf{T}_{(\tau,\tau)} = \mathsf{T}_{(\tau,\tau)} \circ \pi \quad \text{for all } \tau \in \mathcal{X}. \tag{4.33}$$

If a map $\pi$ is $(\mathcal{X} \times \mathcal{X})$-translation equivariant, then it is also diagonally translation equivariant. The class of DTE functions is therefore larger than the class of $(\mathcal{X} \times \mathcal{X})$-TE functions. Whereas the previous section characterised functions on data sets which are TE, this section characterises functions on data sets which are DTE.

Diagonal translation equivariance is intimately related to $(\mathcal{X} \times \mathcal{X})$-translation equivariance. In technical terms, the group of diagonal translations is a *subgroup* of the group of $(\mathcal{X} \times \mathcal{X})$-translations. We will use this relationship to derive a characterisation of DTE functions from Theorem 4.8. The key idea is as follows. Consider a DTE map $\pi \colon A \to B$ between $(\mathcal{X} \times \mathcal{X})$-translation spaces $A$ and $B$. If $\pi$ were also $(\mathcal{X} \times \mathcal{X})$-TE, then we could apply Theorem 4.8, and we would be done. However, $\pi$ is not $(\mathcal{X} \times \mathcal{X})$-TE; it is just DTE. The idea is to *make $\pi$ $(\mathcal{X} \times \mathcal{X})$-TE* by *defining* what happens for non-diagonal translations:

$$\pi(\mathsf{T}_{(\tau_1,\tau_2)} a) := \mathsf{T}_{(\tau_1,\tau_2)} \pi(a) \quad \text{for all } \tau_1, \tau_2 \in \mathcal{X}, \tau_1 \neq \tau_2. \tag{4.34}$$

This *almost* works. The problem is that (4.34) might clash with the original definition of $\pi$: for some $a \in A$ and $\tau_1, \tau_2 \in \mathcal{X}$, $\mathsf{T}_{(\tau_1,\tau_2)} a$ might equal another $a' \in A$, and it might be that $\mathsf{T}_{(\tau_1,\tau_2)} \pi(a) \neq \pi(a')$. Fortunately, this problem can be circumvented, as we will now explain.

Let $C$ be another $(\mathcal{X} \times \mathcal{X})$-translation space, this time topological. Let $\mathbf{e}_\perp = (1, -1)$. Consider $c \in C$ such that, for all sequences $(\boldsymbol{\tau}_i)_{i \geq 1} \subseteq \mathcal{X}_{\text{dbl}}$, the sequence $(\langle \boldsymbol{\tau}_i, \mathbf{e}_\perp \rangle)_{i \geq 1}$ is convergent whenever $(\mathsf{T}_{\boldsymbol{\tau}_i} c)_{i \geq 1}$ is convergent. In particular, then $\mathsf{T}_{\boldsymbol{\tau}_\text{I}} c = \mathsf{T}_{\boldsymbol{\tau}_\text{II}} c$ for $\boldsymbol{\tau}_\text{I}, \boldsymbol{\tau}_\text{II} \in \mathcal{X}$ implies that $\langle \boldsymbol{\tau}_\text{I}, \mathbf{e}_\perp \rangle = \langle \boldsymbol{\tau}_\text{II}, \mathbf{e}_\perp \rangle$.[2] In words, by observing $c$, you can identify the *anti-diagonal* component of a translation. We call such $c$ *anti-diagonal discriminating*. Moreover, assume that $c$ is *diagonally translation invariant* (DTI), by which we mean that $\mathsf{T}_{(\tau,\tau)} c = c$ for all $\tau \in \mathcal{X}$. Then define the map

$$\overline{\pi} \colon A \times \{c\} \to B, \quad \overline{\pi}(a, c) = \pi(a). \tag{4.35}$$

Table 4.1: Comparison of deep sets and convolutional deep sets

|  | Deep sets (Thms 4.4 and 4.5) | Convolutional deep sets (Thms 4.8 and 4.9) |
| --- | --- | --- |
| Model for | Functions $[\mathcal{D}] \to Z$ | Functions $[\mathcal{D}] \to Z$ which are TE |
| Encoder | Multi-layer perceptron | Given; no neural network |
| Encoding | Vector in $\mathbb{R}^{2N}$ | Function in RKHS (Definition 4.6) |
| Decoder | Multi-layer perceptron | Convolutional neural network |

By DTE of $\pi$ and DTI of $c$, this map is still DTE: for all $\tau \in \mathcal{X}$:

$$\begin{aligned}
\overline{\pi}(\mathsf{T}_{(\tau,\tau)} a, \mathsf{T}_{(\tau,\tau)} c) &= \overline{\pi}(\mathsf{T}_{(\tau,\tau)} a, c) & (c \text{ is DTI}) & \quad (4.36)\\
&= \pi(\mathsf{T}_{(\tau,\tau)} a) & (\text{definition of } \overline{\pi}) & \quad (4.37)\\
&= \mathsf{T}_{(\tau,\tau)} \pi(a) & (\pi \text{ is DTE}) & \quad (4.38)\\
&= T_{(\tau,\tau)} \overline{\pi}(a, c). & (\text{definition of } \overline{\pi}) & \quad (4.39)
\end{aligned}$$

However, crucially, $c \neq \mathsf{T}_{(\tau_1, \tau_2)} c$ for any $\tau_1, \tau_2 \in \mathcal{X}$ such that $\tau_1 \neq \tau_2$, because $c$ is anti-diagonal discriminating. Therefore, $\overline{\pi}$ is undefined for any non-diagonal translation! This means that (4.34) now goes through, and we can finally apply Theorem 4.8.

There is one final technical detail to verify, which is that the extension $\overline{\pi}$ needs to be continuous. Appendix B.3 verifies this detail. □

## 4.6 Conclusion

In this chapter, we studied theorems which characterise functions on data sets. In the context of neural processes, we call such theorems *representation theorems*. The starting point was a theorem by Zaheer et al. (2017) and Wagstaff et al. (2019) called deep sets (Theorems 4.4 and 4.5). Deep sets generally characterise functions on data sets. In a deep set, the encoder and decoder can be implemented with MLPs.

---

[2] Consider the sequence $(\boldsymbol{\tau}_i)_{i \geq 1} \subseteq \mathcal{X}$ defined by $\boldsymbol{\tau}_i = \boldsymbol{\tau}_\text{I}$ for $i$ odd and $\boldsymbol{\tau}_i = \boldsymbol{\tau}_\text{II}$ for $i$ even.

The main contribution of this chapter is an extension of deep sets called *convolutional deep sets* (Theorems 4.8 and 4.9). Convolutional deep sets generally characterise functions on data sets which are *translation equivariant*. In a convolutional deep set, the encoder is trivially implemented and only depends on a choice of a positive-definite function. In particular, it does not depend on a neural network. Additionally, the decoder is implemented with a convolutional neural network. Table 4.1 summarises the key differences between deep sets and convolutional deep sets.

Convolutional deep sets have two important limitations. First, the approximation of the encoder relies on a convolutional neural network with convolutions of dimensionality equal to the dimensionality of $\mathcal{X}$ in the case of Theorem 4.8 and twice the dimensionality of $\mathcal{X}$ in the case of Theorem 4.9. One, two, or three-dimensional convolutions have reasonable computational expense. Convolutions of dimension four or higher, however, may be prohibitively expensive to run. This is particularly a problem for Theorem 4.9, where the convolutions have dimensionality twice the dimensionality of $\mathcal{X}$. Theorem 4.8 can therefore only reasonably be applied to data sets with one, two, or three-dimensional inputs and Theorem 4.9 to data sets only with one-dimensional inputs. Second, whereas Theorems 4.8 and 4.9 characterise functions which are translation equivariant, what we might really need is a characterisation of functions which are *approximately translation equivariant*. Ideally, such a result would "interpolate" between deep sets and convolutional deep sets, depending on how translation equivariant the function to represent would be. Van der Wilk et al. (2018) discuss approximate invariance in the setting of Gaussian processes.

Finally, whereas we introduced functional encodings (Definition 4.6) to derive a translation-equivariant version of deep sets, this construction can also be used to provide an alternative proof of deep sets. Theorem 2.4 by Gordon (2020) pursues this approach, which was developed by Jonathan Gordon in collaboration with Andrew Y. K. Foong.

# 5 | Convolutional Neural Processes

**Abstract.** Neural processes approach meta-learning problems by parametrising a prediction map $\pi_\theta \colon \mathcal{D} \to \mathcal{Q}$. In the previous chapter, we discussed *representation theorems*, which are theorems that generally characterise functions on data sets $\mathcal{D}$. The goal of representation theorems is to find practical implementations of such functions. In this chapter, we apply representation theorems to the prediction map.

**Outline.** In Section 5.2, we parametrise the prediction map using deep sets, recovering the original Conditional Neural Process (CNP; Garnelo et al., 2018a). In the remaining sections, we propose three methodological advancements. First, in Section 5.3, we parametrise the prediction map using *convolutional deep sets*. Second, in Section 5.5, we propose to directly parametrise the covariance between target outputs. Finally, in Section 5.6, we propose to deploy CNPs in an autoregressive fashion.

**Attributions and relationship to prior work.** The ConvCNP was originally presented by Gordon, Bruinsma, Foong, Requeima, Dubois, and Turner (2020), the ConvGNP by Markou, Requeima, Bruinsma, Vaughan, and Turner (2022), and the FullConvGNP by Bruinsma, Requeima, Foong, Gordon, and Turner (2021c). The ConvCNP is the result of a collaboration with Jonathan Gordon, Andrew Y. K. Foong, James Requeima, and Yann Dubois. Andrew Y. K. Foong first realised the need for a density channel, and Yann Dubois first proposed to normalise the data channel by the density channel. The ConvGNP is a result of a collaboration with James Requeima and Stratis Markou. James Requeima and Stratis Markou first proposed to model the covariance with a linear kernel. The FullConvGNP is a result of a collaboration with James Requeima, and was later checked by Jonathan Gordon and Andrew Y. K. Foong. The idea of AR CNPs was originally proposed by Richard E. Turner and then tried by Jonathan Gordon and Andrew Y. K. Foong, but they could not get it to work well. All work was supervised by Richard E. Turner.

## 5.1 Introduction

In this chapter, we will construct a new family of neural process models called *convolutional neural processes* (ConvNPs; Section 5.3). ConvNPs improve data efficiency of neural processes by building in *translation equivariance* (Definition 2.4). After constructing the family of ConvNPs, we will build on this family by proposing two additional new families of neural processes: *Gaussian neural processes* (GNPs; Section 5.5) and *autoregressive conditional neural processes* (AR CNPs; Section 5.6). As we will see, GNPs and AR CNPs address the inability of the original class of conditional neural processes (CNPs; Garnelo et al., 2018a) to produce coherent samples. These classes, however, do so by giving up something else. GNPs directly parametrise dependencies between target outputs, but are limited to Gaussian predictions. AR CNPs, on the other hand, propose to deploy CNPs in an autoregressive fashion, but are no longer consistent meta-learning algorithms. Whereas we will focus on GNPs and AR CNPs that are ConvNPs, the classes of GNPs and AR CNPs are more general. The families of ConvNPs, GNPs, and AR CNPs are the three contributions of this chapter.

All models will be derived by applying an appropriate *representation theorem* from Chapter 4 to the prediction map $\pi_\theta\colon \mathcal{D} \to \mathcal{Q}$. Recall that $\mathcal{Q}$ is a collection of stochastic processes called the *variational family*. In the context of neural processes, representations theorems generally characterise functions on data sets $\mathcal{D}$. These theorems can be used to find practical implementations of such functions (Section 4.1). Applying representation theorems to the prediction map forms a general recipe to construct neural processes.

In Chapter 4, we studied functions on $[\mathcal{D}]$ rather than on $\mathcal{D}$. Recall that $[\mathcal{D}]$ is the collection of data sets of all sizes that do not depend on an ordering of the data points. To simplify the notation, we will throughout denote $\mathcal{D}$ instead of $[\mathcal{D}]$, and always assume that functions on $\mathcal{D}$ are permutation invariant (Proposition 4.3).

**A word of caution.** This chapter will feature many existing and new names of neural processes and classes of neural processes. Please see our word of caution on naming conventions in Section 2.2.

Before constructing any new models, we first derive the original Conditional Neural Process (CNP; Garnelo et al., 2018a). We will analyse the strengths and shortcomings of the CNP, propose an improvement that alleviates one shortcoming, and repeat this once more.

## 5.2 Conditional Neural Processes

The class of conditional neural processes (CNPs) was formally defined in Definition 3.22. CNPs choose the variational family $\mathcal{Q}$ to be the collection of Gaussian processes that *do not* model dependencies between target outputs. With this choice, a CNP is identified by its *mean map* $m$ (Definition 3.18) and *variance map* $v$ (Definition 3.20). Recall that the mean map $m$ is the map from a data set to the mean function of the predictive stochastic process:

$$m\colon \mathcal{D} \to \mathcal{Y}^{\mathcal{X}}, \quad m(D) = x \mapsto \mathbb{E}_{\pi(D)}[f(x)]. \tag{5.1}$$

Also recall that the variance map $v$ is the map from a data set to the variance function of the predictive stochastic process:

$$v \colon \mathcal{D} \to [0, \infty)^{\mathcal{X}}, \quad v(D) = x \mapsto \mathrm{var}_{\pi(D)}[f(x)]. \tag{5.2}$$

See Section 3.4 for a more detailed description of the class of CNPs.

The Conditional Neural Process is constructed by applying the deep sets representation theorem (Theorem 4.5) to the mean map $m$ and variance map $v$. In Theorem 4.5, the encoder does not depend on the function that the theorem is applied to, so we can use the same encoder for the mean map and variance map. This gives the following parametrisation:

**Model 5.1** (Conditional Neural Process; CNP; Garnelo et al., 2018a). *The* Conditional Neural Process *(CNP) parametrises*

$$m_\theta(D) = x \mapsto [\mathsf{dec}_\theta(\mathbf{z}, x)]_1, \qquad \mathbf{z} \in \mathbb{R}^K,$$
$$v_\theta(D) = x \mapsto [\mathsf{dec}_\theta(\mathbf{z}, x)]_2, \qquad \mathbf{z} = \underbrace{\sum_{(x,y) \in D} \phi_\theta(x, y)}_{\mathsf{enc}_\theta(D)},$$

*where* $\phi_\theta \colon \mathcal{X} \times \mathcal{Y} \to \mathbb{R}^K$ *and* $\mathsf{dec}_\theta \colon \mathbb{R}^K \times \mathcal{X} \to \mathbb{R} \times [0, \infty)$ *are multi-layer perceptrons.*

As discussed in Section 2.4, parametrising a prediction map presents two challenges: the data set $D$ is of variable dimensionality, and the neural process should not depend on the order of the data points in $D$. The CNP addresses these challenges by decomposing the architecture into an *encoder* $\mathsf{enc}_\theta$ and *decoder* $\mathsf{dec}_\theta$, an *encoder–decoder architecture*. By summing over an encoding $\phi_\theta(x, y)$ of every data point $(x, y) \in D$, the encoder of the CNP effectively handles data sets of varying sizes. Morever, since addition is commutative, the encoding $\mathsf{enc}_\theta(D)$ does not depend on the order of the data points in $D$. After the encoder produces the encoding $\mathbf{z}$, the decoder takes in $\mathbf{z}$ and a target input $x$ and produces the marginal mean $[\mathsf{dec}_\theta(\mathbf{z}, x)]_1$ and marginal variance $[\mathsf{dec}_\theta(\mathbf{z}, x)]_2$ of the CNP's prediction at $x$. The decoder is the more heavyweight component of the CNP and responsible for the majority of the representational capacity of the model.

## 5.3 Convolutional Conditional Neural Processes

In Figure 2.1 in Section 2.5, we saw that the Conditional Neural Process breaks down when it is presented data outside of the *training range*, the part of the input space where the model has seen data during training. In a sense, that the CNP malfunctions outside of the training range is reasonable, because it has not seen any indicators of the behaviour of the data outside the training range. For all the model knows, the data might suddenly start exhibiting wildly erratic behaviour when it crosses the edge of the training range. We have not baked any assumptions into the CNP that exclude such a possibility, however unlikely it may sound.

Our goal is to improve this behaviour of the CNP, to tell the model that the data outside of the training range behaves just like inside the training range. To this end, we will specify that the ground-truth stochastic process $f$ behaves similarly on all parts of the input space. In particular, the appropriate and natural assumption is that $f$ be a *stationary* process. Recall that a process is stationary if its distribution is unaffected by shifts: $f \stackrel{\mathrm{d}}{=} \mathsf{T}_\tau f$ for all shifts $\tau \in \mathcal{X}$. It turns out that stationarity of the prior has a simple characterisation in terms of the posterior prediction map.

**Proposition 5.2.** *The ground-truth stochastic process $f$ is stationary if and only if the posterior prediction map $\pi_f$ is translation equivariant (Definition 2.4).*

*Proof.* See Appendix C.1. □

To build in the assumption that the data behaves similarly on all parts of the input space, Proposition 5.2 reveals that the appropriate assumption is that the prediction map $\pi_\theta \colon \mathcal{D} \to \mathcal{Q}$ be *translation equivariant* (Definition 2.4). Our proposed improvement of the CNP is therefore a translation-equivariant parametrisation of $\pi_\theta$. Translation equivariance carries over to the mean map $m$ and variance map $v$, as the following proposition says.

**Proposition 5.3.** *If a prediction map $\pi$ is translation equivariant, then the mean map $m_\pi$ and variance map $v_\pi$ are also translation equivariant. Conversely, suppose that $\pi$ a conditional neural process in the sense of Definition 3.22. If $m_\pi$ is TE and $v_\pi$ is TE, then $\pi$ is translation equivariant.*

$$\pi \text{ is TE} \quad \underset{\pi \text{ is a CNP}}{\overset{\Longrightarrow}{\Longleftarrow}} \quad m_\pi \text{ is TE and } v_\pi \text{ is TE}. \tag{5.3}$$

*Proof.* See Appendix C.1. □

The CNP used deep sets (Theorem 4.5), a general characterisation of functions on data sets $\mathcal{D}$, to parametrise the mean map $m$ and variance map $v$. In our case, $m$ and $v$ are now translation equivariant, which means that we instead require a general characterisation of functions on data sets $\mathcal{D}$ which are translation equivariant. This is exactly what convolutional deep sets offer (Theorem 4.8)! Using convolutional deep sets to parametrise the mean map $m$ and variance map $v$ constructs a new neural process that we call the *Convolutional Conditional Neural Process* (ConvCNP). In Theorem 4.8, we choose the continuous positive-definite function to be a simple Gaussian kernel.

**Model 5.4** (Convolutional Conditional Neural Process; ConvCNP). *The Convolutional Conditional Neural Process (ConvCNP) parametrises*

$$\mathbf{z}\colon \mathcal{X} \to \mathbb{R}^2,$$

$$m_\theta(D) = [\text{dec}_\theta(\mathbf{z})]_1, \qquad \mathbf{z}(\,\cdot\,) = \underbrace{\sum_{(x,y)\in D} \begin{bmatrix} y \\ 1 \end{bmatrix} e^{-\frac{1}{2\ell^2}(\,\cdot\,-x)^2}}_{\text{enc}_\ell(D)} \quad \begin{array}{l} \text{(data channel)} \\ \text{(density channel)} \end{array}$$

$$v_\theta(D) = [\text{dec}_\theta(\mathbf{z})]_2,$$

*where $\ell > 0$ is a length scale and $\text{dec}_\theta\colon C(\mathcal{X}, \mathbb{R}^2) \to C(\mathcal{X}, \mathbb{R} \times [0, \infty))$ is a translation-equivariant map implemented by a convolutional neural network.*

Although the ConvCNP looks similar to the CNP, there are two important differences. First, whereas in the CNP the encoding is a vector $\mathbf{z} \in \mathbb{R}^K$, in the ConvCNP the encoding is a vector-valued *function* $\mathbf{z}\colon \mathcal{X} \to \mathbb{R}^2$ (Definition 4.6). This function $\mathbf{z}$ is then passed through the decoder $\text{dec}_\theta$, which is a *map between function spaces*. The result $\text{dec}_\theta(\mathbf{z})$ is two new functions, which become the mean and variance function of the prediction. Second, whereas in the CNP the encoder and the decoder are implemented with multi-layer perceptrons (MLPs), the ConvCNP does not require a neural network for the encoder at all and implements the decoder with a *convolutional neural network* (CNN). Compared to MLPs, CNNs offer a massive reduction in parameter count. Therefore, in scenerios where translation equivariance is appropriate, the ConvCNP can be vastly more parameter efficient than the CNP.

In the ConvCNP, the encoding $\mathbf{z}$ is a vector-valued function. We call the functions in this vector different *channels*. The first function $z_1$ is called the *data channel* and the second function $z_2$ is called the *density channel*:

$$\text{data channel:} \quad z_1(\,\cdot\,) = \sum_{(x,y)\in D} y \cdot e^{-\frac{1}{2\ell^2}(\,\cdot\,-x)^2} \tag{5.4}$$

$$\text{density channel:} \quad z_2(\,\cdot\,) = \sum_{(x,y)\in D} 1 \cdot e^{-\frac{1}{2\ell^2}(\,\cdot\,-x)^2} \tag{5.5}$$

Intuitively, the data channel constructs a function by placing Gaussian bumps at the location of the data points. This communicates the values of the observations to the model. However, the data channel can be ambiguous: consider observing no observations $D = \varnothing$ and observing a single observation $D = \{(1, 0)\}$ with $y$-value equal to zero. In both cases, the data channel $z_1 = 0$ is zero! That is, the data channel is unable to distinguish between observing no observation or observing $y = 0$. This is exactly where the density channel comes in. The density channel places a unity-weight Gaussian bump at the location of every observation. Whenever the data channel is zero, the presence of a bump in the density channel reveals whether there was no observation or whether $y = 0$, thereby breaking the ambiguity.

To implement the ConvCNP, unlike the CNP, one additional approximation is required. Namely, in Model 5.4, $\mathbf{z}$ is a whole function and $\text{dec}_\theta$ a mapping between function spaces, and it is not clear how these objects can be implemented on a computer. We propose an approximation by discretisation:

**Procedure 5.5** (Discretisation). *Let* $\text{map} \colon C(\mathcal{X}, \mathbb{R}^{N_1}) \to C(\mathcal{X}, \mathbb{R}^{N_2})$ *be a translation-equivariant map between function spaces, like the decoder of the ConvCNP. Consider an input* $\mathbf{z} \colon \mathcal{X} \to \mathbb{R}^{N_1}$ *for* $\text{map}$. *Then approximate the output* $\text{map}(\mathbf{z}) \colon \mathcal{X} \to \mathbb{R}^{N_2}$ *with the following four steps:*

$$\mathbf{z}(\cdot) \overset{\text{\textcircled{\raisebox{-0.9pt}{2}}}}{\mapsto} \begin{bmatrix} \mathbf{z}(u_1) \\ \vdots \\ \mathbf{z}(u_K) \end{bmatrix} \overset{\text{\textcircled{\raisebox{-0.9pt}{3}} CNN}}{\mapsto} \begin{bmatrix} \mathbf{z}^{(\text{out})}(u_1) \\ \vdots \\ \mathbf{z}^{(\text{out})}(u_K) \end{bmatrix} \overset{\text{\textcircled{\raisebox{-0.9pt}{4}}}}{\mapsto} \sum_{k=1}^{K} \mathbf{z}^{(\text{out})}(u_k) e^{-\frac{1}{2\ell^2}(\cdot - u_k)^2} \approx \text{map}(\mathbf{z})(\cdot)$$

**❶** *Choose a uniformly spaced grid* $\mathbf{u} = (u_1, \ldots, u_K) \in \mathcal{X}^K$ *spanning a region of interest.*

**❷** *Discretise* $\mathbf{z}(\cdot)$ *on* $\mathbf{u}$.

**❸** *Run this discretisation of* $\mathbf{z}(\cdot)$ *through a CNN. The CNN does the job of* $\text{map}$.

**❹** *Interpolate the output of the CNN to any value in* $\mathcal{X}$ *using a Gaussian kernel.*

*Call the grid in* **❶** *the* internal discretisation *or more simply the* discretisation.

That $\text{dec}_\theta$ can be approximated in this way is theoretically motivated by Theorem 3.1 from Yarotsky (2022), although their setting and discretisation method are slightly different.

The discretisation is the most important parameter of Procedure 5.5, and must be chosen right. In practice, we choose the discretisation to be a uniformly spaced grid spanning at least all context and target inputs. Intuitively, the grid spacing determines the smallest length scale that the model can capture. If the grid too coarse, then the model might underfit the data. On the other hand, if the grid is too fine, then the discretisation may consume a lot of memory and unnecessarily waste compute. Our recommendation is to choose the interpoint spacing of the grid half or one-fourth of the smallest length scale present in the data.

In the ConvCNP, note that the division of the data channel by the density channel constructs the *Nadaraya–Watson estimator* $\hat{m}(\cdot)$ (Nadaraya, 1964; Watson, 1964), a nonparametric estimator of the mean function:

$$\hat{m}(\,\cdot\,) := \frac{\sum_{(x,y)\in D} y \cdot e^{-\frac{1}{2\ell^2}(\cdot - x)^2}}{\sum_{(x,y)\in D} 1 \cdot e^{-\frac{1}{2\ell^2}(\cdot - x)^2}} = \frac{z_1(\,\cdot\,)}{z_2(\,\cdot\,)}. \tag{5.6}$$

To help to model fit quicker and better, the architecture can be tweaked by manually dividing the data channel by the density channel before feeding the channels to the decoder:

$$\mathbf{z}(\,\cdot\,) = \begin{bmatrix} z_1 \\ z_2 \end{bmatrix} \mapsto \begin{bmatrix} z_1/z_2 \\ z_2 \end{bmatrix} \mapsto \cdots \quad \begin{array}{l} \text{(data channel)} \\ \text{(density channel)} \end{array} \tag{5.7}$$

With this change, after the division, the data channel is already a reasonable estimator of the mean function.

CNPs bear similarities to methods from the literature of point cloud modelling. The CNP can be compared to PointNet (Qi et al., 2017); and the ConvCNP, a translation-equivariant extension of the CNP, to PointConv (Wu et al., 2019), a translation-equivariant extension of PointNet. A key difference between the ConvCNP and PointConv is that signals in the ConvCNP live on a predefined grid (the discretisation in Procedure 5.5) and convolutions are standard convolutions, whereas signals in PointConv live on the positions of (a subset of) the points and convolutions are performed using weights parametrised with an MLP. In this sense, PointConv is similar to SchNet (Schütt et al., 2017), which uses continuous-filter convolutional layers, or more generally equivariant message passing (Satorras et al., 2021).

## 5.4 Translation Equivariance and Generalisation

In Section 2.2, we proposed building translation equivariance into the prediction map as a way of improving generalisation performance. In the previous section, translation equivariance was derived from stationarity of the ground-truth stochastic process $f$. This makes a strong case for translation equivariance. In this section, we reinforce this case by showing that translation equivariance can indeed improve generalisation performance.

The key to understanding how translation equivariance and generalisation performance connect is a notion called the *receptive field*. The notion of a receptive field originates from the field of biology (Sherrington, 1906), but nowadays it also used to describe a property of convolutional neural networks (Luo et al., 2016). Intuitively, the receptive field of a convolutional neural network is the following. Consider a CNN mapping an image $\mathbf{X}$ to another image $\mathbf{Y}$. If a particular input pixel $X_{i,j}$ is perturbed, then, due to the finite size of the convolutional filters, only a certain region $R$ of output pixels is affected. This region

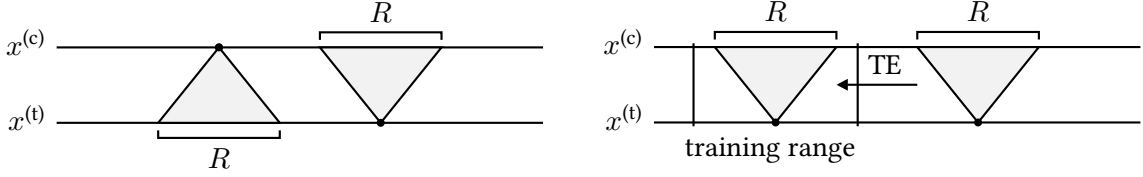

(a) For a model with receptive field $R > 0$ (Def. 5.6), a context point at $x^{(c)}$ influences predictions at target inputs only limitedly far away. Conversely, a prediction at a target input $x^{(t)}$ is influenced by context points only limitedly far away.

(b) If a model is translation equivariant (TE; Def. 2.4), then all context points and targets inputs can simultaneously shifted left or right without changing the output of the model. Intuitively, this means that triangles in the figures can just be "shifted left or right".

Figure 5.1: Translation equivariance in combination with a limited receptive field (see (a)) can help generalisation performance. Consider a translation-equivariant (TE) model which performs well within a *training range* (see (b)). Consider a prediction for a target input outside the training range (right triangle in (b)). If the model has receptive field $R > 0$ and the training range range is bigger than $R$, then TE can be used to "shift that prediction back into the training range" (see (b)). Since the model performs well within the training range, the model also performs well for the target input outside the training range.

$R$ is the receptive field. Usually, however, the receptive field is defined just the other way around. Due to the finite size of the convolutional filters, every output pixel $Y_{i,j}$ is only affected by a certain region $R$ of input pixels, and this region $R$ is called the receptive field. By running the network backwards and noting that this does not influence what affects what, one can see that these definitions are equivalent. Rather than specifying a region, the receptive field $R$ is sometimes just a scalar, referring to the width of the interval $R$ for one-dimensional CNNs, the side length of the square $R$ for two-dimensional CNNs, the side length of the cube $R$ for three-dimensional CNNs, *et cetera*.

We now define the receptive field more carefully. For $\mathbf{x}, \mathbf{y} \in I$, let $\mathbf{y}|_{\mathbf{x},R}$ be the subvector of $\mathbf{y}$ with elements at most distance $R$ from $\mathbf{x}$. Similarly, for $D \in \mathcal{D}$ and $\mathbf{x} \in \mathcal{X}^N$, let $D|_{\mathbf{x},R}$ denote the subcollection of input–output pairs with inputs at most distance $R$ from $\mathbf{x}$.

**Definition 5.6** (Receptive field)**.** *Let $R > 0$.*

- *A $\mathcal{Y}$-valued stochastic process $f$ on $\mathcal{X}$ has* receptive field size $R$ *if, for all $\mathbf{x}, \mathbf{y} \in I$,*

$$f(\mathbf{x}) \,|\, f(\mathbf{y}) \stackrel{\mathrm{d}}{=} f(\mathbf{x}) \,|\, f(\mathbf{y}|_{\mathbf{x},\frac{1}{2}R}). \tag{5.8}$$

- *A prediction map $\pi \colon \mathcal{D} \to \mathcal{P}$ has* receptive field size $R$ *if, for all $D \in \mathcal{D}$ and $\mathbf{x} \in I$,*

$$P_{\mathbf{x}}\pi(D) = P_{\mathbf{x}}\pi(D|_{\mathbf{x},\frac{1}{2}R}). \tag{5.9}$$

Figure 5.1a illustrates how the receptive field of a neural process influences predictions the predictions of the model. If the ground-truth stochastic process $f$ has receptive field size $R > 0$, then it is immediate from the definitions that the posterior prediction map $\pi_f$ has receptive field $R$; and for any $D \in \mathcal{D}$, $\pi_f(D)$ has receptive field $R$. For example, if $f$ is a stationary Gaussian process with a compactly supported kernel $k \colon \mathcal{X} \to \mathbb{R}$, then $f$ has finite receptive field size. More precisely, if $k(\tau) = 0$ for $|\tau| \geq \frac{1}{2}R$, then $f$ has receptive field size $R$.

The main insight of this section is that a limited receptive field in combination with translation equivariance can help generalisation performance. This insight is illustrated and explained in Figure 5.1 and formalised in the following theorem.

**Theorem 5.7.** *Let $\pi_1, \pi_2 \colon \mathcal{D} \to \mathcal{P}$ be translation-equivariant prediction maps with receptive field $R > 0$. Assume that, for all $D \in \mathcal{D}$, $\pi_1(D)$ and $\pi_2(D)$ also have receptive field $R > 0$. Let $\varepsilon > 0$ and fix $N \in \mathbb{N}$. Assume that, for all $\mathbf{x} \in \bigcup_{n=1}^{N}[0, 2R]^n$ and $D \in \mathcal{D} \cap \bigcup_{n=0}^{\infty}([0, 2R] \times \mathbb{R})^n$,*

$$\mathrm{KL}(P_{\mathbf{x}}\pi_1(D), P_{\mathbf{x}}\pi_2(D)) \leq \varepsilon. \tag{5.10}$$

*Then, for all $M > 0$, $\mathbf{x} \in \bigcup_{n=1}^{N}[0, M]^n$, and $D \in \mathcal{D} \cap \bigcup_{n=0}^{\infty}([0, M] \times \mathbb{R})^n$,*

$$\mathrm{KL}(P_{\mathbf{x}}\pi_1(D), P_{\mathbf{x}}\pi_2(D)) \leq \lceil 2M/R \rceil \varepsilon. \tag{5.11}$$

*Proof.* See Appendix C.2. □

If a prediction map is well approximated on an interval of at least twice the receptive field size, then Theorem 5.7 says that it is well approximated on any bigger interval, where the approximation error is proportional to the size of the bigger interval. Since we can compute the receptive field of a CNN architecture, Theorem 5.7 offers practical guidance: during training, it suffices to sample the context sets from an interval at least twice the receptive field size.

In this section, our definition of the receptive field (Definition 5.6) has been rather strict: an output depends on all inputs within the receptive field and on no inputs outside the receptive field. A more nuanced perspective is offered by Luo et al. (2016), who argue that not all inputs within the receptive field contribute equally, investigating the concept of an *effective receptive field*. Therefore, a more realistic definition might be that outputs depend more on inputs nearby and less on inputs far away. A formal definition could prescribe some kind of decay. We leave an investigation of this for future work.

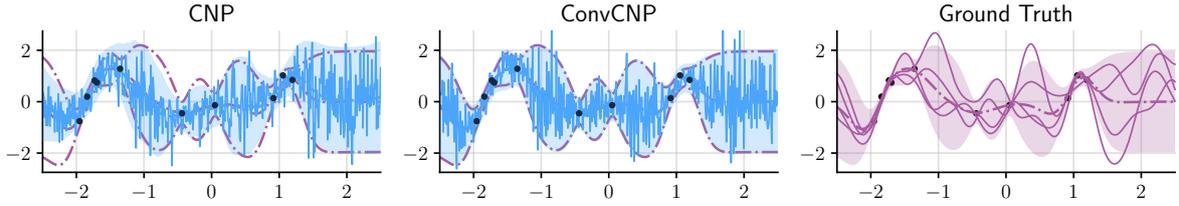

Figure 5.2: Comparison of samples from a trained CNP and ConvCNP to samples from the ground truth. Shows predictions by the models in dashed blue and predictions by the ground truth in dot-dashed purple. Filled regions are central 95%-credible regions. The CNP and ConvCNP are taken from the experiment in Section 6.2.

## 5.5 Gaussian Neural Processes

The ConvCNP (Model 5.4) builds translation equivariance into the CNP, improving generalisation performance whenever the ground-truth stochastic process $f$ is stationary. In this section, we address another shortcoming: the CNP's inability to model dependencies between different target outputs. This deficiency is best understood visually. Figure 5.2 shows predictions and samples from a well-trained CNP and ConvCNP. Due the lack of dependencies in the prediction, the samples look incredibly noisy and not at all like samples from the ground truth.

We address the inability to produce coherent samples simply by changing the variational family $\mathcal{Q}$. We change $\mathcal{Q}$ from the collection of Gaussian processes that *do not* model dependencies between target outputs to the collection of Gaussian processes that *do* model dependencies between target outputs. We call this class *Gaussian neural processes* (GNPs). The class of GNPs was formally defined in Definition 3.23. With this new choice of variational family $\mathcal{Q}$, a Gaussian neural process $\pi_\theta \colon \mathcal{D} \to \mathcal{Q}$ is identified by its mean map $m$ and its *kernel map* $k$ (Definition 3.19). Recall that the kernel map $k$ is the map from a data set to the covariance function of the predictive stochastic process:

$$k \colon \mathcal{D} \to \mathbb{R}^{\mathcal{X} \times \mathcal{X}}, \qquad k(D) = (x, y) \mapsto \mathrm{cov}_{\pi(D)}(f(x), f(y)). \tag{5.12}$$

See Section 3.4 for a more detailed description of the class of GNPs. To construct a Gaussian neural process, the main challenge is to come up with a parametrisation of the kernel map. Our first two constructions are based on the *eigenmap* of a kernel map.

**Definition 5.8** (Eigenmap). *For a kernel map $k$ and Hilbert space $(\mathbb{H}, \langle \,\cdot\, , \,\cdot\, \rangle_{\mathbb{H}})$, call a function $\psi$ an* eigenmap *on $\mathbb{H}$ if it is a function $\psi \colon \mathcal{D} \to \mathbb{H}^{\mathcal{X}}$ such that*

$$k(D)(x, y) = \langle \psi(D)(x), \psi(D)(y) \rangle_{\mathbb{H}} \quad \text{for all } D \in \mathcal{D} \text{ and } x, y \in \mathcal{X}. \tag{5.13}$$

In our first member of the GNP class, called the *Gaussian Neural Process* (GNP), we parametrise the kernel map $k$ by parametrising the eigenmap $\psi$ with a deep set (Theorem 4.5). In particular, we assume that the kernel map $k$ has an eigenmap $\psi$ on some $R$-dimensional Euclidean space $\mathbb{R}^R$ (mnemonic: $R$ for rank). This gives the following parametrisation:

**Model 5.9** (Gaussian Neural Process; GNP). *The* Gaussian Neural Process *(GNP) parametrises*

$$m_\theta(D) = \quad x \mapsto [\mathrm{dec}_\theta(\mathbf{z}, x)]_1,$$
$$k_\theta(D) = (x, y) \mapsto \langle [\mathrm{dec}_\theta(\mathbf{z}, x)]_{2:1+R}, [\mathrm{dec}_\theta(\mathbf{z}, y)]_{2:1+R} \rangle,$$

$$\mathbf{z} \in \mathbb{R}^K,$$
$$\mathbf{z} = \underbrace{\sum_{(x,y) \in D} \phi_\theta(x, y)}_{\mathrm{enc}_\theta(D)},$$

*where* $\phi_\theta \colon \mathcal{X} \times \mathcal{Y} \to \mathbb{R}^K$ *and* $\mathrm{dec}_\theta \colon \mathbb{R}^K \times \mathcal{X} \to \mathbb{R}^{1+R}$ *are multi-layer perceptrons.*

The GNP is a generalisation of the CNP that models dependencies between target outputs. Indeed, comparing Models 5.1 and 5.9, the two parametrisations differ only in the parametrisation of the uncertainty: the CNP parametrises the variance map $v$ with an MLP, whereas the GNP parametrises the eigenmap $\psi$ with an MLP.

The construction of the GNP comes with two technical caveats. First, recall the Hilbert space $(\ell^2, \langle \cdot, \cdot \rangle_{\ell^2})$ given by $\ell^2 = \{(x_n)_{n \geq 1} \subseteq \mathbb{R} : \sum_{n=1}^\infty x_n^2 < \infty\}$ with $\langle x, y \rangle_{\ell^2} = \sum_{n=1}^\infty x_n y_n$. If $\mathcal{X}$ is compact, then, by Mercer's theorem (Theorem 3.3.1; Adler, 1981), every kernel map $k$ has an eigenmap on $\ell^2$. To justify the assumption that $k$ has an eigenmap on some $R$-dimensional Euclidean space $\mathbb{R}^R$, we can truncate the summation of the $\ell^2$–inner product in (5.13). This truncation, however, will need to conform to the topology on the collection of kernel maps. In Section 3.5, we argued that a suitable topology is the one induced by the supremum norm. Therefore, to make the assumption that $k$ has an eigenmap on some $\mathbb{R}^R$ precise, we will need to argue that the $\ell^2$–inner product in (5.13) can be truncated uniformly over $D \in \mathcal{D}$ and $x, y \in \mathcal{X}$. We leave this verification for future work. Second, to approximate the eigenmap $\psi$ with an MLP, we need that it is continuous. We just argued that every kernel map has an eigenmap. However, we did not argue that this eigenmap is continuous. We also leave this verification for future work.

The second model that we construct in this section is a generalisation of the ConvCNP that models dependencies between target outputs. This model is called the *Convolutional Gaussian Neural Process* (ConvGNP). Like for the ConvCNP, we again assume that the ground-truth stochastic process $f$ is stationary, so the posterior prediction map $\pi_f$ is translation equivariant (Proposition 5.2). For GNPs, translation equivariance of the prediction map carries over to the kernel map $k$ in a slightly different way.

**Definition 5.10** (Diagonal translation equivariance of kernel map; DTE). *Call a kernel map $k\colon \mathcal{D} \to \mathbb{R}^{\mathcal{X} \times \mathcal{X}}$ diagonally translation equivariant (DTE) if*

$$k \circ \mathsf{T}_\tau = \mathsf{T}_{(\tau,\tau)} \circ k \quad \text{for all } \tau \in \mathcal{X}. \tag{5.14}$$

**Proposition 5.11.** *If a prediction map $\pi$ is translation equivariant, then the mean map $m_\pi$ is translation equivariant and the kernel map $k_\pi$ is diagonally translation equivariant. Conversely, suppose that $\pi$ is a Gaussian neural process in the sense of Definition 3.23. If $m_\pi$ is TE and $k_\pi$ is DTE, then $\pi$ is translation equivariant. In formulas,*

$$\pi \text{ is TE} \quad \underset{\pi \text{ is a GNP}}{\overset{\Longrightarrow}{\Longleftarrow}} \quad m_\pi \text{ is TE and } k_\pi \text{ is DTE.} \tag{5.15}$$

*Proof.* See Appendix C.3. □

Like for the GNP, assume that the kernel map $k$ has an eigenmap $\psi$ on some $R$-dimensional Euclidean space $\mathbb{R}^R$. The construction of the ConvGNP follows the same progression as Section 5.3, where we replaced the deep set parametrisation with convolutional deep sets. Hence, rather than parametrising the eigenmap $\psi$ with deep sets, we now parametrise $\psi$ with convolutional deep sets (Theorem 4.8):

**Model 5.12** (Convolutional Gaussian Neural Process; ConvGNP). *The Convolutional Gaussian Neural Process (ConvGNP) parametrises*

$$\mathbf{z}\colon \mathcal{X} \to \mathbb{R}^2,$$

$$m_\theta(D) = [\mathsf{dec}_\theta(\mathbf{z})]_1,$$
$$k_\theta(D) = \langle [\mathsf{dec}_\theta(\mathbf{z})]_{2:1+R}, [\mathsf{dec}_\theta(\mathbf{z})]_{2:1+R} \rangle,$$
$$\mathbf{z}(\,\cdot\,) = \underbrace{\sum_{(x,y) \in D} \begin{bmatrix} y \\ 1 \end{bmatrix} e^{-\frac{1}{2\ell^2}(\,\cdot\,-x)^2}}_{\mathsf{enc}_\ell(D)} \quad \begin{array}{l} \text{(data ch.)} \\ \text{(density ch.)} \end{array}$$

*where $\ell > 0$ is a length scale and $\mathsf{dec}_\theta \colon C(\mathcal{X}, \mathbb{R}^2) \to C(\mathcal{X}, \mathbb{R}^{1+R})$ is a translation-equivariant map implemented by a convolutional neural network.*

The caveats for the GNP also apply to the ConvGNP. The ConvGNP, however, comes with an important additional caveat. By Proposition 5.11, we require a general parametrisation of a DTE kernel map. To this end, the ConvGNP parametrised a TE eigenmap. Indeed, if the eigenmap is TE, then the kernel map is DTE. However, if a kernel map is DTE, then it is not necessarily true that the eigenmap is TE! This means that restricting the eigenmap to be TE can potentially come at the loss of representational capacity. We leave a theoretical investigation of this issue for future work, but we mention one significant practical consequence. If $D = \varnothing$, then $\psi(\varnothing)$ will be a constant function, so $k_\theta(\varnothing)$ will be a constant function too. Since conditioning on no data ($D = \varnothing$) recovers the prior, this means that the ConvGNP is unable to model the prior! This substantial flaw of the ConvGNP underlines the importance fully basing neural process architectures on representation theorems, which guarantee that such representational capacity issues cannot occur. For the ConvGNP, we momentarily deviated from this recipe by assuming a TE model for the eigenmap, and we immediately ran into problems.

The GNP and ConvGNP parametrise the kernel map $k$ by parametrising the eigenmap $\psi$. For the third and last model that we construct in the section, we directly parametrise the kernel map $k$ without going through the eigenmap $\psi$. In Section 4.5, we investigated the notation of diagonal translation equivariance. In particular, we developed Theorem 4.9, an extension of convolutional deep sets (Theorem 4.8) to functions on data sets $\mathcal{D}$ which are translation equivariant in the sense of Definition 5.10. The *Fully Convolutional Gaussian Neural Process* (FullConvGNP) uses Theorem 4.8 to parametrise the mean map and Theorem 4.9 to parametrise the kernel map. As suggested just below Theorem 4.9, for Theorem 4.9, we choose $c$ to be a Gaussian which decays with the distance from the diagonal.

**Model 5.13** (Fully Convolutional Gaussian Neural Process; FullConvGNP). *The* Fully Convolutional Gaussian Neural Process *(FullConvGNP) parametrises*

$$\mathbf{z}^{(\mathrm{m})} \colon \mathcal{X} \to \mathbb{R}^2,$$

$$m_\theta(D) = \mathsf{dec}_\theta^{(\mathrm{m})}(\mathbf{z}^{(\mathrm{m})}), \quad \mathbf{z}^{(\mathrm{m})}(\,\cdot\,) = \underbrace{\sum_{(x,y)\in D} \begin{bmatrix} y \\ 1 \end{bmatrix} e^{-\frac{1}{2\ell^2}(\,\cdot\,-x)^2}}_{\mathsf{enc}_\ell^{(\mathrm{m})}(D)}, \quad \begin{array}{l} \text{(data channel)} \\ \text{(density channel)} \end{array}$$

$$\mathbf{z}^{(\mathrm{k})} \colon \mathcal{X} \times \mathcal{X} \to \mathbb{R}^3,$$

$$k_\theta(D) = \mathsf{dec}_\theta^{(\mathrm{k})}(\mathbf{z}^{(\mathrm{k})}), \quad \mathbf{z}^{(\mathrm{k})}(\,\cdot\,) = \underbrace{\begin{bmatrix} \sum_{(x,y)\in D} \begin{bmatrix} y \\ 1 \end{bmatrix} e^{-\frac{1}{2\ell^2}\|\,\cdot\,-(x,x)\|_2^2} \\ e^{-\frac{1}{2\ell^2}\langle\,\cdot\,,(1,-1)\rangle^2} \end{bmatrix}}_{\mathsf{enc}_\ell^{(\mathrm{k})}(D)}, \quad \begin{array}{l} \text{(data channel)} \\ \text{(density channel)} \\ \text{(source channel)} \end{array}$$

*where $\ell > 0$ are length scales and*

$$\mathsf{dec}_\theta^{(\mathrm{m})} \colon C(\mathcal{X}, \mathbb{R}^2) \to C(\mathcal{X}, \mathbb{R}), \tag{5.16}$$

$$\mathsf{dec}_\theta^{(\mathrm{k})} \colon C(\mathcal{X} \times \mathcal{X}, \mathbb{R}^3) \to C^{\mathrm{p.s.d}}(\mathcal{X} \times \mathcal{X}, \mathbb{R}) \tag{5.17}$$

*are translation-equivariant maps implemented by convolutional neural networks.*

The application of Theorem 4.9 yields an architecture which is more complicated than the architectures we have seen so far. To begin with, in Model 5.13, the encoding for the kernel map is a function on $\mathcal{X} \times \mathcal{X}$ rather than on $\mathcal{X}$. For a fixed vector of target inputs $\mathbf{x}$, the kernel map generates *kernel matrices*. This encoding on $\mathcal{X} \times \mathcal{X}$, which you can intuit as an image, should be interpreted as the foundation for these kernel matrices. Whereas the data previously placed bumps at the data point locations, for the encoding on $\mathcal{X} \times \mathcal{X}$, the data points also place bumps, but now *on the diagonal of this image*. Moreover, in addition to the data channel and density channel, the encoding now involves a third channel called the *source channel*. Intuitively, the source channel represents uncorrelated noise, forming the basis for a correlated response. For example, to sample from a Gaussian $\mathbf{x} \sim \mathcal{N}(\boldsymbol{\mu}, \boldsymbol{\Sigma})$, one typically sets $\mathbf{x} = \mathbf{L}\boldsymbol{\varepsilon} + \boldsymbol{\mu}$ where $\mathbf{L} = \mathrm{chol}(\boldsymbol{\Sigma})$ is the Cholesky decomposition and $\boldsymbol{\varepsilon} \sim \mathcal{N}(\mathbf{0}, \mathbf{I})$ is noise; this procedure is called the *reparametrisation trick* (Kingma et al., 2013). Intuitively, the role of the source channel is similar to the role of the noise in the reparametrisation trick. The source channel plays a crucial role in the generality of the architecture. For the ConvGNP, we found that a translation-equivariant parametrisation of the eigenmap of the kernel map limits the representational capacity of the model. For the FullConvGNP, there are no such concerns, because Theorem 4.9 guarantees that Model 5.13 is general. For example, if $D = \varnothing$ and the source channel were absent, then the encoding would be the zero function, so $k_\theta(\varnothing)$ would be a constant function, just like what happens for the ConvGNP. Therefore, in the case of no context data, the presence of source channel enables the architecture to generate a non-constant prior covariance; in other words, the source channel enables the FullConvGNP to model the prior.

The FullConvGNP is called "Fully Convolutional" because, in addition to convolutions on $\mathcal{X}$, it also includes convolutions on $\mathcal{X} \times \mathcal{X}$. One distinctive property of Model 5.13 is that the architectures for the mean map and kernel map are entirely separate. This gives increased flexibility, because these architectures can be configured separately, presenting increased control over the model's computational requirements. On the other hand, separating the architectures for the mean map and kernel map prevents any parameter sharing, possibly hurting the model's performance.

To implement the parametrisation of the kernel map, we propose the same discretisation approach (Procedure 5.5) that we discussed earlier. There are two additional implementation details. First, for small length scales $\ell$, the off-diagonal entries of the source channel become zero. We may therefore more simply implement the source channel with an identity matrix. Second, (5.17) requires that the output is a positive-definite function. We implement this by applying the matrix transformation $\mathbf{Z} \mapsto \mathbf{Z}\mathbf{Z}^\mathsf{T}$ between steps ❸ and ❹ in Procedure 5.5.

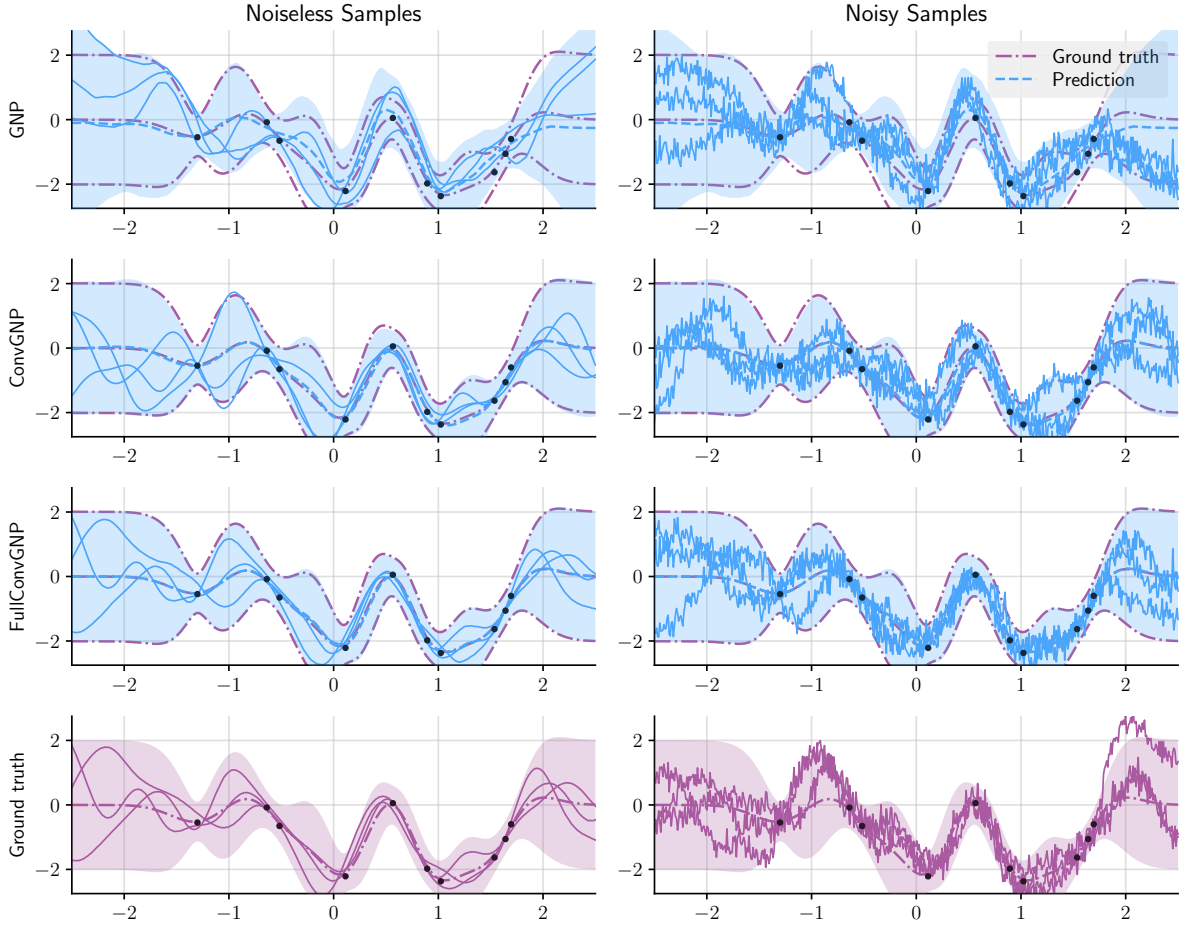

Figure 5.3: Comparison of noiseless (left) and noisy (right) samples from the GNP, ConvGNP, and FullConvGNP to noisy and noiseless sample of the ground truth. Shows predictions by the models in dashed blue and predictions by the ground truth in dot-dashed purple. Filled regions are 95%-credible regions. The GNP, ConvGNP, and FullConvGNP are taken from the experiment in Section 6.2.

The GNP, ConvGNP, and FullConvGNP all model dependencies between target outputs by directly parametrising the covariance between target points. Since the predictions of GNPs are Gaussian, the empirical neural process objective (Definition 2.2) can be evaluated exactly. Therefore, GNPs model dependencies between target outputs and can be trained without approximations.

Figure 5.3 shows noiseless and noisy samples from a well-trained GNP, ConvGNP, and FullConvGNP. Unlike the samples from the CNP and ConvCNP in Figure 5.2, the noiseless samples look smooth and much more like samples from the ground truth. Additionally, note that the amount of noise allocated by the models looks like the amount allocated by the ground truth. This demonstrates the ability of GNPs to separate epistemic and aleatoric uncertainty. See Section 3.4 for a discussion about separation of epistemic and aleatoric uncertainty.

Table 5.1: Overview how all models in this chapter are constructed. For the mean of the predictions, all models parametrise the mean map $m$ (Definition 3.18). However, for the uncertainty of the predictions, models parametrise either the variance map $m$ (Definition 3.20), eigenmap (Definition 5.8), or kernel map (Definition 3.19); and these maps may admit a symmetry. "TE" stands for translation equivariance (Definition 2.4) and "DTE" stands for diagonal translation equivariance (Definition 4.10).

| Model | How uncertainty? | Symmetry | Representation theorem |
| --- | --- | --- | --- |
| CNP (Mod. 5.1) | Variance map $v$ | None | Deep set (Thm 4.5) |
| ConvCNP (Mod. 5.4) | Variance map $v$ | TE | Conv. deep set (Thm 4.5) |
| GNP (Mod. 5.9) | Eigenmap $\psi$ | None | Deep set (Thm 4.5) |
| ConvGNP (Mod. 5.4) | Eigenmap $\psi$ | TE | Conv. deep set (Thm 4.8) |
| FullConvGNP (Mod. 5.13) | Kernel map $k$ | DTE | Conv. deep set (Thm 4.9) |

Table 5.1 provides an overview of all models that we have seen so far.

## 5.6 Autoregressive Conditional Neural Processes

In the previous section, we proposed the class of Gaussian neural processes (GNPs). Gaussian neural processes address the inability of conditional neural process to produce coherent samples by directly parametrising covariances between target outputs. Although GNPs can successfully produce coherent samples and can be trained without approximations, GNPs are limited to Gaussian predictions. Another class of neural processes that can successfully produce coherent samples is the class of *latent-variable neural processes* (LNPs; Garnelo et al., 2018b). LNPs use a latent variable to model dependencies between target outputs and can even produce non-Gaussian predictions. Training LNPs with the neural process objective, however, requires additional approximations.

In this section, we propose a third approach to modelling dependencies between target outputs. Like GNPs, this approach can be trained without additional approximations; and like LNPs, this approach is be able to produce non-Gaussian predictions. Instead, what we give up is *consistency* (Section 2.1): we will no longer produce a prediction that is a stochastic process.

The approach that we propose involves *no modifications* to the model or training procedure. Instead, after training, we propose to *deploy the model in a different way*. Suppose that $\pi \colon \mathcal{D} \to \mathcal{Q}$ is a neural process trained as usual, with the neural process objective. Suppose that we wish to deploy the neural process $\pi$ for a context set $D$ and some target inputs $\mathbf{x}$. According to our developments so far, we would take $P_{\mathbf{x}}^\sigma \pi(D)$ as the prediction for the corresponding target outputs $\mathbf{y}$. The proposal of this section is to instead roll out the CNP in an *autoregressive fashion*, as follows. Recall that $\mathbf{x} \oplus \mathbf{y}$ concatenates $\mathbf{x}$ and $\mathbf{y}$.

**Procedure 5.14** (Autoregressive application of noisy prediction map). *For a noisy prediction map $(\pi, \sigma) \in \mathcal{M}$, context set $(\mathbf{x}^{(c)}, \mathbf{y}^{(c)}) \in \mathcal{D}$, and target inputs $\mathbf{x}^{(t)} \in I$, let $\mathrm{AR}_{\mathbf{x}^{(t)}}^{\sigma}(\pi, D)$ be the distribution defined by the following procedure:*

$$y_1^{(t)} \sim P_{x_1^{(t)}}^{\sigma} \pi(\mathbf{x}^{(c)}, \mathbf{y}^{(c)}), \tag{AR-1}$$

$$y_2^{(t)} \sim P_{x_2^{(t)}}^{\sigma} \pi(\mathbf{x}^{(c)} \oplus x_1^{(t)}, \mathbf{y}^{(c)} \oplus y_1^{(t)}),$$

$$\vdots$$

$$y_N^{(t)} \sim P_{x_N^{(t)}}^{\sigma} \pi(\mathbf{x}^{(c)} \oplus \mathbf{x}_{1:(N-1)}^{(t)}, \mathbf{y}^{(c)} \oplus \mathbf{y}_{1:(N-1)}^{(t)}) \tag{AR-$N$}$$

*where $N$ is the number of target inputs.*

Because earlier samples $y_i$ feed back into later applications of $\pi$, the whole sample $\mathbf{y}$ is correlated. Crucially, this is the case *even if $\pi$ does not model dependencies between target outputs!* This means that we can get correlated samples out of a CNP by rolling out the model in an autoregressive fashion.

Suppressing the target inputs, (AR-1) through (AR-$N$) are inspired by the following application of the product rule:

$$p(y_1, \ldots, y_N \mid D) = p(y_1 \mid D) p(y_2 \mid y_1, D) \cdots p(y_N \mid y_{N-1}, \ldots, y_1, D). \tag{5.18}$$

However, compared to this application of the product rule, there is one very important difference. In (5.18), we could have decomposed the joint $p(y_1, \ldots, y_N \mid D)$ in a different way, leading to a different order of the conditionals; for example,

$$p(y_1, \ldots, y_N \mid D) = p(y_2 \mid D) p(y_{11} \mid y_2, D) p(y_5 \mid y_{11}, y_2, D) \cdots \tag{5.19}$$

Even though the ordering of the conditionals is different, the resulting samples are always samples from the joint $p(y_1, \ldots, y_N \mid D)$. This consistency property tells us that it does not matter in which way we order the conditionals: the resulting samples will always be samples from the same, well-defined joint distribution. Critically, for (AR-1) through (AR-$N$), this consistency property is no longer true! It matters whether we first $y_1 \sim P_{x_1}^{\sigma} \pi(D)$ and then $y_2 \sim P_{x_2}^{\sigma} \pi(D \oplus (x_1, y_1))$; or first $y_2 \sim P_{x_2}^{\sigma} \pi(D)$ and then $y_1 \sim P_{x_1}^{\sigma} \pi(D \oplus (x_2, y_2))$. In other words, (AR-1) through (AR-$N$) require us to choose an ordering of the target inputs, and the quality of the resulting predictions for the target outputs depends on this ordering! Similarly, the quality of the resulting predictions also depend on the number of target inputs. In terms of the terminology of Section 2.1, rolling out a neural process in an autoregressive fashion is a no longer a consistent probabilistic meta-learning algorithm. The predictions therefore no longer define a stochastic process.

For a task with context set $D^{(c)}$, target inputs $\mathbf{x}^{(t)}$, and target outputs $\mathbf{y}^{(t)}$, call the log-probability of $\mathbf{y}^{(t)}$ under $\mathrm{AR}^\sigma_{\mathbf{x}^{(t)}}(\pi, D^{(c)})$ the *autoregressive log-likelihood*. To be clear and avoid ambiguous language, we will call the log-probability of $\mathbf{y}^{(t)}$ under $P^\sigma_{\mathbf{x}^{(t)}} \pi(D^{(c)})$ the *usual log-likelihood*:

$$\text{usual log-likelihood:} \quad \log q_\theta(\mathbf{y}^{(t)} \,|\, \mathbf{x}^{(t)}, D^{(c)}), \tag{5.20}$$

$$\text{autoregressive log-likelihood:} \quad \sum_{n=1}^{N} \log q_\theta(y_n^{(t)} \,|\, x_n^{(t)}, D^{(c)} \oplus (\mathbf{x}^{(t)}_{1:(n-1)}, \mathbf{y}^{(t)}_{1:(n-1)})). \tag{5.21}$$

In some cases, the autoregressive log-likelihood can be a much better estimate of the true log-probability of the data than the usual log-likelihood. By averaging the autoregressive likelihood for all tasks in the meta–data set, we define the *autoregressive neural process objective* $\mathcal{L}_M^{(\mathrm{AR})}$:

$$\mathcal{L}_M(\pi, \sigma) = \frac{1}{M} \sum_{m=1}^{M} \log q_\theta(\mathbf{y}_m^{(t)} \,|\, \mathbf{x}_m^{(t)}, D_m^{(c)}), \tag{5.22}$$

$$\mathcal{L}_M^{(\mathrm{AR})}(\pi, \sigma) = \frac{1}{M} \sum_{m=1}^{M} \sum_{n=1}^{N_m} \log q_\theta(y_{m,n}^{(t)} \,|\, x_{m,n}^{(t)}, D_m^{(c)} \oplus (\mathbf{x}^{(t)}_{m,1:(n-1)}, \mathbf{y}^{(t)}_{m,1:(n-1)})). \tag{5.23}$$

Similarly, in some cases, the autoregressive neural process objective $\mathcal{L}_M^{(\mathrm{AR})}$ can be a much better estimate of the log-probability of a meta–data set than the usual neural process objective $\mathcal{L}_M$.

Using Procedure 5.14, to make a prediction for a task with $N$ target points, the neural process has to be run forward $N$ times; that is, (AR-1) through (AR-$N$) require $N$ applications of the neural process $\pi$. Therefore, whilst any neural process can be rolled out autoregressively, we will focus on the computationally cheapest class of neural processes: conditional neural processes (CNPs). We will use the term *autoregressive conditional neural processes* (AR CNPs) to generally mean rolling out CNPs according to (AR-1) through (AR-$N$). In the remainder of this section, we will discuss the strengths and shortcomings of AR CNPs.

**Strengths.** The first and foremost advantage of AR CNPs over CNPs is that AR CNPs produce predictions which model dependencies between target outputs. In addition, these predictions are *non-Gaussian*: in (AR-1) through (AR-$N$), the samples $y_i$ pass through the prediction map $\pi$, which is implemented with highly nonlinear neural networks. This puts AR CNPs on the same level of flexibility as LNPs. However, although the predictions of AR CNPs are non-Gaussian and flexible, every sample $y_i \sim P^\sigma_{x_i} \pi(\cdots)$ in (AR-1) through (AR-$N$) is still *conditionally Gaussian*, so the predictions are still restricted in some way.

Second, training AR CNPs is as cheap as training CNPs. Namely, to train an AR CNPs, we just train a CNP as we would train it normally. The only difference between AR CNPs and CNPs is how the model is deployed at test time. This is a big advantage compared other models which model dependencies between target outputs, such as LNPs and GNPs, which can be substantially more expensive to train.

Third, even though the autoregressive neural process objective $\mathcal{L}_M^{(\text{AR})}$ can be a much better estimate of the log-probability of a meta–data set than the usual neural process objective $\mathcal{L}_M$, it is *not* necessary to train with $\mathcal{L}_M^{(\text{AR})}$. To see this, consider the derived meta–data set by splitting every task with $N$ target inputs into $N$ tasks with one target point:

$$\underbrace{\{(D_m^{(\text{c})}, \mathbf{x}_m^{(\text{t})}, \mathbf{y}_m^{(\text{t})})\}_{m=1}^M}_{\text{original meta–data set}} \text{ becomes } \underbrace{\bigcup_{m=1}^M \left\{ (D_m^{(\text{c})} \oplus (\mathbf{x}_{m,1:(n-1)}^{(\text{t})}, \mathbf{y}_{m,1:(n-1)}^{(\text{t})}), x_{m,n}^{(\text{t})}, y_{m,n}^{(\text{t})}) \right\}_{n=1}^{N_m}}_{\text{derived meta–data set}}.$$

Then, up to a normalisation factor, $\mathcal{L}_M^{(\text{AR})}$ with the original data set is equal to $\mathcal{L}_M$ with the derived data set: in (5.23), the double summation becomes one summation over the derived data set. For $\mathcal{L}_{\text{NP}}$ with the derived data set, the characterisation of the conditional neural process approximation (CNPA; Definition 3.24) by Proposition 3.26 still applies, because this characterisation works for target set sizes of any fixed size, including size one. Therefore, for the class of CNPs, $\mathcal{L}_M$ and $\mathcal{L}_M^{(\text{AR})}$ are two different objectives for the same solution. Since the autoregressive neural process objective $\mathcal{L}_M^{(\text{AR})}$ is substantially more expensive, the more reasonable choice is to train with the usual neural process objective $\mathcal{L}_M$.

Fourth, in the limit of infinite data, AR CNPs are guaranteed to perform better than GNPs. We formalise this in the following proposition, which is a statement about the conditional neural process approximation (CNPA; Definition 3.24) and Gaussian neural process approximation (GNPA; Definition 3.24). Recall that the CNPA and GNPA are what CNPs and GNPs approximate in the limit of infinite data; see Section 3.5 for a discussion of convergence in the limit of infinite data.

**Proposition 5.15** (Advantage of AR CNPs). *Let $(\pi_\text{C}, \sigma_\text{C})$ be a CNPA and let $(\pi_\text{G}, \sigma_\text{G})$ be the GNPA. Then, for all $\mathbf{x} \in I$ and $D \in \widetilde{\mathcal{D}}$ (see Section 3.2),*

$$\text{KL}(P_\mathbf{x}^{\sigma_f} \pi_f(D), \text{AR}_\mathbf{x}^{\sigma_\text{C}}(\pi_\text{C}, D)) \leq \text{KL}(P_\mathbf{x}^{\sigma_f} \pi_f(D), P_\mathbf{x}^{\sigma_\text{G}} \pi_\text{G}(D)). \quad (5.24)$$

*Proof.* See Appendix C.4. □

**Weaknesses.** Despite the many strengths of AR CNPs, the class also has a few significant weaknesses. The biggest weakness is one which we already discussed: the quality of the predictions of AR CNPs depends on the number and order of the target points. This means that AR CNPs are no longer consistent probabilistic meta-learning algorithms and therefore no longer define stochastic processes (Section 2.1). The lack of consistency has many consequences. One important consequence is that AR CNPs cannot sample successfully sample at arbitrary target locations. What goes wrong is that (AR-1) through (AR-$N$) evaluate the neural process $\pi$ at more and more context points. At some point, the neural process $\pi$ will be evaluated at a number of context points that the model has not seen during training, and the predictions may start to break down. For AR ConvCNPs, due to the spatial locality of the data channel and density channel, what matters is the *density* of the target inputs rather than the number target inputs.

Second, although training an AR CNPs is as cheap as training a CNP, rolling out the CNP according to (AR-1) through (AR-$N$) requires $N$ applications of the neural process $\pi$ rather than just one. Consequently, for large numbers of target points, AR CNPs incur a significant computational cost. One possible workaround is to use an autoregressive GNP instead of an autoregressive CNP. Then, when (AR-1) through (AR-$N$) has nearly exhausted the available computational budget, *e.g.* at $n = N_{\text{budget}} - 1$, the GNP $\pi$ may produce a correlated prediction for the remainder of the target points at once:

$$y_1 \sim P^\sigma_{x_1} \pi(D), \tag{5.25}$$

$$y_2 \sim P^\sigma_{x_2} \pi(D \oplus (x_1, y_1)), \tag{5.26}$$

$$\vdots$$

$$y_{N_{\text{budget}}-1} \sim P^\sigma_{x_{N_{\text{budget}}-1}} \pi\bigl(D \oplus \bigl(\mathbf{x}_{1:(N_{\text{budget}}-2)}, \mathbf{y}_{1:(N_{\text{budget}}-2)}\bigr)\bigr), \tag{5.27}$$

$$\mathbf{y}_{N_{\text{budget}}:N} \sim P^\sigma_{\mathbf{x}_{N_{\text{budget}}:N}} \pi\bigl(D \oplus \bigl(\mathbf{x}_{1:(N_{\text{budget}}-1)}, \mathbf{y}_{1:(N_{\text{budget}}-1)}\bigr)\bigr). \tag{5.28}$$

In this way, it is possible to obtain flexible, non-Gaussian predictions over large numbers of target points without paying the computational cost of running (AR-1) through (AR-$N$) all the way. Note that many more strategies are possible; for example, one can also sample the target outputs in blocks. This sizeable increase in the design space is a consequence of the unfortunate lack of consistency of AR NPs. Note that (5.25) through (5.28) can also be used when there so many target points that consistency becomes an issue, as discussed in the previous paragraph.

Third, like CNPs, but unlike GNPs and LNPs, AR CNPs cannot separate epistemic and aleatoric uncertainty. That is, samples of AR CNPs cannot be decomposed into a smooth component, a component which represents uncertainty about the ground-truth stochastic process, and a noise component. See Section 3.4 for a discussion about separation of epistemic and aleatoric uncertainty. The following proposition provides a partial remedy.

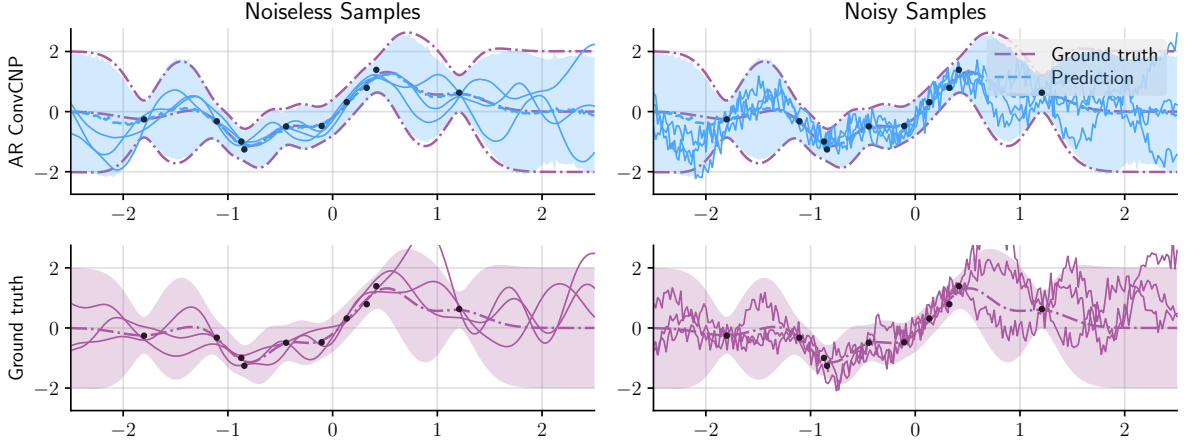

Figure 5.4: Comparison of noiseless (left) and noisy (right) samples from the AR ConvCNP to noisy and noiseless sample of the ground truth. Shows predictions by the models in dashed blue and predictions by the ground truth in dot-dashed purple. Filled regions are 95%-credible regions. The ConvCNP is taken from the experiment in Section 6.2.

**Proposition 5.16** (Recovery of smooth samples). *Let $\mathcal{X} \subseteq \mathbb{R}$ be compact, and let $f$ be a stochastic process with surely continuous sample paths and $\sup_{x \in \mathcal{X}} \|f(x)\|_{L^2} < \infty$. Let $(\varepsilon_n)_{n \geq 0}$ be i.i.d. random variables such that $\mathbb{E}[\varepsilon_0] = 0$ and $\mathrm{var}(\varepsilon_0) < \infty$. Consider any sequence $(x_n)_{n \geq 1} \subseteq \mathcal{X}$, and let $x^* \in \mathcal{X}$ be a limit point of $(x_n)_{n \geq 1}$, assuming that a limit point exists. If $y(x^*) = f(x^*) + \varepsilon_0$ and $y_n = f(x_n) + \varepsilon_n$ are noisy observations of $f$, then*

$$\lim_{n \to \infty} \mathbb{E}[y(x^*) \mid y_1, \ldots, y_n] = f(x^*) \quad \text{almost surely.} \tag{5.29}$$

*Proof.* See Appendix C.4. □

If we assume that a sample of an AR CNP is the sum of a smooth component and independent noise, then Proposition 5.16 says that we can take the noisy sample as a context set $D_{\text{sample}}$ and that the mean of the prediction $m(D_{\text{sample}})$ will approximate the smooth component. This interpretation of Proposition 5.16 assumes that the CNP is able to approximate the true conditional expectation; and that the AR CNP is sampled at sufficiently many inputs. In the limit of infinite data, the former assumption can be true (Proposition 3.26 and Section 3.5). The latter assumption, however, may pose an issue: due to first weakness, the lack of consistency, AR CNPs cannot be evaluated at arbitrarily many target points.

Figure 5.4 shows noiseless and noisy samples from a well-trained AR ConvCNP. In Figure 5.4, the noiseless samples are generated using Proposition 5.16. Note that the amount of noise allocated by the AR ConvCNP looks like the amount allocated by the ground truth. This shows that Proposition 5.16 can successfully be used to separate epistemic and aleatoric uncertainty.

Table 5.2: Comparison of various classes of neural processes. Shows how the two classes of neural processes proposed in this chapter, Gaussian neural processes (GNPs; Section 5.5) and autoregressive conditional neural processes (AR CNPs; Section 5.6), fit in with conditional neural processes (CNPs) and latent-variable neural processes (LNPs). Shows whether a class produces consistent predictions ("Consistent"; Section 2.1); models dependencies in predictions ("Dependencies"); can produce non-Gaussian preditions ("Non-Gaussian"); and can be trained without approximations ("Objective"). For CNPs, even though the presentation by Garnelo et al. (2018a) assumes Gaussian predictions, it is simple relax this Gaussianity assumption; this is not the case for GNPs.

| Class | Consistent | Dependencies | Non-Gaussian | Objective |
| --- | --- | --- | --- | --- |
| AR CNPs (Section 5.6) | ✗ | ✓ | ✓ | ✓ |
| CNPs (Garnelo et al., 2018a) | ✓ | ✗ | ✓ | ✓ |
| GNPs (Section 5.5) | ✓ | ✓ | ✗ | ✓ |
| LNPs (Garnelo et al., 2018b) | ✓ | ✓ | ✓ | ✗ |

**Choice of ordering.** To sample from an AR CNP in practice, one must choose an ordering of the target points. Our recommendation is to choose a different random ordering for every sample. For the first steps in the AR sampling process, the assumption of Gaussian predictions by the CNP may poorly fit the data, so these first few samples may be distorted. However, once you go along in the sampling process and the context set grows large, predictions become more concentrated, and the assumption of Gaussian predictions tends to become less and less restrictive. Therefore, a sample from an AR CNP usually shows the biggest distortions at the target inputs that are sampled first. By choosing a different random ordering for every sample, these distortions are spread out over the input space. This minimises the overall impact of the distortions on the samples and statistics computed from the samples.

Table 5.2 compares the main properties of the classes CNPs, GNPs, LNPs, and AR CNPs.

## 5.7 Conclusion

In this chapter, we proposed a family of neural process models based on convolutional neural networks: the ConvCNP (Model 5.4), the ConvGNP (Model 5.12), and the FullConvGNP (Model 5.13). We call these models *convolutional neural processes* (ConvNPs).

The first neural process was proposed by Garnelo et al. (2018a) and called the Conditional Neural Process (CNP). The CNP is constructed by generally parametrising the prediction map using deep sets (Theorem 4.5). In particular, the CNP makes no assumptions about the underlying ground-truth stochastic process $f$. The family of convolutional neural processes assumes that the ground-truth stochastic process $f$ is stationary. By assuming that $f$ is stationary, ConvNPs may parametrise a prediction map which is *translation* *equivariant* (Proposition 5.2). To parametrise a translation-equivariant prediction map, we used convolutional deep sets (Theorems 4.8 and 4.9) instead of deep sets (Theorem 4.5), resulting in architectures based on convolutional neural networks (CNNs) rather than on multi-layer perceptrons (MLPs). By building translation equivariance in a neural process, we enable the model to better generalise in scenarios where stationarity of the prior $f$ is appropriate (Theorem 5.7).

Convolutional neural processes inherit two important limitations from convolutional deep sets (Theorems 4.8 and 4.9; Section 4.6). First, the implementation of ConvNPs involves discretising a function and passing this discretisation through a CNN (Section 5.3). Importantly, for the ConvCNP and ConvGNP, the dimensionality of this CNN is equal to the dimensionality of the inputs of the data points; and for the FullConvGNP, equal to twice the dimensionality of the inputs of the data points. This means that the ConvCNP and ConvGNP can only feasibly be applied to data with one, two, and three-dimensional; and the FullConvGNP only to data with one-dimensional inputs. Moreover, if the discretisation is fine, then the models may require large amounts of memory and could become computationally too expensive. Second, although ConvNPs successfully parametrise general translation-equivariant prediction maps, perhaps we require a parametrisation which is only *approximately* translation equivariant. Equivalently, perhaps the ground-truth stochastic process $f$ is not perfectly stationary, but only *approximately* stationary. In such scenarios, ideally we would require a model which could "interpolate" between non-convolutional and convolutional architectures, automatically exploiting translation equivariance insofar that is appropriate for the data.

In addition to the above two limitations, we have no understanding of what the CNNs inside convolutional neural processes could be doing. In Appendix D, we perform a preliminary exploration of what could be happening inside a ConvCNP. In this exploration, we explicitly construct a ConvCNP that approximates the predictive mean of a Gaussian process.

We also proposed the class of *Gaussian neural process* (GNPs): the GNP (Model 5.9), the ConvGNP (also a ConvNP; Model 5.12), and the FullConvGNP (also a ConvNP; Model 5.13). Gaussian neural processes address the inability of conditional neural processes to generate coherent samples by directly parametrising covariances between target outputs. In addition, GNPs can be trained without approximations, maintaining a simple objective. The main limitation of GNPs is that these models can only produce Gaussian predictions. Another class that models dependencies between target outputs is the class of latent-variable neural processes (LNPs; Garnelo et al., 2018b). Compared to GNPs, LNP can even produce non-Gaussian predictions. Training LNPs with the neural process objective (Definition 2.2), however, requires additional approximations.

Finally, we argued that a conditional neural process can be trained without any modifications to the model or training procedure and, at test time, can be applied in an autoregressive fashion (Procedure 5.14). We call CNPs deployed in this way *autoregressive CNPs* (AR CNPs). Like LNPs, AR CNPs can produce flexible non-Gaussian predictions; and like GNPs, AR CNPs can be trained without additional approximations. In the limit of infinite data, AR CNPs are even guaranteed to perform better than GNPs (Proposition 5.15). Instead, what we give up is that AR CNPs are no longer consistent probabilistic meta-learning algorithms (Section 2.1), which means that the predictions of AR CNPs are no longer stochastic processes. This lack of consistency, unfortunately, comes with a wide array of new issues. For example, the quality of the predictions depends on the ordering and the number of target points. See Section 5.6 for a more detailed discussion. Another downside of AR CNPs is that making predictions for $N$ target points now requires $N$ forward passes of the neural process, incurring substantial computational cost. AR CNPs equip the neural process framework with a new knob where modelling complexity and computational expense at training time can be traded for computational expense at test time.

In addition to the models proposed in this chapter, we remark that a plethora of other approaches are possible. For example, we could construct GNPs by parametrising the kernel map not through the eigenmap nor using Theorem 4.9. Moreover, we could consider mixtures of CNPs and GNPs, or CNPs and GNPs could be turned into non-Gaussian models by transforming the marginals using an invertible transform. We could even consider using GNPs—or AR CNPs, if one is courageous enough—as encoders and/or decoders inside LNPs, attempting to combine the benefits of different classes of neural processes.

# 6 | Convolutional Neural Processes in Practice

**Abstract.** In the previous chapter, we proposed a variety of new neural process models. In this chapter, we put these models to the test. We establish the general strengths and weaknesses of the various classes of neural processes, and we demonstrate that neural processes can be deployed in a variety of applications.

**Outline.** Section 6.2 performs a large-scale bake-off between a variety of neural process models. In Section 6.3, we apply neural processes to perform *sim-to-real transfer*. Afterwards, in Section 6.4, we challenge the models on an electroencephalography data set. Finally, in Section 6.5, we use neural processes for *statistical downscaling*.

**Attributions and relationship to prior work.** The results in this chapter are not published. The synthetic experiments in Section 6.3 build on the setup by Gordon, Bruinsma, Foong, Requeima, Dubois, and Turner (2020), Foong, Bruinsma, Gordon, Dubois, Requeima, and Turner (2020), Bruinsma, Requeima, Foong, Gordon, and Turner (2021c), and Markou, Requeima, Bruinsma, Vaughan, and Turner (2022). The predator–prey experiment in Section 6.3 builds on the setup by Gordon et al. (2020) and Markou et al. (2022). The EEG experiment in Section 6.4 builds on the setup by Gordon et al. (2020) and Markou et al. (2022). The climate downscaling experiment in Section 6.5 follows the setup by Markou et al. (2022). Although the results in this chapter are new, the original climate experiment was performed by Anna Vaughan and Stratis Markou. Some sentences or parts of sentences are copied verbatim from Markou et al. (2022). The MLP ConvGNP was originally developed by Anna Vaughan and Stratis Markou, and the multiscale architecture for the AR ConvCNP was developed in collaboration with Anna Vaughan. For the climate downscaling experiment in this chapter, Anna Vaughan prepared the 25 ERA-Interim reanalysis variables and the $1\,\text{km}$–resolution elevation data. Generally, Anna Vaughan contributed substantially by always being available for discussions and helping out with the data whenever the author got stuck. All work was supervised by Richard E. Turner.

## 6.1 Introduction

In the previous chapter, we proposed a number of new neural process models: the Convolutional Conditional Neural Process (ConvCNP; Model 5.4), the Gaussian Neural Process (GNP; Model 5.9), the Convolutional Gaussian Neural Process (ConvGNP; Model 5.12), and the Fully Convolutional Gaussian Neural Process (FullConvGNP; Model 5.13). Moreover, we proposed that the original Conditional Neural Process (CNP; Garnelo et al., 2018a) and the ConvCNP can be deployed in an autoregressive fashion (Procedure 5.14 and Section 5.6); we abbreviated this application of the CNP and ConvCNP by AR CNP and AR ConvCNP respectively. In this chapter, we put these new models and approaches to the test. The goal of this chapter is twofold: to establish the strengths and weaknesses of these newly proposed models, especially in relation to existing neural processes; and to demonstrate that neural processes can be deployed in a variety of applications.

The new models constitute three new classes of neural processes. First, the class of *convolutional neural processes* (ConvNPs; Section 5.3) embeds translation equivariance (Definition 2.4) into the models. The class of ConvNPs consists of the ConvCNP, ConvGNP, FullConvGNP, and AR ConvCNP. Second, the class of *Gaussian neural processes* (GNPs; Section 5.5) directly parametrises covariances between target outputs. The class of GNPs consists of the GNP, ConvGNP, and FullConvGNP. Third, the class of *autoregressive conditional neural processes* (AR CNPs; Section 5.6) train a conditional neural process as usual but deploy the model in an autoregressive fashion. The class of AR CNPs consists of the AR CNP and AR ConvCNP.

We will benchmark the proposed models against the following existing neural process models: the original CNP, the Neural Process (NP; Garnelo et al., 2018b), the Attentive Neural Process (ANP; Kim et al., 2019), and the Convolutional Neural Process (ConvNP; Foong et al., 2020). The ConvNP is the natural latent variable variant of the ConvCNP, but the ConvNP is not presented in this thesis. In addition to these four models we throw three additional variants of the ANP into the mix: the Conditional Attentive Neural Process (ACNP), the autoregressive ACNP (AR ACNP), and the Attentive Gaussian Neural Process (AGNP). The ACNP is the ANP, but without the latent variable. The AGNP is the extension of the ACNP that directly parametrises covariances between target outputs, exactly like how Model 5.9 extends Model 5.1; see Appendix E.1. Table 6.1 shows an overview of the existing models and the newly proposed models.

In Sections 6.2 to 6.5, we will analyse the properties of these neural process models and demonstrate that neural processes can be applied in a variety of different settings. To these ends, we will conduct four experiments. First, in Section 6.2, we run all models in Table 6.1 in 60 different subexperiments involving synthetically generated data. The purpose of this experiment is to evaluate the models in cases where the ground truth is known. Afterwards, in Section 6.3, we investigate the ability of neural processes to perform in a setting called *sim-to-real transfer*. In the setting of sim-to-real transfer, the neural process is trained on data generated by a data simulator. After the model is trained on this synthetic source of data, the neural process is applied to real data. In this manner, the neural process facilitates a way in which data simulators can be used to make predictions. In Section 6.4, we apply neural processes to a challenging electroencephalography data set (Zhang et al., 1995), pushing the limits of the models. Last, in Section 6.5, we use neural processes to perform a climate science application called *statistical downscaling* (Maraun et al., 2018).

Table 6.1: Overview of all neural process models evaluated in experiments in this chapter. These models can be divided into the groups of conditional neural processes (CNP; Garnelo et al., 2018a), autoregressive conditional neural processes (AR CNPs; Section 5.6), and neural processes (NPs; Garnelo et al., 2018b). Alternatively, these models can be divided into the groups of deep set–based neural processes (Garnelo et al., 2018a), attentive neural processes (Kim et al., 2019), and convolutional neural processes (Section 5.3). Starred models are models proposed in this thesis.

|  | Deep set based (Garnelo et al., 2018a) | Attentive (Kim et al., 2019) | Convolutional (Section 5.3) |
| --- | --- | --- | --- |
| CNPs (Garnelo et al., 2018a) | CNP [1] | ACNP [5] | ConvCNP* [9] |
| AR CNPs (Section 5.6) | CNP (AR)* [2] | ACNP (AR)* [6] | ConvCNP (AR)* [10] |
| GNPs (Section 5.5) | GNP* [3] | AGNP* [7] | ConvGNP* [11] |
|  |  |  | FullConvGNP* [12] |
| NPs (Garnelo et al., 2018b) | NP [4] | ANP [8] | ConvNP [13] |

[1] Conditional Neural Process (Garnelo et al., 2018a)
[2] Autoregressive Conditional Neural Process (Section 5.6)
[3] Gaussian Neural Process (Model 5.9; Section 5.5)
[4] Neural Process (Garnelo et al., 2018b)
[5] Attentive Conditional Neural Process (Kim et al., 2019)
[6] Attentive Autoregressive Conditional Neural Process (Section 5.6)
[7] Attentive Gaussian Neural Process (Appendix E.1; Section 5.5)
[8] Attentive Neural Process (Kim et al., 2019)
[9] Convolutional Conditional Neural Process (Model 5.4; Section 5.3)
[10] Convolutional Autoregressive Conditional Neural Process (Section 5.6)
[11] Convolutional Gaussian Neural Process (Model 5.12; Section 5.5)
[12] Fully Convolutional Gaussian Neural Process (Model 5.13; Section 5.5)
[13] Convolutional Neural Process (Foong et al., 2020)

For each experiment, we will detail the data and the experimental setup. A precise description of the architectures of the models, however, is deferred to Appendix E.1. We only remark that the convolutional neural processes use a U-Net architecture (Ronneberger et al., 2015), and that all GNPs, in addition to a covariance matrix over the target points, also produce heterogeneous observation noise. All models are configured in a way that makes comparisons fair and ensures that every model has sufficient capacity. We train the models on NVIDIA Tesla V100s with $16\,\mathrm{GB}$ memory. The general training, cross-validation, and evaluation protocols are described in Appendix E.2. Sometimes we reduce the size of the architecture of a model to make sure that the model fits in memory and takes reasonable time to train; these changes are documented in Appendices E.3 to E.6.

In all experiments, we will measure performance with the average log-probability of the target set under the prediction given the context set normalised by the number of target points. For this metric, a small increase $x$ can roughly be interpreted as tightening the model's predictions across the board by a fraction $x$ on average. More precisely, suppose that, for every target point, the prediction by the ground-truth stochastic process is the uniform distribution over some interval. Also suppose that the predictions by some model are uniform distributions over some intervals. Then, for small $x$, an $x$ increase in the metric corresponds to, on average, shrinking the model's predicted intervals by roughly a fraction $x$, whilst at all times ensuring that the ground-truth intervals are entirely contained in the model's predictions. We now state that the smallest such shrinkage that we deem practically significant is $1\%$. Hence, the smallest increase in the metric that we deem practically significant is roughly $0.01$. We therefore decide to show all log-likelihoods up to two decimal places.

All experiments can be reproduced by appropriately calling `train.py` in the root of https://github.com/wesselb/neuralprocesses. Please see the documentation of the repository and `train.py --help` for more detailed instructions.

## 6.2 Synthetic Experiments

In the first experiment, we determine the strengths and weaknesses of all models in Table 6.1 by performing a large-scale bake-off. For five synthetically generated data sets with four configurations each, we evaluate every model in three different ways, totalling 60 subexperiments per model. We first detail the experimental setup and then describe the results.

We synthetically generate data sets by randomly sampling from five different choices for the ground-truth stochastic process $f$. Let the inputs be $d_x$-dimensional. Then define the following stochastic processes:

EQ: a Gaussian process with an exponentiated quadratic (EQ) kernel:

$$f \sim \mathcal{GP}(0, \exp(-\tfrac{1}{2\ell^2}\|\mathbf{x} - \mathbf{x}'\|_2^2)) \tag{6.1}$$

where $\ell > 0$ is a length scale;

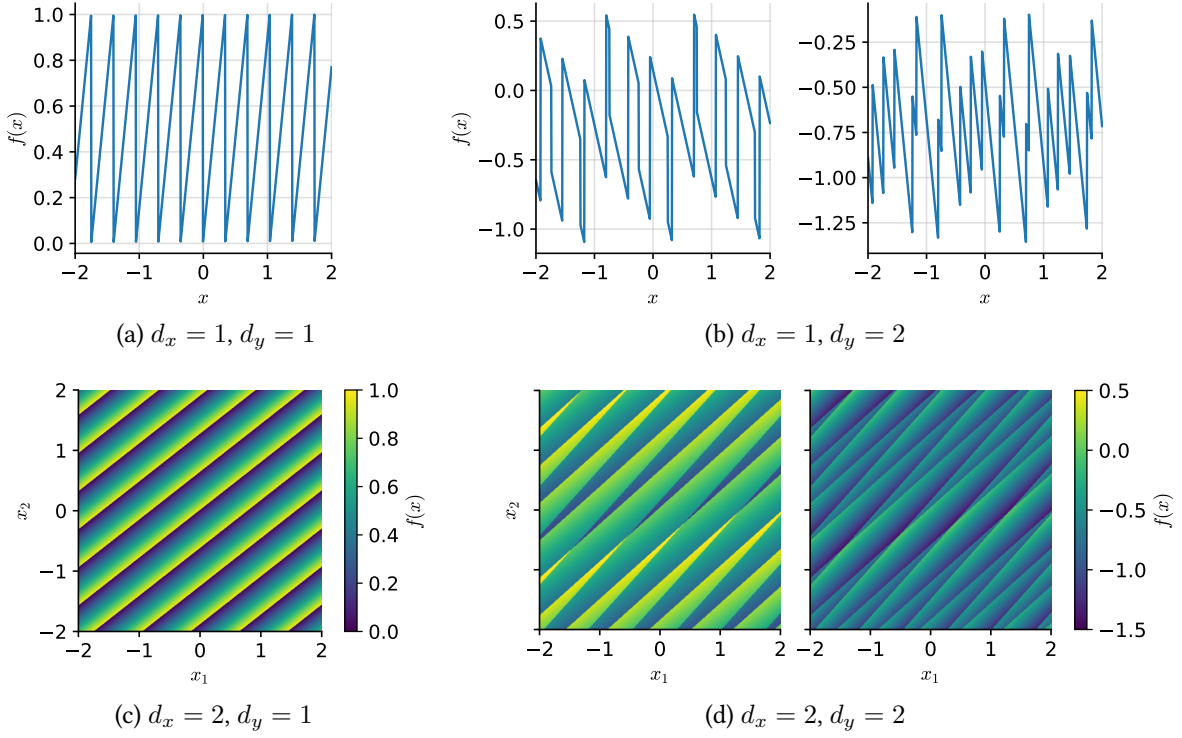

Figure 6.1: Samples from the sawtooth data process with one and two-dimensional inputs ($d_x = 1$ and $d_x = 2$) and one and two-dimensional outputs ($d_y = 1$ and $d_y = 2$)

Matérn–$\frac{5}{2}$: a Gaussian process with a Matérn–$\frac{5}{2}$ kernel:

$$f \sim \mathcal{GP}(0, k(\tfrac{1}{\ell}\|\mathbf{x} - \mathbf{x}'\|_2)) \tag{6.2}$$

where $k(r) = (1 + \sqrt{5}r + \tfrac{5}{3}r^2)e^{-r}$ and $\ell > 0$ is a length scale;

weakly periodic: a Gaussian process with a weakly periodic kernel:

$$f \sim \mathcal{GP}(0, \exp(-\tfrac{1}{2\ell_{\mathrm{d}}^2}\|\mathbf{x} - \mathbf{x}'\|_2^2 - \tfrac{2}{\ell_{\mathrm{p}}^2}\|\sin(\tfrac{\pi}{p}(\mathbf{x} - \mathbf{x}'))\|_2^2)) \tag{6.3}$$

where $\ell_{\mathrm{d}} > 0$ is a length scale specifying how quickly the periodic pattern changes, $\ell_{\mathrm{p}} > 0$ a length scale of the periodic pattern, and $p > 0$ the period; and where the application of $\sin$ is elementwise;

sawtooth: a sawtooth process with a random frequency, direction, and phase:

$$f = \omega\langle\mathbf{x}, \mathbf{u}\rangle_2 + \phi \mod 1 \tag{6.4}$$

where $\omega \sim \mathrm{Unif}(\Omega)$ is the frequency of the sawtooth wave, $\mathbf{u} \sim \mathrm{Unif}(\{\mathbf{x} \in \mathbb{R}^{d_x} : \|\mathbf{x}\|_2 = 1\})$ the direction, and $\phi \sim \mathrm{Unif}([0, 1])$ the phase;

mixture: with equal probability, sample $f$ from the EQ process, Matérn–$\frac{5}{2}$ process, weakly periodic process, or sawtooth process.

We will call these stochastic processes the *data processes*. The data processes are stochastic processes with $d_x$-dimensional inputs and one-dimensional outputs. We will turn them into processes with $d_y$-dimensional outputs according to the following procedure: sample from the one-dimensional-output prior $d_y$ times; and, for these $d_y$ samples, take $d_y$ different linear combinations. The coefficients of these linear combinations are once independently drawn from $\mathcal{N}(0, 1)$ and then fixed. We will consider one ($d_x = 1$) and two-dimensional inputs ($d_x = 2$); and one ($d_y = 1$) and two-dimensional outputs ($d_y = 2$); yielding four configurations per data process. Whereas thus far we only considered one-dimensional inputs and outputs, these configurations will generate context and target sets with potentially multidimensional inputs and outputs. We choose the parameters of the data processes based on the input dimensionality $d_x$:

$$\ell = c \cdot \tfrac{1}{4}, \qquad \ell_\text{d} = c \cdot \tfrac{1}{2}, \qquad \ell_\text{s} = c, \qquad p = c \cdot \tfrac{1}{4}, \qquad \Omega = [c^{-1} \cdot 2, c^{-1} \cdot 4] \qquad (6.5)$$

with $c = \sqrt{d_x}$. Scaling with the input dimensionality aims to roughly ensure that data with one-dimensional inputs and data with two-dimensional inputs are equally difficult. Figure 6.1 illustrates the sawtooth data process in all four configurations.

We will construct data sets by sampling inputs uniformly at random from $\mathcal{X} = [-2, 2]^{d_x}$ and then sampling outputs from one of the data processes. We will colloquially call $\mathcal{X}$ the *training range*. For the EQ, Matérn–$\frac{5}{2}$, and weakly periodic process, but not for the sawtooth process[1], we also add independent Gaussian noise with variance $0.05$. The numbers of context and target points are as follows. For the EQ, Matérn–$\frac{5}{2}$, and weakly periodic process, the number of context points is chosen uniformly at random from $\{0, \ldots, 30 \cdot d_x\}$ and the number of targets points is fixed to $50 \cdot d_x$. For the sawtooth and mixture process, the number of context points is chosen uniformly at random from $\{0, \ldots, 75 \cdot d_x\}$ and the number of targets points is fixed to $100 \cdot d_x$. In the case of a multidimensional-output data process, we separately sample the number and positions of the context and target inputs for every output dimension.

For every data process and each of the four configurations, we train every model from Table 6.1 and evaluate the model in three different ways. First, we evaluate the model on data generated exactly like the training data. This task is called *interpolation* and abbreviated

---

[1] The sawtooth process is already challenging enough.

"int." in the tables of results. The interpolation task measures how well a model fits the data and is the primary measure of performance. Second, we evaluate the model on data with inputs sampled from $[2, 6]^{d_x}$. This task is called *out-of-input-distribution (OOID) interpolation* and abbreviated "OOID" in the tables of results. OOID interpolation measures how well a model generalises to data sampled from other regions of the input space. Third, we evaluate the model on data with context inputs sampled from $[-2, 2]^{d_x}$ and target inputs sampled from $[2, 6]^{d_x}$. This task is called *extrapolation* and abbreviated "ext." in the tables of results. The extrapolation task measures how well predictions based on data in the training range generalise to other regions of the input space.

Further details on the setup and execution of this experiment can be found in Appendices E.1 to E.3. Appendix E.1 describes the general architectures of the models, Appendix E.2 describes the general training, cross-validation, and evaluation protocols, and Appendix E.3 describes details specific to this experiment.

**Results.** In the description of the results, we call the experiments with the EQ, Matérn–$\frac{5}{2}$, and weakly periodic processes the *Gaussian experiments* and the experiments with the sawtooth and mixture processes the *non-Gaussian experiments*. On the page after next page, Table 6.2 summarises the results of the Gaussian experiments, and Table 6.3 summarises the results of the non-Gaussian experiments. After these summaries, the next ten pages show the full results: Tables 6.4 and 6.5 tabulate the results for the EQ process, Tables 6.6 and 6.7 tabulate the results for the Matérn–$\frac{5}{2}$ process, Tables 6.8 and 6.9 tabulate the results for the weakly periodic process, Tables 6.10 and 6.11 tabulate the results for the sawtooth process, and Tables 6.12 and 6.13 tabulate the results for the mixture process.

**A word of caution.** After this page, we will proceed to present all aforementioned tables. The description of the results started in the next paragraph will continue after these tables. In the tables, every LNP occurs three times, corresponding to different combinations of training and evaluation objectives; see Appendix E.2. As this thesis is primarily concerned with non-latent-variable neural processes, we will consider the performance of an LNP to simply be the best of the three numbers.

**Performance of CNPs in Gaussian experiments.** In the Gaussian experiments, the ground-truth $f$ is a Gaussian process, so we can compute the conditional neural process approximation (CNPA; Definition 3.24). Recall that the CNPA is what a CNP converges to in the limit of infinite data; the CNPA provides an upper bound on the performance of a CNP. The CNPA is given by taking the mean map (Definition 3.18) and variance map (Definition 3.20) of the posterior prediction map (Proposition 3.26). Taking the variance map rather than the kernel map (Definition 3.19) discards correlations, so the CNPA is, in some sense, a *diagonalised* version of the posterior prediction map $\pi_f$. This diagonalised version

Table 6.2: For the Gaussian experiments, average Kullback–Leibler divergences of the posterior prediction map $\pi_f$ with respect to the model normalised by the number of target points. Shows for one-dimensional inputs (1D; $d_x = 1$) and two-dimensional inputs (2D; $d_y = 2$) the performance for interpolation within the range $[-2, 2]^{d_x}$ where the models were trained ("Int."); interpolation within the range $[2, 6]^{d_x}$ which the models have never seen before ("OOID"); and extrapolation from the range $[-2, 2]^{d_x}$ to the range $[2, 6]^{d_x}$ ("Ext."). Models are ordered by interpolation performance for one-dimensional inputs. The latent variable models are trained and evaluated with the ELBO objective (ELBO); trained and evaluated with the ML objective (ML); and trained with the ELBO objective and evaluated with the ML objective (ELBO–ML; E.–M.). Diagonal GP refers to predictions by the ground-truth Gaussian processes without correlations. Trivial refers to predicting the empirical means and standard deviation of the test data. Errors indicate the central 95%-confidence interval. Numbers which are significantly best ($p < 0.05$) are boldfaced. Numbers which are very large are marked as failed with "F". Numbers which are missing could not be run.

| Model | Int. (1D) | OOID (1D) | Ext. (1D) | Int. (2D) | OOID (2D) | Ext. (2D) |
| --- | --- | --- | --- | --- | --- | --- |
| FullConvGNP | **0.01** ±0.00 | **0.01** ±0.00 | **0.00** ±0.00 | | | |
| ConvCNP (AR) | 0.03 ±0.00 | 0.03 ±0.00 | 0.02 ±0.00 | **0.03** ±0.00 | **0.03** ±0.00 | **0.02** ±0.00 |
| ConvGNP | 0.04 ±0.00 | 0.04 ±0.00 | 1.75 ±0.12 | 0.12 ±0.00 | 0.12 ±0.00 | 0.71 ±0.03 |
| AGNP | 0.10 ±0.00 | 4.34 ±0.17 | 5.45 ±0.23 | 0.17 ±0.00 | 0.62 ±0.01 | 0.39 ±0.01 |
| ConvNP (ELBO) | 0.19 ±0.01 | 0.19 ±0.01 | 0.29 ±0.03 | 0.39 ±0.01 | 0.39 ±0.01 | 0.36 ±0.01 |
| ACNP (AR) | 0.24 ±0.01 | 1.08 ±0.02 | 0.86 ±0.01 | 0.13 ±0.00 | 0.57 ±0.01 | 0.40 ±0.01 |
| GNP | 0.25 ±0.01 | F | F | 0.25 ±0.01 | 0.75 ±0.01 | 0.57 ±0.00 |
| ConvNP (ML) | 0.31 ±0.01 | 0.31 ±0.01 | 0.64 ±0.01 | 0.28 ±0.01 | 0.28 ±0.01 | 0.36 ±0.01 |
| *Diagonal GP* | 0.42 ±0.02 | 0.42 ±0.02 | 0.84 ±0.01 | 0.29 ±0.01 | 0.29 ±0.01 | 0.40 ±0.01 |
| ANP (ML) | 0.43 ±0.01 | 1.03 ±0.02 | 0.78 ±0.01 | 0.31 ±0.01 | 0.55 ±0.01 | 0.39 ±0.01 |
| ConvCNP | 0.43 ±0.02 | 0.43 ±0.02 | 0.84 ±0.01 | 0.30 ±0.01 | 0.30 ±0.01 | 0.40 ±0.01 |
| CNP (AR) | 0.46 ±0.01 | F | F | 0.36 ±0.01 | F | F |
| NP (ELBO) | 0.51 ±0.01 | F | 4.34 ±0.76 | 0.40 ±0.01 | 3.03 ±1.68 | 0.60 ±0.01 |
| NP (ELBO–ML) | 0.52 ±0.02 | F | 2.39 ±0.33 | 0.39 ±0.01 | 2.35 ±1.06 | 0.57 ±0.01 |
| ANP (ELBO–ML) | 0.53 ±0.01 | 1.12 ±0.03 | 0.85 ±0.02 | 0.42 ±0.01 | 0.78 ±1.72 | 0.41 ±0.01 |
| ACNP | 0.54 ±0.02 | 1.11 ±0.02 | 0.84 ±0.01 | 0.34 ±0.01 | 0.57 ±0.01 | 0.40 ±0.01 |
| ANP (ELBO) | 0.54 ±0.01 | 1.60 ±0.06 | 1.25 ±0.03 | 0.43 ±0.01 | 1.06 ±3.04 | 0.42 ±0.01 |
| NP (ML) | 0.59 ±0.01 | F | 1.13 ±0.01 | 0.41 ±0.01 | 0.88 ±0.03 | 0.52 ±0.01 |
| CNP | 0.63 ±0.01 | F | 1.08 ±0.02 | 0.43 ±0.01 | 1.16 ±0.45 | 0.52 ±0.01 |
| *Trivial* | 1.08 ±0.01 | 1.08 ±0.01 | 0.85 ±0.01 | 0.57 ±0.01 | 0.57 ±0.01 | 0.40 ±0.00 |
| ConvNP (E.–M.) | 2.01 ±0.11 | 2.01 ±0.11 | 5.95 ±0.16 | 0.44 ±0.01 | 0.44 ±0.01 | 0.47 ±0.01 |

Table 6.3: For the non-Gaussian experiments, average log-likelihoods normalised by the number of target points. Shows for one-dimensional inputs (1D; $d_x = 1$) and two-dimensional inputs (2D; $d_y = 2$) the performance for interpolation within the range $[-2, 2]^{d_x}$ where the models were trained ("Int."); interpolation within the range $[2, 6]^{d_x}$ which the models have never seen before ("OOID"); and extrapolation from the range $[-2, 2]^{d_x}$ to the range $[2, 6]^{d_x}$ ("Ext."). Models are ordered by interpolation performance for one-dimensional inputs. The latent variable models are trained and evaluated with the ELBO objective (ELBO); trained and evaluated with the ML objective (ML); and trained with the ELBO objective and evaluated with the ML objective (ELBO–ML; E.–M.). Trivial refers to predicting the empirical means and standard deviation of the test data. Errors indicate the central 95%-confidence interval. Numbers which are significantly best ($p < 0.05$) are boldfaced. Numbers which are very large are marked as failed with "F". Numbers which are missing could not be run.

| Model | Int. (1D) | OOID (1D) | Ext. (1D) | Int. (2D) | OOID (2D) | Ext. (2D) |
|---|---|---|---|---|---|---|
| ConvCNP (AR) | **1.62** ±0.04 | **1.62** ±0.04 | **1.33** ±0.04 | **0.56** ±0.03 | **0.56** ±0.03 | **0.29** ±0.03 |
| ConvNP (ELBO) | **1.62** ±0.05 | **1.61** ±0.05 | 0.92 ±0.04 | 0.06 ±0.03 | 0.06 ±0.03 | −0.62 ±0.04 |
| ConvNP (ML) | 1.57 ±0.06 | 1.57 ±0.06 | −0.26 ±0.03 | 0.26 ±0.04 | 0.26 ±0.04 | −0.70 ±0.02 |
| ConvGNP | 1.43 ±0.08 | 1.44 ±0.06 | −0.96 ±0.12 | 0.23 ±0.04 | 0.23 ±0.04 | −0.79 ±0.02 |
| FullConvGNP | 1.36 ±0.08 | 1.37 ±0.07 | −0.15 ±0.02 | | | |
| ConvCNP | 1.19 ±0.08 | 1.20 ±0.06 | −0.68 ±0.02 | 0.18 ±0.05 | 0.18 ±0.05 | −0.86 ±0.03 |
| ACNP (AR) | 0.21 ±0.03 | −0.86 ±0.03 | −0.84 ±0.02 | −0.53 ±0.02 | −1.52 ±0.10 | −1.51 ±0.09 |
| ANP (ELBO–ML) | 0.13 ±0.04 | −0.97 ±0.04 | −0.88 ±0.04 | −0.67 ±0.03 | −1.06 ±0.62 | −0.70 ±0.03 |
| ANP (ELBO) | 0.11 ±0.04 | F | −3.04 ±0.08 | −0.68 ±0.03 | −2.59 ±7.64 | −0.75 ±0.03 |
| AGNP | −0.12 ±0.03 | −1.18 ±0.07 | −1.64 ±0.12 | −0.55 ±0.02 | −0.79 ±0.02 | −0.76 ±0.03 |
| ACNP | −0.17 ±0.03 | −0.84 ±0.03 | −0.86 ±0.03 | −0.60 ±0.02 | −1.50 ±0.10 | −0.73 ±0.03 |
| ANP (ML) | −0.17 ±0.03 | −0.69 ±0.03 | −0.68 ±0.03 | −0.53 ±0.02 | −0.74 ±0.04 | −0.69 ±0.02 |
| NP (ELBO–ML) | −0.29 ±0.02 | F | F | −0.66 ±0.02 | F | −0.96 ±0.03 |
| NP (ELBO) | −0.30 ±0.02 | F | F | −0.66 ±0.02 | F | F |
| GNP | −0.37 ±0.02 | F | F | −0.69 ±0.02 | −0.74 ±0.04 | −0.70 ±0.03 |
| NP (ML) | −0.48 ±0.02 | −2.90 ±0.39 | −0.81 ±0.03 | −0.62 ±0.02 | −1.50 ±0.11 | −0.75 ±0.03 |
| CNP (AR) | −0.65 ±0.02 | −4.14 ±7.34 | −1.23 ±0.13 | −0.69 ±0.02 | −1.05 ±0.07 | −0.72 ±0.03 |
| CNP | −0.66 ±0.02 | −1.65 ±0.25 | −0.75 ±0.03 | −0.69 ±0.02 | −1.05 ±0.08 | −0.71 ±0.03 |
| *Trivial* | −0.82 ±0.00 | −0.82 ±0.00 | −0.82 ±0.00 | −0.82 ±0.00 | −0.82 ±0.00 | −0.82 ±0.00 |
| ConvNP (E.–M.) | −6.57 ±3.76 | −6.68 ±3.76 | F | −0.04 ±0.05 | −0.04 ±0.05 | −1.47 ±0.87 |

Table 6.4: For the EQ synthetic experiments with one-dimensional inputs, average Kullback–Leibler divergences of the posterior prediction map $\pi_f$ with respect to the model normalised by the number of target points. Shows for one-dimensional outputs ($d_y = 1$) and two-dimensional outputs ($d_y = 2$) the performance for interpolation within the range $[-2, 2]$ where the models where trained ("Int."); interpolation within the range $[2, 6]$ which the models have never seen before ("OOID"); and extrapolation from the range $[-2, 2]$ to the range $[2, 6]$ ("Ext."). Models are ordered by interpolation performance. The latent variable models are trained and evaluated with the ELBO objective (ELBO); trained and evaluated with the ML objective (ML); and trained with the ELBO objective and evaluated with the ML objective (ELBO–ML; E.–M.). Diagonal GP refers to predictions by the ground-truth Gaussian processes without correlations. Trivial refers to predicting the empirical means and standard deviation of the test data. Errors indicate the central 95%-confidence interval. Numbers which are significantly best ($p < 0.05$) are boldfaced. Numbers which are very large are marked as failed with "F". Numbers which are missing could not be run.

| EQ $d_x=1, d_y=1$ | Int. | OOID | Ext. | EQ $d_x=1, d_y=2$ | Int. | OOID | Ext. |
| --- | --- | --- | --- | --- | --- | --- | --- |
| FullConvGNP | **0.00** ±0.00 | **0.00** ±0.00 | **0.00** ±0.00 | FullConvGNP | **0.00** ±0.00 | **0.00** ±0.00 | **0.00** ±0.00 |
| ConvGNP | 0.01 ±0.00 | 0.01 ±0.00 | 3.46 ±0.08 | ConvGNP | 0.01 ±0.00 | 0.01 ±0.00 | 1.73 ±0.05 |
| ConvCNP (AR) | 0.01 ±0.00 | 0.01 ±0.00 | 0.01 ±0.00 | ConvCNP (AR) | 0.01 ±0.00 | 0.01 ±0.00 | 0.01 ±0.00 |
| AGNP | 0.03 ±0.00 | 4.28 ±0.08 | 7.38 ±0.13 | AGNP | 0.04 ±0.00 | 7.71 ±0.10 | 7.87 ±0.10 |
| ConvNP (ELBO) | 0.06 ±0.00 | 0.06 ±0.00 | 0.11 ±0.01 | ACNP (AR) | 0.07 ±0.00 | 1.31 ±0.01 | 1.09 ±0.01 |
| ACNP (AR) | 0.07 ±0.00 | 1.19 ±0.01 | 0.98 ±0.01 | ConvNP (ELBO) | 0.08 ±0.00 | 0.08 ±0.00 | 0.13 ±0.00 |
| GNP | 0.08 ±0.00 | F | F | GNP | 0.13 ±0.00 | F | F |
| ConvNP (ML) | 0.25 ±0.01 | 0.25 ±0.01 | 0.67 ±0.01 | ConvNP (ML) | 0.36 ±0.01 | 0.36 ±0.01 | 0.88 ±0.00 |
| CNP (AR) | 0.28 ±0.00 | F | F | ANP (ML) | 0.41 ±0.01 | 1.23 ±0.01 | 0.99 ±0.00 |
| ANP (ML) | 0.31 ±0.01 | 1.04 ±0.01 | 0.84 ±0.01 | CNP (AR) | 0.42 ±0.00 | F | F |
| NP (ELBO) | 0.34 ±0.01 | F | 1.34 ±0.01 | *Diagonal GP* | 0.47 ±0.01 | 0.47 ±0.01 | 1.06 ±0.00 |
| NP (ELBO–ML) | 0.37 ±0.01 | F | 1.27 ±0.01 | ConvCNP | 0.48 ±0.01 | 0.48 ±0.01 | 1.06 ±0.00 |
| *Diagonal GP* | 0.40 ±0.01 | 0.40 ±0.01 | 0.95 ±0.01 | ACNP | 0.51 ±0.01 | 1.38 ±0.01 | 1.06 ±0.01 |
| ConvCNP | 0.41 ±0.01 | 0.41 ±0.01 | 0.95 ±0.01 | ANP (ELBO–ML) | 0.52 ±0.01 | 1.47 ±0.02 | 1.07 ±0.01 |
| ANP (ELBO–ML) | 0.42 ±0.01 | 1.18 ±0.01 | 0.94 ±0.01 | ANP (ELBO) | 0.53 ±0.01 | 3.79 ±0.05 | 2.84 ±0.02 |
| ANP (ELBO) | 0.44 ±0.01 | 1.32 ±0.01 | 1.25 ±0.01 | NP (ELBO) | 0.54 ±0.01 | F | 1.47 ±0.01 |
| ACNP | 0.45 ±0.01 | 1.22 ±0.01 | 0.95 ±0.01 | NP (ELBO–ML) | 0.56 ±0.01 | F | 1.42 ±0.01 |
| NP (ML) | 0.49 ±0.01 | 1.54 ±0.01 | 1.45 ±0.01 | NP (ML) | 0.64 ±0.00 | F | 1.52 ±0.00 |
| CNP | 0.54 ±0.01 | F | 1.41 ±0.01 | CNP | 0.66 ±0.01 | F | 1.28 ±0.00 |
| ConvNP (E.–M.) | 0.90 ±0.04 | 0.90 ±0.04 | 4.05 ±0.06 | *Trivial* | 1.31 ±0.00 | 1.31 ±0.00 | 1.07 ±0.00 |
| *Trivial* | 1.19 ±0.00 | 1.19 ±0.00 | 0.96 ±0.00 | ConvNP (E.–M.) | 2.14 ±0.06 | 2.14 ±0.06 | 9.30 ±0.08 |

Table 6.5: For the EQ synthetic experiments with two-dimensional inputs, average Kullback–Leibler divergences of the posterior prediction map $\pi_f$ with respect to the model normalised by the number of target points. Shows for one-dimensional outputs ($d_y = 1$) and two-dimensional outputs ($d_y = 2$) the performance for interpolation within the range $[-2, 2]^2$ where the models where trained ("Int."); interpolation within the range $[2, 6]^2$ which the models have never seen before ("OOID"); and extrapolation from the range $[-2, 2]^2$ to the range $[2, 6]^2$ ("Ext."). Models are ordered by interpolation performance. The latent variable models are trained and evaluated with the ELBO objective (ELBO); trained and evaluated with the ML objective (ML); and trained with the ELBO objective and evaluated with the ML objective (ELBO–ML; E.–M.). Diagonal GP refers to predictions by the ground-truth Gaussian processes without correlations. Trivial refers to predicting the empirical means and standard deviation of the test data. Errors indicate the central 95%-confidence interval. Numbers which are significantly best ($p < 0.05$) are boldfaced. Numbers which are very large are marked as failed with "F". Numbers which are missing could not be run.

| EQ $d_x=2, d_y=1$ | Int. | OOID | Ext. | EQ $d_x=2, d_y=2$ | Int. | OOID | Ext. |
|---|---|---|---|---|---|---|---|
| ConvCNP (AR) | **0.01** ±0.00 | **0.01** ±0.00 | **0.01** ±0.00 | ConvCNP (AR) | **0.03** ±0.00 | **0.03** ±0.00 | **0.02** ±0.00 |
| ConvGNP | 0.08 ±0.00 | 0.08 ±0.00 | 1.92 ±0.02 | ACNP (AR) | 0.11 ±0.00 | 0.79 ±0.00 | 0.56 ±0.00 |
| AGNP | 0.09 ±0.00 | 0.70 ±0.00 | 0.50 ±0.00 | ConvGNP | 0.19 ±0.00 | 0.19 ±0.00 | 0.74 ±0.01 |
| ACNP (AR) | 0.09 ±0.00 | 0.72 ±0.00 | 0.51 ±0.00 | AGNP | 0.22 ±0.00 | 0.87 ±0.01 | 0.57 ±0.00 |
| GNP | 0.19 ±0.00 | 1.01 ±0.00 | 0.80 ±0.00 | GNP | 0.38 ±0.00 | 1.06 ±0.00 | 0.75 ±0.00 |
| ConvNP (ML) | 0.34 ±0.00 | 0.34 ±0.00 | 0.47 ±0.00 | ConvNP (ML) | 0.39 ±0.00 | 0.39 ±0.00 | 0.52 ±0.00 |
| ANP (ML) | 0.34 ±0.00 | 0.70 ±0.00 | 0.51 ±0.00 | *Diagonal GP* | 0.40 ±0.00 | 0.40 ±0.00 | 0.56 ±0.00 |
| *Diagonal GP* | 0.36 ±0.00 | 0.36 ±0.00 | 0.51 ±0.00 | ConvCNP | 0.41 ±0.00 | 0.41 ±0.00 | 0.56 ±0.00 |
| ConvCNP | 0.37 ±0.00 | 0.37 ±0.00 | 0.51 ±0.00 | ANP (ML) | 0.42 ±0.00 | 0.79 ±0.00 | 0.54 ±0.00 |
| ACNP | 0.40 ±0.00 | 0.72 ±0.00 | 0.51 ±0.00 | ACNP | 0.44 ±0.00 | 0.79 ±0.00 | 0.56 ±0.00 |
| ConvNP (ELBO) | 0.41 ±0.00 | 0.41 ±0.00 | 0.46 ±0.00 | CNP (AR) | 0.52 ±0.00 | F | F |
| CNP (AR) | 0.41 ±0.00 | 0.90 ±0.00 | 0.71 ±0.00 | NP (ELBO–ML) | 0.56 ±0.00 | 1.95 ±0.03 | 0.72 ±0.00 |
| NP (ELBO–ML) | 0.46 ±0.00 | 0.99 ±0.01 | 0.65 ±0.00 | ANP (ELBO–ML) | 0.56 ±0.00 | 1.90 ±1.72 | 0.55 ±0.00 |
| NP (ELBO) | 0.48 ±0.00 | 1.04 ±0.01 | 0.67 ±0.00 | NP (ELBO) | 0.57 ±0.00 | 1.99 ±0.03 | 0.72 ±0.00 |
| ConvNP (E.–M.) | 0.48 ±0.00 | 0.48 ±0.00 | 0.59 ±0.01 | ANP (ELBO) | 0.57 ±0.00 | 3.51 ±3.04 | 0.56 ±0.00 |
| ANP (ELBO–ML) | 0.49 ±0.01 | 0.72 ±0.00 | 0.51 ±0.00 | NP (ML) | 0.59 ±0.00 | 1.17 ±0.01 | 0.75 ±0.00 |
| ANP (ELBO) | 0.50 ±0.01 | 0.73 ±0.00 | 0.52 ±0.00 | CNP | 0.60 ±0.00 | 3.08 ±0.42 | 0.66 ±0.00 |
| NP (ML) | 0.51 ±0.00 | 0.92 ±0.00 | 0.72 ±0.00 | *Trivial* | 0.79 ±0.00 | 0.79 ±0.00 | 0.56 ±0.00 |
| CNP | 0.52 ±0.00 | 0.92 ±0.00 | 0.72 ±0.00 | ConvNP (E.–M.) | 0.79 ±0.00 | 0.79 ±0.00 | 0.56 ±0.00 |
| *Trivial* | 0.72 ±0.00 | 0.72 ±0.00 | 0.51 ±0.00 | ConvNP (ELBO) | 0.79 ±0.00 | 0.79 ±0.00 | 0.56 ±0.00 |
| FullConvGNP | | | | FullConvGNP | | | |

Table 6.6: For the Matérn–$\frac{5}{2}$ synthetic experiments with one-dimensional inputs, average Kullback–Leibler divergences of the posterior prediction map $\pi_f$ with respect to the model normalised by the number of target points. Shows for one-dimensional outputs ($d_y = 1$) and two-dimensional outputs ($d_y = 2$) the performance for interpolation within the range $[-2, 2]$ where the models where trained ("Int."); interpolation within the range $[2, 6]$ which the models have never seen before ("OOID"); and extrapolation from the range $[-2, 2]$ to the range $[2, 6]$ ("Ext."). Models are ordered by interpolation performance. The latent variable models are trained and evaluated with the ELBO objective (ELBO); trained and evaluated with the ML objective (ML); and trained with the ELBO objective and evaluated with the ML objective (ELBO–ML; E.–M.). Diagonal GP refers to predictions by the ground-truth Gaussian processes without correlations. Trivial refers to predicting the empirical means and standard deviation of the test data. Errors indicate the central 95%-confidence interval. Numbers which are significantly best ($p < 0.05$) are boldfaced. Numbers which are very large are marked as failed with "F". Numbers which are missing could not be run.

| Matérn–$\frac{5}{2}$ $d_x=1$, $d_y=1$ | Int. | OOID | Ext. | Matérn–$\frac{5}{2}$ $d_x=1$, $d_y=2$ | Int. | OOID | Ext. |
|---|---|---|---|---|---|---|---|
| FullConvGNP | **0.00** ±0.00 | **0.00** ±0.00 | **0.00** ±0.00 | FullConvGNP | **0.00** ±0.00 | **0.00** ±0.00 | **0.00** ±0.00 |
| ConvCNP (AR) | 0.00 ±0.00 | 0.00 ±0.00 | 0.00 ±0.00 | ConvCNP (AR) | 0.01 ±0.00 | 0.01 ±0.00 | 0.01 ±0.00 |
| ConvGNP | 0.01 ±0.00 | 0.01 ±0.00 | 2.32 ±0.06 | ConvGNP | 0.02 ±0.00 | 0.02 ±0.00 | 1.71 ±0.04 |
| AGNP | 0.03 ±0.00 | 4.53 ±0.08 | 7.22 ±0.12 | AGNP | 0.04 ±0.00 | 6.14 ±0.08 | 7.03 ±0.09 |
| ACNP (AR) | 0.04 ±0.00 | 1.08 ±0.01 | 0.87 ±0.01 | ACNP (AR) | 0.05 ±0.00 | 1.18 ±0.01 | 0.96 ±0.01 |
| GNP | 0.09 ±0.00 | F | F | GNP | 0.13 ±0.00 | F | F |
| ConvNP (ELBO) | 0.13 ±0.00 | 0.13 ±0.00 | 0.31 ±0.02 | ConvNP (ELBO) | 0.16 ±0.00 | 0.16 ±0.00 | 0.29 ±0.00 |
| ConvNP (ML) | 0.26 ±0.01 | 0.26 ±0.01 | 0.58 ±0.00 | ConvNP (ML) | 0.36 ±0.00 | 0.36 ±0.00 | 0.76 ±0.00 |
| ANP (ML) | 0.30 ±0.00 | 0.98 ±0.01 | 0.78 ±0.01 | ANP (ML) | 0.40 ±0.00 | 1.10 ±0.01 | 0.88 ±0.00 |
| CNP (AR) | 0.34 ±0.01 | 1.81 ±0.04 | 1.32 ±0.02 | CNP (AR) | 0.45 ±0.00 | F | F |
| NP (ELBO) | 0.36 ±0.00 | F | 1.31 ±0.01 | *Diagonal GP* | 0.46 ±0.01 | 0.46 ±0.01 | 0.93 ±0.00 |
| NP (ELBO–ML) | 0.37 ±0.01 | F | 1.14 ±0.00 | ConvCNP | 0.46 ±0.01 | 0.46 ±0.01 | 0.93 ±0.00 |
| *Diagonal GP* | 0.40 ±0.01 | 0.40 ±0.01 | 0.84 ±0.01 | ACNP | 0.49 ±0.01 | 1.23 ±0.01 | 0.93 ±0.00 |
| ConvCNP | 0.40 ±0.01 | 0.40 ±0.01 | 0.84 ±0.01 | ANP (ELBO–ML) | 0.51 ±0.01 | 1.28 ±0.01 | 0.99 ±0.01 |
| ANP (ELBO–ML) | 0.41 ±0.01 | 1.13 ±0.01 | 0.84 ±0.01 | ANP (ELBO) | 0.51 ±0.01 | 1.43 ±0.02 | 1.10 ±0.01 |
| ACNP | 0.42 ±0.01 | 1.10 ±0.01 | 0.84 ±0.01 | NP (ELBO–ML) | 0.54 ±0.00 | F | 1.24 ±0.00 |
| ANP (ELBO) | 0.43 ±0.01 | 1.15 ±0.01 | 0.87 ±0.01 | NP (ELBO) | 0.54 ±0.00 | F | 1.79 ±0.01 |
| NP (ML) | 0.51 ±0.00 | 1.87 ±0.02 | 1.30 ±0.01 | NP (ML) | 0.63 ±0.00 | 2.33 ±0.02 | 1.23 ±0.00 |
| CNP | 0.54 ±0.01 | 1.47 ±0.02 | 1.11 ±0.01 | CNP | 0.65 ±0.00 | 7.72 ±0.69 | 1.23 ±0.00 |
| *Trivial* | 1.08 ±0.00 | 1.08 ±0.00 | 0.85 ±0.00 | *Trivial* | 1.18 ±0.00 | 1.18 ±0.00 | 0.94 ±0.00 |
| ConvNP (E.–M.) | 1.37 ±0.04 | 1.36 ±0.04 | 4.30 ±0.06 | ConvNP (E.–M.) | 3.07 ±0.06 | 3.06 ±0.06 | 9.83 ±0.09 |

Table 6.7: For the Matérn–$\frac{5}{2}$ synthetic experiments with two-dimensional inputs, average Kullback–Leibler divergences of the posterior prediction map $\pi_f$ with respect to the model normalised by the number of target points. Shows for one-dimensional outputs ($d_y = 1$) and two-dimensional outputs ($d_y = 2$) the performance for interpolation within the range $[-2, 2]^2$ where the models where trained ("Int."); interpolation within the range $[2, 6]^2$ which the models have never seen before ("OOID"); and extrapolation from the range $[-2, 2]^2$ to the range $[2, 6]^2$ ("Ext."). Models are ordered by interpolation performance. Diagonal GP refers to predictions by the ground-truth Gaussian processes without correlations. Trivial refers to predicting the empirical means and standard deviation of the test data. Errors indicate the central 95%-confidence interval. Numbers which are significantly best ($p < 0.05$) are boldfaced. Numbers which are very large are marked as failed with "F". Numbers which are missing could not be run.

| Matérn–$\frac{5}{2}$ $d_x=2$, $d_y=1$ | Int. | OOID | Ext. | Matérn–$\frac{5}{2}$ $d_x=2$, $d_y=2$ | Int. | OOID | Ext. |
|---|---|---|---|---|---|---|---|
| ConvCNP (AR) | **0.01** ±0.00 | **0.01** ±0.00 | **0.00** ±0.00 | ConvCNP (AR) | **0.01** ±0.00 | **0.01** ±0.00 | **0.01** ±0.00 |
| ACNP (AR) | 0.05 ±0.00 | 0.54 ±0.00 | 0.38 ±0.00 | ACNP (AR) | 0.06 ±0.00 | 0.58 ±0.00 | 0.41 ±0.00 |
| AGNP | 0.08 ±0.00 | 0.83 ±0.01 | 0.37 ±0.00 | ConvGNP | 0.14 ±0.00 | 0.14 ±0.00 | 0.64 ±0.01 |
| ConvGNP | 0.08 ±0.00 | 0.08 ±0.00 | 0.60 ±0.01 | AGNP | 0.17 ±0.00 | 0.58 ±0.00 | 0.40 ±0.00 |
| GNP | 0.16 ±0.00 | 0.90 ±0.00 | 0.75 ±0.00 | GNP | 0.28 ±0.00 | 0.78 ±0.00 | 0.60 ±0.00 |
| ConvNP (ML) | 0.25 ±0.00 | 0.25 ±0.00 | 0.34 ±0.00 | ConvNP (ML) | 0.29 ±0.00 | 0.29 ±0.00 | 0.38 ±0.00 |
| ANP (ML) | 0.26 ±0.00 | 0.51 ±0.00 | 0.37 ±0.00 | ANP (ML) | 0.29 ±0.00 | 0.56 ±0.00 | 0.40 ±0.00 |
| *Diagonal GP* | 0.28 ±0.00 | 0.28 ±0.00 | 0.39 ±0.00 | *Diagonal GP* | 0.30 ±0.00 | 0.30 ±0.00 | 0.41 ±0.00 |
| ConvCNP | 0.28 ±0.00 | 0.28 ±0.00 | 0.39 ±0.00 | ConvCNP | 0.30 ±0.00 | 0.30 ±0.00 | 0.41 ±0.00 |
| ACNP | 0.29 ±0.00 | 0.54 ±0.00 | 0.39 ±0.00 | ACNP | 0.32 ±0.00 | 0.58 ±0.00 | 0.41 ±0.00 |
| CNP (AR) | 0.31 ±0.00 | 0.69 ±0.00 | 0.52 ±0.00 | ConvNP (ELBO) | 0.36 ±0.00 | 0.36 ±0.00 | 0.37 ±0.00 |
| ConvNP (ELBO) | 0.32 ±0.00 | 0.32 ±0.00 | 0.30 ±0.00 | CNP (AR) | 0.37 ±0.00 | F | 0.88 ±0.17 |
| NP (ELBO–ML) | 0.34 ±0.00 | 1.07 ±0.01 | 0.69 ±0.00 | NP (ELBO–ML) | 0.41 ±0.00 | 2.29 ±0.05 | 0.59 ±0.00 |
| NP (ELBO) | 0.35 ±0.00 | 1.25 ±0.01 | 0.72 ±0.00 | NP (ELBO) | 0.41 ±0.00 | 2.36 ±0.05 | 0.60 ±0.00 |
| ConvNP (E.–M.) | 0.36 ±0.00 | 0.36 ±0.00 | 0.43 ±0.00 | ANP (ELBO–ML) | 0.42 ±0.00 | 0.61 ±0.00 | 0.41 ±0.00 |
| NP (ML) | 0.37 ±0.00 | 0.75 ±0.01 | 0.51 ±0.00 | ANP (ELBO) | 0.42 ±0.00 | 0.61 ±0.00 | 0.41 ±0.00 |
| ANP (ELBO–ML) | 0.39 ±0.00 | 0.65 ±0.01 | 0.43 ±0.00 | NP (ML) | 0.43 ±0.00 | 0.68 ±0.00 | 0.53 ±0.00 |
| CNP | 0.39 ±0.00 | 0.67 ±0.00 | 0.54 ±0.00 | CNP | 0.44 ±0.00 | 0.86 ±0.17 | 0.59 ±0.00 |
| ANP (ELBO) | 0.41 ±0.01 | 0.67 ±0.01 | 0.44 ±0.00 | ConvNP (E.–M.) | 0.49 ±0.00 | 0.49 ±0.00 | 0.61 ±0.00 |
| *Trivial* | 0.55 ±0.00 | 0.55 ±0.00 | 0.39 ±0.00 | *Trivial* | 0.58 ±0.00 | 0.58 ±0.00 | 0.41 ±0.00 |
| FullConvGNP | | | | FullConvGNP | | | |

Table 6.8: For the weakly periodic synthetic experiments with one-dimensional inputs, average Kullback–Leibler divergences of the posterior prediction map $\pi_f$ with respect to the model normalised by the number of target points. Shows for one-dimensional outputs ($d_y = 1$) and two-dimensional outputs ($d_y = 2$) the performance for interpolation within the range $[-2, 2]$ where the models where trained ("Int."); interpolation within the range $[2, 6]$ which the models have never seen before ("OOID"); and extrapolation from the range $[-2, 2]$ to the range $[2, 6]$ ("Ext."). Models are ordered by interpolation performance. The latent variable models are trained and evaluated with the ELBO objective (ELBO); trained and evaluated with the ML objective (ML); and trained with the ELBO objective and evaluated with the ML objective (ELBO–ML; E.–M.). Diagonal GP refers to predictions by the ground-truth Gaussian processes without correlations. Trivial refers to predicting the empirical means and standard deviation of the test data. Errors indicate the central 95%-confidence interval. Numbers which are significantly best ($p < 0.05$) are boldfaced. Numbers which are very large are marked as failed with "F". Numbers which are missing could not be run.

| Weakly Periodic $d_x=1, d_y=1$ | Int. | OOID | Ext. | Weakly Periodic $d_x=1, d_y=2$ | Int. | OOID | Ext. |
|---|---|---|---|---|---|---|---|
| FullConvGNP | **0.02** ±0.00 | **0.02** ±0.00 | **0.00** ±0.00 | FullConvGNP | **0.03** ±0.00 | **0.03** ±0.00 | **0.00** ±0.00 |
| ConvCNP (AR) | 0.05 ±0.00 | 0.05 ±0.00 | 0.04 ±0.00 | ConvCNP (AR) | 0.09 ±0.00 | 0.09 ±0.00 | 0.06 ±0.00 |
| ConvGNP | 0.05 ±0.00 | 0.05 ±0.00 | 0.56 ±0.02 | ConvGNP | 0.12 ±0.00 | 0.12 ±0.00 | 0.72 ±0.01 |
| AGNP | 0.22 ±0.00 | 1.25 ±0.02 | 1.25 ±0.02 | AGNP | 0.25 ±0.00 | 2.17 ±0.02 | 1.95 ±0.02 |
| ConvNP (ML) | 0.28 ±0.00 | 0.28 ±0.00 | 0.43 ±0.00 | ConvNP (ML) | 0.38 ±0.00 | 0.38 ±0.00 | 0.54 ±0.00 |
| ConvNP (ELBO) | 0.34 ±0.00 | 0.33 ±0.00 | 0.45 ±0.02 | ConvNP (ELBO) | 0.39 ±0.00 | 0.39 ±0.00 | 0.44 ±0.00 |
| *Diagonal GP* | 0.38 ±0.01 | 0.38 ±0.01 | 0.59 ±0.01 | *Diagonal GP* | 0.42 ±0.00 | 0.42 ±0.00 | 0.65 ±0.00 |
| ConvCNP | 0.40 ±0.01 | 0.40 ±0.01 | 0.60 ±0.01 | ConvCNP | 0.46 ±0.00 | 0.46 ±0.00 | 0.65 ±0.00 |
| ANP (ML) | 0.53 ±0.00 | 0.77 ±0.01 | 0.57 ±0.01 | GNP | 0.50 ±0.00 | 1.02 ±0.01 | 0.76 ±0.00 |
| ACNP (AR) | 0.57 ±0.01 | 0.82 ±0.01 | 0.61 ±0.01 | ANP (ML) | 0.62 ±0.00 | 1.04 ±0.01 | 0.64 ±0.00 |
| GNP | 0.59 ±0.01 | 1.31 ±0.02 | 0.62 ±0.01 | ACNP (AR) | 0.63 ±0.00 | 0.89 ±0.01 | 0.66 ±0.00 |
| CNP (AR) | 0.59 ±0.01 | 2.33 ±0.27 | 1.46 ±0.05 | CNP (AR) | 0.67 ±0.00 | 2.52 ±0.07 | 1.21 ±0.01 |
| NP (ELBO–ML) | 0.60 ±0.01 | F | 4.09 ±0.28 | NP (ELBO–ML) | 0.68 ±0.00 | F | 5.18 ±0.18 |
| ANP (ELBO–ML) | 0.60 ±0.01 | 0.78 ±0.01 | 0.59 ±0.01 | NP (ML) | 0.69 ±0.00 | 1.26 ±0.01 | 0.68 ±0.01 |
| NP (ML) | 0.60 ±0.01 | 0.80 ±0.01 | 0.62 ±0.01 | NP (ELBO) | 0.69 ±0.00 | F | F |
| NP (ELBO) | 0.61 ±0.01 | F | 9.91 ±0.70 | ANP (ELBO–ML) | 0.70 ±0.00 | 0.85 ±0.01 | 0.64 ±0.00 |
| ANP (ELBO) | 0.62 ±0.01 | 1.01 ±0.01 | 0.71 ±0.01 | ACNP | 0.71 ±0.00 | 0.89 ±0.01 | 0.66 ±0.00 |
| ACNP | 0.65 ±0.01 | 0.82 ±0.01 | 0.61 ±0.01 | ANP (ELBO) | 0.72 ±0.00 | 0.93 ±0.01 | 0.72 ±0.00 |
| CNP | 0.67 ±0.01 | 1.45 ±0.03 | 0.68 ±0.01 | CNP | 0.74 ±0.00 | 1.27 ±0.01 | 0.77 ±0.01 |
| *Trivial* | 0.82 ±0.00 | 0.82 ±0.00 | 0.61 ±0.00 | *Trivial* | 0.89 ±0.00 | 0.89 ±0.00 | 0.67 ±0.00 |
| ConvNP (E.–M.) | 1.58 ±0.03 | 1.57 ±0.03 | 2.85 ±0.04 | ConvNP (E.–M.) | 3.02 ±0.03 | 3.02 ±0.03 | 5.40 ±0.04 |

Table 6.9: For the weakly periodic synthetic experiments with two-dimensional inputs, average Kullback–Leibler divergences of the posterior prediction map $\pi_f$ with respect to the model normalised by the number of target points. Shows for one-dimensional outputs ($d_y = 1$) and two-dimensional outputs ($d_y = 2$) the performance for interpolation within the range $[-2, 2]^2$ where the models where trained ("Int."); interpolation within the range $[2, 6]^2$ which the models have never seen before ("OOID"); and extrapolation from the range $[-2, 2]^2$ to the range $[2, 6]^2$ ("Ext."). Models are ordered by interpolation performance. The latent variable models are trained and evaluated with the ELBO objective (ELBO); trained and evaluated with the ML objective (ML); and trained with the ELBO objective and evaluated with the ML objective (ELBO–ML; E.–M.). Diagonal GP refers to predictions by the ground-truth Gaussian processes without correlations. Trivial refers to predicting the empirical means and standard deviation of the test data. Errors indicate the central 95%-confidence interval. Numbers which are significantly best ($p < 0.05$) are boldfaced. Numbers which are very large are marked as failed with "F". Numbers which are missing could not be run.

| Weakly Periodic $d_x = 2$, $d_y = 1$ | Int. | OOID | Ext. | Weakly Periodic $d_x = 2$, $d_y = 2$ | Int. | OOID | Ext. |
|---|---|---|---|---|---|---|---|
| ConvCNP (AR) | **0.05** ±0.00 | **0.05** ±0.00 | **0.03** ±0.00 | ConvCNP (AR) | **0.08** ±0.00 | **0.08** ±0.00 | **0.05** ±0.00 |
| ConvGNP | 0.10 ±0.00 | 0.10 ±0.00 | 0.19 ±0.00 | ConvGNP | 0.13 ±0.00 | 0.13 ±0.00 | 0.18 ±0.00 |
| ConvNP (ML) | 0.18 ±0.00 | 0.18 ±0.00 | 0.21 ±0.00 | *Diagonal GP* | 0.20 ±0.00 | 0.20 ±0.00 | 0.28 ±0.00 |
| *Diagonal GP* | 0.19 ±0.00 | 0.19 ±0.00 | 0.27 ±0.00 | ConvNP (ML) | 0.21 ±0.00 | 0.21 ±0.00 | 0.25 ±0.00 |
| ConvCNP | 0.20 ±0.00 | 0.20 ±0.00 | 0.27 ±0.00 | ConvCNP | 0.23 ±0.00 | 0.23 ±0.00 | 0.28 ±0.00 |
| ConvNP (ELBO) | 0.22 ±0.00 | 0.22 ±0.00 | 0.22 ±0.00 | AGNP | 0.24 ±0.00 | 0.39 ±0.00 | 0.27 ±0.00 |
| AGNP | 0.23 ±0.00 | 0.35 ±0.00 | 0.26 ±0.00 | ACNP (AR) | 0.25 ±0.00 | 0.40 ±0.00 | 0.28 ±0.00 |
| ACNP (AR) | 0.23 ±0.00 | 0.37 ±0.00 | 0.26 ±0.00 | GNP | 0.25 ±0.00 | 0.38 ±0.00 | 0.25 ±0.00 |
| ConvNP (E.–M.) | 0.23 ±0.00 | 0.23 ±0.00 | 0.29 ±0.00 | ConvNP (ELBO) | 0.26 ±0.00 | 0.26 ±0.00 | 0.27 ±0.00 |
| GNP | 0.23 ±0.00 | 0.35 ±0.00 | 0.24 ±0.00 | CNP (AR) | 0.27 ±0.00 | 3.21 ±0.14 | 0.77 ±0.02 |
| CNP (AR) | 0.25 ±0.00 | 0.82 ±0.01 | 0.72 ±0.01 | ANP (ML) | 0.28 ±0.00 | 0.39 ±0.00 | 0.28 ±0.00 |
| ANP (ML) | 0.25 ±0.00 | 0.36 ±0.00 | 0.25 ±0.00 | NP (ML) | 0.29 ±0.00 | 0.42 ±0.00 | 0.31 ±0.00 |
| NP (ML) | 0.26 ±0.00 | 1.32 ±0.03 | 0.33 ±0.00 | NP (ELBO–ML) | 0.29 ±0.00 | 1.52 ±0.04 | 0.36 ±0.00 |
| NP (ELBO–ML) | 0.26 ±0.00 | 6.28 ±1.06 | 0.44 ±0.01 | NP (ELBO) | 0.30 ±0.00 | 1.60 ±0.05 | 0.39 ±0.00 |
| NP (ELBO) | 0.27 ±0.00 | 9.91 ±1.68 | 0.52 ±0.01 | ConvNP (E.–M.) | 0.30 ±0.00 | 0.30 ±0.00 | 0.34 ±0.00 |
| ACNP | 0.28 ±0.00 | 0.37 ±0.00 | 0.27 ±0.00 | ACNP | 0.30 ±0.00 | 0.40 ±0.00 | 0.29 ±0.00 |
| CNP | 0.29 ±0.00 | 0.81 ±0.01 | 0.29 ±0.00 | CNP | 0.31 ±0.00 | 0.63 ±0.01 | 0.33 ±0.00 |
| ANP (ELBO–ML) | 0.31 ±0.00 | 0.39 ±0.00 | 0.26 ±0.00 | ANP (ELBO–ML) | 0.36 ±0.00 | 0.43 ±0.00 | 0.28 ±0.00 |
| ANP (ELBO) | 0.32 ±0.00 | 0.42 ±0.00 | 0.29 ±0.00 | ANP (ELBO) | 0.36 ±0.00 | 0.44 ±0.00 | 0.29 ±0.00 |
| *Trivial* | 0.38 ±0.00 | 0.38 ±0.00 | 0.27 ±0.00 | *Trivial* | 0.40 ±0.00 | 0.40 ±0.00 | 0.28 ±0.00 |
| FullConvGNP | | | | FullConvGNP | | | |

Table 6.10: For the sawtooth synthetic experiments with one-dimensional inputs, average log-likelihoods normalised by the number of target points. Shows for one-dimensional outputs ($d_y = 1$) and two-dimensional outputs ($d_y = 2$) the performance for interpolation within the range $[-2, 2]$ where the models where trained ("Int."); interpolation within the range $[2, 6]$ which the models have never seen before ("OOID"); and extrapolation from the range $[-2, 2]$ to the range $[2, 6]$ ("Ext."). Models are ordered by interpolation performance. The latent variable models are trained and evaluated with the ELBO objective (ELBO); trained and evaluated with the ML objective (ML); and trained with the ELBO objective and evaluated with the ML objective (ELBO–ML; E.–M.). "Conv" is abbreviated with "Cv". Trivial refers to predicting the empirical means and standard deviation of the test data. Errors indicate the central 95%-confidence interval. Numbers which are significantly best ($p < 0.05$) are boldfaced. Numbers which are very large are marked as failed with "F". Numbers which are missing could not be run.

| Sawtooth $d_x = 1$, $d_y = 1$ | Int. | OOID | Ext. | Sawtooth $d_x = 1$, $d_y = 2$ | Int. | OOID | Ext. |
| --- | --- | --- | --- | --- | --- | --- | --- |
| CvNP (ELBO) | **3.71** ±0.01 | **3.71** ±0.01 | 2.79 ±0.02 | CvNP (ML) | **2.49** ±0.02 | **2.49** ±0.02 | 0.05 ±0.01 |
| CvGNP | **3.70** ±0.04 | **3.71** ±0.04 | 0.19 ±0.11 | CvCNP (AR) | 2.41 ±0.01 | 2.40 ±0.01 | **1.93** ±0.01 |
| CvCNP (AR) | 3.65 ±0.01 | 3.66 ±0.01 | **3.36** ±0.01 | CvNP (ELBO) | 2.32 ±0.01 | 2.32 ±0.01 | 1.65 ±0.01 |
| CvNP (ML) | 3.61 ±0.03 | 3.61 ±0.03 | 0.84 ±0.01 | FullCvGNP | 2.03 ±0.06 | 2.05 ±0.04 | 0.18 ±0.00 |
| FullCvGNP | 3.24 ±0.04 | 3.25 ±0.04 | 0.25 ±0.01 | CvCNP | 1.85 ±0.02 | 1.85 ±0.02 | −0.24 ±0.00 |
| CvCNP | 3.17 ±0.03 | 3.17 ±0.03 | 0.06 ±0.01 | CvGNP | 1.76 ±0.05 | 1.80 ±0.03 | −0.24 ±0.00 |
| ANP (E.–M.) | 1.01 ±0.02 | −0.18 ±0.00 | −0.30 ±0.01 | ACNP (AR) | 1.31 ±0.01 | −0.48 ±0.01 | −0.45 ±0.01 |
| ANP (ELBO) | 1.01 ±0.02 | F | −7.55 ±0.06 | CvNP (E.–M.) | 0.98 ±0.08 | 0.97 ±0.08 | −6.48 ±0.04 |
| NP (E.–M.) | 0.26 ±0.01 | F | F | ACNP | 0.55 ±0.02 | −0.44 ±0.01 | −0.50 ±0.01 |
| NP (ELBO) | 0.26 ±0.01 | F | F | ANP (E.–M.) | 0.54 ±0.02 | −0.70 ±0.01 | −0.36 ±0.00 |
| ANP (ML) | 0.20 ±0.00 | −0.18 ±0.00 | −0.18 ±0.00 | ANP (ELBO) | 0.53 ±0.02 | −0.73 ±0.01 | −0.37 ±0.00 |
| NP (ML) | −0.18 ±0.00 | −0.18 ±0.00 | −0.18 ±0.00 | AGNP | 0.47 ±0.01 | −0.38 ±0.01 | −0.70 ±0.08 |
| *Trivial* | −0.18 ±0.00 | −0.18 ±0.00 | −0.18 ±0.00 | GNP | −0.02 ±0.00 | F | F |
| ACNP (AR) | −0.18 ±0.00 | −0.18 ±0.00 | −0.18 ±0.00 | NP (E.–M.) | −0.08 ±0.01 | F | −1.37 ±0.00 |
| GNP | −0.18 ±0.00 | −0.18 ±0.00 | −0.18 ±0.00 | NP (ELBO) | −0.08 ±0.01 | F | −1.37 ±0.00 |
| ACNP | −0.18 ±0.00 | −0.18 ±0.00 | −0.18 ±0.00 | ANP (ML) | −0.21 ±0.00 | −0.33 ±0.00 | −0.33 ±0.00 |
| CNP | −0.18 ±0.00 | −0.18 ±0.00 | −0.18 ±0.00 | CNP (AR) | −0.29 ±0.00 | F | −2.40 ±0.13 |
| CNP (AR) | −0.18 ±0.00 | −0.18 ±0.00 | −0.18 ±0.00 | CNP | −0.29 ±0.00 | −3.97 ±0.25 | −0.36 ±0.00 |
| AGNP | −0.18 ±0.00 | −0.18 ±0.00 | −0.18 ±0.00 | *Trivial* | −0.32 ±0.00 | −0.32 ±0.00 | −0.32 ±0.00 |
| CvNP (E.–M.) | F | F | F | NP (ML) | −0.32 ±0.00 | −0.32 ±0.00 | −0.32 ±0.00 |

Table 6.11: For the sawtooth synthetic experiments with two-dimensional inputs, average log-likelihoods normalised by the number of target points. Shows for one-dimensional outputs ($d_y = 1$) and two-dimensional outputs ($d_y = 2$) the performance for interpolation within the range $[-2, 2]^2$ where the models where trained ("Int."); interpolation within the range $[2, 6]^2$ which the models have never seen before ("OOID"); and extrapolation from the range $[-2, 2]^2$ to the range $[2, 6]^2$ ("Ext."). Models are ordered by interpolation performance. The latent variable models are trained and evaluated with the ELBO objective (ELBO); trained and evaluated with the ML objective (ML); and trained with the ELBO objective and evaluated with the ML objective (ELBO–ML; E.–M.). "Conv" is abbreviated with "Cv". Trivial refers to predicting the empirical means and standard deviation of the test data. Errors indicate the central 95%-confidence interval. Numbers which are significantly best ($p < 0.05$) are boldfaced. Numbers which are very large are marked as failed with "F". Numbers which are missing could not be run.

| Sawtooth $d_x = 2, d_y = 1$ | Int. | OOID | Ext. | Sawtooth $d_x = 2, d_y = 2$ | Int. | OOID | Ext. |
|---|---|---|---|---|---|---|---|
| CvCNP (AR) | **2.59** ±0.01 | **2.59** ±0.01 | **2.10** ±0.01 | CvCNP (AR) | **0.38** ±0.00 | **0.38** ±0.00 | **0.18** ±0.00 |
| CvNP (ML) | 2.07 ±0.02 | 2.08 ±0.02 | −0.17 ±0.00 | CvNP (ML) | 0.31 ±0.01 | 0.31 ±0.01 | −0.32 ±0.00 |
| CvCNP | 1.93 ±0.04 | 1.94 ±0.03 | −0.18 ±0.00 | CvGNP | 0.26 ±0.01 | 0.26 ±0.01 | −0.33 ±0.00 |
| CvGNP | 1.90 ±0.04 | 1.91 ±0.03 | −0.18 ±0.00 | CvCNP | 0.12 ±0.01 | 0.12 ±0.01 | −0.32 ±0.00 |
| CvNP (ELBO) | 1.77 ±0.02 | 1.77 ±0.02 | 0.33 ±0.02 | CvNP (ELBO) | 0.04 ±0.00 | 0.04 ±0.00 | −0.30 ±0.00 |
| CvNP (E.–M.) | 1.71 ±0.04 | 1.72 ±0.04 | −2.30 ±0.87 | CvNP (E.–M.) | −0.07 ±0.01 | −0.07 ±0.01 | −0.48 ±0.00 |
| *Trivial* | −0.18 ±0.00 | −0.18 ±0.00 | −0.18 ±0.00 | *Trivial* | −0.32 ±0.00 | −0.32 ±0.00 | −0.32 ±0.00 |
| CNP (AR) | −0.18 ±0.00 | −0.18 ±0.00 | −0.18 ±0.00 | ANP (ML) | −0.32 ±0.00 | −0.32 ±0.00 | −0.32 ±0.00 |
| CNP | −0.18 ±0.00 | −0.18 ±0.00 | −0.18 ±0.00 | CNP (AR) | −0.32 ±0.00 | −0.32 ±0.00 | −0.32 ±0.00 |
| GNP | −0.18 ±0.00 | −0.18 ±0.00 | −0.18 ±0.00 | CNP | −0.32 ±0.00 | −0.32 ±0.00 | −0.32 ±0.00 |
| NP (ML) | −0.18 ±0.00 | −0.18 ±0.00 | −0.18 ±0.00 | NP (ML) | −0.32 ±0.00 | −0.32 ±0.00 | −0.32 ±0.00 |
| AGNP | −0.18 ±0.00 | −0.18 ±0.00 | −0.18 ±0.00 | ACNP | −0.32 ±0.00 | −0.32 ±0.00 | −0.32 ±0.00 |
| ANP (ML) | −0.18 ±0.00 | −0.18 ±0.00 | −0.18 ±0.00 | ACNP (AR) | −0.32 ±0.00 | −0.32 ±0.00 | −0.32 ±0.00 |
| ACNP (AR) | −0.18 ±0.00 | −0.18 ±0.00 | −0.18 ±0.00 | GNP | −0.32 ±0.00 | −0.32 ±0.00 | −0.32 ±0.00 |
| ACNP | −0.18 ±0.00 | −0.18 ±0.00 | −0.18 ±0.00 | AGNP | −0.32 ±0.00 | −0.32 ±0.00 | −0.32 ±0.00 |
| NP (E.–M.) | −0.19 ±0.00 | F | −0.86 ±0.02 | NP (E.–M.) | −0.33 ±0.00 | −0.54 ±0.00 | −0.34 ±0.00 |
| NP (ELBO) | −0.19 ±0.00 | F | F | NP (ELBO) | −0.33 ±0.00 | −0.54 ±0.00 | −0.34 ±0.00 |
| ANP (E.–M.) | −0.20 ±0.01 | −0.18 ±0.00 | −0.18 ±0.00 | ANP (E.–M.) | −0.36 ±0.00 | −0.33 ±0.00 | −0.33 ±0.00 |
| ANP (ELBO) | −0.20 ±0.01 | −0.71 ±0.00 | −0.33 ±0.00 | ANP (ELBO) | −0.36 ±0.00 | −0.33 ±0.00 | −0.33 ±0.00 |
| FullCvGNP | | | | FullCvGNP | | | |

Table 6.12: For the mixture synthetic experiments with one-dimensional inputs, average log-likelihoods normalised by the number of target points. Shows for one-dimensional outputs ($d_y = 1$) and two-dimensional outputs ($d_y = 2$) the performance for interpolation within the range $[-2, 2]$ where the models where trained ("Int."); interpolation within the range $[2, 6]$ which the models have never seen before ("OOID"); and extrapolation from the range $[-2, 2]$ to the range $[2, 6]$ ("Ext."). Models are ordered by interpolation performance. The latent variable models are trained and evaluated with the ELBO objective (ELBO); trained and evaluated with the ML objective (ML); and trained with the ELBO objective and evaluated with the ML objective (ELBO–ML; E.–M.). "Conv" is abbreviated with "Cv". Trivial refers to predicting the empirical means and standard deviation of the test data. Errors indicate the central 95%-confidence interval. Numbers which are significantly best ($p < 0.05$) are boldfaced. Numbers which are very large are marked as failed with "F". Numbers which are missing could not be run.

| Mixture $d_x=1$, $d_y=1$ | Int. | OOID | Ext. | Mixture $d_x=1$, $d_y=2$ | Int. | OOID | Ext. |
|---|---|---|---|---|---|---|---|
| CvCNP (AR)  | **0.53** ±0.04  | **0.53** ±0.04  | **0.31** ±0.04  | CvNP (ELBO) | **0.15** ±0.03  | **0.15** ±0.03  | −0.46 ±0.03 |
| FullCvGNP   | 0.30 ±0.04      | 0.30 ±0.04      | −0.44 ±0.01     | CvNP (ML)   | 0.02 ±0.03      | 0.02 ±0.03      | −1.06 ±0.02 |
| CvNP (ELBO) | 0.28 ±0.03      | 0.28 ±0.03      | −0.32 ±0.03     | CvGNP       | 0.01 ±0.02      | 0.01 ±0.02      | −2.45 ±0.06 |
| CvGNP       | 0.23 ±0.04      | 0.23 ±0.04      | −1.32 ±0.03     | CvCNP (AR)  | −0.10 ±0.01     | −0.10 ±0.01     | **−0.28** ±0.02 |
| CvNP (ML)   | 0.16 ±0.03      | 0.16 ±0.03      | −0.85 ±0.02     | FullCvGNP   | −0.13 ±0.01     | −0.13 ±0.01     | −0.59 ±0.01 |
| CvCNP       | 0.10 ±0.06      | 0.12 ±0.04      | −1.20 ±0.01     | ACNP (AR)   | −0.23 ±0.02     | −1.46 ±0.02     | −1.42 ±0.02 |
| ACNP (AR)   | −0.07 ±0.02     | −1.32 ±0.01     | −1.32 ±0.01     | CvCNP       | −0.34 ±0.02     | −0.34 ±0.02     | −1.33 ±0.01 |
| ANP (ML)    | −0.25 ±0.02     | −1.04 ±0.02     | −1.04 ±0.02     | ANP (ML)    | −0.43 ±0.02     | −1.21 ±0.02     | −1.19 ±0.02 |
| AGNP        | −0.26 ±0.02     | −0.95 ±0.02     | −2.70 ±0.07     | AGNP        | −0.51 ±0.01     | −3.23 ±0.07     | −3.00 ±0.06 |
| ANP (E.–M.) | −0.36 ±0.02     | −1.23 ±0.01     | −1.11 ±0.02     | ACNP        | −0.58 ±0.02     | −1.43 ±0.02     | −1.53 ±0.03 |
| ANP (ELBO)  | −0.39 ±0.02     | −1.56 ±0.02     | −2.18 ±0.03     | GNP         | −0.59 ±0.01     | −1.68 ±0.03     | −1.63 ±0.02 |
| CvNP (E.–M.)| −0.45 ±0.11     | −0.43 ±0.10     | −3.39 ±0.06     | ANP (E.–M.) | −0.66 ±0.02     | −1.76 ±0.04     | −1.74 ±0.04 |
| ACNP        | −0.47 ±0.02     | −1.31 ±0.01     | −1.25 ±0.01     | ANP (ELBO)  | −0.69 ±0.02     | −2.14 ±0.05     | −2.06 ±0.05 |
| NP (E.–M.)  | −0.56 ±0.01     | F               | −1.79 ±0.01     | NP (E.–M.)  | −0.77 ±0.01     | F               | −1.65 ±0.02 |
| NP (ELBO)   | −0.60 ±0.01     | F               | −1.99 ±0.01     | NP (ELBO)   | −0.79 ±0.01     | F               | −1.78 ±0.01 |
| NP (ML)     | −0.64 ±0.01     | −1.64 ±0.01     | −1.32 ±0.02     | NP (ML)     | −0.80 ±0.01     | −9.46 ±0.39     | −1.41 ±0.02 |
| GNP         | −0.69 ±0.02     | −2.38 ±0.06     | −1.56 ±0.04     | CNP (AR)    | −1.15 ±0.02     | −1.28 ±0.02     | −1.22 ±0.02 |
| CNP (AR)    | −1.00 ±0.02     | −1.18 ±0.02     | −1.10 ±0.02     | CNP         | −1.16 ±0.02     | −1.29 ±0.03     | −1.29 ±0.02 |
| CNP         | −1.04 ±0.02     | −1.16 ±0.02     | −1.17 ±0.02     | CvNP (E.–M.)| −1.29 ±0.07     | −1.31 ±0.07     | −5.28 ±0.07 |
| *Trivial*   | −1.32 ±0.00     | −1.32 ±0.00     | −1.32 ±0.00     | *Trivial*   | −1.46 ±0.00     | −1.46 ±0.00     | −1.46 ±0.00 |

Table 6.13: For the mixture synthetic experiments with two-dimensional inputs, average log-likelihoods normalised by the number of target points. Shows for one-dimensional outputs ($d_y = 1$) and two-dimensional outputs ($d_y = 2$) the performance for interpolation within the range $[-2, 2]^2$ where the models where trained ("Int."); interpolation within the range $[2, 6]^2$ which the models have never seen before ("OOID"); and extrapolation from the range $[-2, 2]^2$ to the range $[2, 6]^2$ ("Ext."). Models are ordered by interpolation performance. The latent variable models are trained and evaluated with the ELBO objective (ELBO); trained and evaluated with the ML objective (ML); and trained with the ELBO objective and evaluated with the ML objective (ELBO–ML; E.–M.). "Conv" is abbreviated with "Cv". Trivial refers to predicting the empirical means and standard deviation of the test data. Errors indicate the central 95%-confidence interval. Numbers which are significantly best ($p < 0.05$) are boldfaced. Numbers which are very large are marked as failed with "F". Numbers which are missing could not be run.

| Mixture $d_x = 2$, $d_y = 1$ | Int. | OOID | Ext. | Mixture $d_x = 2$, $d_y = 2$ | Int. | OOID | Ext. |
|---|---|---|---|---|---|---|---|
| CvCNP (AR) | **−0.10** ±0.03 | **−0.10** ±0.03 | **−0.34** ±0.03 | CvCNP (AR) | **−0.62** ±0.01 | **−0.62** ±0.01 | **−0.79** ±0.01 |
| CvCNP | −0.49 ±0.03 | −0.49 ±0.03 | −1.45 ±0.02 | CvGNP | −0.74 ±0.01 | −0.74 ±0.01 | −1.43 ±0.02 |
| CvGNP | −0.50 ±0.02 | −0.50 ±0.02 | −1.24 ±0.02 | CvNP (ML) | −0.78 ±0.02 | −0.79 ±0.02 | −1.25 ±0.02 |
| CvNP (ML) | −0.57 ±0.02 | −0.57 ±0.02 | −1.07 ±0.02 | CvCNP | −0.85 ±0.01 | −0.85 ±0.01 | −1.50 ±0.02 |
| CvNP (ELBO) | −0.63 ±0.02 | −0.63 ±0.02 | −1.08 ±0.02 | ACNP (AR) | −0.85 ±0.01 | −4.30 ±0.09 | −4.24 ±0.09 |
| ANP (ML) | −0.73 ±0.02 | −1.04 ±0.02 | −1.05 ±0.02 | ANP (ML) | −0.88 ±0.01 | −1.41 ±0.03 | −1.21 ±0.02 |
| CvNP (E.–M.) | −0.76 ±0.02 | −0.76 ±0.02 | −1.37 ±0.02 | CvNP (ELBO) | −0.92 ±0.01 | −0.92 ±0.01 | −1.41 ±0.02 |
| ACNP (AR) | −0.77 ±0.01 | −1.28 ±0.01 | −1.30 ±0.01 | AGNP | −0.93 ±0.01 | −1.34 ±0.01 | −1.21 ±0.02 |
| AGNP | −0.78 ±0.01 | −1.32 ±0.02 | −1.32 ±0.02 | ACNP | −0.99 ±0.02 | −4.19 ±0.10 | −1.27 ±0.02 |
| ACNP | −0.91 ±0.02 | −1.29 ±0.01 | −1.17 ±0.02 | CvNP (E.–M.) | −1.05 ±0.01 | −1.05 ±0.01 | −1.73 ±0.03 |
| NP (ML) | −0.91 ±0.02 | −1.44 ±0.04 | −1.19 ±0.02 | NP (ML) | −1.07 ±0.01 | −4.04 ±0.11 | −1.31 ±0.02 |
| NP (E.–M.) | −0.92 ±0.02 | −1.51 ±0.02 | −1.38 ±0.02 | ANP (E.–M.) | −1.11 ±0.02 | −2.65 ±0.62 | −1.23 ±0.02 |
| NP (ELBO) | −0.93 ±0.02 | −1.74 ±0.03 | −1.44 ±0.02 | ANP (ELBO) | −1.12 ±0.02 | −8.22 ±7.64 | −1.26 ±0.02 |
| ANP (E.–M.) | −1.00 ±0.02 | −1.07 ±0.02 | −1.08 ±0.02 | GNP | −1.20 ±0.02 | −1.39 ±0.03 | −1.21 ±0.02 |
| ANP (ELBO) | −1.03 ±0.02 | −1.08 ±0.02 | −1.09 ±0.02 | CNP (AR) | −1.20 ±0.02 | −2.61 ±0.07 | −1.32 ±0.02 |
| CNP (AR) | −1.06 ±0.02 | −1.07 ±0.02 | −1.08 ±0.02 | CNP | −1.20 ±0.02 | −2.60 ±0.08 | −1.23 ±0.02 |
| GNP | −1.06 ±0.02 | −1.09 ±0.02 | −1.08 ±0.02 | NP (E.–M.) | −1.20 ±0.02 | −3.23 ±0.10 | −1.24 ±0.02 |
| CNP | −1.07 ±0.02 | −1.09 ±0.02 | −1.10 ±0.02 | NP (ELBO) | −1.20 ±0.02 | −3.25 ±0.10 | −1.24 ±0.02 |
| *Trivial* | −1.32 ±0.00 | −1.32 ±0.00 | −1.32 ±0.00 | *Trivial* | −1.46 ±0.00 | −1.46 ±0.00 | −1.46 ±0.00 |
| FullCvGNP | | | | FullCvGNP | | | |

of $\pi_f$, which is equal to the CNPA, is listed as "*Diagonal GP*" in Tables 6.2 and 6.4 to 6.9.

Table 6.2 summarises the results of the Gaussian experiments. For the task of interpolation, the ConvCNP is, for all six summary statistics, 0.00–0.01 away from the CNPA. This means that the ConvCNP is able to consistently recover the CNPA. In contrast, the ACNP comes close, but is not quite able to converge onto the CNPA; and the CNP is not able to come close at all. For OOID interpolation, the performance of the ACNP and CNP deteriorates. The ConvCNP, on the other hand, performs equally well in the interpolation and OOID interpolation tasks. This clearly demonstrates the benefits of translation equivariance, which allows ConvCNP to generalise to unseen parts of the input space.

Amongst the Gaussian experiments, the weakly periodic process appears to be the more challenging one. For the EQ and weakly periodic process, Figure 6.2 compares predictions from the CNP, ACNP, and ConvCNP, showing that only the ConvCNP consistently recovers the CNPA and only the ConvCNP successfully generalises beyond the training range.

**Performance of GNPs in Gaussian experiments.** Since the ground-truth stochastic process $f$ is Gaussian, the Gaussian neural process approximation (GNPA; Definition 3.25) is equal to the posterior prediction map $\pi_f$. Therefore, in the limit of infinite data, GNPs should have the capacity the recover the posterior prediction map, the best possible outcome. Table 6.2 shows that the FullConvGNP consistently recovers the posterior prediction map $\pi_f$ for one-dimensional inputs. For two-dimensional inputs, the FullConvGNP is computationally too expensive, unfortunately, so it cannot be run (Section 5.7). In terms of interpolation performance, the ConvGNP performs second best, followed by the AGNP, and the GNP comes last. Note, however, that no GNP is able to consistently recover the posterior prediction map in all cases.

In Table 6.2, all GNPs perform better than the CNPA, which means that all GNPs successfully exploit correlations to achieve performances unachievable by models that do not model correlations. In comparison, the LNPs struggle to outperform the CNPA. Comparing GNPs and LNPs in the more detailed results, Tables 6.4 to 6.9 show that GNPs consistently perform better than their latent-variable counterparts, demonstrating the benefits of GNPs when the ground truth really is Gaussian. In the most difficult Gaussian experiment, the weakly periodic process with two-dimensional inputs and outputs, Table 6.9 shows that the ConvGNP is the only GNP to consistently outperform the CNPA; and no other GNP and no LNP is able to outperform the CNPA. In this most difficult case, however, even though ConvGNP performs best amongst all GNPs and LNPs, the ConvGNP is not able to come close to the posterior prediction map.

In terms of generalisation performance, Table 6.2 shows that the ConvGNP and FullCon-vGNP perform equally well in the interpolation and OOID interpolation tasks, whereas the performance of the GNP and AGNP deteriorate for OOID interpolation. This behaviour is expected: the ConvGNP and FullConvGNP have a built-in mechanism to generalise to unseen parts of the input space, but the GNP and AGNP do not.

However, in terms of generalisation performance, there is one important anomaly: the ConvGNP performs well on OOID interpolation, but the model's performance for extrapolation is consistently poor. This shows in Table 6.2 as well as in the more detailed results, Tables 6.4 to 6.9. What is going wrong is that the ConvGNP implements the eigenmap (Definition 5.8) with the convolutional deep set from Theorem 4.8, and the output of this convolutional deep set becomes constant when it is far away from the context inputs. This problem is exactly what the FullConvGNP's additional source channel solves, so this problem is not present for the FullConvGNP. We might hope to fix this flaw of the ConvGNP by also incorporating a source channel, but it is not clear what a source channel would look like for the ConvGNP. On a deeper level, this flaw of the ConvGNP is a consequence of that a DTE kernel map might not have a TE eigenmap. See Section 5.5 for a more detailed discussion of problems with the representational capacity of the ConvGNP. We again emphasise that these issues—theoretical in Section 5.5, but now practical here—underline the importance of entirely basing neural process architectures on representation theorems.

For the EQ data process and weakly-periodic data process, Figure 6.3 compares predictions of LNPs and GNPs.

**Performance of AR CNPs in Gaussian experiments.** Having analysed the performance of CNPs and GNPs, we add AR CNPs to the mix. In terms of interpolation performance, Table 6.2 shows that the AR CNP performs better than the CNP; that the AR ACNP performs better than the ACNP; and that the AR ConvCNP performs better than the ConvCNP. In fact, whereas the ACNP and ConvCNP perform no better and cannot perform better than the CNPA, the AR ACNP and AR ConvCNP significantly outperform the CNPA. This demonstrates that rolling out a CNP autoregressively can improve performance by successfully modelling dependencies between target outputs.

For Gaussian data, an important question is which of the classes CNPs, GNPs, LNPs, and AR CNPs should preferably be used. For the deep-set-based and attentive families, Table 6.2 shows that the GNPs perform best overall. For the convolutional family, however, the AR ConvCNP performs best. Amongst all neural process models, for one-dimensional inputs, the FullConvGNP performs best, with the AR ConvCNP a close second; and for two-dimensional inputs, the AR ConvCNP performs best. For two-dimensional inputs, the AR ConvCNP is the only model able to recover the posterior prediction map, an impressive achievement.

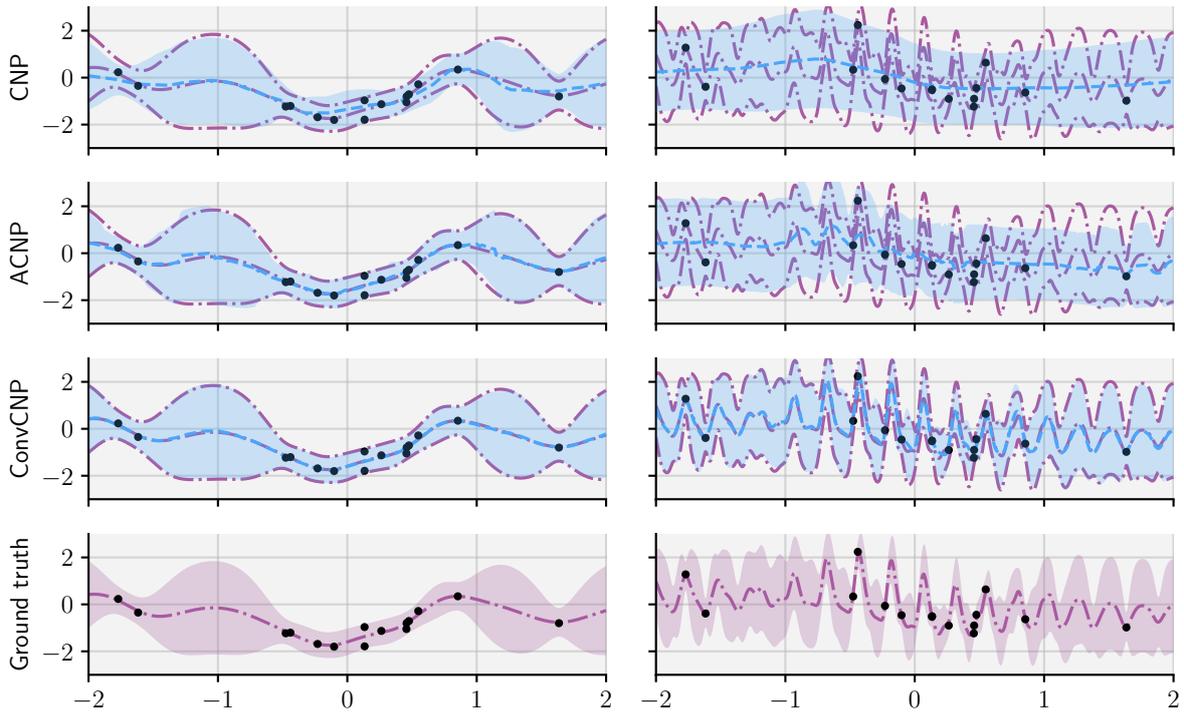

(a) Predictions and observations inside the training range

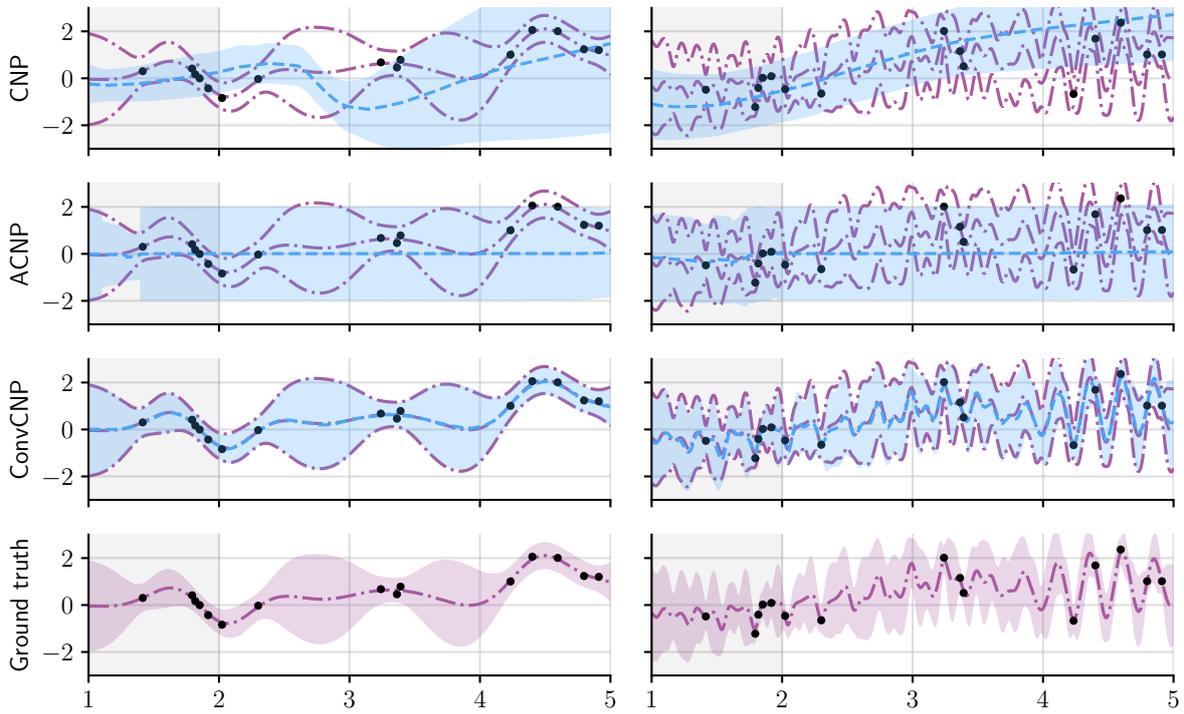

(b) Predictions and observations partly inside and partly outside the training range

Figure 6.2: Predictions by CNPs in the Gaussian synthetic experiments. *Left column:* data sampled from the EQ data process. *Right column:* data sampled from the weakly-periodic data process. The training range is shaded grey. Shows predictions by the models in dashed blue and predictions by the ground truth in dot-dashed purple. Filled regions are central 95%-credible regions.

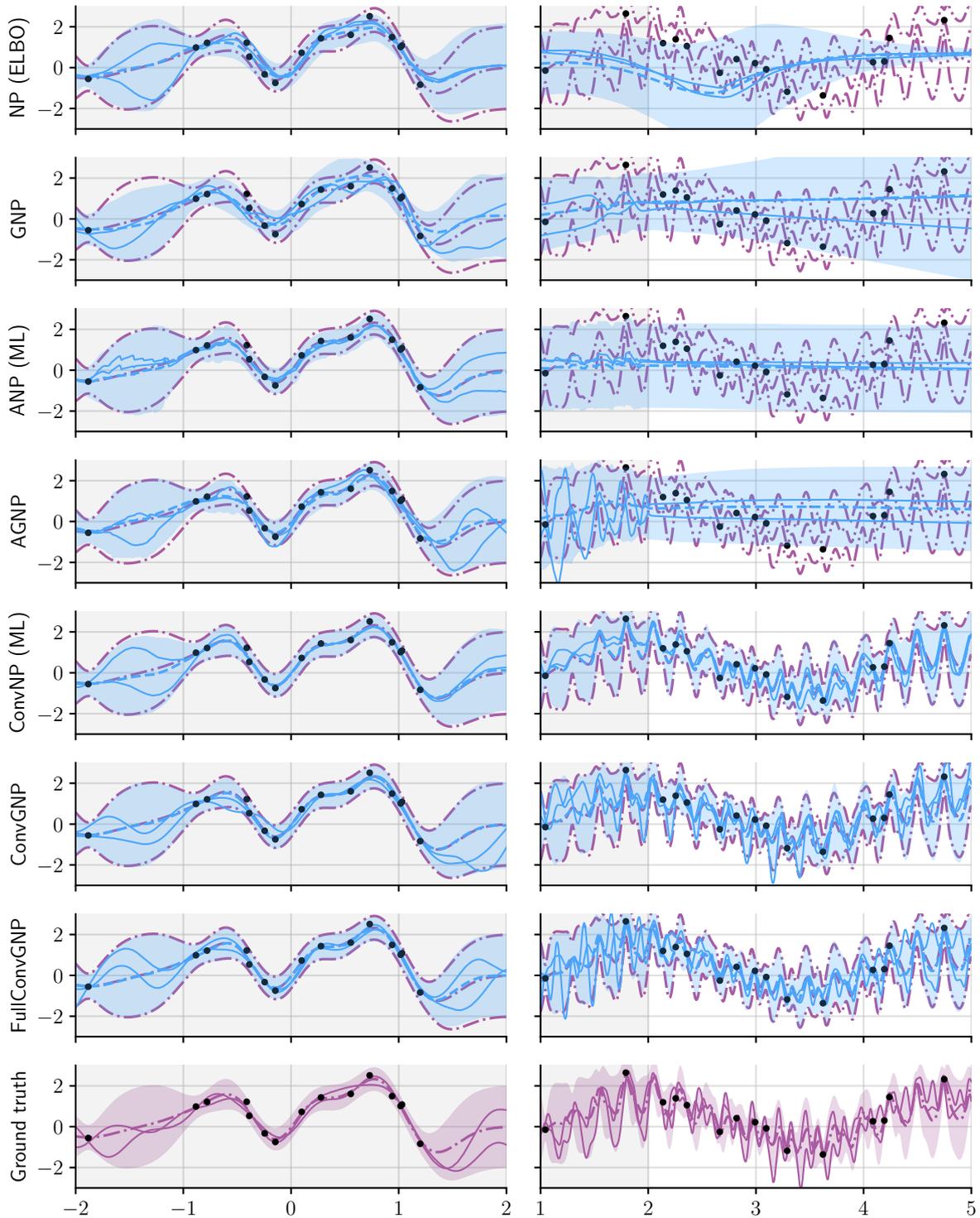

Figure 6.3: Predictions by LNPs and GNPs in the Gaussian synthetic experiments. *Left column:* data sampled from the EQ data process. *Right column:* data sampled from the weakly-periodic data process. The training range is shaded grey. Shows predictions by the models in dashed blue and predictions by the ground truth in dot-dashed purple. Filled regions are central 95%-credible regions.

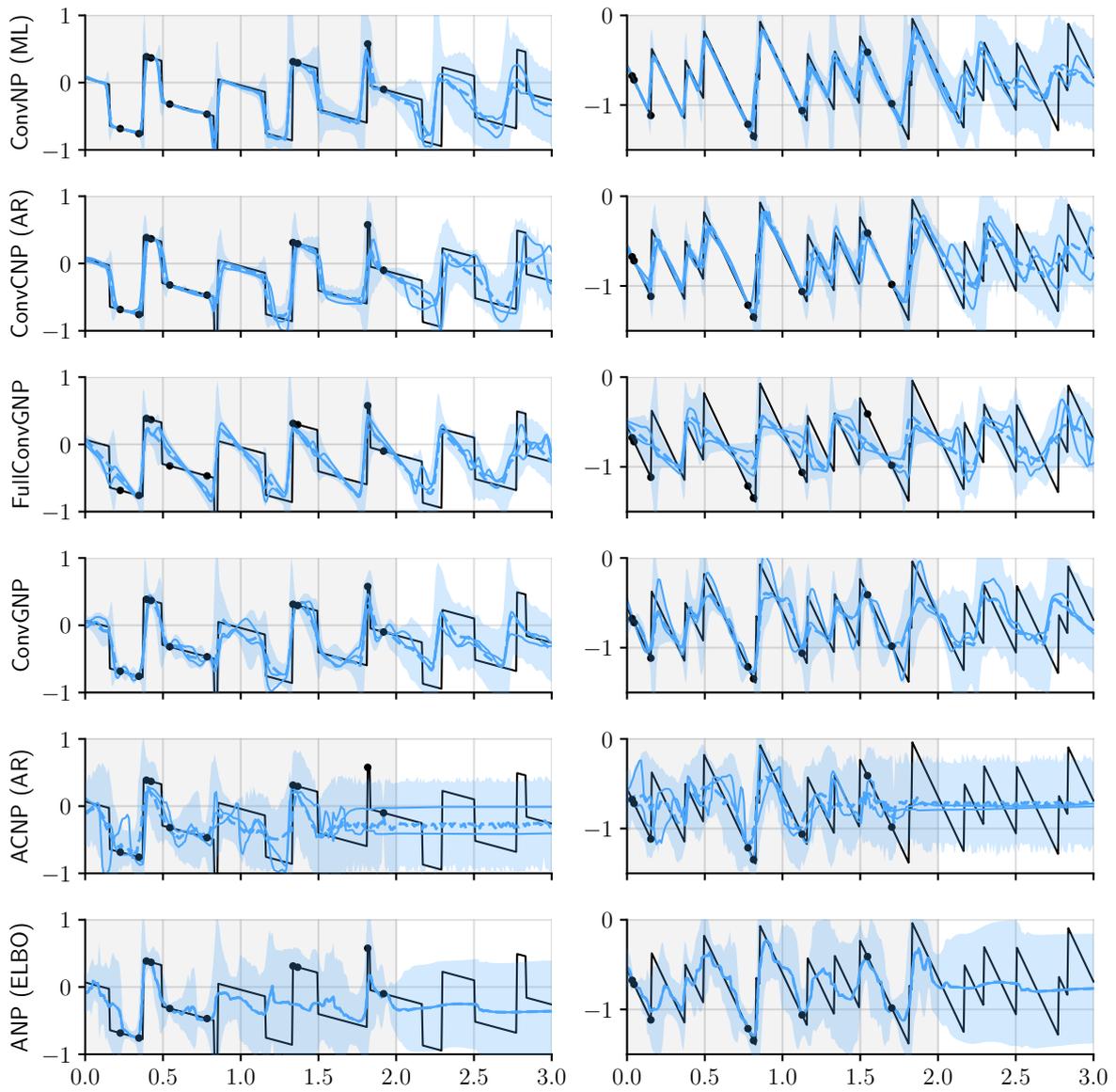

Figure 6.4: Predictions by the best-performing models in the sawtooth synthetic experiment with one-dimensional inputs and two-dimensional outputs (Table 6.10). For both outputs, the models observe 20 data points randomly sampled from $[-2, 2]$; the plots only show the observations on $[0, 2]$. *Left column:* first output of the sawtooth process. *Right column:* second output of the sawtooth process. The training range is shaded grey. Filled regions are central 95%-credible regions.

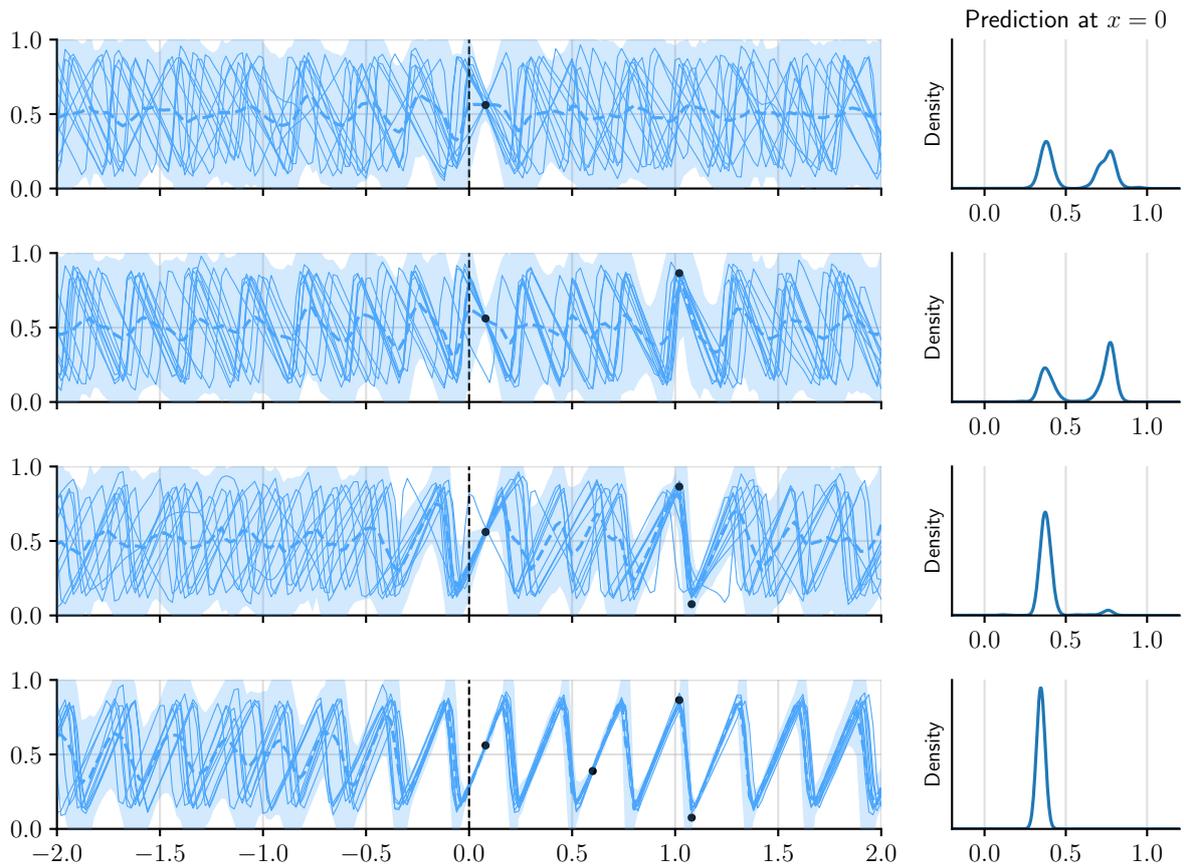

Figure 6.5: Multi-modality of predictions by the AR ConvCNP. Shows four observations sampled from the sawtooth process. In the four rows, these four observations are revealed one data point at a time. Every row also shows a kernel density estimate of the prediction at $x = 0$. Note that the prediction is bimodal for one and two observations, and collapses to a single mode upon observing the third observation. Filled regions are central 95%-credible regions.

In these synthetic experiments, the number of target points is at most 200. It might be that this number is low enough for the inconsistency of AR CNPs (Section 5.6) to not be a problem. It is possible that, if this number is increased, the performance of AR CNPs deteriorates, whereas the performance of CNPs, GNPs, and LNPs should remain the same. We do not perform evaluation with more than 200 target points, because, at 200 target points, a single evaluation of an AR CNP already takes around two hours in the worst case.

**Performance in non-Gaussian experiments.** Whereas the FullConvGNP dominated the leaderboard in the Gaussian experiments, in the non-Gaussian experiments, the models' relative performances are different. Table 6.3 summarises the non-Gaussian experiments. Note that Table 6.3 shows log-likelihoods rather than Kullback–Leibler divergences: the posterior prediction map is intractable for non-Gaussian ground-truth stochastic processes.

To begin with, for one-dimensional inputs, the top position is now shared by the AR ConvCNP and ConvNP. Notably, the AR ConvCNP and ConvNP are both non-Gaussian models. Previously, in the Gaussian experiments, the best performing GNPs, ConvGNP and FullConvGNP, outperformed all LNPs. Now, in the non-Gaussian experiments, the best performing LNP, the ConvNP, outperforms all GNPs. This shows that the Gaussianity assumption of GNPs helps when the ground truth really is Gaussian, but hurts when the data is non-Gaussian. For two-dimensional inputs, the AR ConvCNP strongly outperforms all other models.

For the sawtooth data with one-dimensional inputs and outputs, Table 6.10 shows that the non-convolutional CNPs and GNPs completely failed to capture any structure of the data. Only the non-convolutional neural processes using a latent variable were able to achieve non-trivial performance. Remarkably, when the output dimension is then increased to two, all models except the CNP achieve non-trivial performance. This could be due to an increase in number of target points, or perhaps due to increased model capacity; namely, the architectures of non-convolutional CNPs and GNPs are slightly different for one and two-dimensional outputs (Appendix E.1). For the sawtooth data with two-dimensional inputs, Table 6.11 shows that only the convolutional neural processes achieves non-trivial performance. All non-convolutional neural processes were unsuccessful in fitting any non-trivial aspect of the data.

Compared to the Gaussian experiments, the results of the non-Gaussian experiments are generally more variable. For example, whereas the AR ConvCNP is amongst the best in most non-Gaussian experiments, Table 6.12 shows that it ranks fourth in on the mixture data with one-dimensional inputs and two-dimensional outputs, ranking even behind the ConvGNP. This variability could be due to imperfect training, due to the fact that we are comparing log-likelihoods rather than Kullback–Leibler divergences, or even due to increased variability between the non-Gaussian data processes.

Finally, we remark that the AR ConvCNP exhibits unparalleled extrapolation performance. For example, for two-dimensional inputs, Table 6.3 the AR ConvCNP achieves log-likelihood $0.29$ on average, whereas the second best, the ConvNP, achieves $-0.62$ on average, a difference of $0.91$! We hypothesise that this exceptional performance is due to the AR ConvCNP's ability to "renew" the receptive field, which we explain now. The non-autoregressive convolutional models are all limited by the receptive field of the CNN architecture (Definition 5.6): due to the finite size of the convolutional filters, signals deriving from the context set carry only limitedly far (Section 5.4). This limits the model's ability to extrapolate. When the ConvCNP is sampled, however, and this sample is fed back into the architecture (Procedure 5.14), this produces a new signal starting at the input of that sample. Hence, at every autoregressive sampling step, the AR ConvCNP effectively "renews" the receptive field.

Figure 6.4 compares predictions by the best-performing models in the sawtooth synthetic experiment with one-dimensional inputs ($d_x = 1$) and two-dimensional outputs ($d_y = 2$); see Table 6.10. Moreover, Figure 6.5 illustrates the that, although the predictions of the ConvCNP are Gaussian, the AR ConvCNP can produce appropriate multi-modal predictions.

## 6.3 Sim-to-Real Transfer with the Lotka–Volterra Equations

In this second experiment, we investigate the ability of neural processes to perform sim-to-real transfer. The goal of sim-to-real transfer is to use a data simulator to make predictions for real data. There are no requirements on the data simulator: it can be arbitrarily complicated and involve niche expert knowledge. Because of the complexity of the data simulator, it usually is not possible to directly use the simulator to make predictions. The strategy of sim-to-real transfer is to train a model on data generated with the data simulator. After the model is trained on this synthetic source of data, the model is applied to real data. In this way, the model facilitates a mechanism in which complex data simulators can be used to make predictions.

Our goal will be to make predictions for the famous hare–lynx data set. The hare–lynx data set is a time series from 1845 to 1935 recording yearly population counts of a population of Snowshoe hares and a population of Canadian lynx (MacLulich, 1937). A digital version extracted from the original graph by MacLulich (1937) is available at `http://people.whitman.edu/~hundledr/courses/M250F03/LynxHare.txt` (Hundley, 2022). Hundley (2022), the author of this digital source, says that other authors caution that the hare–lynx data is actually

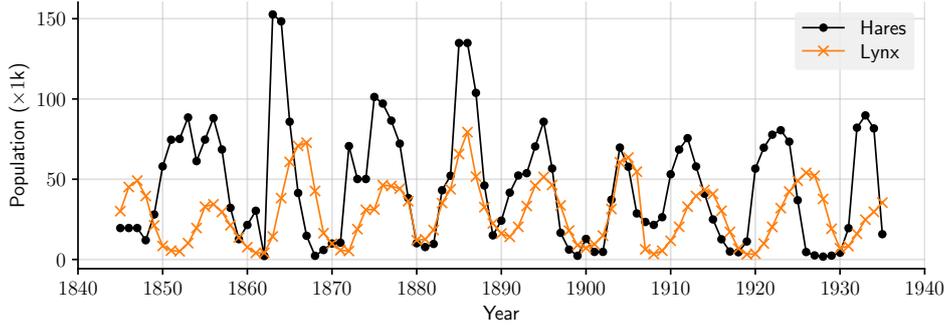

(a) Illustration of the hare–lynx data set

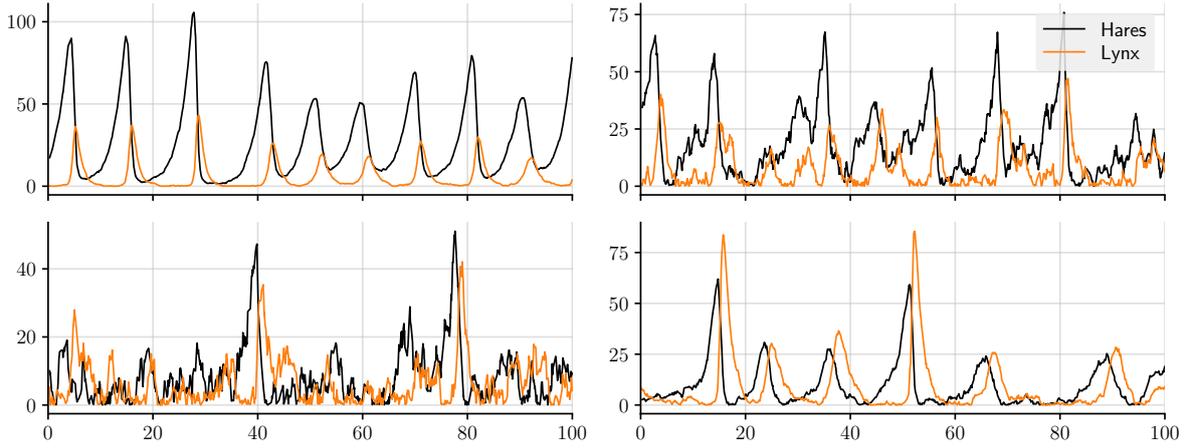

(b) Four samples from the proposed stochastic version of the Lotka–Volterra equations (6.8) and (6.9). The parameters of (6.8) and (6.9) are sampled according to Table 6.14.

Figure 6.6: Hare–lynx data set and proposed stochastic simulator

a composition of multiple time series, and presents the data with that caution. We, therefore, also present the data with this caution. Figure 6.6a visualises the hare–lynx data set.

To make predictions for the hare–lynx data set, we will use the Lotka–Volterra equations (Lotka, 1910; Volterra, 1926), also called the predator–prey equations. The Lotka–Volterra equations are an idealised mathematical model for the population counts of a prey population and a predator population:

$$\text{prey population:} \qquad x'(t) = \alpha x(t) - \beta x(t)y(t), \tag{6.6}$$

$$\text{predator population:} \qquad y'(t) = -\delta y(t) + \gamma x(t)y(t). \tag{6.7}$$

These differential equations say that the prey population naturally grows exponentially with rate $\alpha$, and that the predator population naturally decays exponentially with rate $\delta$. In addition, the predators hunt the prey. The resulting additional growth in the predator population and the resulting additional decrease in the prey population are both proportional to the product of the densities. In this idealised mathematical form, the population counts converge to a smooth, noiseless limit cycle and then perfectly track this limit cycle ever after. This is unlike real-world predator–prey population counts, which exhibit noisy behaviour and imperfect cycles. We therefore consider a stochastic version of the Lotka–Volterra equations, given by the following two coupled stochastic differential equations:

$$\mathrm{d}X_t = \alpha X_t \, \mathrm{d}t - \beta Y_t X_t \, \mathrm{d}t + \sigma X_t^\nu \, \mathrm{d}W_t^{(1)}, \tag{6.8}$$

$$\mathrm{d}Y_t = -\gamma X_t \, \mathrm{d}t + \delta Y_t X_t \, \mathrm{d}t + \sigma Y_t^\nu \, \mathrm{d}W_t^{(2)} \tag{6.9}$$

where $W^{(1)}$ and $W^{(2)}$ are two independent Brownian motions. Compared to the Lotka–Volterra equations, (6.8) and (6.9) have two additional terms, $\sigma X_t^\nu \, \mathrm{d}W_t^{(1)}$ and $\sigma Y_t^\nu \, \mathrm{d}W_t^{(2)}$, which introduce noisy behaviour. In these terms, multiplying by $X_t^\nu$ and $Y_t^\nu$ makes the noise go to zero when $X_t$ and $Y_t$ become small, ensuring that $X_t$ and $Y_t$ remain positive. In addition, we multiply by a parameter $\sigma > 0$ to control the magnitude of the noise, and we raise $X_t$ and $Y_t$ to a power $\nu > 0$ to control how quickly the noise grows as $X_t$ and $Y_t$ grow. Namely, $X_t$ naturally grows exponentially, so, by adding noise of magnitude proportional to $X_t$, we risk large spikes in the prey population. To moderate this behaviour, we choose $\nu$ to be strictly less than one. In particular, after simulating from (6.8) and (6.9) a few times, we settle on $\nu = \frac{1}{6}$. For the remainder of the parameters, we simply manually play around with (6.8) and (6.9), settle on parameter ranges that look reasonable, and randomly sample parameters from those intervals. Table 6.14 summarises the sampling distributions for all parameters of (6.8) and (6.9). Figure 6.6b shows four samples from the proposed stochastic model.

Table 6.14: Sampling distributions for the parameters of the stochastic version of the Lotka–Volterra equations (6.8) and (6.9). These equations are simulated on a dense grid spanning $[-10, 100]$. The table also shows the distribution for the initial conditions at $t = -10$. To not depend too heavily on these initial conditions, the simulation results on $[-10, 0]$ are discarded.

| Parameter | Distribution |
| --- | --- |
| Initial condition $X_{-10}$ | $\text{Unif}([5, 100])$ |
| Initial condition $Y_{-10}$ | $\text{Unif}([5, 100])$ |
| $\alpha$ | $\text{Unif}([0.2, 0.8])$ |
| $\beta$ | $\text{Unif}([0.04, 0.08])$ |
| $\gamma$ | $\text{Unif}([0.8, 1.2])$ |
| $\delta$ | $\text{Unif}([0.04, 0.08])$ |
| $\nu$ | Fixed to $1/6$ |
| $\sigma$ | $\text{Unif}([0.5, 10])$ |

We have constructed a simulator that can generate data which looks roughly like the hare–lynx data. Following the sim-to-real paradigm, we will generate a meta–data set with this simulator, train a neural process on the meta–data set, and finally apply the neural process to the real hare–lynx data.

To generate a meta–data set, we simulate (6.8) and (6.9) on a dense grid spanning 110 years, throw away the first 10 years, and retain between 150 and 250 data points for $X_t$ and $Y_t$. The numbers of data points and the locations of the data points are sampled separately for $X_t$ and $Y_t$. Hence, whereas the hare–lynx data is regularly spaced and the populations are always simultaneously observed, our simulator generates data at arbitrary and nonsimultaneous points in time. We split these data sets into context and target sets in three different ways. First, for $X_t$ and $Y_t$ separately, randomly designate 100 points to be the target points and let the remainder be the context points. This task is called *interpolation* and is the primary measure of performance. Second, select a random time between 25 years and 75 years, let everything before that time be the context set, and let everything after that time be the target set. This task is called *forecasting* and measures the model's ability extrapolate the cyclic behaviour. Third, randomly choose $X_t$ or $Y_t$ and, for the choice, split the data in two exactly like for forecasting. For $X_t$ or $Y_t$ that was not chosen, append all data to the context set. This task is called *reconstruction* and measures the model's ability to infer one population count from the other. To train the models, for every batch, we randomly choose one of these tasks by rolling a three-sided die. We will also perform these tasks on the real hare–lynx data; in that case, for interpolation, we let the number of target points per output be between one and fifteen. The tasks on simulated and real data are similar, but slightly differ in the number of context and target points.

We compare the strongest newly proposed models against the strongest existing baselines. Based on Section 6.2, we choose to compare the ConvCNP, ConvGNP, FullConvGNP, and AR ConvCNP against the ACNP, ANP, ConvNP, and AR ACNP. To deal with the positivity of population counts, we transform the marginals of all models to distributions on $(0, \infty)$ by pushing the marginals through $x \mapsto \log(1 + x)$. Appendix E.1 describes the general architectures of the models, Appendix E.2 describes the general training, cross-validation, and evaluation protocols, and Appendix E.4 describes details specific to this experiment.

**Results.** For simulated data, Table 6.15 shows the results for interpolation, forecasting, and reconstruction, and the left column of Figure 6.7 illustrates predictions by the models. The FullConvGNP and AR ConvCNP take a shared top position in the results. For interpolation, the ConvNP, ConvGNP, and ConvCNP come a close second; and for reconstruction, the AR ConvCNP outperforms the FullConvGNP. Note that, for interpolation, the ConvGNP fails to outperform the ConvCNP, which indicates that ConvGNP was unable to successfully exploit dependencies between target outputs in this task. Also note that the AR ConvCNP outperforms the ConvCNP, ConvGNP, and the ConvNP; and that the AR ACNP outperforms the ACNP and the ANP. This again demonstrates that running CNPs in AR mode can improve performance and can even outperform strong GNPs and LNPs.

Table 6.15: Normalised log-likelihoods in the predator–prey experiments. Shows the performance for interpolation ("Int."), forecasting ("For."), and reconstruction ("Rec.") on simulated ("S") and real ("R") data. Models are ordered by interpolation performance on simulated data. See Section 6.3 for a more detailed description. The latent variable models are trained and evaluated with the ELBO objective (ELBO); trained and evaluated with the ML objective (ML); and trained with the ELBO objective and evaluated with the ML objective (ELBO–ML; E.–M.). Errors indicate the central 95%-confidence interval. Numbers which are significantly best ($p < 0.05$) are boldfaced. Numbers which are very large are marked as failed with "F". Numbers which are missing could not be run.

| Model | Int. (S) | For. (S) | Rec. (S) | Int. (R) | For. (R) | Rec. (R) |
| --- | --- | --- | --- | --- | --- | --- |
| FullConvGNP | $-\mathbf{2.23}_{\pm 0.02}$ | $-\mathbf{2.39}_{\pm 0.02}$ | $-2.68_{\pm 0.02}$ | $-\mathbf{4.56}_{\pm 0.09}$ | $-5.61_{\pm 0.24}$ | $-\mathbf{4.35}_{\pm 0.03}$ |
| ConvCNP (AR) | $-\mathbf{2.23}_{\pm 0.02}$ | $-\mathbf{2.38}_{\pm 0.02}$ | $-\mathbf{2.50}_{\pm 0.02}$ | $-4.69_{\pm 0.11}$ | $-4.80_{\pm 0.11}$ | $-4.47_{\pm 0.06}$ |
| ConvNP (ML) | $-2.35_{\pm 0.02}$ | $-2.76_{\pm 0.02}$ | $-3.32_{\pm 0.01}$ | $-4.73_{\pm 0.10}$ | $-5.18_{\pm 0.15}$ | $-4.47_{\pm 0.01}$ |
| ConvGNP | $-2.39_{\pm 0.02}$ | $-2.57_{\pm 0.02}$ | $-3.03_{\pm 0.02}$ | $-5.26_{\pm 0.17}$ | $-7.63_{\pm 0.51}$ | $-4.63_{\pm 0.04}$ |
| ConvCNP | $-2.39_{\pm 0.02}$ | $-2.94_{\pm 0.02}$ | $-3.70_{\pm 0.01}$ | $-4.84_{\pm 0.13}$ | $-6.04_{\pm 0.27}$ | $-4.70_{\pm 0.02}$ |
| ConvNP (ELBO) | $-2.71_{\pm 0.02}$ | $-2.78_{\pm 0.02}$ | $-3.06_{\pm 0.01}$ | $-8.54_{\pm 0.30}$ | F | $-9.97_{\pm 0.18}$ |
| ACNP (AR) | $-2.83_{\pm 0.01}$ | $-2.98_{\pm 0.02}$ | $-3.25_{\pm 0.02}$ | $-\mathbf{4.60}_{\pm 0.09}$ | $-4.69_{\pm 0.06}$ | $-4.76_{\pm 0.05}$ |
| ANP (ELBO–ML) | $-2.86_{\pm 0.02}$ | $-3.16_{\pm 0.02}$ | $-3.63_{\pm 0.01}$ | $-5.28_{\pm 0.17}$ | $-5.70_{\pm 0.20}$ | $-4.75_{\pm 0.02}$ |
| ANP (ELBO) | $-2.95_{\pm 0.03}$ | $-3.21_{\pm 0.02}$ | $-3.66_{\pm 0.01}$ | $-5.84_{\pm 0.24}$ | $-5.77_{\pm 0.16}$ | $-4.83_{\pm 0.03}$ |
| ACNP | $-2.95_{\pm 0.02}$ | $-3.27_{\pm 0.02}$ | $-3.73_{\pm 0.01}$ | $-4.65_{\pm 0.10}$ | $-5.90_{\pm 0.19}$ | $-4.73_{\pm 0.02}$ |
| ANP (ML) | $-2.97_{\pm 0.01}$ | $-3.11_{\pm 0.02}$ | $-3.44_{\pm 0.01}$ | $-\mathbf{4.54}_{\pm 0.04}$ | $-\mathbf{4.41}_{\pm 0.04}$ | $-4.69_{\pm 0.02}$ |
| ConvNP (E.–M.) | $-5.87_{\pm 0.05}$ | $-7.82_{\pm 0.13}$ | F | F | F | F |

After having the trained and evaluated the models on data generated with our stochastic version of the Lotka–Volterra equations, we finally apply the models to the real hare–lynx data set. Table 6.15 again shows the results for all three tasks, and the right column of Figure 6.7 illustrates predictions by the models. Whereas the FullConvGNP shared the top position with the AR ConvCNP for the simulated data, the FullConvGNP now shares the top position with the AR ACNP and ANP. For the simulated data in Figure 6.7, predictions appear appropriately calibrated. For the real hare–lynx data in Figure 6.7, however, the predicted uncertainties look much less well calibrated, and sometimes even fail to capture the data. We conclude that the neural processes trained on our simulator can be used to produce acceptable predictions. However, these predictions are not perfectly calibrated, so we also conclude that there is a statistical mismatch between the simulator and the real hare–lynx data. That there is a statistical mismatch is corroborated by the observation that the log-likelihoods for the real data are generally much lower than for the simulated data.

To alleviate accidental overconfidence due to mismatch between the simulator and the real data, it works in a model's favour to generally produce predictions with bigger uncertainty. Figure 6.7 shows that the convolutional models tightly fit the simulated data, whereas the AR ACNP and ANP generally produce bigger uncertainties. Consequently, the convolutional model comparably perform worse on the real data, and the AR ACNP and ANP comparably

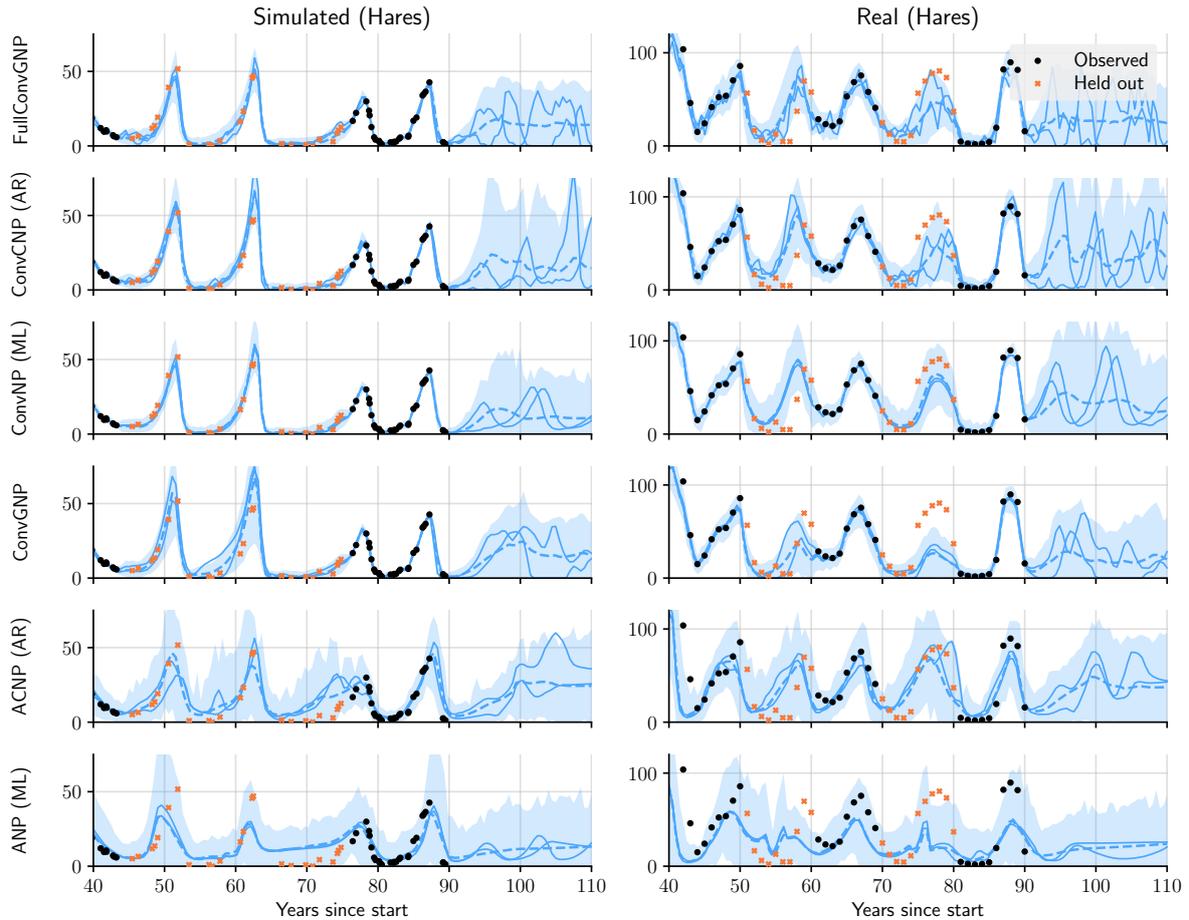

Figure 6.7: Predictions by the best performing models in the predator–prey experiment. *S*hows predictions for data generated from the simulator (left) and predictions for the real hare–lynx data (right). The models observe all lynx population counts and some hare population counts. The plots show predictions for some observed and some unobserved hare population counts. Filled regions are central 95%-credible regions.

perform better.

## 6.4  Electroencephalography Experiments

In the third experiment, we explore an electroencephalography data set collected from 122 subjects (Begleiter, 2022). There are two groups of subjects: alcoholic and control. Every subject was subjected to a single stimulus or two stimuli, and their response was measured with 64 electrodes placed on a subject's scalp. These measurements are in terms of *trials*, where a trial consists of 256 samples of the electrodes spanning one second. The data sets contains up to 120 trials for each subject. The data is available at https://archive.ics.uci.edu/ml/datasets/eeg+database and the collection is described in detail by Zhang et al. (1995). In this experiment, we focus only on seven frontal electrodes: FZ, F1, F2, F3, F4, F5, and F6. Figure 6.8 illustrates a trial of a subject, showing the samples for these seven electrodes.

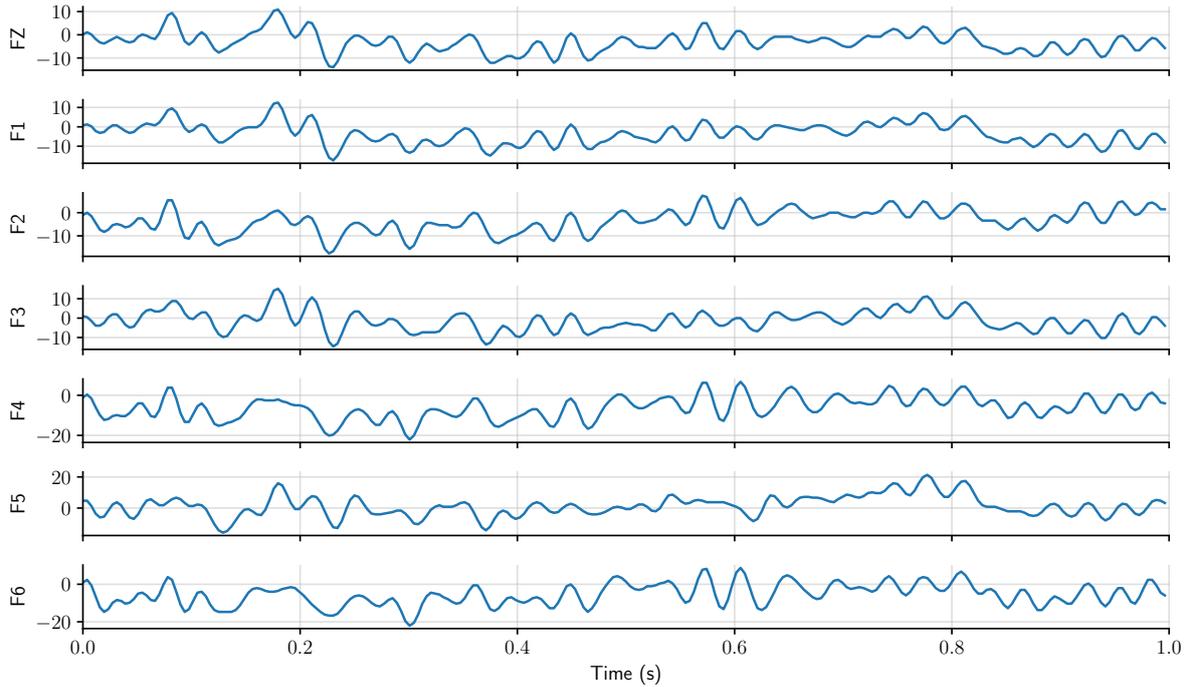

Figure 6.8: Example of trial in the EEG data set. Note that the signals for all electrodes appear correlated, but are subtly different.

We randomly split all subjects into three sets: an evaluation set consisting of ten subjects, a cross-validation set consisting of ten other subjects, and a training set consisting of all remaining subjects. For each of these sets, we create a meta–data set by aggregating the trials for all subjects. We split every trial into a context and target set in the same three ways as for the predator–prey experiment (Section 6.3). First, for all seven electrodes separately, randomly designate between 50 and 100 points to be the target points and let the remainder be the context points. This task is called *interpolation* and is the primary measure of performance. Second, select a random time between $0.5\,\text{s}$ and $0.75\,\text{s}$, let everything before that time be the context set, and let everything after that time be the target set. This task is called *forecasting* and measures the model's ability extrapolate. Third, randomly choose one of the seven electrodes and, for that choice, split the data in two exactly like for forecasting. For all other electrodes, append all data to the context set. This task is called *reconstruction* and measures the model's ability to infer a signal for one electrode from the others. To train the models, for every batch, we randomly choose one of the tasks by flipping a three-sided coin.

Like for the predator–prey experiment (Section 6.3), we compare the ConvCNP, ConvGNP, FullConvGNP, and AR ConvCNP against the ACNP, ANP, ConvNP, and AR ACNP. Appendix E.1 describes the general architectures of the models, Appendix E.2 describes the general training, cross-validation, and evaluation protocols, and Appendix E.5 describes details specific to this experiment. We remark that architecture of the ANP required modi-fications in order to scale the model, and that the training of the FullConvGNP and ANP were cut short after having run for 45 hours; see Appendix E.5 for more details.

Table 6.16: Normalised log-likelihoods in the EEG experiments. Shows the performance for interpolation, forecasting, and reconstruction. Models are ordered by interpolation performance. See Section 6.4 for a more detailed description. The latent variable models are trained and evaluated with the ELBO objective (ELBO); trained and evaluated with the ML objective (ML); and trained with the ELBO objective and evaluated with the ML objective (ELBO–ML; E.–M.). Errors indicate the central 95%-confidence interval. Numbers which are significantly best ($p < 0.05$) are boldfaced. Numbers which are very large are marked as failed with "F". Numbers which are missing could not be run.

| Model | Interpolation | Forecasting | Reconstruction |
| --- | --- | --- | --- |
| FullConvGNP | $-\mathbf{0.39}_{\pm 0.01}$ | $-\mathbf{0.52}_{\pm 0.01}$ | $-\mathbf{0.67}_{\pm 0.00}$ |
| ConvCNP (AR) | $-0.74_{\pm 0.01}$ | $-1.32_{\pm 0.02}$ | $-1.52_{\pm 0.01}$ |
| ConvGNP | $-0.83_{\pm 0.01}$ | $-0.57_{\pm 0.01}$ | $-1.84_{\pm 0.01}$ |
| ConvCNP | $-1.10_{\pm 0.01}$ | $-2.08_{\pm 0.04}$ | $-3.42_{\pm 0.02}$ |
| ConvNP (ML) | $-1.23_{\pm 0.01}$ | $-1.59_{\pm 0.02}$ | $-3.01_{\pm 0.01}$ |
| ConvNP (ELBO) | $-1.58_{\pm 0.01}$ | $-2.54_{\pm 0.04}$ | $-2.44_{\pm 0.01}$ |
| ANP (ML) | $-1.71_{\pm 0.01}$ | $-1.72_{\pm 0.02}$ | $-3.00_{\pm 0.01}$ |
| ACNP (AR) | $-1.79_{\pm 0.01}$ | $-2.04_{\pm 0.02}$ | $-2.18_{\pm 0.01}$ |
| ANP (ELBO–ML) | $-1.82_{\pm 0.01}$ | $-2.31_{\pm 0.03}$ | $-3.06_{\pm 0.01}$ |
| ANP (ELBO) | $-1.84_{\pm 0.01}$ | $-2.41_{\pm 0.03}$ | $-3.06_{\pm 0.01}$ |
| ACNP | $-1.96_{\pm 0.01}$ | $-2.33_{\pm 0.04}$ | $-3.42_{\pm 0.02}$ |
| ConvNP (E.–M.) | $-3.06_{\pm 0.04}$ | $-7.11_{\pm 0.70}$ | F |

**Results.** Table 6.16 presents the results for interpolation, forecasting, and reconstruction, and Figure 6.9 illustrates predictions by the models. On all three tasks, the FullConvGNP outperforms all other models. For interpolation, the AR ConvCNP comes second; and, for forecasting, the ConvGNP is a close second. For reconstruction, however, no other model comes close. Note that the AR ConvCNP outperforms the ConvGNP in the interpolation and reconstruction tasks, but the ConvGNP performs substantially better in the forecasting task. This is a rather strong performance by the ConvGNP. Except for the ConvGNP, for strictly all models, the forecasting and reconstruction log-likelihoods are worse than the interpolation log-likelihood. For the ConvGNP, the forecasting log-likelihood is best and considerably improves over the interpolation log-likelihood. This strong performance of the ConvGNP is confirmed by Figure 6.9, where the prediction of the ConvGNP in the forecasting task is indeed similar to that of the FullConvGNP.

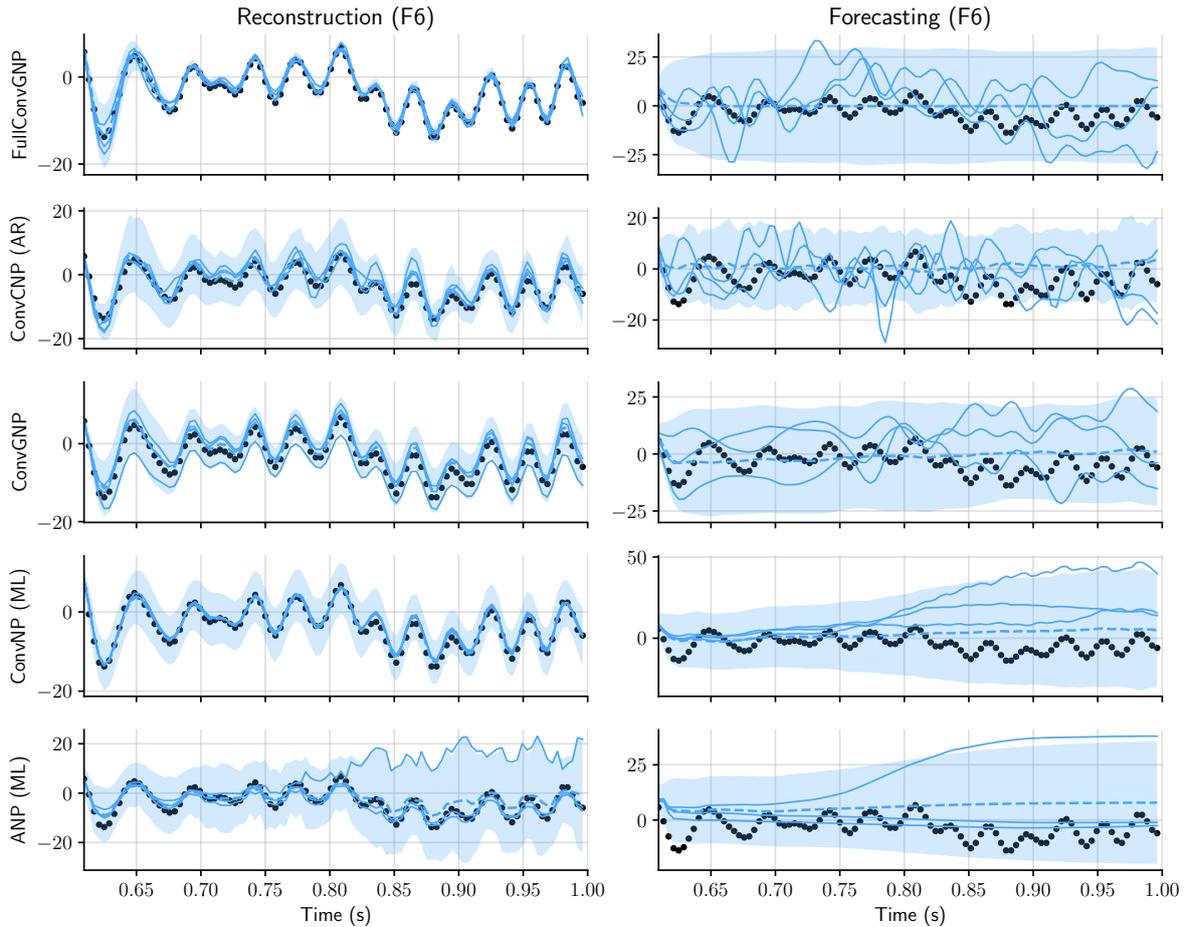

Figure 6.9: Predictions by the best performing models in the EEG experiment. The data in this example experiment is the sample trial from Figure 6.8. In both tasks, the models predict the last 100 samples of electrode F6. In the reconstruction task, the models observe for electrodes FZ and F1–F5 the entire signal and for electrode F6 only the first 156 samples. In the forecasting task, the models observe for all electrodes only the first 156 samples. Filled regions are central 95%-credible regions.

## 6.5 Climate Downscaling

In the last experiment, we explore a climate science application. In climate modelling, future projections are generated by integrating the atmospheric equations of motion forward in time using a numerical solver. Unfortunately, computational constraints typically limit the spatial resolution of those projections to around $100\,\text{km}$ (Eyring et al., 2016), which is insufficient to resolve extreme events and produce local predictions (Stocker et al., 2013; Maraun et al., 2017). Statistical downscaling methods attempt to address this issue by refining the coarse-grained outputs into more fine-grained predictions (Maraun et al., 2018). While numerous data-driven approaches exist, statistical downscaling methods are often limited to making predictions at a fixed set of points at which historical observations are available (Vandal et al., 2017; Bhardwaj et al., 2018; Misra et al., 2018; Sachindra et al., 2018;

Vandal et al., 2018; Vandal et al., 2019; Pan et al., 2019; A. Singh et al., 2019; Baño-Medina et al., 2020; Y. Liu et al., 2020).

Recently, Vaughan et al. (2022) used the ConvCNP to perform statistical downscaling. Compared to existing downscaling methods, a notable feature of the ConvCNP is that the model is not limited to a fixed set of points, but can make fine-grained predictions anywhere. However, as we discussed in Section 5.5, the ConvCNP does not model dependencies between target outputs, and is consequently unable to produce coherent samples. This limits the applicability of Vaughan et al.'s approach, because the ability to model dependencies is important for downstream applications. For example, to estimate the maximum temperature of a region, as opposed to single point location, it is necessary to model spatial dependencies. In this experiment, we extend Vaughan et al.'s approach to the ConvGNP and AR ConvCNP. Importantly, the ConvGNP and AR ConvCNP model spatial dependencies, so we broaden the downstream applicability of Vaughan et al.'s approach.

Thus far, context sets and target sets have consisted of data from the same domain. Specifically, in Section 3.1, we proposed a motivating generative model where the context and target set both consist of noisy observations of the same ground-truth stochastic process $f$. However, nowhere in our developments have we actually used or required that the outputs of the context and target sets live in the same space! In fact, all models that we have considered so far also work when the outputs of the context sets are vectors of a different dimensionality than the outputs of the target set. In the CNP (Model 5.1) and GNP (Model 5.9), this simply means that the MLP $\phi_\theta$ of the encoder takes in more of fewer $y$-dimensions; and in the ConvCNP (Model 5.4), ConvGNP (Model 5.12), and FullConvGNP (Model 5.13), this simply means that the CNN takes in more or fewer data channels—no other modifications are necessary. This generalisation might not have been immediately apparent earlier, because we only considered one-dimensional outputs for notational simplicity. Letting the outputs of the context and target sets be from different domains says, very simply, to make predictions for *some data* based on *other data*.[2] In a way, this formulation is even simpler than what we have considered so far. Vaughan et al.'s (2022) approach is based on this interpretation of neural processes. Before we explain the setup of the ConvCNP and the extension to the ConvGNP and AR ConvCNP, we first detail the experimental setup.

In this experiment, we will consider two statistical downscaling tasks: the *VALUE task* and the *Germany task*. The VALUE task follows the VALUE framework (Maraun et al., 2015), which provides results for a large ensemble of frequently used downscaling methods on a standardised set of tasks. More specifically, for the VALUE task, we consider experiment 1a from the VALUE framework and estimate the maximum daily temperature at 86 weather stations around Europe. To estimate these temperatures, we follow Vaughan et al. (2022) and use 25 coarse-grained ERA-Interim reanalysis variables (Dee et al., 2011) in combination with $1\,\text{km}$–resolution elevation data (Earth Resources Observation and Science Center, U.S. Geological Survey, U.S. Department of the Interior, 1997). We consider the following 25 ERA-Interim reanalysis variables:

- at the surface: maximum temperature, mean temperature, northward wind, and eastward wind;

- in the upper atmosphere at $850\,\text{hPa}$, $700\,\text{hPa}$, and $500\,\text{hPa}$: specific humidity, temperature, northward wind, and eastward wind; and

- invariant features: angle of sub-grid-scale orography, anisotropy of sub-grid-scale orography, standard deviation of filtered sub-grid-scale orography, standard deviation of orography, geopotential, longitude, latitude, and fractional position in the year $t$ transformed encoded with $t \mapsto (\cos(2\pi t), \sin(2\pi t))$.

---

[2] For translation equivariance (Definition 2.4) to make sense, we do require that the inputs of the context and target set live in the same space. However, should we not be concerned with translation equivariance, then the inputs need not live in the same space either.

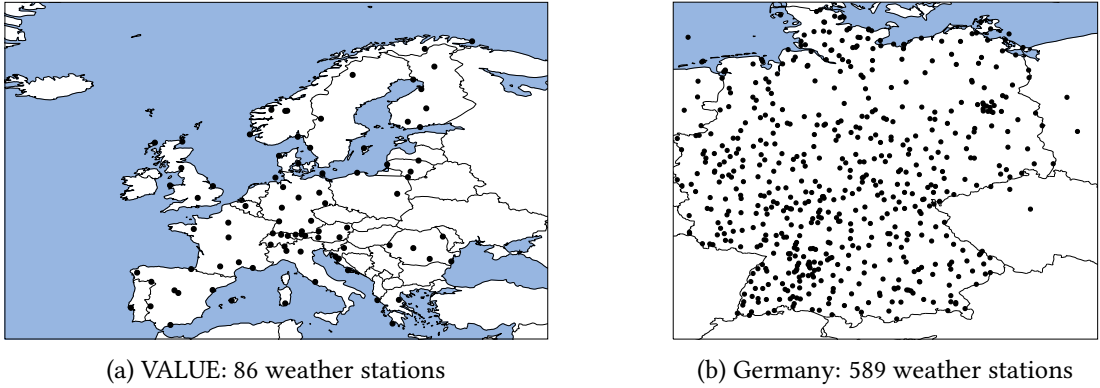

| (a) VALUE: 86 weather stations | (b) Germany: 589 weather stations |

Figure 6.10: Locations of the weather stations in the VALUE and Germany downscaling and fusion experiments

These variables are bilinearly interpolated to a $2°$-resolution grid. The temperature data is provided by the VALUE framework at http://www.value-cost.eu/data and the ERA-Interim reanalysis data is available at https://www.ecmwf.int/en/forecasts/datasets/reanalysis-datasets/era-interim. The $1\,\text{km}$–resolution elevation data is available at https://doi.org/10.5066/F7DF6PQS. The locations of the weather stations around Europe are visualised in Figure 6.10a.

For the Germany task, we estimate the maximum daily temperature at 589 weather stations around Germany using the same 25 coarse-grained ERA-Interim reanalysis variables and the same $1\,\text{km}$–resolution elevation data. In this case, however, the ERA-Interim reanalysis variables are used at their native $0.75°$-resolution grid. The weather station data are a subselection of the European Climate Assessment & Dataset (Tank et al., 2002) and are available at https://www.ecad.eu; we use the blended data. The locations of the weather stations around Germany are visualised in Figure 6.10b.

**Part 1: downscaling with the MLP ConvCNP and MLP ConvGNP.** This experiment consists of two parts. In this first part, we perform the VALUE and Germany downscaling tasks by setting up the context and target sets as follows. For every day, let the context set be the 25 coarse-grained ERA-Interim reanalysis variables and let the target set be the observations of the weather stations—we will shortly explain how the $1\,\text{km}$–resolution elevation data is incorporated. As we previously explained, there is no problem in letting the context and target data be from different domains. For the ConvCNP and ConvGNP, we set the internal discretisation (Procedure 5.5) equal to the $2°$ or $0.75°$-resolution grid of the ERA-Interim reanalysis variables. Moreover, following Vaughan et al. (2022), rather than using a U-Net architecture, we use a residual convolutional neural network (He et al., 2016) with depthwise-separable convolutions (Chollet, 2017). To incorporate the $1\,\text{km}$–resolution elevation data, one approach would be to also include this data in the context set. However, we would then have to increase the resolution of the discretisation from $2°$ or $0.75°$ to $1\,\text{km}$; otherwise, much of the detail in the elevation data would be lost. Unfortunately, making the discretisation this much finer would come with a substantial increase in computational cost. Instead, Vaughan et al. propose a different approach. After the CNN architecture, insert a pointwise multilayer perceptron (MLP) which, at every target input, fuses the local elevation with the output of the CNN architecture:

**Definition 6.1** (Pointwise MLP; Vaughan et al., 2022). *Consider a neural process with a decoder* $\text{dec}_\theta$ *operating on a functional encoding (Definition 4.6):* $\text{dec}_\theta\colon C(\mathcal{X}, \mathbb{R}^{K_1}) \to C(\mathcal{X}, \mathbb{R}^{K_2})$. *Let* $\text{aux}\colon \mathcal{X} \to \mathbb{R}^{K_3}$ *be auxiliary information. We say that the neural process incorporates the auxiliary information with a pointwise MLP if the decoder is of the form* $\text{dec}_\theta = \text{fuse}_\theta \circ \text{dec}'_\theta$ *where* $\text{dec}'_\theta\colon C(\mathcal{X}, \mathbb{R}^{K_1}) \to C(\mathcal{X}, \mathbb{R}^{\widetilde{K}_2})$ *and*

$$\text{fuse}_\theta\colon C(\mathcal{X}, \mathbb{R}^{\widetilde{K}_2}) \to C(\mathcal{X}, \mathbb{R}^{K_2}), \quad \text{fuse}_\theta(\mathbf{z}(\,\cdot\,)) = \text{MLP}_\theta(\mathbf{z}(\,\cdot\,), \text{aux}(\,\cdot\,)) \qquad (6.10)$$

*with* $\text{MLP}_\theta\colon \mathbb{R}^{\widetilde{K}_2+K_3} \to \mathbb{R}^{K_2}$ *a multi-layer perceptron called the* pointwise MLP.

We let the ConvCNP and ConvGNP incorporate the $1\,\text{km}$–resolution elevation data using a pointwise MLP in the sense of Definition 6.1; specifically, aux produces the elevation at any input using bilinear interpolation. We call these variations of the ConvCNP and ConvGNP the *MLP ConvCNP* and *MLP ConvGNP* respectively.

Table 6.17: Normalised log-likelihoods and mean absolute errors (MAEs) in the climate downscaling experiments. Shows the performance on the VALUE and Germany downscaling tasks averaged over the five folds from the VALUE protocol (Maraun et al., 2015). The MAE is the mean of the MAE per station. The MAE of the VALUE models are computed from results published at http://www.value-cost.eu/validationportal. VALUE models refer to the models used the comparison of 52 downscaling approaches (http://www.value-cost.eu; Maraun et al., 2015; Gutiérrez et al., 2019). "PP" stands for perfect prognosis and refers to the class of VALUE models which are comparable to the models in this chapter (Maraun et al., 2010). Errors indicate the central 95%-confidence interval. Numbers which are significantly best ($p < 0.05$) are boldfaced.

| Model | VALUE Log-lik. | MAE | Germany Log-lik. | MAE |
|---|---|---|---|---|
| ConvCNP (MLP) | $-1.66 \pm 0.00$ | $\mathbf{1.04} \pm 0.04$ | $-1.62 \pm 0.01$ | $\mathbf{1.00} \pm 0.05$ |
| ConvGNP (MLP) | $\mathbf{-1.63} \pm 0.00$ | $\mathbf{1.01} \pm 0.04$ | $\mathbf{-1.45} \pm 0.00$ | $1.14 \pm 0.06$ |
| Best VALUE PP model | | $1.37 \pm 0.05$ | | |
| Best VALUE model | | $1.20 \pm 0.08$ | | |

We perform the VALUE and Germany downscaling tasks for all days of the years 1979–2008. Following the VALUE framework (Maraun et al., 2015), we split these years into five folds. On each of the five folds, we evaluate the neural process, using the four other folds for training and holding out the last 1000 days of the training folds for cross-validation. Appendix E.2 describes the general training, cross-validation, and evaluation protocols, and Appendix E.6 describes the architectures and further details specific to this experiment.

**Results.** The VALUE framework includes results for 52 downscaling approaches (http://www.value-cost.eu/; Maraun et al., 2015; Gutiérrez et al., 2019). Amongst these 52 approaches, the category that is comparable to our neural process approach is called *perfect prognosis* (PP) (Maraun et al., 2010). Table 6.17 shows the results of the MLP ConvCNP and MLP ConvGNP on the VALUE and Germany downscaling tasks and compares these results to the overall best and the best PP approach from the VALUE framework. Compared to all 52 downscaling approaches, the MLP ConvCNP and MLP ConvGNP achieve 24%–26% lower MAE than the best comparable PP model and 18%–20% lower MAE than the overall best performing approach. In addition, compared to the ConvCNP MLP, the ConvGNP MLP achieves a moderate but statistically significant improvement in log-likelihood.

In the VALUE task, most weather stations are geographically far apart, so daily maximum temperatures at these weather stations are only moderately correlated. For this reason, the benefits of the ConvGNP MLP over the ConvCNP MLP are less pronounced. On the other hand, in the Germany task, the weather stations are geographically more nearby, so the daily maximum temperatures in this task are more strongly correlated. This is reflected in Table 6.17, which shows that, in the Germany task, the MLP ConvGNP gives a bigger im-provement in log-likelihood over the MLP ConvCNP than in the VALUE task. Note, however, that the MAE of the MLP ConvGNP is worse than that of the MLP ConvCNP. This is reasonable, as the MLP ConvGNP must spend model capacity on modelling dependencies, whereas the MLP ConvCNP can fully focus on the marginals. Figure 6.11 illustrates two samples from the MLP ConvGNP over Germany. The samples exhibit a realistic degree of variability.

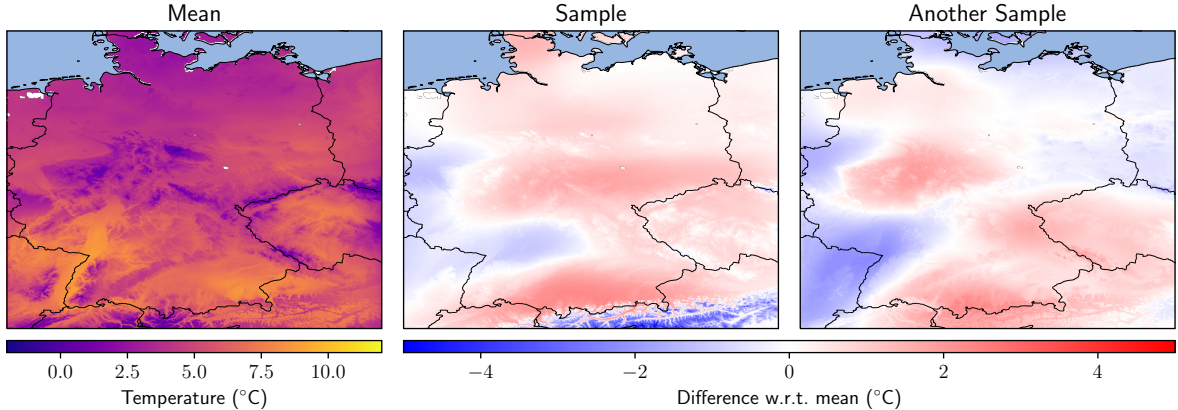

Figure 6.11: Mean prediction by and two samples from the MLP ConvGNP for an arbitrary day in the Germany downscaling experiment. For the samples, shows the difference with respect to the mean.

**Part 2: downscaling and fusion with the AR ConvCNP.** We have demonstrated that the MLP ConvGNP can be used to successfully model dependencies between outputs in a statistical downscaling task, improving log-likelihoods over the MLP ConvCNP (Table 6.17) and enabling coherent samples (Figure 6.11). In this second part of this experiment, we demonstrate that the AR ConvCNP can also be used for this purpose. Deploying the AR ConvCNP in this task, however, comes with a significant challenge. In the autoregressive sampling procedure (Procedure 5.14), samples from the model will be fed back into the model. If samples of the AR ConvCNP are to be of the same level of quality as the MLP ConvGNP and, in particular, are to vary on the same spatial scale, then the AR ConvCNP must handle context data which varies on this spatial scale. Since the elevation data is on a $1\,\mathrm{km}$–resolution grid, samples of the MLP ConvGNP will also vary on roughly this spatial scale. Consequently, the discretisation of the AR ConvCNP must also be a $1\,\mathrm{km}$–resolution grid, and we previously argued that such a discretisation would be prohibitively expensive! We must therefore innovate on AR ConvCNP design to come up with a convolutional architecture that can handle such a fine discretisation at reasonable computational expense.

The architecture that we propose is a *multiscale architecture* operating on multiple spatial length scales. Let us divide the context set $D = D_{\mathrm{lr}} \cup D_{\mathrm{mr}} \cup D_{\mathrm{hr}}$ into a low-resolution component $D_{\mathrm{lr}}$, a medium-resolution component $D_{\mathrm{mr}}$, and a high-resolution component $D_{\mathrm{hr}}$. Let the low-resolution component $D_{\mathrm{lr}}$ consist of features of the context set that vary on a *long* spatial length scale, the medium-resolution component $D_{mr}$ of features that vary on a *medium-range* spatial length scale, and the high-resolution component $D_{hr}$ of features that vary on a *short* spatial length scale. The central assumption of the architecture is that predictions for target points depend on precise short-length-scale details $D_{hr}$ nearby, but that this dependence weakens as we move away from the target point, starting to depend more on broad-stroke long-length-scale components $D_{lr}$. For example, predictions might depend on detailed orographic information nearby, but more on general orographic shapes farther away.

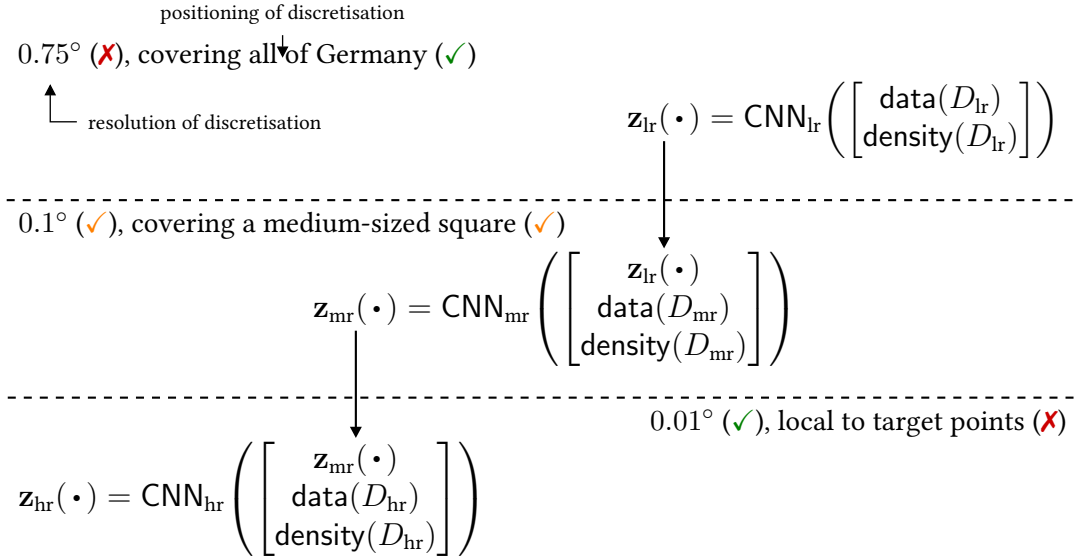

Figure 6.12: Multiscale architecture for the AR ConvCNP. A cascade of three convolutional deep sets (Theorem 4.8) representing a low-resolution, medium-resolution, and high-resolution component. Shows the resolution and positioning of the internal discretisation for every convolutional deep set. The context set $D = D_{lr} \cup D_{mr} \cup D_{hr}$ is also divided into a low-resolution $D_{lr}$, medium-resolution $D_{mr}$, and high-resolution component $D_{hr}$. The low-resolution context data $D_{lr}$ consists of the 25 coarse-grained ERA-Interim reanalysis variables. The medium-resolution $D_{mr}$ and high-resolution context data $D_{hr}$ both consist of the station observations and the $1\,\text{km}$–resolution elevation data. See Section 6.5 for more details. The functions $\text{data}(D)$ and $\text{density}(D)$ produce respectively the data channel and density channel for context data $D$; see Model 5.4 and equations (5.4) and (5.5). The maps $\text{CNN}_{lr}$, $\text{CNN}_{mr}$, and $\text{CNN}_{hr}$ are translation-equivariant maps between functions on $\mathcal{X}$. In practice, these maps are all implemented with convolutional neural networks (CNN) using the discretisation approach outlined in Procedure 5.5. For $\text{CNN}_{lr}$, the internal discretisation is the $0.75°$-resolution grid corresponding to the 25 coarse-grained ERA-Interim reanalysis variables. For $\text{CNN}_{mr}$, the internal discretisation is a $0.1°$-resolution grid spanning $5°$ more than the most extremal target inputs; the discretisation does *not* depend on the context set. For $\text{CNN}_{hr}$, the internal discretisation is a $0.01°$-resolution grid spanning $0.25°$ more than the most extremal target inputs; the discretisation also does *not* depend on the context set.

Figure 6.12 depicts the multiscale architecture. The architecture is a cascade of three convolutional deep sets, parametrised by three CNNs; please see the caption. The low-resolution CNN handles the context data $D_{lr}$ with a long spatial length scale. Because these features have a long spatial length scale, the CNN can get away with a low-resolution discretisation. The output of the low-resolution CNN then feeds into a medium-resolution CNN. The medium-resolution CNN handles the context data $D_{\text{mr}}$ with a medium spatial length scale and has a medium-resolution discretisation. Finally, the output of the medium-resolution CNN feeds into a high-resolution CNN. This CNN handles the context data $D_{\text{hr}}$ with a short spatial length scale and has a high-resolution discretisation.

The key to the computational efficiency of this architecture is that we construct the high-resolution discretisation only locally to the target points: a small square covering $0.25°$ more than the most extremal target points. If the target points are confined to a small region, then the high-resolution grid will also be small, covering only $0.25°$ more than that region. Crucially, the high-resolution grid will not be constructed over all of Germany, like it would if we were to more naively apply the ConvCNP with a high-resolution discretisation, incurring prohitive computational cost. Even though the high-resolution grid is only constructed locally to the target points, the model can still capture long-range dependencies via the medium-resolution and low-resolution grids. Namely, the medium-resolution grid is a square covering $5°$ more than the most extremal target points, and the low-resolution grid covers all of Germany; see Figure 6.12. To utilise this computational gain, the target points must be confined to a small region. This perfectly synergises with the autoregressive sampling procedure (Procedure 5.14), because this procedure evaluates the model one target point at a time. The training procedure, however, must be adapted. During training, we subsample the target points to ensure that the target set is always confined to a small square. See Appendix E.6 for more details.

During the autoregressive sampling procedure, the AR ConvCNP takes in earlier AR samples from the model. Currently, these is no natural context data to which these samples can be appended. Therefore, in addition to the ERA-Interim reanalysis variables and the elevation data, we also let the AR ConvCNP take in observed weather stations as context data. We will append the earlier AR samples to these weather station context data. To have the model make appropriate use of the weather station context set, we must randomly divide the weather stations observations over the context and target set. We let the low-resolution context data $D_{\text{lr}}$ consist of the 25 coarse-grained ERA-Interim reanalysis variables, and let the medium-resolution $D_{\text{mr}}$ and high-resolution context data $D_{\text{hr}}$ both consist of the weather station observations (and earlier AR samples) and the $1\,\text{km}$–resolution elevation data. When the $1\,\text{km}$–resolution data is fed to the medium-resolution CNN, the data loses some detail, because the internal discretisation of the medium-resolution CNN is coarser than the data; however, when it is fed to the high-resolution CNN, the data retains its detail. The same holds for the weather station observations (and earlier AR samples).

**Results.** We run the AR ConvCNP only on the Germany task and only on the fifth fold. In addition to the climate downscaling task, we also perform a *data fusion task*. Data fusion is a ubiquitous problem in environmental and climate science, as observations of a variable are often available from multiple stations, satellite platforms, and model outputs. The data fusion task is toy experiment to showcase the potential of the ConvCNP for these types of problems. For the data fusion task, we randomly divide the weather station observations into a context set and a target set. The model must learn to fuse the observed weather stations with the ERA-Interim reanalysis variables and elevation data to produce the best predictions possible for the unobserved weather stations. The MLP ConvCNP and MLP ConvGNP have no mechanism to incorporate the observed weather stations; hence, these models simply ignore this data. Table 6.18 shows the results. For the downscaling task the AR ConvCNP achieves the same log-likelihood and MAE as the MLP ConvGNP, demonstrating that autoregressive CNPs can be successfully applied to perform statistical downscaling. For the data fusion task, the AR ConvCNP achieves better log-likelihood and MAE than the MLP ConvCNP and MLP ConvGNP. On the one hand, this is according to expectations, because only the AR ConvCNP makes use of the observed weather stations. On the other hand, the AR ConvCNP is a vastly more involved model than the MLP ConvCNP and MLP ConvGNP, so it is encouraging to see that the model is functioning correctly and beats the strong performance of the MLP ConvGNP in this task.

Table 6.18: Normalised log-likelihoods and mean absolute errors (MAEs) in the climate downscaling and fusion experiments. Shows the performance on the Germany task for the fifth fold of the VALUE protocol (Maraun et al., 2015). The MAE is the mean of the MAE per station. Errors indicate the central 95%-confidence interval. Numbers which are significantly best ($p < 0.05$) are boldfaced.

| Model | Downscaling Log-lik. | MAE | Fusion Log-lik. | MAE |
| --- | --- | --- | --- | --- |
| ConvCNP (MLP) | $-1.55$ ±0.01 | **0.94** ±0.03 | $-1.55$ ±0.01 | $0.94$ ±0.03 |
| ConvGNP (MLP) | $\mathbf{-1.36}$ ±0.01 | $1.09$ ±0.09 | $-1.38$ ±0.01 | $1.09$ ±0.09 |
| ConvCNP (AR) | $\mathbf{-1.36}$ ±0.01 | $1.04$ ±0.04 | $\mathbf{-1.31}$ ±0.01 | **0.85** ±0.05 |

## 6.6 Conclusion

In this chapter, we conducted four experiments that evaluate and test the limits of existing and newly proposed neural process models.

First, in Section 6.2, we performed a large-scale bake-off between five existing and eight newly proposed models (Table 6.1), evaluating every model in a total of 60 different subexperiments. These subexperiments involved data synthetically generated from Gaussian and non-Gaussian stationary processes. For the Gaussian experiments, the FullConvGNP outper-formed all other methods, but only for one-dimensional inputs; the FullConvGNP could not be scaled to two-dimensional inputs. For two-dimensional inputs, the AR ConvCNP outperformed all other methods. For the non-Gaussian experiments with one-dimensional inputs, the AR ConvCNP and ConvNP performed best; and for two-dimensional inputs, the AR ConvCNP outperformed all other methods by a large margin. The ConvCNP was the only CNP able to consistently approach the conditional neural process approximation (CNPA; Definition 3.24) and the FullConvGNP was the only GNP able to consistently approach the Gaussian neural process approximation (GNPA; Definition 3.25). Overall, between convolutional, attentive, and deep set–based models, convolutional models demonstrated much better in-distribution performance and superior generalisation performance. This is not surprising, because the data was generated from stationary processes, matching the inductive bias of ConvNPs (Proposition 5.2). In addition, on the whole, GNPs outperformed LNPs in Gaussian experiments, whereas LNPs showed an edge in non-Gaussian experiments.

There are two interesting additional takeaways from the synthetic experiments. To begin with, while the ConvGNP showed good interpolation and OOID performance, the model's extrapolation performance was poor. The problem is that the ConvGNP uses a translation-equivariant parametrisation of the eigenmap, but not every eigenmap might admit a translation-equivariant parametrisation (Section 5.5). Consequently, the ConvGNP suffers from limited representational capacity. Poor extrapolation performance is a manifestation of this issue. This observation underlines the importance of fully basing neural process architecture on representation theorems (Chapter 4), which guarantee that such representational capacity problems cannot occur. Moreover, we saw that the AR ConvCNP exhibited strong performance across the board, and in some cases even outperformed all other methods by a large margin. This is surprising, because the ConvCNP is a much simpler model than, for example, the FullConvGNP or ConvNP. This demonstrates that CNPs, despite being the least sophisticated neural process, actually do have the capacity to compete with much more sophisticated approaches.

Second, in Section 6.3, we demonstrated that neural process can be used to perform sim-to-real transfer. In the experiment, we trained neural processes on a stochastic version of the Lotka–Volterra equations, and then deployed the models on real data: the famous hare–lynx data set (MacLulich, 1937). On the simulated data, the FullConvGNP and AR ConvCNP performed best, and convolutional models generally outperformed non-convolutional models. However, when evaluated on the real data, the FullConvGNP was paralleled by the AR ACNP and ANP. This change in ranking was primarily attributed to mismatch between the real data and data simulator. Secondarily, the AR ACNP and ANP generally produced more uncertain predictions, and increasing uncertainty helps prevent accidental overconfidence in the case of model mismatch. On the whole, predictions for the real data looked reasonable. This experiment therefore shows that neural processes can successfully perform sim-to-real transfer, with the caveat that the models perform only as well as the simulator matches the real data. Moreover, the experiment shows that calibration in the case of mismatch can be improved by appropriately inflating uncertainty. Therefore, in the sim-to-real setting, post-hoc uncertainty calibration using a small part of the real data might be beneficial. Post-hoc uncertainty calibration has been shown to help for classification with neural networks (Guo et al., 2017; Tomani et al., 2021).

Third, in Section 6.4, we pushed the limits of neural process models in an experiment with electroencephalography data. The FullConvGNP performed best, by a large margin, with the AR ConvCNP ranking second. Convolutional models again generally outperformed non-convolutional models.

In the last experiment, we explored the ability of neural processes to perform statistical downscaling. We extended Vaughan et al.'s (2022) approach (a pointwise MLP; Definition 6.1) for the ConvCNP to the ConvGNP. We first reproduced Vaughan et al.'s baseline experiment, showing that the ConvCNP and ConvGNP outperform all models in an established comparison of 52 downscaling approaches (http://www.value-cost.eu; Maraun et al., 2015; Gutiérrez et al., 2019). We then performed another downscaling experiment involving temperatures across Germany. Compared to the ConvCNP, the ConvGNP demonstrated improved log-likelihoods. In addition, unlike the ConvCNP, the ConvGNP was able to produce samples showing a realistic degree of variability (Figure 6.11), which is especially important in downstream applications. Afterwards, we considered the AR ConvCNP, which required a novel multiscale convolutional architecture (Figure 6.12) to overcome prohibitive computational cost. The AR ConvCNP demonstrated performance on par with the ConvGNP. In a variation on the downscaling experiment, involving data fusion, the AR ConvCNP was even able to outperform the ConvCNP and ConvGNP. We remark, however, that Vaughan et al.'s approach for the ConvCNP and ConvGNP was not designed for this data fusion setup. These climate downscaling and fusion experiments demonstrate the considerable performance and flexibility of convolutional deep sets (Theorem 4.8) and more generally the neural process framework.

All in all, when stationarity is a reasonable assumption, convolutional neural process models demonstrate strong in-distribution and generalisation performance and outperform their non-convolutional counterparts, at least in experiments in this chapter. If stationarity is not reasonable, it might be possible to include features that render the data stationary, like in the climate experiments; otherwise, a non-convolutional approach or a even combination of a convolutional and non-convolutional approach may be the better choice.

In terms of the overall best performing models, when dependencies between target outputs are not important, the ConvCNP emerged as the clear winner. If dependencies are important and the data is roughly Gaussian, the FullConvGNP tended to be the best performing model, whenever it is computationally feasible. Barring the FullConvGNP, for roughly Gaussian data, the AR ConvCNP and ConvGNP generally exhibited strong overall performance at feasible computational cost. For more non-Gaussian data, the AR ConvCNP and ConvNP were the best performing models.

In terms of the various classes of neural processes, if the data is roughly Gaussian, GNPs tended to outperform LNPs. For more non-Gaussian data, LNPs tended to show an edge over GNPs. In either case, AR CNPs showed strong performance and, in some cases, put forward the best performing model. AR CNPs, however, suffer high computational cost at test time and are no longer consistent models. This lack of consistency comes with slew of problems unique to AR CNPs (Section 5.6).

Finally, we remark that all experimental results depend strongly on the precise details of the model architectures and the precise execution of the experiments. Whilst we attempted to optimise the performance of all models and attempted to execute all experiments in a way that is as reproducible, as careful, and as fair as possible, improvements are almost certainly possible. One particular improvement is the use of a cyclical learning rate schedule, which could drastically reduce training time and generally boost performance (Smith, 2017).

# 7 | A Software Framework for Composing Neural Processes

**Abstract.** This chapter presents a software abstraction called *coders* that enables the design of cohesive and loosely coupled building blocks for neural processes. Coders form the basis of a Python package `neuralprocesses` available at https://github.com/wesselb/neuralprocesses. `neuralprocesses` implements all models in this thesis.

**Outline.** Section 7.1 argues why a software abstraction is necessary. In Section 7.2, we explain how models are constructed in `neuralprocesses`; and in Section 7.3, we explain a key principle of `neuralprocesses`. Afterwards, in Section 7.4, we introduce the proposed abstraction *coders*. Finally, in Section 7.5, we illustrate how coders enable the design of neural processes by putting together more elementary building blocks in different ways.

**Attributions and relationship to prior work.** The abstraction presented in this chapter was conceived in collaboration with Jonathan Gordon in `NeuralProcesses.jl` (Bruinsma and Gordon, 2022b). `neuralprocesses` (Bruinsma, Andersson, Markou, and Requeima, 2022a) is primarily developed by the author, but features contributions from Tom Andersson, Stratis Markou, and James Requeima. All work was supervised by Richard E. Turner.

## 7.1 Introduction

In the previous chapter, we put a wide variety of neural process models to the test. Of course, all these models had to be implemented. When implementing a large number models, it is undesirable to reimplement every model from scratch, completely starting over every time again. Without any code reuse, a code base becomes hard to maintain and improve. For example, if a detail of deep sets would need to change, then this change would need to be effected in every model that uses deep sets, an error-prone and tedious undertaking. Intead, one would prefer to make this change in only one place. This can be achieved by carefully crafting modular, composable building blocks, and constructing neural process models by putting together these building blocks in different ways. Then, when a building block is improved, that improvement immediately benefits every model built from that block.

Two desirable qualities of a software design are *low coupling* and *high cohesion* (Stevens et al., 1974). Low coupling means that modules communicate via simple and transparent interfaces and that connections between modules are minimised; and high cohesion means that elements within a module belong together. Succesfully designing a modular implementation of neural processes is difficult, because there are many desirable features which all interact in particular ways: latent variables, attentive mechanisms, functional encodings, multidimensional outputs, *et cetera*. Carelessly isolating these features runs the risk of high coupling. This chapter proposes a simple software abstraction called *coders* that enables the design of cohesive and loosely coupled building blocks for neural processes.

The software abstraction proposed in this chapter forms the basis of a Python (van Rossum, 1995) package called `neuralprocesses` (Bruinsma et al., 2022a), which is available at `https://github.com/wesselb/neuralprocesses`. `neuralprocesses` implements all models presented in this thesis (and more) and was also used to perform all experiments. This chapter, however, will not be a user manual nor a technical description of `neuralprocesses`; for that, we refer the reader to `https://github.com/wesselb/neuralprocesses`. We will solely focus on the underlying abstraction. Before presenting this software abstraction in Section 7.4, we first discuss how models are constructed in `neuralprocesses` (Section 7.2) and introduce a key principle of the package (Section 7.3).

## 7.2 Model Design

Recall that a neural processes parametrises a prediction map $\pi_\theta \colon \mathcal{D} \to \mathcal{Q}$ where $\mathcal{Q}$ is some variational family (Section 2.2). To parametrise a prediction map, we argued that encoder–decoder architectures are a convenient way to break up the model (Section 2.4). This choice is embodied in `neuralprocesses`, where models are compositions of encoders and decoders:

```
import neuralprocesses.torch as nps

encoder = ...
decoder = ...

model = nps.Model(encoder, decoder)  # Composition of `encoder` and `decoder`
```

The above code uses PyTorch (Paszke et al., 2019) as the backend. Other backends can be used by changing the import. For example, to use TensorFlow (Abadi et al., 2016), change the import to `import neuralprocesses.tensorflow as nps`. In fact, the implementation of `neuralprocesses` is backend agnostic and can easily be extended to other frameworks, like JAX (Bradbury et al., 2018). This backend-agnostic implementation is facilitated by the Python package LAB (Bruinsma et al., 2022c).

Once a model is defined, it can be run forward to make a prediction for some context set $D$. If `D = (xc, yc)` and `x = xt`, then

```
pred = model(xc, yc, xt)
```

represents the distribution $P_\mathbf{x}\pi_\theta(D)$. For any model constructed with `nps.Model`, the package provides functionality to automatically compute log-likelihoods, autoregressive log-likelihoods (Section 5.6), the ELBO objective for latent-variable neural processes (LNPs; Garnelo et al., 2018b), the ML objective for LNPs (Foong et al., 2020), all whilst appropriately dealing with multidimensional inputs and outputs, *et cetera*. In this chapter, we will focus on how `encoder` and `decoder` can be constructed in a modular way.

## 7.3 Functions as Intermediate Representations

Our approach will be to implement `encoder` and `decoder` as *compositions of transformations*. To flexibly compose transformations, what comes out of one transformation should be a compatible input for another. We will achieve this by letting all transformations adhere to an agreed-upon interface. Letting all transformations adhere to one interface is challenging, because neural processes deal with a variety of objects: data sets, vector encodings, functional encodings, attentive mechanisms, latent variables, predictions, *et cetera*.

To deal with the challenging variety of objects, we homogenise what is passed between transformations. In particular, we propose to let all intermediate representations be *functions*. Some examples are as follows:

- A *context set* $(\mathbf{x}^{(c)}, \mathbf{y}^{(c)}) \in D$ is represented as the function mapping from the inputs to the outputs:
$$f \colon \{x_1^{(c)}, \ldots, x_N^{(c)}\} \to \{y_1^{(c)}, \ldots, y_N^{(c)}\}, \quad f(x_i^{(c)}) = y_i^{(c)}. \tag{7.1}$$

    Note that this function is a map between *finite sets*.

- A *vector encoding* $\mathbf{x} \in \mathbb{R}^K$ is represented as a function mapping from the empty set to the vector:
$$f \colon \varnothing \to \{\mathbf{x}\}, \quad f() = \mathbf{x}. \tag{7.2}$$

- A *functional encoding* $\mathbf{z}(\,\cdot\,) \in Z^{\mathcal{X}}$ is already a function:

$$f\colon \mathcal{X} \to Z, \quad f(x) = \mathbf{z}(x). \tag{7.3}$$

- An *attentive mechanism* assigns a vector $\mathbf{z}_n^{(t)}$ to every target input $x^{(t)}$, which is represented as function just like context sets:

$$f\colon \{x_1^{(t)}, \ldots, x_N^{(t)}\} \to \{\mathbf{z}_1^{(t)}, \ldots, \mathbf{z}_N^{(t)}\}, \quad f(x_i^{(t)}) = \mathbf{z}_i^{(t)}. \tag{7.4}$$

- A *prediction* at target inputs $\mathbf{x}^{(t)} \in \mathcal{X}^N$ assigns some parameters $\boldsymbol{\theta}_i^{(t)}$ to every target input $x_i^{(t)}$, *e.g.* a marginal mean, a noise variance, and a covariance embedding (Appendix E.1). The prediction is represented as a function just like context sets:

$$f\colon \{x_1^{(t)}, \ldots, x_N^{(t)}\} \to \{\boldsymbol{\theta}_1^{(t)}, \ldots, \boldsymbol{\theta}_N^{(t)}\}, \quad f(x_i^{(t)}) = \boldsymbol{\theta}_i^{(t)}. \tag{7.5}$$

The above list is non-exhaustive, but it should give a sense of how objects can be represented as functions.

Representing everything with functions is convenient for two reasons:

1. Functions are flexible enough to naturally represent most objects of interest.

2. Functions are the right intermediate representation to express symmetries. For example, maps on functions can be translation equivariant (Definition 2.4).

Functional representations can be infinite dimensional. In `neuralprocesses`, however, functions are always maps between finite sets. For example, the functional encoding (Definition 4.6) in convolutional neural processes (ConvNPs; Section 5.3) is approximated with a discretisation (Procedure 5.5), which turns the infinite-dimensional functional encoding into a mapping between finite sets.

In `neuralprocesses`, functions between finite sets are represented as tuples `(xz, z)`. A tuple `(xz, z)` corresponds to the function `xz[i] ↦ z[i]` where `i` is a potentially multidimensional index. We write `(xz, z)` rather than `(x, y)` to emphasise that `(xz, z)` is an intermediate representation and is not necessarily connected to the context or target set.

In summary, in `neuralprocesses`, everything is represented with a function. These functions are always maps between finite sets and are represented by tuples `(xz, z)` where `xz` is the domain of the function and `z` the codomain.

## 7.4 Coders

By representing everything with functions, `encoder` and `decoder` become *transformations of functions*. In `neuralprocesses`, there is no fundamental distinction between `encoder` and `decoder`: both are implemented with the same building blocks. The central abstraction that we propose is called *coders*, which refers to a unification of `encoder` and `decoder`.

On a high level, coders are transformations of functions. Crucially, coders can be composed to make new coders. Coders form the fundamental building blocks which we will use to construct neural processes. In particular, `encoder` and `decoder` are coders and both are a composition of more elementary coders. We now give the definition of a coder.

**Definition 7.1** (Coder). *A coder is a transformation which transforms an* input function $\mathcal{X}_i \to Z_i$ *and a* target domain $\mathcal{X}_t$ *into an* output function $\mathcal{X}_o \to Z_o$.

The most important example of a coder is the map from a data set $D$ and some target inputs $\mathbf{x}$ to the prediction $P_\mathbf{x} \pi_\theta(D)$ for the corresponding target outputs. For this coder, the *input function* is the data set $D$, the *target domain* is the set of target inputs $\mathbf{x}$, and the *output function* is the prediction $P_\mathbf{x} \pi_\theta(D)$ for the target outputs.

In practice, a coder is a function `coder(xz, z, x)` where `(xz, z)` is the input function and `x` is the target domain. Moreover, the output of the coder `xz2, z2 = coder(xz, z, x)` is the output function:

$$\underbrace{\text{xz2, z2}}_{\text{output function}} = \text{coder}(\ \underbrace{\text{xz, z}}_{\text{input function}}\ ,\ \underbrace{\text{x}}_{\text{target domain}}\ ) \tag{7.6}$$

An important property of coders is that they can be composed. This property is what enables a compositional construction of neural processes. Let `coder1` and `coder2` be two coders. Then the composition is defined as follows:

```python
def compostion(xz, z, x):
    xz, z = coder1(xz, z, x)
    xz, z = coder2(xz, z, x)
    return xz, z
```

Note that, in the composition, when `coder2` is applied, `xz, z` are modified by `coder1`, but the target domain `x` is unaffected. In `neuralprocesses`, the composition of `coder1` and `coder2` can be constructed with `nps.Chain(coder1, coder2)`.

To succintly summarise, all building blocks are coders, and coders can be composed.

## 7.5  Building Existing and New Models

Having introduced coders, we now illustrate how this abstraction enables the design of neural processes by putting together building blocks in different ways. In `neuralprocesses`, the blueprint of a model is the following pattern:

```
encoder = nps.Chain(
    coder1,
    coder2,
    ⋮
    nps.⋯Likelihood(),
)
decoder = nps.Chain(
    coder3,
    coder4,
    ⋮
    nps.⋯Likelihood(),
)
model = nps.Model(encoder, decoder)
```

In this blueprint, the encoder and decoder are compositions of more elementary coders. Both compositions end in a so-called *likelihood*. The likelihood is a necessary component of `encoder` and `decoder` and must always be last in the composition. The purpose of the likelihood is to turn the output of the coder into a probabilistic prediction; `nps.⋯Likelihood()` determines the form of that prediction. For example, conditional neural processes (CNPs; Garnelo et al., 2018a) use a fixed, deterministic encoding. In that case, `encoder` uses `nps.DeterministicLikelihood()`. Latent-variable neural processes (LNPs; Garnelo et al., 2018b), on the other hand, use a latent variable. This can be achieved by using, *e.g.*, `nps.HeterogeneousGaussianLikelihood()`. We give two examples of models that illustrate the compositional approach that `neuralprocess` offers.

Our first example is the Conditional Neural Process (CNP; Garnelo et al., 2018a), which can be implemented as follows:

```
encoder = nps.Chain(
    nps.DeepSet(nps.MLP(⋯)),   # Deep set (Theorem 4.5) with φ_θ given by an MLP
    nps.DeterministicLikelihood(),
)
decoder = nps.Chain(
    nps.MLP(⋯),
```

```
    nps.HeterogeneousGaussianLikelihood(),
)
cnp = nps.Model(encoder, decoder)
```

Note that we suppress configuration settings of neural network components. Also note that the multi-layer perceptron (MLP) also functions as a coder. We can transform the CNP into a Neural Process (NP; Garnelo et al., 2018b) simply by replacing the deterministic likelihood with a stochastic one:

```
encoder = nps.Chain(
    nps.DeepSet(nps.MLP(···)),
    nps.MLP(···),
    nps.HeterogeneousGaussianLikelihood(),
)
decoder = nps.Chain(
    nps.MLP(···),
    nps.HeterogeneousGaussianLikelihood(),
)
np = nps.Model(encoder, decoder)
```

As Garnelo et al. (2018b) mention, this design can be extended by, in parallel with the stochastic encoding, also including a deterministic encoding:

```
encoder = nps.Parallel(
    nps.Chain(   # Deterministic encoder
        nps.DeepSet(nps.MLP(···)),
        nps.MLP(···),
        nps.DeterministicLikelihood(),
    ),
    nps.Chain(   # Stochastic encoder
        nps.DeepSet(nps.MLP(···)),
        nps.MLP(···),
        nps.HeterogeneousGaussianLikelihood(),
    ),
)
decoder = nps.Chain(
    nps.Concatenate(),   # Turn the two encodings into one
    nps.MLP(···),
    nps.HeterogeneousGaussianLikelihood(),
)
```

```
np = nps.Model(encoder, decoder)
```

In the deterministic encoder, we could replace `nps.DeepSet` with an attentive mechanism `nps.Attention`, which would create an Attentive Neural Process (ANP; Kim et al., 2019). Alternatively, rather than using two separate deep sets, perhaps one wants to use a single deep set in a two-headed architecture:

```
encoder = nps.Chain(
    nps.DeepSet(nps.MLP(···)),
    nps.MLP(···),
    nps.Splitter(···,···),
    nps.Parallel(
        nps.DeterministicLikelihood(),
        nps.HeterogeneousGaussianLikelihood(),
    ),
)
```

We see that the ability to compose elementary coders allows us to quickly explore various model designs.

Our second example is the Convolutional Conditional Neural Process (ConvCNP; Model 5.4), which can be implemented as follows:

```
encoder = nps.FunctionalCoder(
    # The encoder produces a functional encoding (Definition 4.6), which is
    # discretised (Procedure 5.5).
    nps.Discretisation(···),
    nps.Chain(
        nps.PrependDensityChannel(),
        # `nps.SetConv` performs interpolation with a Gaussian kernel like in
        # Procedure 5.5. The combination of coders in this `nps.Chain` implements
        # the encoder enc_θ of a convolutional deep set (Theorem 4.8).
        nps.SetConv(···),
        nps.DivideByFirstChannel(),
        nps.DeterministicLikelihood(),
    ),
)
decoder = nps.Chain(
    nps.UNet(···),
    nps.SetConv(···),
```

```
        nps.HeterogeneousGaussianLikelihood(),
)
convcnp = nps.Model(encoder, decoder)
```

To turn the ConvCNP into a Convolutional Gaussian Neural Process (ConvGNP; Model 5.12), we simply replace the likelihood of the decoder with `nps.LowRankGaussianLikelihood(···)`. And to turn the ConvCNP into a Convolutional Neural Process (ConvNP; Foong et al., 2020), we simply replace `nps.DeterministicLikelihood()` in the encoder with another `nps.UNet(···)` and `nps.HeterogeneousGaussianLikelihood()`. In Section 5.7, we briefly philosophised that a ConvGNP could even be used as the encoder and/or decoder in a ConvNP. Also this model is now simply explored:

```
encoder = nps.FunctionalCoder(
    nps.Discretisation(···),
    nps.Chain(
        nps.PrependDensityChannel(),
        nps.SetConv(···),
        nps.DivideByFirstChannel(),
        nps.UNet(···),
        nps.LowRankGaussianLikelihood(···),
    ),
)
decoder = nps.Chain(
    nps.UNet(···),
    nps.SetConv(···),
    nps.LowRankGaussianLikelihood(···),
)
novel_convnp = nps.Model(encoder, decoder)
```

Using a ConvGNP as the encoder and/or decoder in a ConvNP might improve predictive uncertainty when trained with the ELBO objective. This model has not yet been proposed in the literature.

## 7.6 Conclusion

In this chapter, we briefly explored a software abstraction called *coders*. In the Python package `neuralproceses`, the abstraction of coders is used to define building blocks that can be put together in different ways to construct neural process models. The simplicity of this compositional approach enables the user to rapidly explore a large model space.

By carefully defining a universal interface that all building blocks adhere to, the goal is to enable an implementation of the components of neural processes that are useful beyond this thesis. In an ideal world, there would only be a single, gold-standard implementation of every neural process component that is highly optimised and used by every project. Such an implementation needs to be flexible enough to suit every project's needs. This flexibility, however, needs to not come at the cost of simplicity of use. The abstraction of coders hopes to strike this balance right.

# 8 | Conclusion

The contributions of this thesis are best interpreted as new tools in the neural process toolbox. These new tools are certainly not meant to replace prior work, but rather to add to existing methods. Every tool has a particular regime where it works well and where it is the right choice. It is the art of the neural process practitioner to correctly identify which tools are best suited for a particular application. To conclude this thesis, we recapitulate the tools put forward in this thesis and provide advice for how the practitioner may decide to use them.

## 8.1 New Tools in the Neural Process Toolbox

The primary tool that this thesis puts forward are convolutional deep sets (Theorems 4.8 and 4.9). We used convolutional deep sets to construct convolutional neural processes (ConvNPs; Section 5.3). In problems where the data is roughly stationary, convolutional neural processes offer strong alternatives to existing neural processes. The models constructed in Section 5.3, however, are just a few of many. The more powerful idea is that convolutional deep sets are a flexible neural network building block that can be generally used in spatial, temporal or spatio–temporal data problems. A great illustration of this idea is Section 6.5, where, following the approach by Vaughan et al. (2022), we used neural processes to perform statistical downscaling. In the setup of the ConvCNP and ConvGNP, convolutional deep sets in combination with a pointwise MLP (Definition 6.1) provided a mechanism to combine coarse-grained ERA-Interim reanalysis variables with high-resolution elevation data (Vaughan et al., 2022); and in the multiscale architecture of the AR ConvCNP, a cascade of convolutional deep sets seamlessly stitched together convolutional neural networks operating at different resolutions. Although convolutional deep sets are flexible and can work well, we emphasise that the technique is not a panacea. The main limitation of convolutional deep sets are demanding computational requirements deriving from the discretisation (Procedure 5.5). Since we first published the Convolutional Conditional Neural Process (ConvCNP; Gordon et al., 2020), the ConvCNP has been extended to symmetries other than translation equivariance (Kawano et al., 2021; Holderrieth et al., 2021) and formed the basis for various models in a variety of applications (Foong et al., 2020; Shysheya, 2020; Petersen et al., 2021; Wang et al., 2021; Vaughan et al., 2021; Pondaven et al., 2022; Vaughan et al., 2022).

The second tool that this thesis proposes is the idea of directly parametrising the covariance between target outputs. This gave rise to Gaussian neural processes (GNPs; Section 5.5). GNP are a contender to latent-variable neural processes (LNPs; Garnelo et al., 2018b) by also modelling dependencies between target outputs. In problems where the data is roughly Gaussian, GNPs are a promising new model class to explore. Like with convolutional deep sets, the idea of directly parametrising the covariance between target outputs is useful beyond the models presented in Chapter 5. For example, we used the technique to enable correlated samples in the climate downscaling experiment (Section 6.5), and we illustrated that a GNP can even be the encoder and/or decoder in a LNP (Section 7.5), potentially combining the benefits of both classes. Besides the GNPs explored in this thesis, many more approaches to parametrising the covariance are possible. Since we first published the Fully Convolutional Gaussian Neural Process (FullConvGNP; Bruinsma et al., 2021c; Markou et al., 2022), we used the FullConvGNP to meta-learn posterior distributions in PAC-Bayes bounds (Foong et al., 2021).

The third and final tool presented by this thesis are autoregressive conditional neural processes (AR CNPs; Section 5.6). Although CNPs are the simplest of neural processes, by deploying CNPs in an autoregressive fashion, CNPs have the capacity to compete with much more sophisticated approaches. Most notably, across all experiments, Gaussian or non-Gaussian, the AR ConvCNP has consistently been amongst the best performing models. AR CNPs equip the neural process framework with a new knob where modelling complexity and computational expense at training time can be traded for computational expense at test time. AR CNPs have not yet been published.

## 8.2 Advice for the Neural Process Practitioner

The neural process framework offers a great variety of models. To choose a model, one should first consider the general strengths and weakness of the various classes of neural processes. Just like there is no free lunch, every class of neural processes gives up something: CNPs do not model dependencies; GNPs only produce Gaussian predictions; LNPs require approximations and consequently produce inferior uncertainty; and AR CNPs are no longer consistent and have high computational cost at test time. See also Table 5.2. Is it the job of the neural process practitioner to decide which shortcoming is most acceptable. Within a class of neural processes, choosing the right model depends on the application, but the summary in Section 6.6 could be used to guide your decision.

Rather than choosing an existing model, the neural process framework can also be used to tailor a solution to a particular data problem. For example, rather than making the choice between convolutional deep sets and an attentive mechanism, one may perfectly well combine the two. In fact, mixing and matching the building blocks of neural processes is the best way to utilise the neural process framework!

To incorporate context data into a neural process architecture, the three main generic approaches are deep sets (Theorem 4.5; Zaheer et al., 2017; Edwards et al., 2017; Garnelo et al., 2018b; Wagstaff et al., 2019), an attentive mechanism (Bahdanau et al., 2015; Vaswani et al., 2017; Kim et al., 2019), or convolutional deep sets (Theorems 4.8 and 4.9). We now give general considerations that may help deciding between the three. Deep sets are the most general and computationally cheapest approach, but are not very parameter efficient and might yield models that underfit. An attentive mechanism often improves over deep sets, generally yielding better performance. Attentive mechanisms, however, are computationally more expensive and do not construct a fixed-dimensional intermediate encoding like deep sets do. If the data is stationary or can be rendered approximately stationary by including more features, then convolutional deep sets can be used. Convolutional deep sets are only suitable for one, two, or three-dimensional inputs, and the computational cost is determined by the discretisation (Procedure 5.5). When convolutional deep sets are appropriate, they often yield superior performance.

To successfully deploy convolutional deep sets, we have a few words of more specific advice. To begin with, the discretisation must not be unnecessarily fine, because that incurs unnecessary computational expensive; nor must it be too coarse, because that loses important detail of the data. Our recommendation is to decide on the smallest length scale in the data that needs to be captured, and to make the interpoint spacing of the discretisation half or one-fourth of that length scale. Second, although theory says that the length scale of the Gaussian kernels may be anything, practice shows that it must be chosen right. Our experience is that models learn more quickly if the length scales are sufficiently small, but not too small. We advice to initialise the length scales of the Gaussian kernels to twice the interpoint spacing of the discretisation. Smaller length scales might introduce visual artefacts, and larger length scales might hamper performance. Third, the receptive field of the convolutional neural network (CNN) should be controlled explicitly. It should be set according to how far observations should influence predictions. For a typical CNN, the receptive field of a CNN is determined by the number of layers and kernel sizes of the convolutional filters. Finally, for a given receptive field, the capacity of the CNN should to be tuned appropriately, and here one can use all tricks in the book. A honourable mention is the U-Net architecture (Ronneberger et al., 2015), which effectively achieves the large receptive fields at reasonable parameter counts.

# A | Proofs for Chapter 3

## A.1 Proofs for Section 3.1

Recall the definition of the posterior prediction map in Definition 3.6. Henceforth, for all $D \in \mathcal{D}$, denote the Radon–Nikodym derivative of $\pi_f(D)$ with respect to $\mu_f$ by $\pi'(D)$:

$$\pi'(D) = \frac{r(\mathbf{y} - f(\mathbf{x}))}{Z(\mathbf{x}, \mathbf{y})} \tag{A.1}$$

where $D = (\mathbf{x}, \mathbf{y})$, $r(\mathbf{x}) = \exp(-\frac{1}{2\sigma_f^2}\|\mathbf{x}\|_2^2)$, and $Z(\mathbf{x}, \mathbf{y}) = \mathbb{E}_f[r(\mathbf{y} - f(\mathbf{x}))]$.

**Lemma A.1.** *Let $(D_i)_{i \geq 1} \subseteq \mathcal{D}$ be a bounded sequence and denote $D_i = (\mathbf{x}_i, \mathbf{y}_i)$. Then*

$$0 < \inf_{i \geq 1} Z(\mathbf{x}_i, \mathbf{y}_i) \leq \sup_{i \geq 1} Z(\mathbf{x}_i, \mathbf{y}_i) < \infty. \tag{A.2}$$

*Proof.* That the supremum is finite follows from that $r(\mathbf{x}) \leq 1$. For the infimum, let $M > 0$, and consider the event $A = \{f : \|f(\mathbf{x}_i)\|_\infty \leq M\}$. Then

$$\mathbb{E}_f[e^{-\frac{1}{2\sigma_f^2}\|\mathbf{y}_i - f(\mathbf{x}_i)\|_2^2}] \geq \mathbb{E}_f[\mathbb{1}_A e^{-\frac{1}{\sigma_f^2}(\|\mathbf{y}_i\|_2^2 + \|f(\mathbf{x}_i)\|_2^2)}] \geq \mathbb{P}(A)\, e^{-\frac{1}{\sigma_f^2}\|\mathbf{y}_i\|_2^2 - \frac{1}{\sigma^2}NM^2}. \tag{A.3}$$

To lower bound the probability, denote $V = \sup_{x \in \mathcal{X}} \|f(x)\|_{L^2}^2$. Note that $V < \infty$ by Assumption 3.5. Estimate

$$\mathbb{P}(A) = 1 - \mathbb{P}(A^c) \geq 1 - N \sup_{x \in \mathcal{X}} \mathbb{P}(|f(x)| \geq M) \geq 1 - \frac{NV}{M^2}. \tag{A.4}$$

Set $M = \sqrt{2NV}$ to find $\mathbb{P}(A) \geq \frac{1}{2}$. Hence

$$\mathbb{E}_f[e^{-\frac{1}{2\sigma_f^2}\|\mathbf{y}_i - f(\mathbf{x}_i)\|_2^2}] \geq \tfrac{1}{2} e^{-\frac{1}{\sigma_f^2}\|\mathbf{y}_i\|_2^2 - \frac{2}{\sigma_f^2}VN^2} \tag{A.5}$$

We conclude that the infimum is non-zero: $(\|\mathbf{y}_i\|_2^2)_{i \geq 1}$ is bounded, because $(D_i)_{i \geq 1}$ is bounded. □

**Proposition 3.8** (Regularity of posterior prediction map, part one).

(1) *For all data sets $D \in \mathcal{D}$, there exists a constant $c_D > 0$ such that*

$$\|f(x) - f(y)\|_{L^p(\pi_f(D))} \leq c_D |x-y|^\beta \quad \text{whenever} \quad |x-y| < r. \tag{A.6}$$

*In addition, $\|f(x)\|_{L^p(\pi_f(D))} < \infty$ for all $x \in \mathcal{X}$.*

(2) *$\pi_f$ is continuous.*

*Proof.* (1): Let $D \in \mathcal{D}$ and denote $D = (\mathbf{x}, \mathbf{y})$. Then use that

$$\|\cdot\|_{L^p(\pi_f(D))} = \mathbb{E}_f[\pi'(D)(f)(\cdot)^p]^{1/p} \leq \frac{1}{(Z(\mathbf{x},\mathbf{y}))^{1/p}} \|\cdot\|_{L^p}, \tag{A.7}$$

and we already noted that $Z(\mathbf{x}, \mathbf{y}) > 0$.

(2): Let $D_i \to D$ and let $L \colon C_b(\mathcal{X}, \mathcal{Y}) \to \mathbb{R}$ be a continuous and bounded function. To begin with, $(\mathbf{x}, \mathbf{y}) \mapsto Z(\mathbf{x}, \mathbf{y})$ is continuous by pathwise continuity of $f$ in combination with bounded convergence, and $(\mathbf{x}, \mathbf{y}) \mapsto r(\mathbf{y} - f(\mathbf{x}))$ is almost surely continuous also by pathwise continuity of $f$. Hence, for almost every $f$, $(D, f) \mapsto \pi'(D)(f)$ is continuous in $D$. In addition, $(D_i)_{i \geq 1}$ is bounded, so $\sup_{i \geq 1, f \in \mathcal{Y}^\mathcal{X}} \pi'(D_i)(f) < \infty$ by Lemma A.1. Therefore, by bounded convergence,

$$\lim_{i \to \infty} \mathbb{E}_{\pi_f(D_i)}[L(f)] = \lim_{i \to \infty} \mathbb{E}_f[\pi'(D_i)(f) L(f)] \tag{A.8}$$
$$= \mathbb{E}_f[\pi'(D)(f) L(f)] \tag{A.9}$$
$$= \mathbb{E}_{\pi_f(D)}[L(f)]. \tag{A.10}$$

Since $L$ was arbitrary, we conclude that $\pi_f(D_i) \rightharpoonup \pi_f(D)$. $\square$

## A.2 Proofs for Section 3.3

**Proposition 3.11.** *The neural process objective $\mathcal{L}_{\mathrm{NP}}$ is well defined.*

*Proof.* Let $(\pi, \sigma) \in \overline{\mathcal{M}}$ and $D \in \mathcal{D}$. We have $\pi_f(D) \in \mathcal{P}$, so there exists a measurable $A_1 \subseteq \widetilde{I}$ such that $\mathbf{x} \mapsto P_\mathbf{x}^{\sigma_f} \pi_f(D)$ is weakly continuous at all $\mathbf{x} \in A_1$ and $\int_{A_1} p(\mathbf{x}) \, \mathrm{d}\mathbf{x} = 1$. Similarly, $\pi(D) \in \mathcal{P}$, so there exists another measurable $A_2 \subseteq \widetilde{I}$ such that $\mathbf{x} \mapsto P_\mathbf{x}^\sigma \pi(D)$ is weakly continuous at all $\mathbf{x} \in A_2$ and $\int_{A_2} p(\mathbf{x}) \, \mathrm{d}\mathbf{x} = 1$. Set $A = A_1 \cap A_2$. Then still $\int_A p(\mathbf{x}) \, \mathrm{d}\mathbf{x} = 1$. Denote $h \colon \widetilde{I} \to \mathbb{R} \cup \{\infty\}$, $h(\mathbf{x}) = \mathrm{KL}(P_\mathbf{x}^{\sigma_f} \pi_f(D), P_\mathbf{x}^\sigma \pi(D))$. Let $\tau$ be all open sets of $\widetilde{I}$. By weak lower semi-continuity of the Kullback–Leibler divergence (Posner, 1975), the restriction $h|_A$ of $h$ to $A$ is a lower semi-continuous function with respect to the subspace topology $(A, A \cap \tau)$. In particular, $h|_A$ is measurable with respect to the Borel $\sigma$-algebra on $(A, A \cap \tau)$, which is equal to $\sigma(A \cap \tau) = A \cap \sigma(\tau) = A \cap \mathcal{B}(\widetilde{I})$. Since $A$ has probability one, this suggests that $h$ is also measurable, which we now argue. Let $y \in \mathbb{R}$. Decompose

$$\{\mathbf{x} \in \widetilde{I} : h(\mathbf{x}) \leq y\} = \{\mathbf{x} \in A : h(\mathbf{x}) \leq y\} \cup \{\mathbf{x} \in \widetilde{I} \setminus A : h(\mathbf{x}) \leq y\}. \tag{A.11}$$

By $A \cap \mathcal{B}(I)$–measurability of $h|_A$, $\{\mathbf{x} \in A : h(\mathbf{x}) \leq y\} \in A \cap \mathcal{B}(\widetilde{I}) \subseteq \mathcal{B}(\widetilde{I})$. In addition, $\{\mathbf{x} \in \widetilde{I} \setminus A : h(\mathbf{x}) \leq y\} \subseteq \widetilde{I} \setminus A$, so it is measurable by completeness of $p(\mathbf{x})$ (Assumption 3.2). Therefore, $\{\mathbf{x} \in \widetilde{I} : h(\mathbf{x}) \leq y\}$ is measurable. Since $y \in \mathbb{R}$ was arbitary, $h$ is measurable. We conclude that the expectation $\mathbb{E}_{p(\mathbf{x})}[\mathrm{KL}(P_{\mathbf{x}}^{\sigma_f}\pi_f(D), P_{\mathbf{x}}^\sigma \pi(D))]$ is well defined.

By Proposition 3.8.(2), the posterior prediction map $\pi_f$ is continuous. Note that $\pi$ is also continuous. Then, for any $\mathbf{x} \in I$ and $\sigma > 0$, $D \mapsto (P_{\mathbf{x}}^{\sigma_f}\pi_f(D), P_{\mathbf{x}}^\sigma\pi(D))$ is weakly continuous, so $D \mapsto \mathrm{KL}(P_{\mathbf{x}}^{\sigma_f}\pi_f(D), P_{\mathbf{x}}^\sigma\pi(D))$ is lower semi-continuous by weak lower semi-continuity of the Kullback–Leibler divergence (Posner, 1975). By Fatou's lemma, $D \mapsto \mathbb{E}_{p(\mathbf{x})}[\mathrm{KL}(P_{\mathbf{x}}^{\sigma_f}\pi_f(D), P_{\mathbf{x}}^\sigma\pi(D))]$ is then also lower semi-continuous. In particular, it is measurable, which means that $\mathbb{E}_{p(D)p(\mathbf{x})}[\mathrm{KL}(P_{\mathbf{x}}^{\sigma_f}\pi_f(D), P_{\mathbf{x}}^\sigma\pi(D))]$ is well defined. $\square$

**Proposition 3.12.** *The neural process objective $\mathcal{L}_{\mathrm{NP}}$ is lower semi-continuous.*

*Proof.* Consider $(\pi, \sigma) \mapsto P_{\mathbf{x}}^\sigma \pi(D)$. This function is weakly continuous; recall that the topology on $\overline{\mathcal{M}}$ is defined in Definition 3.9. The result then follows from Fatou's lemma in combination with weak lower semi-continuity of the Kullback–Leibler divergence (Posner, 1975). $\square$

**Lemma A.2.** *Consider $\mu_1, \mu_2 \in \mathcal{P}_{\mathrm{c}}$. Let $I' \subseteq I$ be dense. Then $P_{\mathbf{x}}^{\sigma_1}\mu_1 = P_{\mathbf{x}}^{\sigma_2}\mu_2$ for all $\mathbf{x} \in I'$ if and only if $\mu_1 = \mu_2$ and $\sigma_1 = \sigma_2$.*

*Proof.* The "if" is clear, so we show the "only if". Let $f^{(1)} \sim \mu_1$ and $f^{(2)} \sim \mu_2$. Fix some $\mathbf{x} \in I$, and let $(\mathbf{x}_1^{(\ell)}, \ldots, \mathbf{x}_\ell^{(\ell)})_{\ell \geq 1} \subseteq I$ be a sequence of tuples of vectors such that (1) every concatenation $\mathbf{x}_1^{(\ell)} \oplus \cdots \oplus \mathbf{x}_\ell^{(\ell)}$ is in $I'$; and (2) as $m \to \infty$, it holds that $\sup |\mathbf{x} - \mathbf{x}_k^{(\ell)}| \to 0$, where the supremum is over $\ell \geq m$ and $k = m, \ldots, \ell$. To construct such a sequence, take any sequence $(\mathbf{z}_\ell)_{\ell \geq 1} \subseteq I$ convergent to $\mathbf{x}$ and use density of $I'$ to find, for every $\ell \geq 1$, an element $\mathbf{x}_1^{(\ell)} \oplus \cdots \oplus \mathbf{x}_\ell^{(\ell)} \in I'$ which is at most distance $1/\ell$ from $\mathbf{z}_1 \oplus \cdots \oplus \mathbf{z}_\ell$.

For all $k \geq 1$, let $\boldsymbol{\varepsilon}_k^{(1)} \sim \mathcal{N}(\mathbf{0}, \sigma_1^2 \mathbf{I})$ and $\boldsymbol{\varepsilon}_k^{(2)} \sim \mathcal{N}(\mathbf{0}, \sigma_2^2 \mathbf{I})$. Set

$$\mathbf{y}_\ell^{(1)} = \frac{1}{\ell}\sum_{k=1}^\ell f^{(1)}(\mathbf{x}_k^{(\ell)}) + \frac{1}{\ell}\sum_{k=1}^\ell \boldsymbol{\varepsilon}_k^{(1)}, \quad \mathbf{y}_\ell^{(2)} = \frac{1}{\ell}\sum_{k=1}^\ell f^{(2)}(\mathbf{x}_k^{(\ell)}) + \frac{1}{\ell}\sum_{k=1}^\ell \boldsymbol{\varepsilon}_k^{(2)}. \tag{A.12}$$

By almost sure uniform continuity of $f^{(1)}$ and $f^{(2)}$ ($\mathcal{X}$ is compact) and the strong law of large numbers,
$$\mathbf{y}_\ell^{(1)} \to f^{(1)}(\mathbf{x}) \quad \text{and} \quad \mathbf{y}_\ell^{(2)} \to f^{(2)}(\mathbf{x}) \tag{A.13}$$
almost surely and therefore weakly. But, by assumption,
$$P^{\sigma_1}_{\mathbf{x}_1^{(\ell)} \oplus \ldots \oplus \mathbf{x}_\ell^{(\ell)}} \mu_1 = P^{\sigma_2}_{\mathbf{x}_1^{(\ell)} \oplus \ldots \oplus \mathbf{x}_\ell^{(\ell)}} \mu_2 \quad \text{for all } \ell \geq 1, \tag{A.14}$$
so $\mathbf{y}_\ell^{(1)} \stackrel{d}{=} \mathbf{y}_\ell^{(2)}$ for all $\ell \geq 1$. Therefore, since weak limits are unique, $f^{(1)}(\mathbf{x}) \stackrel{d}{=} f^{(2)}(\mathbf{x})$. Since $\mathbf{x} \in I$ was arbitrary, we conclude that $f^{(1)} \stackrel{d}{=} f^{(2)}$.

To find that also $\sigma_1 = \sigma_2$, repeat the above construction with $\mathbf{x}$ instead fixed to some $\tilde{\mathbf{x}} \in I'$. Denote $\tilde{\mathbf{y}}^{(1)} = f^{(1)}(\tilde{\mathbf{x}}) + \tilde{\boldsymbol{\varepsilon}}^{(1)}$ and $\tilde{\mathbf{y}}^{(2)} = f^{(2)}(\tilde{\mathbf{x}}) + \tilde{\boldsymbol{\varepsilon}}^{(2)}$. Then
$$(\mathbf{y}_\ell^{(1)}, \tilde{\mathbf{y}}^{(1)}) \to (f^{(1)}(\tilde{\mathbf{x}}), \tilde{\mathbf{y}}^{(1)}) \quad \text{and} \quad (\mathbf{y}_\ell^{(2)}, \tilde{\mathbf{y}}^{(2)}) \to (f^{(2)}(\tilde{\mathbf{x}}), \tilde{\mathbf{y}}^{(2)}) \tag{A.15}$$
weakly and $(\mathbf{y}_\ell^{(1)}, \tilde{\mathbf{y}}^{(1)}) \stackrel{d}{=} (\mathbf{y}_\ell^{(2)}, \tilde{\mathbf{y}}^{(2)})$ for all $\ell \geq 1$, so $(f^{(1)}(\tilde{\mathbf{x}}), \tilde{\mathbf{y}}^{(1)}) \stackrel{d}{=} (f^{(2)}(\tilde{\mathbf{x}}), \tilde{\mathbf{y}}^{(2)})$. Finally apply $(\mathbf{f}, \mathbf{y}) \mapsto \mathbf{y} - \mathbf{f}$ to find that $\tilde{\boldsymbol{\varepsilon}}^{(1)} \stackrel{d}{=} \tilde{\boldsymbol{\varepsilon}}^{(2)}$, so $\sigma_1 = \sigma_2$. □

**Proposition 3.13.** *Assume that $\widetilde{I} \subseteq I$ is dense. Then $(\pi, \sigma) \in \arg\min_{\overline{\mathcal{M}_c}} \mathcal{L}_{NP}$ if and only if*
$$\pi_f|_{\widetilde{\mathcal{D}}} = \pi|_{\widetilde{\mathcal{D}}} \quad \text{and} \quad \sigma = \sigma_f. \tag{A.16}$$

*Proof.* It is clear that $\mathcal{L}_{NP}(\pi_f, \sigma_f) = 0$. For the converse, assume that $\mathcal{L}_{NP}(\pi, \sigma) = 0$ for some $(\pi, \sigma) \in \overline{\mathcal{M}_c}$. Then there exists a measurable $A \subseteq \widetilde{\mathcal{D}}$ such that $\int_A p(D) \, dD = 1$ and
$$\mathbb{E}_{p(\mathbf{x})}[\text{KL}(P^{\sigma_f}_\mathbf{x} \pi_f(D), P^\sigma_\mathbf{x} \pi(D))] = 0 \quad \text{for all } D \in A. \tag{A.17}$$

Fix such a $D \in A$. Then there exists a measurable $B \subseteq \widetilde{I}$ such that $\int_B p(\mathbf{x}) \, d\mathbf{x} = 1$ and $P^{\sigma_f}_\mathbf{x} \pi_f(D) = P^\sigma_\mathbf{x} \pi(D)$ for all $\mathbf{x} \in B$. But $B$ is dense in $\widetilde{I}$, because $p(\mathbf{x})$ assigns positive probability to every open set (Assumption 3.2), which means that $B$ is also dense in $I$. Therefore, by Lemma A.2, $\pi_f(D) = \pi(D)$ and $\sigma_f = \sigma$. Similarly, $A$ is dense in $\widetilde{\mathcal{D}}$, because $p(D)$ assigns positive probability to every open set. Since $\pi_f$ and $\pi$ are two continuous functions that agree on a dense subset of $\widetilde{\mathcal{D}}$, $\pi_f$ and $\pi$ must agree on all of $\widetilde{\mathcal{D}}$. □

## A.3 Proofs for Section 3.4

**Proposition 3.27** (Characterisation of GNPA)**.** *Assume that $\inf_{\mathcal{Q}_G} \mathcal{L}_{NP} < \infty$. Also assume that $\widetilde{I}$ is dense in $I_N$ for some $N \geq 2$. Then a noisy prediction map $(\pi, \sigma) \in \mathcal{Q}_G$ is a GNPA if and only if*
$$m_\pi|_{\widetilde{\mathcal{D}}} = m_f|_{\widetilde{\mathcal{D}}}, \quad k_\pi|_{\widetilde{\mathcal{D}}} = k_f|_{\widetilde{\mathcal{D}}}, \quad \text{and} \quad \sigma = \sigma_f. \tag{A.18}$$

*Proof.* Let $\mathcal{P}_\lambda^N$ be the collection of distributions on $\mathbb{R}^N$ that (a) have a density with respect to the Lebesgue measure and (b) have a covariance matrix which is strictly positive definite. Let $\mathcal{P}_{\lambda,\mathrm{G}}^N \subseteq \mathcal{P}_\lambda^N$ be subcollection of distributions which are Gaussian. Consider $\mu \in \mathcal{P}_\lambda^N$ and $\nu \in \mathcal{P}_{\lambda,\mathrm{G}}^N$ such that $\inf_{\nu_\mathrm{G} \in \mathcal{P}_{\lambda,\mathrm{G}}^N} \mathrm{KL}(\mu, \nu_\mathrm{G}) < \infty$. Define

$$\mathrm{G}(\mu, \nu) = \mathrm{KL}(\mu, \nu) - \inf_{\nu_\mathrm{G} \in \mathcal{P}_{\lambda,\mathrm{G}}^N} \mathrm{KL}(\mu, \nu_\mathrm{G}) \tag{A.19}$$

Note that $\mathrm{G}(\mu, \nu) \geq 0$, because $\nu \in \mathcal{P}_{\lambda,\mathrm{G}}^N$. Moreover, a computation shows that $\mathrm{G}(\mu, \nu) = \mathrm{KL}(\mathcal{N}(\mu), \nu)$ where we denote $\mathcal{N}(\mu) = \mathcal{N}(\boldsymbol{\mu}, \boldsymbol{\Sigma})$ with $\boldsymbol{\mu}$ the mean vector of $\mu$ and $\boldsymbol{\Sigma}$ the covariance matrix of $\mu$. Therefore,

$$\mathrm{KL}(\mu, \nu) = \mathrm{G}(\mu, \nu) + \inf_{\nu_\mathrm{G} \in \mathcal{P}_{\lambda,\mathrm{G}}^N} \mathrm{KL}(\mu, \nu_\mathrm{G}) \tag{A.20}$$

and $\mathrm{G}(\mu, \nu) = 0$ if and only if $\nu = \mathcal{N}(\mu)$.

Consider $(\pi, \sigma) \in \mathcal{Q}_\mathrm{G}$. Let $\mathbf{x} \in \widetilde{I}$ and $D \in \widetilde{\mathcal{D}}$. Note that $P_\mathbf{x}^{\sigma_f} \pi_f(D) \in \mathcal{P}_\lambda^N$ and $P_\mathbf{x}^\sigma \pi(D) \in \mathcal{P}_{\lambda,\mathrm{G}}^N$. Moreover, using that $\sigma_f > 0$, $\inf_{\nu_\mathrm{G} \in \mathcal{P}_{\lambda,\mathrm{G}}^N} \mathrm{KL}(P_\mathbf{x}^{\sigma_f} \pi_f(D), \nu_\mathrm{G}) < \infty$, so

$$\begin{aligned}\mathrm{KL}(P_\mathbf{x}^{\sigma_f} \pi_f(D), P_\mathbf{x}^\sigma \pi(D)) \\ = \mathrm{G}(P_\mathbf{x}^{\sigma_f} \pi_f(D), P_\mathbf{x}^\sigma \pi(D)) + \inf_{\nu_\mathrm{G} \in \mathcal{P}_{\lambda,\mathrm{G}}^N} \mathrm{KL}(P_\mathbf{x}^{\sigma_f} \pi_f(D), \nu_\mathrm{G}).\end{aligned} \tag{A.21}$$

We therefore have the decomposition

$$\mathbb{E}_{p(D)p(\mathbf{x})}[\mathrm{KL}(P_\mathbf{x}^{\sigma_f} \pi_f(D), P_\mathbf{x}^\sigma \pi(D))] \tag{A.22}$$
$$= \underbrace{\mathbb{E}_{p(D)p(\mathbf{x})}[\mathrm{G}(P_\mathbf{x}^{\sigma_f} \pi_f(D), P_\mathbf{x}^\sigma \pi(D))]}_{\text{(i)}} + \underbrace{\mathbb{E}_{p(D)p(\mathbf{x})}\left[\inf_{\nu_\mathrm{G} \in \mathcal{P}_{\lambda,\mathrm{G}}^N} \mathrm{KL}(P_\mathbf{x}^{\sigma_f} \pi_f(D), \nu_\mathrm{G})\right]}_{\text{(ii)}}.$$

It is clear that (ii) $\leq \inf_{\mathcal{Q}_\mathrm{G}} \mathcal{L}_\mathrm{NP} < \infty$, so (ii) is finite. Hence, $(\pi, \sigma)$ minimises $\mathcal{L}_\mathrm{NP}$ over $\mathcal{Q}_\mathrm{G}$ if and only if $(\pi, \sigma)$ minimises (i) over $\mathcal{Q}_\mathrm{G}$.

Suppose that (i) is zero. Then there exists a measurable $A \subseteq \widetilde{\mathcal{D}}$ such that $\int_A p(D) \, \mathrm{d}D = 1$ and

$$\mathbb{E}_{p(\mathbf{x})}[\mathrm{G}(P_\mathbf{x}^{\sigma_f} \pi_f(D), P_\mathbf{x}^\sigma \pi(D))] = 0 \quad \text{for all } D \in A. \tag{A.23}$$

Fix such a $D \in A$. Then there exists a measurable $B \subseteq \widetilde{I}$ such that $\int_B p(\mathbf{x}) \, \mathrm{d}\mathbf{x} = 1$ and

$$\mathrm{G}(P_\mathbf{x}^{\sigma_f} \pi_f(D), P_\mathbf{x}^\sigma \pi(D)) = 0 \quad \text{for all } \mathbf{x} \in B. \tag{A.24}$$

Hence, $P_{\mathbf{x}}^{\sigma}\pi(D) = \mathcal{N}(P_{\mathbf{x}}^{\sigma_f}\pi_f(D))$ for all $\mathbf{x} \in B$, which means that

$$m_\pi(D)(x') = m_f(D)(x') \tag{A.25}$$
$$k_\pi(D)(x', y') + \sigma^2 \mathbb{1}(x' = y') = k_f(D)(x', y') + \sigma_f^2 \mathbb{1}(x' = y') \tag{A.26}$$

for all $\mathbf{x} \in B$ and $x', y' \in \{x_1, \ldots, x_{|\mathbf{x}|}\}$. But $B$ is dense in $\widetilde{I}$, because $p(\mathbf{x})$ assigns positive probability to every open set (Assumption 3.2). Additionally, by assumption, $\widetilde{I}$ is dense in $\mathcal{X}^N$ for some $N \geq 2$. Therefore, (A.25) and (A.26) hold for a collection of $(x', y')$ dense in $\mathcal{X}^2$. The condition that $N \geq 2$ is necessary, otherwise always $x' = y'$, so (A.26) would not necessarily hold for $x' \neq y'$. Taking appropriate limits, using that $m_\pi$, $m_f$, $k_\pi$, and $k_f$ are continuous, we hence find that

$$m_\pi|_B = m_f|_B, \qquad k_\pi|_B = k_f|_B, \qquad \text{and} \qquad \sigma = \sigma_f. \tag{A.27}$$

Finally, using density of $B$ in $\widetilde{\mathcal{D}}$ in combination with continuity of $\pi_f$ and $\pi$ gives

$$m_\pi|_{\widetilde{\mathcal{D}}} = m_f|_{\widetilde{\mathcal{D}}}, \qquad k_\pi|_{\widetilde{\mathcal{D}}} = k_f|_{\widetilde{\mathcal{D}}}, \qquad \text{and} \qquad \sigma = \sigma_f, \tag{A.28}$$

which is the desired condition. Conversely, suppose that (A.28) is satisfied. Then (A.25) and (A.26) are satisfied for all $x', y' \in \mathcal{X}$ and $D \in \widetilde{\mathcal{D}}$, so (i) is zero. □

**Proposition 3.26** (Characterisation of CNPA). *Assume that $\inf_{\mathcal{Q}_{\mathrm{G,MF}}} \mathcal{L}_{\mathrm{NP}} < \infty$. Also assume that $\widetilde{I}$ is dense in $I_N$ for some $N \geq 1$. Then a noisy prediction map $(\pi, \sigma) \in \mathcal{Q}_{\mathrm{G,MF}}$ is a CNPA if and only if*

$$m_\pi|_{\widetilde{\mathcal{D}}} = m_f|_{\widetilde{\mathcal{D}}} \quad \text{and} \quad v_\pi|_{\widetilde{\mathcal{D}}} + \sigma^2 = v_f|_{\widetilde{\mathcal{D}}} + \sigma_f^2, \tag{A.29}$$

*Proof.* The proof goes exactly like the proof of Proposition 3.27, with the following modifications. Additionally assume that the distributions in $\mathcal{P}_{\lambda,\mathrm{G}}^N$ factorise, *i.e.* that the dimensions are independent. Then $\mathrm{G}(\mu, \nu) = 0$ if and only if $\nu = \mathcal{N}_{\mathrm{d}}(\mu)$ where $\mathcal{N}_{\mathrm{d}}(\mu) = \mathcal{N}(\boldsymbol{\mu}, d(\boldsymbol{\Sigma}))$ with $d(\boldsymbol{\Sigma})$ the diagonal matrix with diagonal equal to $\mathrm{diag}(\boldsymbol{\Sigma})$. The key difference with respect to the proof of Proposition 3.27 is that $k_\pi(D)$ in (A.25) and (A.26) is now only evaluated at $x' = y'$, which yields a condition on the variance map. Because the variance map is a function of one argument, rather than requiring $N \geq 2$, the requirement $N \geq 1$ suffices. Since always $x' = y'$, the indicator functions in (A.25) and (A.26) cannot be used any more to find $\sigma = \sigma_f$; we now only find $v_\pi|_{\widetilde{\mathcal{D}}} + \sigma^2 = v_f|_{\widetilde{\mathcal{D}}} + \sigma_f^2$. □

**Proposition 3.28** (Regularity of GNPA). *Let $(\pi, \sigma) \in \mathcal{Q}_{\mathrm{G}}$ be a Gaussian neural process approximation. Then, for all data sets $D \in \widetilde{\mathcal{D}}$,*

$$\|f(x) - f(y)\|_{L^2(\pi(D))} \leq c_D |x - y|^\beta \quad \text{whenever} \quad |x - y| < r, \tag{A.30}$$

where $c_D > 0$ is the constant from Proposition 3.8.(1) and the Hölder exponent $\beta \in (\frac{1}{p}, 1]$ and radius $r > 0$ are from Assumption 3.5.

*Proof.* Let $D \in \widetilde{\mathcal{D}}$. By Proposition 3.27, $m_{\pi_\mathrm{G}}(D) = m_f(D)$ and $k_{\pi_\mathrm{G}}(D) = k_f(D)$. Therefore, for any $x, y \in \mathcal{X}$,

$$\|f(x) - f(y)\|_{L^2(\pi_\mathrm{G}(D))} = \|f(x) - f(y)\|_{L^2(\pi_f(D))}. \tag{A.31}$$

The result then follows from $\|\cdot\|_{L^2(\pi_f(D))} \leq \|\cdot\|_{L^p(\pi_f(D))}$ ($p \geq 2$; see Assumption 3.5) in combination with Proposition 3.8.(1). $\square$

## A.4 Proofs for Section 3.5

**Lemma A.3.**
$$|r(\mathbf{x}_1) - r(\mathbf{x}_2)| \leq \frac{1}{\underline{\sigma}} \|\mathbf{x}_1 - \mathbf{x}_2\|_2. \tag{A.32}$$

*Proof.* For $x_1, x_2 \in \mathbb{R}$, using the fundamental theorem of calculus,

$$|e^{-\frac{1}{2}x_1^2} - e^{-\frac{1}{2}x_2^2}| = \left|\int_{x_1}^{x_2} xe^{-\frac{1}{2}x^2} \, dx\right| \leq \|x \mapsto xe^{-\frac{1}{2}x^2}\|_\infty |x_1 - x_2|, \tag{A.33}$$

and $\|x \mapsto xe^{-\frac{1}{2}x^2}\|_\infty \leq 1$. Set $x_1 = \frac{1}{\sigma}\|\mathbf{x}_1\|_2$ and $x_2 = \frac{1}{\sigma}\|\mathbf{x}_2\|_2$, so

$$|r(\mathbf{x}_1) - r(\mathbf{x}_2)| \leq \frac{1}{\sigma}|\|\mathbf{x}_1\|_2 - \|\mathbf{x}_2\|_2| \leq \frac{1}{\underline{\sigma}}\|\mathbf{x}_1 - \mathbf{x}_2\|_2, \tag{A.34}$$

which concludes. $\square$

**Lemma A.4.** *There exist universal $c_\ell$ and $c_u$ such that, for all $(\mathbf{x}, \mathbf{y}) \in \widetilde{\mathcal{D}}$,*

$$0 < c_\ell \leq Z(\mathbf{x}, \mathbf{y}) \leq c_u < \infty. \tag{A.35}$$

*In particular, $\frac{1}{2}e^{-\frac{1}{\underline{\sigma}^2} B_{\widetilde{\mathcal{D}}}^2 (1 + 2B_f^2)} \leq c_\ell$ and $c_u \leq 1$.*

*Proof.* The proof is similar to Lemma A.1. For the lower bound on $c_\ell$, see (A.5). $\square$

For $f \in C(\mathcal{X}, \mathcal{Y})$, define the *modulus of continuity* by

$$\omega_f(h) = \sup_{x,y \in \mathcal{X} : |x-y| < h} |f(x) - f(y)|. \tag{A.36}$$

**Proposition A.5.** *Under Assumption 3.5,*

$$\mathbb{P}(\omega_f(h) \geq \varepsilon) \leq \frac{c_{\beta,p}}{\varepsilon} h^{\beta - \frac{1}{p}} \qquad (A.37)$$

*where $c_{\beta,p} > 0$ is a constant only depending on $c$, $\beta$, and $p$.*

*Proof.* This follows directly from Proposition A.8. Proposition A.8 is a stronger result which we will prove later. □

**Proposition A.6.** *There exists a universal $c_Z > 0$ such that, for all $D_1, D_2 \in \mathcal{D}$,*

$$|Z(\mathbf{x}_1, \mathbf{y}_1) - Z(\mathbf{x}_2, \mathbf{y}_2)| \leq c_Z \, d_\mathcal{D}(D_1, D_2)^{\frac{1}{2}(\beta - \frac{1}{p})} \quad \text{whenever} \quad d_\mathcal{D}(D_1, D_2) < 1 \qquad (A.38)$$

*where $D_1 = (\mathbf{x}_1, \mathbf{y}_1)$ and $D_2 = (\mathbf{x}_2, \mathbf{y}_2)$.*

*Proof.* Suppose that $D_1, D_2 \in \mathcal{D}^N$. Consider the event $A = \{f : \omega_f(\|\mathbf{x}_1 - \mathbf{x}_2\|_\infty) < \varepsilon\}$. We have

$$|Z(\mathbf{x}_1, \mathbf{y}_1) - Z(\mathbf{x}_2, \mathbf{y}_2)| \leq |\mathbb{E}_f[(r(\mathbf{y}_1 - f(\mathbf{x}_1)) - r(\mathbf{y}_2 - f(\mathbf{x}_2)))\mathbb{1}_A]| + 2\mathbb{P}(A^c). \qquad (A.39)$$

We bound the two terms separately, starting with the second term:

$$\mathbb{P}(A^c) = \mathbb{P}(\omega_f(\|\mathbf{x}_1 - \mathbf{x}_2\|_\infty) \geq \varepsilon) \leq \frac{c_{\beta,p}}{\varepsilon} \|\mathbf{x}_1 - \mathbf{x}_2\|_\infty^{\beta - \frac{1}{p}}. \qquad (A.40)$$

For the first term, simply take a supremum and use Lemma A.3:

$$|\mathbb{E}_f[\mathbb{1}_A(r(\mathbf{y}_1 - f(\mathbf{x}_1)) - r(\mathbf{y}_2 - f(\mathbf{x}_2)))]|$$

$$\leq \sup_{f \in A} |r(\mathbf{y}_1 - f(\mathbf{x}_1)) - r(\mathbf{y}_2 - f(\mathbf{x}_2))| \qquad (A.41)$$

$$\leq \frac{1}{\underline{\sigma}} \sup_{f \in A} (\|f(\mathbf{x}_1) - f(\mathbf{x}_2)\|_2 + \|\mathbf{y}_1 - \mathbf{y}_2\|_2) \qquad (A.42)$$

$$\leq \frac{1}{\underline{\sigma}} \sup_{f \in A} (\sqrt{N}\omega_f(\|\mathbf{x}_1 - \mathbf{x}_2\|_\infty) + \|\mathbf{y}_1 - \mathbf{y}_2\|_2) \qquad (A.43)$$

$$\leq \frac{1}{\underline{\sigma}} (\sqrt{N}\varepsilon + \|\mathbf{y}_1 - \mathbf{y}_2\|_2), \qquad (A.44)$$

where the last inequality follows from that $\omega_f(\|\mathbf{x}_1 - \mathbf{x}_2\|_\infty) \leq \varepsilon$ for all $f \in A$. Putting the bounds for the two terms together, we have

$$|Z(\mathbf{x}_1, \mathbf{y}_1) - Z(\mathbf{x}_2, \mathbf{y}_2)| \leq \frac{2c_{\beta,p}}{\varepsilon} \|\mathbf{x}_1 - \mathbf{x}_2\|_\infty^{\beta - \frac{1}{p}} + \frac{1}{\underline{\sigma}}(\sqrt{N}\varepsilon + \|\mathbf{y}_1 - \mathbf{y}_2\|_2). \qquad (A.45)$$

This holds for any choice of $\varepsilon > 0$, so we may also take an infimum over $\varepsilon > 0$. For $a, b \geq 0$,

note that the infimum of $a/\varepsilon + b$ over $\varepsilon > 0$ is given by $2\sqrt{ab}$. Therefore,

$$|Z(\mathbf{x}_1, \mathbf{y}_1) - Z(\mathbf{x}_2, \mathbf{y}_2)| \leq \frac{2\sqrt{2}}{\sqrt{\underline{\sigma}}} (\sqrt{N} c_{\beta,p})^{\frac{1}{2}} \|\mathbf{x}_1 - \mathbf{x}_2\|_{\infty}^{\frac{1}{2}(\beta - \frac{1}{p})} + \frac{1}{\underline{\sigma}} \|\mathbf{y}_1 - \mathbf{y}_2\|_2. \quad (A.46)$$

Finally note that $\|\mathbf{x}_1 - \mathbf{x}_2\|_\infty < 1$ and $\|\mathbf{y}_1 - \mathbf{y}_2\|_2 < 1$ whenever $d_\mathcal{D}(D_1, D_2) < 1$ to find the result. □

**Proposition A.7.** *Let $D_1, D_2 \in \mathcal{D}_N \cap \widetilde{\mathcal{D}}$ and denote $D_1 = (\mathbf{x}_1, \mathbf{y}_1)$ and $D_2 = (\mathbf{x}_2, \mathbf{y}_2)$. Let $F \colon C(\mathcal{X}, \mathcal{Y}) \to \mathbb{R}$ be a continuous functional such that, for some $\gamma' > 0$, $B_F = \|F(f)\|_{L^{1+\gamma'}} < \infty$. Then there exists a function $L \colon [0, \infty) \to [0, \infty)$ such that*

$$|\mathbb{E}_{\pi(D_1)}[F(f)] - \mathbb{E}_{\pi(D_2)}[F(f)]| \leq L(d(D_1, D_2)). \quad (A.47)$$

*Moreover, $L$ only depends on the universal parameters, $\gamma'$, and $B_F$; and $L$ goes to zero as $d(D_1, D_2)$ goes to zero.*

*Proof.* Consider the event $A = \{f : \omega_f(\|\mathbf{x}_1 - \mathbf{x}_2\|_\infty) < \varepsilon\}$. Denote $g(\mathbf{x}, \mathbf{y}) = r(\mathbf{y} - f(\mathbf{x}))/Z(\mathbf{x}, \mathbf{y})$ and note that $g(\mathbf{x}_1, \mathbf{y}_1) \leq 1/c_\ell$ and $g(\mathbf{x}_2, \mathbf{y}_2) \leq 1/c_\ell$, where $c_\ell$ is defined in Lemma A.4. Then

$$|\mathbb{E}_{\pi(D_1)}[F(f)] - \mathbb{E}_{\pi(D_2)}[F(f)]|$$
$$= |\mathbb{E}_f[(g(\mathbf{x}_1, \mathbf{y}_1) - g(\mathbf{x}_2, \mathbf{y}_2))F(f)]| \quad (A.48)$$
$$\leq |\mathbb{E}_f[\mathbb{1}_A(g(\mathbf{x}_1, \mathbf{y}_1) - g(\mathbf{x}_2, \mathbf{y}_2))F(f)]| + \frac{2}{c_\ell} \mathbb{E}_f[\mathbb{1}_{A^c}|F(f)|]. \quad (A.49)$$

We bound the two terms separately, starting with the second term:

$$\mathbb{E}[\mathbb{1}_{A^c}|F(f)|] \leq \|F(f)\|_{L^{1+\gamma'}} \mathbb{P}(A^c)^{\frac{\gamma'}{1+\gamma'}} \leq \|F(f)\|_{L^{1+\gamma'}} \left[ \frac{c_{\beta,p}}{\varepsilon} \|\mathbf{x}_1 - \mathbf{x}_2\|_\infty^{\beta - \frac{1}{p}} \right]^{\frac{\gamma'}{1+\gamma'}}. \quad (A.50)$$

For the first term, observe that, for any two functions $f$ and $g$ such that $0 < c_\ell \leq g$ and $f, g \leq c_u < \infty$,

$$\left| \frac{f(x_1)}{g(x_1)} - \frac{f(x_2)}{g(x_2)} \right| \leq \frac{c_u}{c_\ell^2} (|f(x_1) - f(x_2)| + |g(x_1) - g(x_2)|). \quad (A.51)$$

Therefore, taking a supremum,

$$|\mathbb{E}_f[\mathbb{1}_A(g(\mathbf{x}_1, \mathbf{y}_1) - g(\mathbf{x}_2, \mathbf{y}_2))F(f)]| \quad (A.52)$$
$$\leq \frac{c_u}{c_\ell^2} \|F(f)\|_{L^1} \sup_{f \in A} \left[ |r(\mathbf{y}_1 - f(\mathbf{x}_1)) - r(\mathbf{y}_2 - f(\mathbf{x}_2))| + |Z(\mathbf{x}_1, \mathbf{y}_1) - Z(\mathbf{x}_2, \mathbf{y}_2)| \right].$$

Here, using Lemma A.3 and Proposition A.6,

$$|r(\mathbf{y}_1 - f(\mathbf{x}_1)) - r(\mathbf{y}_2 - f(\mathbf{x}_2))| \le \frac{1}{\underline{\sigma}}(\|f(\mathbf{x}_1) - f(\mathbf{x}_2)\|_2 + \|\mathbf{y}_1 - \mathbf{y}_2\|_2) \quad (A.53)$$

$$\le \frac{1}{\underline{\sigma}}(\sqrt{N}\omega_f(\|\mathbf{x}_1 - \mathbf{x}_2\|_\infty) + \|\mathbf{y}_1 - \mathbf{y}_2\|_2), \quad (A.54)$$

$$|Z(\mathbf{x}_1, \mathbf{y}_1) - Z(\mathbf{x}_2, \mathbf{y}_2)| \le L_Z(d(D_1, D_2)), \quad (A.55)$$

so, using that $\omega_f(\|\mathbf{x}_1 - \mathbf{x}_2\|_\infty) \le \varepsilon$ for $f \in A$,

$$|\mathbb{E}_f[\mathbb{1}_A(g(\mathbf{x}_1, \mathbf{y}_1) - g(\mathbf{x}_2, \mathbf{y}_2))F(f)]|$$
$$\le \frac{c_u}{c_\ell^2}\|F(f)\|_{L^1}\left[\frac{1}{\underline{\sigma}}(\sqrt{N}\varepsilon + \|\mathbf{y}_1 - \mathbf{y}_2\|_2) + L_Z(d(D_1, D_2))\right]. \quad (A.56)$$

Putting the bounds for the two terms together, we have

$$|\mathbb{E}_{\pi(D_1)}[F(f)] - \mathbb{E}_{\pi(D_2)}[F(f)]|$$
$$\le \frac{2}{c_\ell}\|F(f)\|_{L^{1+\gamma'}}\left[\frac{c_{\beta,p}}{\varepsilon}\|\mathbf{x}_1 - \mathbf{x}_2\|_\infty^{\beta - \frac{1}{p}}\right]^{\frac{\gamma'}{1+\gamma'}}$$
$$+ \frac{c_u}{c_\ell^2}\|F(f)\|_{L^1}\left[\frac{1}{\underline{\sigma}}(\sqrt{N}\varepsilon + \|\mathbf{y}_1 - \mathbf{y}_2\|_2) + L_Z(d(D_1, D_2))\right], \quad (A.57)$$

which we simplify to

$$\frac{c_\ell}{2}\|F(f)\|_{L^{1+\gamma'}}^{-1}|\mathbb{E}_{\pi(D_1)}[F(f)] - \mathbb{E}_{\pi(D_2)}[F(f)]| \quad (A.58)$$
$$\le \left[\frac{c_{\beta,p}}{\varepsilon}\|\mathbf{x}_1 - \mathbf{x}_2\|_\infty^{\beta - \frac{1}{p}}\right]^{\frac{\gamma'}{1+\gamma'}} + \frac{c_u}{c_\ell}\left[\frac{1}{\underline{\sigma}}(\sqrt{N}\varepsilon + \|\mathbf{y}_1 - \mathbf{y}_2\|_2) + L_Z(d(D_1, D_2))\right].$$

This holds for any choice of $\varepsilon > 0$, so we may also take an infimum over $\varepsilon > 0$. Temporarily denote $\rho = \frac{\gamma'}{1+\gamma'}$. For $a, b \ge 0$ and $\rho > 0$, note that the infimum of $(a/\varepsilon)^\rho + b\varepsilon$ over $\varepsilon > 0$ is given by $c_\rho(ab)^{\frac{\rho}{1+\rho}}$ with $c_\rho = \rho^{\frac{1}{1+\rho}} + \rho^{-\frac{\rho}{1+\rho}}$. Therefore,

$$\frac{c_\ell}{2}\|F(f)\|_{L^{1+\gamma'}}^{-1}|\mathbb{E}_{\pi(D_1)}[F(f)] - \mathbb{E}_{\pi(D_2)}[F(f)]| \quad (A.59)$$
$$\le c_\rho\left[\frac{c_u}{c_\ell}\frac{\sqrt{N}}{\underline{\sigma}}c_{\beta,p}\|\mathbf{x}_1 - \mathbf{x}_2\|_\infty^{\beta - \frac{1}{p}}\right]^{\frac{\gamma'}{1+2\gamma'}} + \frac{c_u}{c_\ell}\left[\frac{1}{\underline{\sigma}}\|\mathbf{y}_1 - \mathbf{y}_2\|_2 + L_Z(d(D_1, D_2))\right],$$

noting that $\frac{\rho}{1+\rho} = \frac{\gamma'}{1+2\gamma'}$. $\square$

**Proposition 3.33** (Regularity of posterior prediction map, part two).

(1) *There exist universal constants $c_m > 0$ and $c_k > 0$ such that, for any two $D_1, D_2 \in \widetilde{\mathcal{D}}$,*

*whenever* $d_{\mathcal{D}}(D_1, D_2) < 1$, *then*

$$\sup_{x \in \mathcal{X}} |\mathbb{E}_{\pi_f(D_1)}[f(x)] - \mathbb{E}_{\pi_f(D_2)}[f(x)]| \leq c_m \, d_{\mathcal{D}}(D_1, D_2)^{\frac{2+2\gamma}{3+2\gamma}(\beta - \frac{1}{p})}, \quad \text{(A.60)}$$

$$\sup_{x,y \in \mathcal{X}} |\mathbb{E}_{\pi_f(D_1)}[f(x)f(y)] - \mathbb{E}_{\pi_f(D_2)}[f(x)f(y)]| \leq c_k \, d_{\mathcal{D}}(D_1, D_2)^{\frac{\gamma}{1+\gamma}(\beta - \frac{1}{p})}. \quad \text{(A.61)}$$

(2) *Consequently, for any* $D_1, D_2 \in \widetilde{\mathcal{D}}$ *and* $x, y \in \mathcal{X}$, *whenever* $d_{\mathcal{D}}(D_1, D_2) < 1$, $|x_1 - x_2| < r$, *and* $|y_1 - y_2| < r$, *then*

$$|\mathbb{E}_{\pi_f(D_1)}[f(x_1)] - \mathbb{E}_{\pi_f(D_2)}[f(x_2)]| \leq c_m \, d_{\mathcal{D}}(D_1, D_2)^{\frac{2+2\gamma}{3+2\gamma}(\beta - \frac{1}{p})} + \frac{c}{c_\ell}|x_1 - x_2|^\beta \quad \text{(A.62)}$$

*and*

$$|\mathbb{E}_{\pi_f(D_1)}[f(x_1)f(y_1)] - \mathbb{E}_{\pi_f(D_2)}[f(x_2)f(y_2)]|$$
$$\leq c_k \, d_{\mathcal{D}}(D_1, D_2)^{\frac{\gamma}{1+\gamma}(\beta - \frac{1}{p})} + B_f \frac{c}{c_\ell^2}|x_1 - x_2|^\beta + B_f \frac{c}{c_\ell^2}|y_1 - y_2|^\beta, \quad \text{(A.63)}$$

*where* $c_\ell > 0$ *is a universal constant from Lemma A.4.*

*Proof.* (1): For the mean, choose some $x \in \mathcal{X}$, apply Proposition A.7 with $F(f) = f(x)$ and $\gamma' = 1 + \gamma$, and verify that

$$\|f(x)\|_{L^{1+\gamma'}} \leq \sup_{x \in \mathcal{X}} \|f(x)\|_{L^{2+\gamma}} = B_f, \quad \text{(A.64)}$$

which only depends on the universal parameters. For the covariance, choose two $x, y \in \mathcal{X}$, apply Proposition A.7 with $F(f) = f(x)f(y)$ and $\gamma' = \frac{1}{2}\gamma$, and verify that

$$\|f(x)f(y)\|_{L^{1+\gamma'}} \leq \|f(x)\|_{L^{2+2\gamma'}}\|f(y)\|_{L^{2+2\gamma'}} \leq \sup_{x \in \mathcal{X}} \|f(x)\|_{L^{2+\gamma}}^2 = B_f^2 \quad \text{(A.65)}$$

which again only depends on the universal parameters.

(2): We will repeatedly use the observation that, for any $q \geq 1$ and $D \in \widetilde{\mathcal{D}}$,

$$\|\cdot\|_{L^q(\pi(D))} \leq \frac{1}{c_\ell^{1/q}} \|\cdot\|_{L^q} \quad \text{(A.66)}$$

where $c_\ell$ is from Lemma A.4. We first prove the inequality for the mean:

$$|\mathbb{E}_{\pi(D_1)}[f(x_1)] - \mathbb{E}_{\pi(D_2)}[f(x_2)]|$$
$$\leq |\mathbb{E}_{\pi(D_1)}[f(x_1)] - \mathbb{E}_{\pi(D_2)}[f(x_1)]| + |\mathbb{E}_{\pi(D_2)}[f(x_1)] - \mathbb{E}_{\pi(D_2)}[f(x_2)]| \quad \text{(A.67)}$$
$$\leq L_m(d(D_1, D_2)) + \|f(x_1) - f(x_2)\|_{L^1(\pi(D_2))} \quad \text{(A.68)}$$
$$\leq L_m(d(D_1, D_2)) + \frac{c}{c_\ell}|x_1 - x_2|^\beta \quad \text{(A.69)}$$

where the second inequality follows from part (2) and the third inequality from the observation combined with Assumption 3.5. For the covariance,

$$|\mathbb{E}_{\pi(D_1)}[f(x_1)f(y_1)] - \mathbb{E}_{\pi(D_2)}[f(x_2)f(y_2)]|$$
$$\leq |\mathbb{E}_{\pi(D_1)}[f(x_1)f(y_1)] - \mathbb{E}_{\pi(D_2)}[f(x_1)f(y_1)]|$$
$$\quad + |\mathbb{E}_{\pi(D_2)}[f(x_1)f(y_1)] - \mathbb{E}_{\pi(D_2)}[f(x_1)f(y_2)]|$$
$$\quad + |\mathbb{E}_{\pi(D_2)}[f(x_1)f(y_2)] - \mathbb{E}_{\pi(D_2)}[f(x_2)f(y_2)]|. \quad \text{(A.70)}$$

From part (2), we have

$$|\mathbb{E}_{\pi(D_1)}[f(x_1)f(y_1)] - \mathbb{E}_{\pi(D_2)}[f(x_1)f(y_1)]| \leq L_k(d(D_1, D_2)). \quad \text{(A.71)}$$

We also have

$$|\mathbb{E}_{\pi(D_2)}[f(x_1)(f(y_1) - f(y_2))]| \leq \|f(x_1)\|_{L^2(\pi(D_2))} \|f(y_1) - f(y_2)\|_{L^2(\pi(D_2))} \quad \text{(A.72)}$$
$$\leq \frac{1}{c_\ell}\|f(x_1)\|_{L^2} \|f(y_1) - f(y_2)\|_{L^2} \quad \text{(A.73)}$$
$$\leq B_f \frac{c}{c_\ell}|y_1 - y_2|^\beta. \quad \text{(A.74)}$$

Therefore,

$$|\mathbb{E}_{\pi(D_1)}[f(x_1)f(y_1)] - \mathbb{E}_{\pi(D_2)}[f(x_2)f(y_2)]|$$
$$\leq L_k(d(D_1, D_2)) + B_f \frac{c}{c_\ell}|x_1 - x_2|^\beta + B_f \frac{c}{c_\ell}|y_1 - y_2|^\beta, \quad \text{(A.75)}$$

which concludes. $\square$

**Proposition 3.34** (Consistency of CNPA). *Assume that $\widetilde{I}$ is dense in $I_N$ for some $N \geq 1$, and assume that $\sup_{\mathbf{x} \in \widetilde{I}} |\mathbf{x}| < \infty$. Let $(\pi_M, \sigma_M) \in \widetilde{\mathcal{Q}}_{\text{G,MF}}$ be such that*

$$\mathcal{L}_M(\pi_M, \sigma_M) \leq \inf_{\widetilde{\mathcal{Q}}_{\text{G,MF}}} \mathcal{L}_M + o_{\mathbb{P}}(1). \quad \text{(A.76)}$$

*Then, as $M \to \infty$, the distance of $(\pi_M, \sigma_M)$ to the closest CNPA converges to zero in probability.*

*Proof.* The result follows directly from Proposition 3.26 in combination with Theorem 5.4.1 by van der Vaart (1998). It remains to verify the conditions for Theorem 5.4.1. Convergence in the metric on $\widetilde{\mathcal{Q}}_{\text{G,MF}}$ (see (3.28)) implies convergence of all means and variances, which is stronger than the required continuity condition. For the integrability condition, combine the following three facts. First, by assumption, the maximum target set size is bounded. Second, the universal Hölder condition implies that $\sup_{\pi \in \widetilde{\mathcal{Q}}_{\text{G,MF}}, D \in \widetilde{\mathcal{D}}, x \in \mathcal{X}} \mathbb{E}_{\pi(D)}[f^2(x)] < \infty$, which bounds all means and covariances. Third, by Assumption 3.30 and the following paragraph, every noise variance $\sigma$ is in $[\underline{\sigma}, \overline{\sigma}]$. □

**Proposition 3.35** (Consistency of GNPA). *Assume that $\widetilde{I}$ is dense in $I_N$ for some $N \geq 2$, and assume that $\sup_{\mathbf{x} \in \widetilde{I}} |\mathbf{x}| < \infty$. Let $(\pi_M, \sigma_M) \in \widetilde{\mathcal{Q}}_G$ be such that*

$$\mathcal{L}_M(\pi_M, \sigma_M) \leq \inf_{\widetilde{\mathcal{Q}}_G} \mathcal{L}_M + o_{\mathbb{P}}(1). \tag{A.77}$$

*Then, as $M \to \infty$, the distance of $(\pi_M, \sigma_M)$ to the GNPA converges to zero in probability.*

*Proof.* The proof goes exactly like the proof of Proposition 3.34. □

**Proposition A.8.** *Consider $(\mu_i)_{i \geq 1} \subseteq \mathcal{P}_c$. Suppose there exist $p > 0$, $\beta > \frac{1}{p}$, a constant $c > 0$, and a radius $r > 0$ such that*

$$\sup_{i \geq 1} \|f(x) - f(y)\|_{L^p(\mu_i)} \leq c|x-y|^\beta \quad \text{whenever} \quad |x-y| < r. \tag{A.78}$$

*Then, for all $\varepsilon > 0$,*

$$\sup_{i \geq 1} \mu_i(\omega_f(h) \geq \varepsilon) \leq \frac{c}{\varepsilon} \frac{2}{1 - 2^{-(\beta - \frac{1}{p})}} h^{\beta - \frac{1}{p}}. \tag{A.79}$$

*Proof.* Choose $k \in \mathbb{N}$ such that $2^{-(k+1)} \leq h \leq 2^{-k}$. The proof strategy mimics the proof of Theorem 4.2.1 by Norris (2018). Let $i \in \mathbb{N}$. For $n \in \mathbb{N}$, denote $\mathbb{D}_n = \{0, 2^{-n}, 2 \cdot 2^{-n}, \ldots, 1\}$. Consider $f \in C([0,1], \mathcal{Y})$. Set

$$K_n = \sup_{x \in \mathbb{D}_n \setminus \{1\}} = |f(x + 2^{-n}) - f(x)|. \tag{A.80}$$

Overestimate the supremum by a sum and use $L^p$-Hölder continuity of $\mu_i$:

$$\mathbb{E}_{\mu_i}[K_n^p] \leq c \sum_{i=0}^{2^n - 1} \mathbb{E}_{\mu_i}[|f(x + 2^{-n}) - f(x)|^p] \leq c \sum_{i=0}^{2^n - 1} 2^{-\beta p n} = 2^{-(\beta p - 1)n}. \tag{A.81}$$

For any $x, y \in \bigcup_{n=1}^\infty \mathbb{D}_n$ such that $x < y < x + 2^{-k}$, note that the interval $[x, y)$ is the finite, disjoint union of intervals $[r, r + 2^{-n})$ with $r \in \mathbb{D}_n$ for $n \geq k+1$ where no three intervals have the same length (proof of Theorem 4.2.1; Norris, 2018). Therefore, for such $x$ and $y$, by continuity of $f$,

$$\omega_f(2^{-k}) = \sup_{x,y \in \bigcup_{n=1}^{\infty} \mathbb{D}_n : |x-y| < 2^{-k}} |f(x) - f(y)| \leq 2 \sum_{n=k+1}^{\infty} K_n. \tag{A.82}$$

Hence,

$$\mathbb{E}_{\mu_i}[\omega_f(2^{-k})] \leq 2 \sum_{n=k+1}^{\infty} \mathbb{E}_{\mu_i}[K_n^p]^{\frac{1}{p}} \leq 2c \sum_{n=k+1}^{\infty} 2^{-(\beta-\frac{1}{p})n} = c \frac{2^{1-(\beta-\frac{1}{p})(k+1)}}{1 - 2^{-(\beta-\frac{1}{p})}}, \tag{A.83}$$

using that $\beta - \frac{1}{p} > 0$, where is some constant $c_{p,q}$ that depends on $p$, $\beta$ and $c$. Then, for any $\varepsilon > 0$, by Markov's inequality,

$$\sup_{i \geq 1} \mu_i(\omega_f(h) \geq \varepsilon) \leq \sup_{i \geq 1} \mu_i(\omega_f(2^{-k}) \geq \varepsilon) \leq \frac{2c}{\varepsilon} \frac{2^{-(\beta-\frac{1}{p})k}}{2^{\beta-\frac{1}{p}} - 1} \leq \frac{2c}{\varepsilon} \frac{(2h)^{\beta-\frac{1}{p}}}{2^{\beta-\frac{1}{p}} - 1} \tag{A.84}$$

which proves the result. $\qquad \square$

# B | Proofs for Chapter 4

## B.1 Proofs for Section 4.2

In what follows, $D, E, F \in \mathcal{D}$ will denote generic data sets, and $A \subseteq \mathcal{D}$ will denote a generic collection of data sets. For $\varepsilon > 0$, denote the open balls of radius $\varepsilon$ by

$$\mathbb{B}_\varepsilon(D) = \{E \in \mathcal{D} : d_\mathcal{D}(D, E) < \varepsilon\}, \quad \mathbb{B}_\varepsilon([D]) = \{E \in \mathcal{D} : d_{[\mathcal{D}]}([D], [E]) < \varepsilon\}. \quad \text{(B.1)}$$

Let $\mathbb{S} = \bigcup_{N \geq 0} \mathbb{S}^N$ be the collection of permutations of any size.

**Proposition 4.2.**

(1) The function $d_{[\mathcal{D}]}$ is a metric on $[\mathcal{D}]$.

(2) If $A \subseteq \mathcal{D}$ is open, then $[A]$ is open in the topology of $d_{[\mathcal{D}]}$.

(3) If $A \subseteq \mathcal{D}$ is closed and permutation invariant, then $[A]$ is closed in the topology of $d_{[\mathcal{D}]}$.

(4) The topology on $[\mathcal{D}]$ induced by $d_{[\mathcal{D}]}$ coincides with the quotient topology.

*Proof.* (1): To begin with, observe that $d_\mathcal{D}(D, E) = d_\mathcal{D}(\sigma D, \sigma E)$ for all $\sigma \in \mathbb{S}^{|D|}$. We call this property *permutation invariance* of $d_\mathcal{D}$. We first show that $d_{[\mathcal{D}]}$ is well defined. To this end, consider $D, D', E, E' \in \mathcal{D}$ such that $D = \sigma_D D'$ and $E = \sigma_E E'$ for some $\sigma_D \in \mathbb{S}^{|D'|}$ and $\sigma_E \in \mathbb{S}^{|E'|}$. In particular, this means that $|D| = |D'|$ and $|E| = |E'|$. Then

$$d_{[\mathcal{D}]}([D], [E]) = \inf_{\sigma_2 \in \mathbb{S}^{|E|}} d_\mathcal{D}(D, \sigma_2 E) \quad \text{(B.2)}$$

$$= \inf_{\sigma_2 \in \mathbb{S}^{|E'|}} d_\mathcal{D}(\sigma_D D', \sigma_2 \sigma_E E') \quad \text{(B.3)}$$

$$= \inf_{\sigma_2 \in \mathbb{S}^{|E'|}} d_\mathcal{D}(D', \sigma_D^{-1} \sigma_2 \sigma_E E') \quad \text{(B.4)}$$

$$= \inf_{\sigma_2 \in \mathbb{S}^{|E'|}} d_\mathcal{D}(D', \sigma_2 E') \quad \text{(B.5)}$$

$$= d_{[\mathcal{D}]}([D'], [E']). \quad \text{(B.6)}$$

We conclude that $d_{[\mathcal{D}]}$ does not depend on the choice of the representative, so it is well defined. It remains to show that $d_{[\mathcal{D}]}$ is a metric. It is clear that $d_{[\mathcal{D}]}([D], [E]) = d_{[\mathcal{D}]}([E], [D])$ and that $d_{[\mathcal{D}]}([D], [E]) = 0$ if and only if $[D] = [E]$. To show the triangle inequality, let $D, E, F \in \mathcal{D}$. We will show that

$$d_{[\mathcal{D}]}([D], [F]) \leq d_{[\mathcal{D}]}([D], [E]) + d_{[\mathcal{D}]}([E], [F]). \tag{B.7}$$

If either of the terms on the right-hand side is infinite, then nothing remains to be shown. Hence, assume that both are finite. In that case, $|D| = |E| = |F|$. Let $\sigma_2, \sigma_2' \in \mathbb{S}^{|D|}$. Then, by the triangle inequality of $d_\mathcal{D}$,

$$d_\mathcal{D}(D, \sigma_2\sigma_2' F) \leq d_\mathcal{D}(D, \sigma_2 E) + d_\mathcal{D}(\sigma_2 E, \sigma_2\sigma_2' F) = d_\mathcal{D}(D, \sigma_2 E) + d_\mathcal{D}(E, \sigma_2' F), \tag{B.8}$$

using permutation invariance of $d_\mathcal{D}$. Take the infimum over $\sigma_2 \in \mathbb{S}^{|D|}$ then $\sigma_2' \in \mathbb{S}^{|D|}$ to conclude.

(2): Let $[D] \in [A]$. Then there exists a $\sigma \in \mathbb{S}$ such that $\sigma D \in A$. Hence, there exists a $\varepsilon > 0$ such that $B_\varepsilon(\sigma D) \subseteq A$. We claim that $\mathbb{B}_\varepsilon([D]) \subseteq [A]$, which shows the result. To show the claim, let $[E] \in \mathbb{B}_\varepsilon([D])$. Then $d_{[\mathcal{D}]}([D], [E]) < \varepsilon$, so $d_\mathcal{D}(D, \sigma' E) < \varepsilon$ for some $\sigma' \in \mathbb{S}$, meaning that $d_\mathcal{D}(\sigma D, \sigma\sigma' E) < \varepsilon$. Hence $\sigma\sigma' E \in B_\varepsilon(\sigma D) \subseteq A$, so $[E] = [\sigma\sigma' E] \in [A]$.

(3): Let $([D_i])_{i \geq 1} \subseteq [A]$ be convergent to some $[D] \in [\mathcal{D}]$. Then $d([D_i], [D]) \to 0$, so there exists a sequence $(\sigma_i)_{i \geq 1}$ such that $d(\sigma_i D_i, D) \to 0$. Since $A$ is permutation invariant, all $\sigma_i D_i \in A$. Hence $D \in A$, because $A$ is closed. We conclude that $[D] \in [A]$, so $[A]$ is also closed.

(4): Denote $p \colon \mathcal{D} \to [\mathcal{D}]$, $p(D) = [D]$. Note that $p$ is $d_\mathcal{D}$-$d_{[\mathcal{D}]}$ continuous. Let $[A] \subseteq [\mathcal{D}]$. If $[A]$ is open in the quotient topology, then, by definition, $p^{-1}([A])$ is open. Hence, by (3), $p(p^{-1}([A]))$ is open in $d_{[\mathcal{D}]}$. Conversely, assume that $[A]$ is open in $d_{[\mathcal{D}]}$. Then, by $d_\mathcal{D}$-$d_{[\mathcal{D}]}$ continuity of $p$, $p^{-1}([A])$ is open in $d_\mathcal{D}$. Hence, $[A]$ is open in the quotient topology. □

**Proposition 4.3.** *Suppose that $f \colon \mathcal{D} \to Z$ is continuous and permutation invariant. Then $[f] \colon [\mathcal{D}] \to Z$ defined by $[f]([D]) = f(D')$ for any $D'$ such that $D' = [D]$ is well defined and continuous. Conversely, suppose that $[f] \colon [\mathcal{D}] \to Z$ is continuous. Then $f \colon \mathcal{D} \to Z$ defined by $f(D) = [f]([D])$ is continuous and permutation invariant.*

*Proof.* That $[f]$ is well defined follows from permutation invariance of $f$, and that $[f]$ is continuous follows from continuity of $f$. Using continuity of the quotient map $D \mapsto [D]$, the converse is immediate. □

## B.2 Proofs for Section 4.4

Call a space $\mathcal{F}$ of functions on $\mathcal{X}$ *interpolating* if, for every $N \in \mathbb{N}$, $\mathbf{x} \in \mathcal{X}^N$, and $\mathbf{y} \in \mathbb{R}^N$, there exists an $f \in \mathcal{F}$ such that $f(x_n) = y_n$ for all $n = 1, \ldots, N$. Note that every reproducing kernel Hilbert space associated to a strictly-positive-definite kernel is interpolating. Throughout, we assume that $\mathcal{Y}$ is compact.

**Lemma B.1.** *Let $[\mathcal{D}'_N] \subseteq [\mathcal{D}_N]$ have multiplicity $K$. Set*

$$\phi \colon \mathcal{Y} \to \mathbb{R}^{K+1}, \quad \phi(y) = (y^0, y^1, \cdots, y^K). \tag{B.9}$$

*Let $k \colon \mathcal{X} \times \mathcal{X} \to \mathbb{R}$ be a continuous strictly-positive-definite kernel. Denote the reproducing kernel Hilbert space of $k$ by $\mathbb{H}$. Endow $\mathbb{H}^{K+1}$ with the inner product $\langle f, g \rangle_{\mathbb{H}^{K+1}} = \sum_{i=1}^{K+1} \langle f_i, g_i \rangle_{\mathbb{H}}$. Define*

$$\mathbb{H}_N = \left\{ \sum_{n=1}^N \phi(y_n) k(\,\cdot\,, x_n) : [(x_n, y_n)_{n=1}^N] \subseteq [\mathcal{D}'_N] \right\} \subseteq \mathbb{H}^{K+1}, \tag{B.10}$$

*Then the embedding*

$$\mathsf{enc}_N \colon [\mathcal{D}'_N] \to \mathbb{H}_N, \quad \mathsf{enc}_N([(x_1, y_1), \ldots, (x_N, y_N)]) = \sum_{n=1}^N \phi(y_n) k(\,\cdot\,, x_n) \tag{B.11}$$

*is injective, hence invertible, and continuous.*

*Proof.* First, we show that $\mathsf{enc}_N$ is injective. Suppose that

$$\sum_{n=1}^N \phi(y_n) k(\,\cdot\,, x_n) = \sum_{n=1}^N \phi(y'_n) k(\,\cdot\,, x'_n). \tag{B.12}$$

Denote $\mathbf{x} = (x_1, \ldots, x_N)$ and $\mathbf{y} = (y_1, \ldots, y_N)$, and denote $\mathbf{x}'$ and $\mathbf{y}'$ similarly. Taking the inner product with any $f \in \mathbb{H}$ on both sides and using the reproducing property of $k$ implies that

$$\sum_{n=1}^N \phi(y_n) f(x_n) = \sum_{n=1}^N \phi(y'_n) f(x'_n) \quad \text{for all } f \in \mathbb{H}. \tag{B.13}$$

In particular, by construction $\phi_1(\,\cdot\,) = 1$, so

$$\sum_{n=1}^N f(x_n) = \sum_{n=1}^N f(x'_n) \tag{B.14}$$

for all $f \in \mathbb{H}$. Using that $\mathbb{H}$ is interpolating, choose an element $\hat{x}$ of $\mathbf{x}$ or $\mathbf{x}'$, and let $f \in \mathbb{H}$ be such that $f(\hat{x}) = 1$ and $f(\,\cdot\,) = 0$ at all other elements of $\mathbf{x}$ and $\mathbf{x}'$. Then

$$\sum_{n\,:\,x_n=\hat{x}} 1 = \sum_{n\,:\,x'_n=\hat{x}} 1, \tag{B.15}$$

so the number of $\hat{x}$ in $\mathbf{x}$ and the number of $\hat{x}$ in $\mathbf{x}'$ are the same. Since this holds for every $\hat{x}$, $\mathbf{x}$ is a permutation of $\mathbf{x}'$: $\mathbf{x} = \sigma \mathbf{x}'$ for some permutation $\sigma \in \mathbb{S}^N$. Plugging in the permutation, we can write

$$\sum_{n=1}^{N} \phi(y_n) f(x_n) = \sum_{n=1}^{N} \phi(y'_n) f(x'_n) \tag{B.16}$$

$$= \sum_{n=1}^{N} \phi(y'_n) f(x_{\pi^{-1}(n)}) \qquad (\mathbf{x}' = \sigma^{-1}\mathbf{x}) \tag{B.17}$$

$$= \sum_{n=1}^{N} \phi(y'_{\pi(n)}) f(x_n). \qquad (n \leftarrow \sigma^{-1}(n)) \tag{B.18}$$

Then, by a similar argument, for any particular $\hat{x}$,

$$\sum_{n\,:\,x_n=\hat{x}} \phi(y_n) = \sum_{n\,:\,x_n=\hat{x}} \phi(y'_{\sigma(n)}). \tag{B.19}$$

Let the number of terms in each sum equal $S$. Since $[\mathcal{D}'_N]$ has multiplicity $K$, $S \leq K$. Consider the first $S$ dimensions of $\phi$ in (B.19). Then, using the definition of $\phi$, Lemma 4 by Zaheer et al. (2017) shows that

$$(y_n)_{n\,:\,x_n=\hat{x}} \quad \text{is a permutation of} \quad (y'_{\sigma(n)})_{n\,:\,x_n=\hat{x}}. \tag{B.20}$$

By $\sigma \mathbf{x} = \mathbf{x}'$, note that $x'_{\sigma(n)} = x_n = \hat{x}$ for all these $n$. Therefore, all these $x_n$ and $x'_{\sigma(n)}$ are equal. Using (B.20), adjust the permutation $\sigma$ to obtain $y_n = y'_{\sigma(n)}$ for all these $n$. Since all these $x_n$ and $x'_{\sigma(n)}$ are equal, still $\mathbf{x} = \sigma \mathbf{x}'$ after the adjustment. Performing this adjustment for all $\hat{x}$, we find that $\mathbf{y} = \sigma \mathbf{y}'$ and $\mathbf{x} = \sigma \mathbf{x}'$.

Second, we show that $E_N$ is continuous. Compute

$$\left\| \sum_{n=1}^{N} \phi(y_n) k(\,\cdot\,, x_n) - \sum_{n=1}^{N} \phi(y'_n) k(\,\cdot\,, x'_n) \right\|_{\mathbb{H}^{K+1}}^{2} \tag{B.21}$$

$$= \sum_{i=1}^{K+1} (\phi^\mathsf{T}(\mathbf{y}) k(\mathbf{x}, \mathbf{x}) \phi_i(\mathbf{y}) - 2\phi^\mathsf{T}(\mathbf{y}) k(\mathbf{x}, \mathbf{x}') \phi_i(\mathbf{y}') + \phi^\mathsf{T}(\mathbf{y}') k(\mathbf{x}', \mathbf{x}') \phi_i(\mathbf{y}')),$$

which, by continuity of $k$, goes to zero if $[(\mathbf{x}', \mathbf{y}')] \to [(\mathbf{x}, \mathbf{y})]$. $\square$

**Lemma B.2.** *Consider Lemma B.1. Suppose that $[\mathcal{D}'_N]$ is also closed and that $k$ also satisfies (1) $k(x,x) = \sigma^2 > 0$, (2) $k \geq 0$, and (3) $k(x,x') \to 0$ as $|x| \to \infty$. Then $\mathbb{H}_N$ is closed in $\mathbb{H}^{K+1}$. Moreover, $\mathsf{enc}_N$ is a homeomorphism.*

*Proof.* Define
$$[\mathcal{D}_J] = [([-J, J] \times \mathcal{Y})^N] \cap [\mathcal{D}'_N], \tag{B.22}$$

which is compact in $[\mathcal{D}_N]$ as a closed subset of the compact set $[([-J, J] \times \mathcal{Y})^N]$. We aim to show that $\mathbb{H}_N$ is closed in $\mathbb{H}^{K+1}$ and that $\mathsf{enc}_N^{-1}$ is continuous. To this end, consider a convergent sequence

$$f^{(m)} = \sum_{n=1}^N \phi(y_n^{(m)}) k(\,\cdot\,, x_n^{(m)}) \to f \in \mathbb{H}^{K+1}. \tag{B.23}$$

Denote $\mathbf{x}^{(m)} = (x_1^{(m)}, \ldots, x_N^{(m)})$ and $\mathbf{y}^{(m)} = (y_1^{(m)}, \ldots, y_N^{(m)})$. Claim: $(\mathbf{x}^{(m)})_{m \geq 1}$ is a bounded sequence, so $(\mathbf{x}^{(m)})_{m \geq 1} \subseteq [-J, J]^N$ for $J$ large enough, which means that $[(\mathbf{x}^{(m)}, \mathbf{y}^{(m)})_{m \geq 1}] \subseteq [\mathcal{D}_J]$.

First, assuming the claim, we show that $\mathbb{H}_N$ is closed. By boundedness of $(\mathbf{x}^{(m)}, \mathbf{y}^{(m)})_{m \geq 1}$, $(f^{(m)})_{m \geq 1}$ is in the image of $\mathsf{enc}_N|_{[\mathcal{D}_J]} : [\mathcal{D}_J] \to \mathbb{H}_N$. By continuity of $\mathsf{enc}_N|_{[\mathcal{D}_J]}$ and compactness of $[\mathcal{D}_J]$, the image of $\mathsf{enc}_N|_{[\mathcal{D}_J]}$ is compact and hence closed. Therefore, the image of $\mathsf{enc}_N|_{[\mathcal{D}_J]}$ contains the limit $f$. Since the image of $\mathsf{enc}_N|_{[\mathcal{D}_J]}$ is included in $\mathbb{H}_N$, we have that $f \in \mathbb{H}_N$, which shows that $\mathbb{H}_N$ is closed.

Second, assuming the claim, we prove that $\mathsf{enc}_N^{-1}$ is continuous. Consider $\mathsf{enc}_N|_{[\mathcal{D}_J]} : [\mathcal{D}_J] \to \mathsf{enc}_N([\mathcal{D}_J])$ restricted to its image. Then $(\mathsf{enc}_N|_{[\mathcal{D}_J]})^{-1}$ is continuous, because every continuous bijection from a compact space to a Hausdorff space is a homeomorphism (Theorem 26.6; Munkres, 2000). Therefore,

$$\mathsf{enc}_N^{-1}(f^{(m)}) = [(\mathbf{x}^{(m)}, \mathbf{y}^{(m)})] = (\mathsf{enc}_N|_{[\mathcal{D}_J]})^{-1}(f^{(m)}) \to (\mathsf{enc}_N|_{[\mathcal{D}_J]})^{-1}(f). \tag{B.24}$$

Denote $(\mathsf{enc}_N|_{[\mathcal{D}_J]})^{-1}(f) = [(\mathbf{x}, \mathbf{y})]$. By continuity and invertibility of $\mathsf{enc}_N$, then $f^{(m)} \to \mathsf{enc}_N([(\mathbf{x}, \mathbf{y})])$, which means that $[(\mathbf{x}, \mathbf{y})] = \mathsf{enc}_N^{-1}(f)$ by uniqueness of limits. We conclude that $\mathsf{enc}_N^{-1}(f^{(m)}) \to \mathsf{enc}_N^{-1}(f)$, so $\mathsf{enc}_N^{-1}$ is continuous.

It remains to show the claim. Let $f_1$ denote the first element of $f$, *i.e.* the *density channel*. Using the reproducing property of $k$,

$$|f_1^{(m)}(x) - f_1(x)| = |\langle k(x,\,\cdot\,), f_1^{(m)} - f_1 \rangle_{\mathbb{H}}| \tag{B.25}$$
$$\leq \|k(x,\,\cdot\,)\|_{\mathbb{H}} \|f_1^{(m)} - f_1\|_{\mathbb{H}} \tag{B.26}$$
$$= \sigma \|f_1^{(m)} - f_1\|_{\mathbb{H}}, \tag{B.27}$$

so $f_1^{(n)} \to f_1$ in $\mathbb{H}$ means that it does so uniformly. Hence, we can let $M \in \mathbb{N}$ be such that $m \geq M$ implies that $|f_1^{(m)}(x) - f_1(x)| < \frac{1}{3}\sigma^2$ for all $x \in \mathcal{X}$. Let $R$ be such that $|k(x, x_n^{(M)})| < \frac{1}{3}\sigma^2/N$ for $|x| \geq R$ and all $n \in \{1, \ldots, N\}$. Then, for $|x| \geq R$,

$$|f_1^{(M)}(x)| \leq \sum_{n=1}^{N} |k(x, x_n^{(M)})| < \tfrac{1}{3}\sigma^2, \tag{B.28}$$

which implies that, for all $|x| \geq R$,

$$|f_1(x)| \leq |f_1^{(M)}(x)| + |f_1^{(M)}(x) - f_1(x)| < \tfrac{2}{3}\sigma^2. \tag{B.29}$$

At the same time, by pointwise nonnegativity of $k$, we have that

$$f_1^{(m)}(x_n^{(m)}) = \sum_{n'=1}^{N} k(x_n^{(n)}, x_{n'}^{(n)}) \geq k(x_n^{(m)}, x_n^{(m)}) = \sigma^2. \tag{B.30}$$

Towards contradiction, suppose that $(\mathbf{x}^{(m)})_{m \geq 1}$ is unbounded. Then $(x_n^{(m)})_{m \geq 1}$ is unbounded for some $n \in \{1, \ldots, N\}$. Therefore, $|x_n^{(m)}| \geq R$ for some $m \geq M$, so

$$\tfrac{2}{3}\sigma^2 > |f_1(x_n^{(m)})| \geq |f_1^{(m)}(x_n^{(m)})| - |f_1^{(m)}(x_n^{(m)}) - f_1(x_n^{(m)})| \geq \sigma^2 - \tfrac{1}{3}\sigma^2 = \tfrac{2}{3}\sigma^2, \tag{B.31}$$

which is a contradiction. $\square$

**Lemma B.3.** *Suppose that $[\mathcal{D}'] \subseteq [\mathcal{D}]$ is closed, has multiplicity $K$, and has maximum data set size $N_{\max} < \infty$. Set*

$$\phi \colon \mathcal{Y} \to \mathbb{R}^{K+1}, \quad \phi(y) = (y^0, y^1, \cdots, y^K). \tag{B.32}$$

*Let $k \colon \mathcal{X} \times \mathcal{X} \to \mathbb{R}$ be a continuous strictly-positive-definite kernel such that (1) $k(x, x) = \sigma^2 > 0$, (2) $k \geq 0$, and (3) $k(x, x') \to 0$ as $|x| \to \infty$. Denote the reproducing kernel Hilbert space of $k$ by $\mathbb{H}$. Endow $\mathbb{H}^{K+1}$ with the inner product $\langle f, g \rangle_{\mathbb{H}^{K+1}} = \sum_{i=1}^{K+1} \langle f_i, g_i \rangle_{\mathbb{H}}$. For $N \in \{0, \ldots, N_{\max}\}$, denote $[\mathcal{D}'_N] = [\mathcal{D}'] \cap [\mathcal{D}_N]$ and define*

$$\mathbb{H}_N = \left\{ \sum_{n=1}^{N} \phi(y_n) k(\,\cdot\,, x_n) : [(x_n, y_n)_{n=1}^{N}] \subseteq [\mathcal{D}'_N] \right\} \subseteq \mathbb{H}^{K+1}. \tag{B.33}$$

*Then $(\mathbb{H}_N)_{N=0}^{N_{\max}}$ are pairwise disjoint. Denote $\mathbb{H}' = \bigcup_{N=0}^{N_{\max}} \mathbb{H}_N$. Then*

$$\mathsf{enc} \colon [\mathcal{D}'] \to \mathbb{H}', \quad \mathsf{enc}([D]) = \mathsf{enc}_N([D]) \quad \text{if} \quad [D] \in [\mathcal{D}_N] \tag{B.34}$$

*is a homeomorphism.*

*Proof.* To begin with, we show that $(\mathbb{H}_N)_{N=0}^{N_{\max}}$ are pairwise disjoint. To this end, suppose that

$$\sum_{n=1}^{N} \phi(y_n) k(\,\cdot\,, x_n) = \sum_{n=1}^{N'} \phi(y'_n) k(\,\cdot\,, x'_n) \tag{B.35}$$

for $N \neq N'$. Then, by arguments like in the proof of Lemma B.1,

$$\sum_{n=1}^{N} \phi(y_n) = \sum_{n=1}^{N'} \phi(y'_n). \tag{B.36}$$

Since $\phi_1(\,\cdot\,) = 1$, this gives $N = N'$, which is a contradiction.

Second, we show that enc is a homeomorphism. Note that $(\mathbb{H}_N)_{N=0}^{N_{\max}}$ are closed and pairwise disjoint, and that $([\mathcal{D}'_N])_{N=0}^{N_{\max}}$ are also closed and pairwise disjoint. By Lemma B.2, every

$$\mathsf{enc}|_{[\mathcal{D}'_N]} \colon [\mathcal{D}'_N] \to \mathbb{H}_N \tag{B.37}$$

is a homeomorphism. Therefore,

$$\mathsf{enc} \colon \bigcup_{N=0}^{N_{\max}} [\mathcal{D}'_N] \to \bigcup_{N=0}^{N_{\max}} \mathbb{H}_N \tag{B.38}$$

is also a homeomorphism: invertibility follows from disjointness, continuity of enc follows from the pasting lemma (Theorem 18.3; Munkres, 2000), and continuity of $\mathsf{enc}^{-1}$ similarly follows from the pasting lemma. □

**Theorem 4.8** (Convolutional deep set). *Let $\mathcal{Y} \subseteq \mathbb{R}$ be compact. Suppose that $[\mathcal{D}'] \subseteq [\mathcal{D}]$ is closed, is closed under translations, has multiplicity $K$, and has maximum data set size $N < \infty$. Let $k \colon \mathcal{X} \to \mathbb{R}$ be a continuous strictly-positive-definite function such that (1) $k(0) = \sigma^2 > 0$ (2) $k \geq 0$, and (3) $k(\tau) \to 0$ as $|\tau| \to \infty$. Denote the reproducing kernel Hilbert space of $k$ by $\mathbb{H}$. Let $Z$ be a translation space. Then a function $\pi \colon [\mathcal{D}'] \to Z$ is continuous and translation equivariant if and only if it is of the form*

$$\pi = \mathsf{dec} \circ \mathsf{enc} \quad \text{where} \quad \mathsf{enc}(D) = \sum_{(x,y) \in D} \phi(y) k(\,\cdot\, - x) \tag{B.39}$$

*with $\mathsf{enc} \colon [\mathcal{D}'] \to \mathbb{H}'$ continuous and translation equivariant, $\mathsf{dec} \colon \mathbb{H}' \to Z$ continuous and translation equivariant, and $\phi(y) = (0, y^1, \ldots, y^K)$. Here $\mathbb{H}' = \mathsf{enc}([\mathcal{D}'])$ is a subspace of $\mathbb{H}^{K+1}$ which is closed and closed under translations.*

*Proof.* Our proof follows the strategy used by Zaheer et al. (2017). If $\pi$ is of the stated form, then it is continuous and translation equivariant as a composition of two continuous and translation-equivariant functions. Note that continuity of enc follows from Lemma B.3, and that translation equivariance of enc is easily verified. Conversely, suppose that $\pi \colon [\mathcal{D}'] \to Z$ is continuous and translation equivariant. Let enc be of the stated form. By Lemma B.3, enc is a homeomorphism. Set $\text{dec} = \pi \circ \text{enc}^{-1}$. Then enc is continuous, and dec is continuous as a composition of two continuous functions. Moreover, enc is translation equivariant by construction, and dec is translation equivariant as a composition of two translation equivariant functions. □

## B.3 Proofs for Section 4.5

**Theorem** 4.9 (Convolutional deep set for DTE). *Let $\mathcal{Y} \subseteq \mathbb{R}$ be compact. Suppose that $[\mathcal{D}'] \subseteq [\mathcal{D}]$ is closed, is closed under translations, has multiplicity $K$, and has maximum data set size $N < \infty$. Let $k \colon \mathcal{X} \times \mathcal{X} \to \mathbb{R}$ be a continuous strictly-positive-definite function such that (1) $k(\mathbf{0}) = \sigma^2 > 0$, (2) $k \geq 0$, and (3) $k(\boldsymbol{\tau}) \to 0$ as $\|\boldsymbol{\tau}\|_2 \to \infty$. Denote the reproducing kernel Hilbert space associated to $k$ by $\mathbb{H}$. Let $Z$ be a topological $(\mathcal{X} \times \mathcal{X})$-translation space. Let $C$ be another topological $(\mathcal{X} \times \mathcal{X})$-translation space, let $c \in C$ be diagonally translation invariant and anti-diagonal discriminating, and denote $C' = \{\mathsf{T}_{\boldsymbol{\tau}} c : \boldsymbol{\tau} \in \mathcal{X} \times \mathcal{X}\}$. Then a function $\pi \colon [\mathcal{D}'] \to Z$ is continuous and diagonally translation equivariant in the sense of*

$$\pi \circ \mathsf{T}_\tau = \mathsf{T}_{(\tau,\tau)} \circ \pi \quad \text{for all } \tau \in \mathcal{X} \tag{B.40}$$

*if and only if it is of the form*

$$\pi = \text{dec} \circ \text{enc} \quad \text{where} \quad \text{enc}(D) = \begin{bmatrix} \sum_{(x,y) \in C} \phi(y) k(\,\cdot\, - (x,x)) \\ c \end{bmatrix} \tag{B.41}$$

*with $\text{enc} \colon [\mathcal{D}'] \to \mathbb{H}' \times C'$ continuous and translation equivariant, $\text{dec} \colon \mathbb{H}^{K+1} \to Z$ continuous and translation equivariant, and $\phi(y) = (0, y^1, \ldots, y^K)$. Here $\mathbb{H}' = \text{enc}([\mathcal{D}'])$ is a subspace of $\mathbb{H}^{K+1}$ which is closed and closed under translations.*

*Addendum to proof.* The proof in Section 4.5 is nearly complete. The only thing which remains is to show that the extension $\overline{\pi} \colon A \times \{\mathsf{T}_{\boldsymbol{\tau}} c : \boldsymbol{\tau} \in \mathcal{X} \times \mathcal{X}\} \to Z$ is continuous. To this end, consider a sequence $(a_i, \mathsf{T}_{\boldsymbol{\tau}_i} c)_{i \geq 1}$ convergent to $(a, \mathsf{T}_{\boldsymbol{\tau}} c)$. Let $\mathbf{e}_\| = (1,1)/\sqrt{2}$ and $\mathbf{e}_\perp = (1,-1)/\sqrt{2}$. Denote $\boldsymbol{\tau}_{i,\|} = \langle \boldsymbol{\tau}_i, \mathbf{e}_\| \rangle \mathbf{e}_\|$ and $\boldsymbol{\tau}_{i,\perp} = \langle \boldsymbol{\tau}_i, \mathbf{e}_\perp \rangle \mathbf{e}_\perp$. Then

$$\overline{\pi}(a_i, \mathsf{T}_{\boldsymbol{\tau}_i} c) \stackrel{(i)}{=} \overline{\pi}(a_i, \mathsf{T}_{\boldsymbol{\tau}_{i,\perp}} c) \stackrel{(ii)}{=} \mathsf{T}_{\boldsymbol{\tau}_{i,\perp}} \overline{\pi}(\mathsf{T}_{-\boldsymbol{\tau}_{i,\perp}} a_i, c) \stackrel{(iii)}{=} \mathsf{T}_{\boldsymbol{\tau}_{i,\perp}} \pi(\mathsf{T}_{-\boldsymbol{\tau}_{i,\perp}} a_i, c) \tag{B.42}$$

using in (i) that $c$ is diagonally translation invariant, in (ii) that $\pi$ is TE, and in (iii) the definition of $\overline{\pi}$. Because $c$ is anti-diagonal discriminating, $(\boldsymbol{\tau}_{i,\perp})_{i \geq 1}$ is convergent. Note that $(\boldsymbol{\tau}, a) \mapsto \mathsf{T}_{-\boldsymbol{\tau}} a$ and $(\boldsymbol{\tau}, z) \mapsto \mathsf{T}_{\boldsymbol{\tau}} z$ are continuous because $A$ and $Z$ are topological $(\mathcal{X} \times \mathcal{X})$-translation spaces. Therefore, using continuity of $\pi$,

$$\lim_{i \to \infty} \bar{\pi}(a_i, \mathsf{T}_{\boldsymbol{\tau}_i} c) = \mathsf{T}_{\boldsymbol{\tau}_\perp} \pi(\mathsf{T}_{-\boldsymbol{\tau}_\perp} a_i, c) = \bar{\pi}(a, \mathsf{T}_{\boldsymbol{\tau}} c) \tag{B.43}$$

where the latter equality reverses the steps in (B.42). $\square$

# C | Proofs for Chapter 5

## C.1 Proofs for Section 5.3

**Proposition 5.2.** *The ground-truth stochastic process $f$ is stationary if and only if the posterior prediction map $\pi_f$ is translation equivariant (Definition 2.4).*

*Proof.* Suppose that the ground-truth stochastic process $f$ is stationary: $f \stackrel{\mathrm{d}}{=} \mathsf{T}_\tau f$ for all $\tau \in \mathcal{X}$. Let $D \in \mathcal{D}$ and $\tau \in \mathcal{X}$. Let $\mu_f$ denote the law of $f$, and let $\pi'(D)$ denote the Radon–Nikodym of $\pi_f(D)$ with respect to $\mu_f$ (Definition 3.6). From Definition 3.6 and stationarity of $f$, it is clear that $\pi'(\mathsf{T}_\tau D) = \pi'(D) \circ \mathsf{T}_{-\tau}$. Let $B$ be a cylinder set. Then, by the changes-of-variables formula for pushforward measures (Theorem 14.1; Schilling, 2005),

$$\int_B \pi'(\mathsf{T}_\tau D) \, \mathrm{d}\mu_f = \int_B \pi'(D) \circ \mathsf{T}_{-\tau} \, \mathrm{d}\mu_f \tag{C.1}$$

$$= \int_{\mathsf{T}_{-\tau}(B)} \pi'(D) \, \mathrm{d}\mathsf{T}_{-\tau}(\mu_f) \quad \text{(change of variables)} \tag{C.2}$$

$$= \int_{\mathsf{T}_\tau^{-1}(B)} \pi'(D) \, \mathrm{d}\mu_f. \quad (f \text{ is stationary and } \mathsf{T}_{-\tau} = \mathsf{T}_\tau^{-1}) \tag{C.3}$$

We conclude that $\pi_f$ is translation equivariant (Definition 2.4). Conversely, if $\pi_f$ is translation equivariant, then

$$\mathsf{T}_\tau \pi_f(\varnothing) = \pi_f(\mathsf{T}_\tau \varnothing) = \pi_f(\varnothing) \quad \text{for all } \tau \in \mathcal{X}. \tag{C.4}$$

Since $\pi_f(\varnothing) = \mu_f$, this means that $f$ is stationary. □

If we just assume that $f$ is stationary and $\pi_f$ is translation equivariant, then we can prove that $\pi'(\mathsf{T}_\tau D) = \pi'(D) \circ \mathsf{T}_{-\tau}$ $\mu_f$–almost surely. Let $B$ be a cylinder set. Then

$$\int_B \pi'(\mathsf{T}_\tau D) \, \mathrm{d}\mu_f = \int_{\mathsf{T}_\tau^{-1}(B)} \pi'(D) \, \mathrm{d}\mu_f \quad (\pi_f \text{ is translation equivariant}) \tag{C.5}$$

$$= \int_{\mathsf{T}_{-\tau}(B)} \pi'(D) \, \mathrm{d}\mathsf{T}_{-\tau}(\mu_f) \quad (\mathsf{T}_\tau^{-1} = \mathsf{T}_{-\tau} \text{ and } f \text{ is stationary}) \tag{C.6}$$

$$= \int_B \pi'_P(D) \circ \mathsf{T}_{-\tau} \, \mathrm{d}\mu_f. \quad \text{(change of variables)} \tag{C.7}$$

Since this holds for all cylinder sets, $\pi'(\mathsf{T}_\tau D) = \pi'(D) \circ \mathsf{T}_{-\tau}$ $\mu_f$–almost surely. Therefore, we have actually proven the stronger statement that $\pi_f$ is stationary if and only if $f$ is stationary and $\pi'(\mathsf{T}_\tau D) = \pi'(D) \circ \mathsf{T}_{-\tau}$ $\mu_f$–almost surely. However, for the current exposition, this generality is unnecessary, because we assume the form of $\pi'$ in Definition 3.6.

**Proposition 5.3.** *If a prediction map $\pi$ is translation equivariant, then the mean map $m_\pi$ and variance map $v_\pi$ are also translation equivariant. Conversely, suppose that $\pi$ a conditional neural process in the sense of Definition 3.22. If $m_\pi$ is TE and $v_\pi$ is TE, then $\pi$ is translation equivariant.*

$$\pi \text{ is TE} \quad \underset{\pi \text{ is a CNP}}{\overset{\Longrightarrow}{\Longleftarrow}} \quad m_\pi \text{ is TE and } v_\pi \text{ is TE}. \tag{C.8}$$

*Proof.* "$\Rightarrow$": Let $D \in \mathcal{D}$ and $\tau \in \mathcal{X}$. Then

$$\begin{aligned}
m_\pi(\mathsf{T}_\tau D)(x) &= \mathbb{E}_{\pi(\mathsf{T}_\tau D)}[f(x)] && \text{(Definition 3.18)} & (\text{C.9})\\
&= \mathbb{E}_{\mathsf{T}_\tau \pi(D)}[f(x)] && \text{(translation equivariance of } \pi\text{)} & (\text{C.10})\\
&= \mathbb{E}_{\pi(D)}[f(x-\tau)] && \text{(change of variables)} & (\text{C.11})\\
&= m_\pi(D)(x-\tau). && \text{(Definition 3.18)} & (\text{C.12})
\end{aligned}$$

The proof for $v_\pi$ is similar. "$\Leftarrow$": A conditional neural process $\pi$ in the sense of Definition 3.22 is characterised by $m_\pi$ and $v_\pi$ (Section 3.4). Namely, let $D \in \mathcal{D}$, $\mathbf{x} \in I_N$. Then

$$P_\mathbf{x} \pi(D) = \mathcal{N}\left( \begin{bmatrix} m(D)(x_1) \\ \vdots \\ m(D)(x_N) \end{bmatrix}, \begin{bmatrix} v(D)(x_1) & & 0 \\ & \ddots & \vdots \\ 0 & & v(D)(x_N) \end{bmatrix} \right). \tag{C.13}$$

Therefore, if $m_\pi$ and $v_\pi$ are translation equivariant, then $\pi(D)$ is translation equivariant. $\square$

## C.2 Proofs for Section 5.4

**Theorem 5.7.** *Let $\pi_1, \pi_2 \colon \mathcal{D} \to \mathcal{P}$ be translation-equivariant prediction maps with receptive field $R > 0$. Assume that, for all $D \in \mathcal{D}$, $\pi_1(D)$ and $\pi_2(D)$ also have receptive field $R > 0$. Let $\varepsilon > 0$ and fix $N \in \mathbb{N}$. Assume that, for all $\mathbf{x} \in \bigcup_{n=1}^N [0, 2R]^n$ and $D \in \mathcal{D} \cap \bigcup_{n=0}^\infty ([0, 2R] \times \mathbb{R})^n$,*

$$\mathrm{KL}(P_\mathbf{x} \pi_1(D), P_\mathbf{x} \pi_2(D)) \leq \varepsilon. \tag{C.14}$$

*Then, for all $M > 0$, $\mathbf{x} \in \bigcup_{n=1}^N [0, M]^n$, and $D \in \mathcal{D} \cap \bigcup_{n=0}^\infty ([0, M] \times \mathbb{R})^n$,*

$$\mathrm{KL}(P_\mathbf{x} \pi_1(D), P_\mathbf{x} \pi_2(D)) \leq \lceil 2M/R \rceil \varepsilon. \tag{C.15}$$

*Proof.* The proof follows the intuition from Figure 5.1. Let $M > 0$, $n \in \{1, \ldots, N\}$,

$\mathbf{x} \in [0, M]^n$, and $D \in \mathcal{D} \cap \bigcup_{n=0}^\infty ([0,M] \times \mathbb{R})^n$. Sort and put the $n$ inputs $\mathbf{x}$ into $B = \lceil M/\frac{1}{2}R \rceil$ buckets $(B_i)_{i=1}^B$ such that $x_j \in [(i-1) \cdot \frac{1}{2}R, i \cdot \frac{1}{2}R]$ for all $j \in B_i$. More concisely written, $\mathbf{x}_{B_i} \in [(i-1) \cdot \frac{1}{2}R, i \cdot \frac{1}{2}R]^{|B_i|}$. Write $C_i = \bigcup_{j=1}^{i-1} B_j$. Let $D_i$ be the subset of $D$ with inputs in $[\min(\mathbf{x}_{B_{i-2}}), \max(\mathbf{x}_{B_{i+1}})]$.

If $\mathbf{y}_1 \oplus \mathbf{y}_2 \sim P^\sigma_{\mathbf{x}_1 \oplus \mathbf{x}_2} \pi(D)$, then denote the distribution of $\mathbf{y}_1 \,|\, \mathbf{y}_2$ by $P^\sigma_{\mathbf{x}_1 \,|\, \mathbf{x}_2} \pi(D)$. Use the chain rule for the KL divergence to decompose

$$\mathrm{KL}(P^{\sigma_1}_\mathbf{x} \pi_1(D), P^{\sigma_2}_\mathbf{x} \pi_2(D))$$
$$= \sum_{i=1}^B \mathbb{E}_{P^{\sigma_1}_{\mathbf{x}_{C_i}} \pi_1(D)} [\mathrm{KL}(P^{\sigma_1}_{\mathbf{x}_{B_i} \,|\, \mathbf{x}_{C_i}} \pi_1(D), P^{\sigma_2}_{\mathbf{x}_{B_i} \,|\, \mathbf{x}_{C_i}} \pi_2(D))]. \quad \text{(C.16)}$$

We focus on the $i^\text{th}$ term in the sum. Using that $\pi_1(D)$ and $\pi_2(D)$ have receptive field $R$, we may drop the dependency on $B_1, \ldots, B_{i-2}$:

$$\mathrm{KL}(P^{\sigma_1}_{\mathbf{x}_{B_i} \,|\, \mathbf{x}_{C_i}} \pi_1(D), P^{\sigma_2}_{\mathbf{x}_{B_i} \,|\, \mathbf{x}_{C_i}} \pi_2(D)) = \mathrm{KL}(P^{\sigma_1}_{\mathbf{x}_{B_i} \,|\, \mathbf{x}_{B_{i-1}}} \pi_1(D), P^{\sigma_2}_{\mathbf{x}_{B_i} \,|\, \mathbf{x}_{B_{i-1}}} \pi_2(D)) \quad \text{(C.17)}$$

Therefore,

$$\mathbb{E}_{P^{\sigma_1}_{\mathbf{x}_{C_i}} \pi_1(D)} [\mathrm{KL}(P^{\sigma_1}_{\mathbf{x}_{B_i} \,|\, \mathbf{x}_{C_i}} \pi_1(D), P^{\sigma_2}_{\mathbf{x}_{B_i} \,|\, \mathbf{x}_{C_i}} \pi_2(D))]$$
$$= \mathbb{E}_{P^{\sigma_1}_{\mathbf{x}_{B_{i-1}}} \pi_1(D)} [\mathrm{KL}(P^{\sigma_1}_{\mathbf{x}_{B_i} \,|\, \mathbf{x}_{B_{i-1}}} \pi_1(D), P^{\sigma_2}_{\mathbf{x}_{B_i} \,|\, \mathbf{x}_{B_{i-1}}} \pi_2(D))] \quad \text{(C.18)}$$
$$\le \mathrm{KL}(P^{\sigma_1}_{\mathbf{x}_{B_i \cup B_{i-1}}} \pi_1(D), P^{\sigma_2}_{\mathbf{x}_{B_i \cup B_{i-1}}} \pi_2(D)). \quad \text{(C.19)}$$

Next, we use that $\pi_1$ and $\pi_2$ also have receptive field $R$, allowing us to replace $D$ with $D_i$:

$$\mathrm{KL}(P^{\sigma_1}_{\mathbf{x}_{B_i \cup B_{i-1}}} \pi_1(D), P^{\sigma_2}_{\mathbf{x}_{B_i \cup B_{i-1}}} \pi_2(D)) = \mathrm{KL}(P^{\sigma_1}_{\mathbf{x}_{B_i \cup B_{i-1}}} \pi_1(D_i), P^{\sigma_2}_{\mathbf{x}_{B_i \cup B_{i-1}}} \pi_2(D_i)). \quad \text{(C.20)}$$

Finally, we use translation equivariance to shift everything back to the origin. Let $\tau_i$ be $\min(\mathbf{x}_{B_{i-2}})$. By translation equivariance of $\pi_1$,

$$P^{\sigma_1}_{\mathbf{x}_{B_i \cup B_{i-1}}} \pi_1(D_i) = P^{\sigma_1}_{\mathbf{x}_{B_i \cup B_{i-1}}} \mathsf{T}_{\tau_i} \pi_1(\mathsf{T}_{-\tau_i} D_i) = P^{\sigma_1}_{\mathbf{x}_{B_i \cup B_{i-1}} - \tau_i} \pi_1(\mathsf{T}_{-\tau_i} D_i) \quad \text{(C.21)}$$

where by $\mathbf{x}_{B_i \cup B_{i-1}} - \tau_i$ we mean subtraction elementwise. Crucially, note that all elements of $\mathbf{x}_{B_i \cup B_{i-1}} - \tau_i$ and all inputs of $\mathsf{T}_{-\tau_i} D_i$ lie in $[0, 4 \cdot \frac{1}{2}R]$. We have a similar equality for $\pi_2$. Putting everything together,

$$\mathbb{E}_{P^{\sigma_1}_{\mathbf{x}_{C_i}} \pi_1(D)} [\mathrm{KL}(P^{\sigma_1}_{\mathbf{x}_{B_i} \,|\, \mathbf{x}_{C_i}} \pi_1(D), P^{\sigma_2}_{\mathbf{x}_{B_i} \,|\, \mathbf{x}_{C_i}} \pi_2(D))]$$
$$\le \mathrm{KL}(P^{\sigma_1}_{\mathbf{x}_{B_i \cup B_{i-1}} - \tau_i} \pi_1(\mathsf{T}_{-\tau_i} D_i), P^{\sigma_2}_{\mathbf{x}_{B_i \cup B_{i-1}} - \tau_i} \pi_2(\mathsf{T}_{-\tau_i} D_i)), \quad \text{(C.22)}$$

which is less than $\varepsilon$ by the assumption of the theorem. The conclusion now follows. $\square$

## C.3 Proofs for Section 5.5

**Proposition 5.11.** *If a prediction map $\pi$ is translation equivariant, then the mean map $m_\pi$ is translation equivariant and the kernel map $k_\pi$ is diagonally translation equivariant. Conversely, suppose that $\pi$ is a Gaussian neural process in the sense of Definition 3.23. If $m_\pi$ is TE and $k_\pi$ is DTE, then $\pi$ is translation equivariant. In formulas,*

$$\pi \text{ is TE} \underset{\pi \text{ is a GNP}}{\overset{\Longrightarrow}{\Longleftarrow}} m_\pi \text{ is TE and } k_\pi \text{ is DTE.} \tag{C.23}$$

*Proof.* "$\Rightarrow$": Follows from "$\Rightarrow$" of Proposition 5.3, noting that TE of $v_\pi$ implies DTE of $k_\pi$. "$\Leftarrow$": Exactly like the proof for "$\Leftarrow$" of Proposition 5.3, now using DTE of $k_\pi$ rather than TE of $v_\pi$. $\square$

## C.4 Proofs for Section 5.6

**Proposition 5.15** (Advantage of AR CNPs). *Let $(\pi_C, \sigma_C)$ be a CNPA and let $(\pi_G, \sigma_G)$ be the GNPA. Then, for all $\mathbf{x} \in I$ and $D \in \widetilde{\mathcal{D}}$ (see Section 3.2),*

$$\mathrm{KL}(P_\mathbf{x}^{\sigma_f} \pi_f(D), \mathrm{AR}_\mathbf{x}^{\sigma_C}(\pi_C, D)) \leq \mathrm{KL}(P_\mathbf{x}^{\sigma_f} \pi_f(D), P_\mathbf{x}^{\sigma_G} \pi_G(D)). \tag{C.24}$$

*Proof.* Use the notation $P_{\mathbf{x}_1 \mid \mathbf{x}_2}^\sigma \pi$ from the proof of Theorem 5.7. We will argue that, for all $n = 1, \ldots, |\mathbf{x}|$,

$$\begin{aligned}\mathrm{KL}(P_{x_n \mid \mathbf{x}_{1:(n-1)}}^{\sigma_f} &\pi_f(D), P_{x_n}^{\sigma_C} \pi_C(D \oplus (\mathbf{x}_{1:(n-1)}, \mathbf{y}_{1:(n-1)}))) \\ &\leq \mathrm{KL}(P_{x_n \mid \mathbf{x}_{1:(n-1)}}^{\sigma_f} \pi_f(D), P_{x_n \mid \mathbf{x}_{1:(n-1)}}^{\sigma_G} \pi_G(D))\end{aligned} \tag{C.25}$$

where $\mathbf{y}_{1:(n-1)}$ is the realisation for inputs $\mathbf{x}_{1:(n-1)}$ on which the KL divergences are conditioned. Assuming this inequality, the result follows from the chain rule for the KL divergence in combination with the definition of $\mathrm{AR}_\mathbf{x}^{\sigma_C}$ (Procedure 5.14). To prove the inequality, note that, conditional on $\mathbf{y}_{1:(n-1)}$,

$$\underset{\mu_G \in \mathcal{P}_{\lambda,G}^1}{\arg\min} \, \mathrm{KL}(P_{x_n \mid \mathbf{x}_{1:(n-1)}}^{\sigma_f} \pi_f(D), \mu) = \mathcal{N}(P_{x_n \mid \mathbf{x}_{1:(n-1)}}^{\sigma_f} \pi_f(D)). \tag{C.26}$$

See the proof of Proposition 3.27 for the definition of $\mathcal{P}^1_{\lambda,\text{G}}$ and the argument for (C.26). But, by Proposition 3.26, this is exactly what $P^{\sigma_\text{C}}_{x_n} \pi_\text{C}(D \oplus (\mathbf{x}_{1:(n-1)}, \mathbf{y}_{1:(n-1)}))$ is! Since $P^{\sigma_\text{G}}_{x_n \mid \mathbf{x}_{1:(n-1)}} \pi_\text{G}(D) \in \mathcal{P}^1_{\lambda,\text{G}}$, the inequality follows. $\square$

**Proposition 5.16** (Recovery of smooth samples). *Let $\mathcal{X} \subseteq \mathbb{R}$ be compact, and let $f$ be a stochastic process with surely continuous sample paths and $\sup_{x \in \mathcal{X}} \|f(x)\|_{L^2} < \infty$. Let $(\varepsilon_n)_{n \geq 0}$ be i.i.d. random variables such that $\mathbb{E}[\varepsilon_0] = 0$ and $\mathrm{var}(\varepsilon_0) < \infty$. Consider any sequence $(x_n)_{n \geq 1} \subseteq \mathcal{X}$, and let $x^* \in \mathcal{X}$ be a limit point of $(x_n)_{n \geq 1}$, assuming that a limit point exists. If $y(x^*) = f(x^*) + \varepsilon_0$ and $y_n = f(x_n) + \varepsilon_n$ are noisy observations of $f$, then*

$$\lim_{n \to \infty} \mathbb{E}[y(x^*) \mid y_1, \ldots, y_n] = f(x^*) \quad \text{almost surely.} \tag{C.27}$$

*Proof.* Consider the increasing filtration $\mathcal{F}_n = \sigma(y_1, \ldots, y_n)$ with limit $\mathcal{F}_\infty = \sigma(\bigcup_{n=1}^\infty \mathcal{F}_n)$. Also let $\mathcal{T}_n = \sigma(\varepsilon_{n+1}, \varepsilon_{n+2}, \ldots)$ and consider the tail $\sigma$-algebra $\mathcal{T} = \bigcap_{n=1}^\infty \mathcal{T}_n$. Let $(x_{n_i})_{i=1}^\infty$ be a subsequence of $(x_n)_{n=1}^\infty$ such that $x_{n_i} \to x^*$. Let $g_n = \frac{1}{n} \sum_{i=1}^n y_i$. Since $g_n$ is a function of $y_1, \ldots, y_n$, it is $\mathcal{F}_n$–measurable and therefore $\mathcal{F}_\infty$–measurable. Note that

$$g_n = \frac{1}{n} \sum_{i=1}^n f(x_{n_i}) + \frac{1}{n} \sum_{i=1}^n \varepsilon_i. \tag{C.28}$$

By sure continuity of $f$, the first term converges to $f(x^*)$ surely. By the strong law of large numbers (Example 5.6.1; Durrett, 2010), the second term converges to zero on a tail event $A \in \mathcal{T}$ of probability one. We conclude that $\mathbb{1}_A f(x^*)$ is $\sigma(\mathcal{F}_\infty, \mathcal{T})$–measurable. Therefore, by almost sure convergence of $L^2$–bounded martingales (Theorem 5.4.5; Durrett, 2010),

$$\begin{aligned}
\lim_{n \to \infty} \mathbb{E}[y(x^*) \mid y_1, \ldots, y_n] &= \lim_{n \to \infty} \mathbb{E}[f(x^*) \mid y_1, \ldots, y_n] & (\mathbb{E}[\varepsilon_0] = 0) & \quad\text{(C.29)} \\
&= \lim_{n \to \infty} \mathbb{E}[f(x^*) \mid \mathcal{F}_n] & (\text{definition of } \mathcal{F}_n) & \quad\text{(C.30)} \\
&= \lim_{n \to \infty} \mathbb{E}[f(x^*) \mid \mathcal{F}_n, \mathcal{T}] & (\sigma(f(x^*), \mathcal{F}_n) \perp \mathcal{T}) & \quad\text{(C.31)} \\
&= \lim_{n \to \infty} \mathbb{E}[\mathbb{1}_A f(x^*) \mid \mathcal{F}_n, \mathcal{T}] & (\mathbb{P}(A) = 1) & \quad\text{(C.32)} \\
&= \mathbb{E}[\mathbb{1}_A f(x^*) \mid \mathcal{F}_\infty, \mathcal{T}] & (L^2\text{-mart. convergence}) & \quad\text{(C.33)} \\
&= \mathbb{1}_A f(x^*) & (\mathbb{1}_A f(x^*) \in \sigma(\mathcal{F}_\infty, \mathcal{T})) & \quad\text{(C.34)} \\
&= f(x^*), & (\mathbb{P}(A) = 1) & \quad\text{(C.35)}
\end{aligned}$$

where all equalities hold almost surely. $\square$

# D | Behind the Scenes of the ConvCNP

**Attributions.** Connecting the discretisation of the ConvCNP to inducing points of a sparse Gaussian process was first suggested by Tom Minka. Tom Minka also suggested using the Jacobi method to approximate the inverse. These suggestions were worked out in more detail by the author in collaboration with Stratis Markou, James Requeima, and Anna Vaughan.

## D.1 Introduction

Although the ConvCNP works well in practice, the model remains a black box. In this chapter, we attempt to improve our understanding of what could be going on inside a ConvCNP. We will explicitly construct a convolutional deep set (Theorem 4.8) that approximates the posterior mean of a Gaussian process with a stationary kernel. This draws a connection between ConvCNPs and Gaussian processes and gives one possible explanation of what could be going on inside.

To approximate the posterior mean of a Gaussian process with a ConvCNP, the idea is to consider a sparse approximation of a Gaussian processes (Titsias, 2009) with inducing point locations equal to the discretisation of the convolutional deep set (Procedure 5.5).

Let $k\colon \mathcal{X} \times \mathcal{X} \to \mathbb{R}$ be a stationary kernel and consider $f \sim \mathcal{GP}(0, k)$. Let $\mathbf{t} \in I$ be target inputs, and let $(\mathbf{x}, \mathbf{y}) \in \mathcal{D}_N$ be context data observed under Gaussian noise with variance one. Consider inducing point locations $\mathbf{z} \in \mathcal{X}^M$. Then the sparse approximation of the posterior mean is given by

$$\mathbb{E}[f(\mathbf{t}) \,|\, \mathbf{y}] \approx \mathbf{K_{tz}}(\mathbf{K_z} + \mathbf{K_{zx}}\mathbf{K_{xz}})^{-1}\mathbf{K_{zx}}\mathbf{y} \tag{D.1}$$

where $\mathbf{K_{tz}} = k(\mathbf{t}, \mathbf{z})$, $\mathbf{K_z} = k(\mathbf{z}, \mathbf{z})$, $\mathbf{K_{zx}} = k(\mathbf{z}, \mathbf{x})$, and $\mathbf{K_{xz}} = k(\mathbf{x}, \mathbf{z})$. See (6) and (10) by Titsias (2009). In this chapter, we will argue that (D.1) can be approximated with a convolutional deep set.

## D.2 Construction of a Convolutional Deep Set

Set the inducing point locations $\mathbf{z}$ to positions equally spaced over some interval $[a, b]$ with interpoint spacing $\Delta$. For simplicity, assume that the inputs of the observations $\mathbf{x}$ are all distinct. Moreover, assume that every input of an observation is equal to some inducing point location; that is, every element of $\mathbf{x}$ is some element of $\mathbf{z}$. Consider a convolutional deep set (Theorem 4.8) implemented according Procedure 5.5. In Procedure 5.5, set the discretisation equal to $\mathbf{z}$. Let $\mathbf{c}_y$ denote the discretised data channel and $\mathbf{c}_1$ the discretised density channel of the convolutional deep set (see (5.4) and (5.5)).

In the encoder enc of the convolutional deep set, assume that the length scale of the Gaussian kernel is infinitesimally small. Then, by the assumptions on the inputs of the observations,

$$c_{y,m} = \begin{cases} y_n & \text{if } z_m = x_n, \\ 0 & \text{otherwise,} \end{cases} \qquad c_{1,m} = \begin{cases} 1 & \text{if } z_m = x_n, \\ 0 & \text{otherwise,} \end{cases} \tag{D.2}$$

for all $m = 1, \ldots, M$. Therefore, $\mathbf{K_{zx}y} = \mathbf{K_z c}_y$. This reveals a first connection between (D.1) and a discretised convolutional deep set:

$$\mathbf{K_{tz}}(\mathbf{K_z} + \mathbf{K_{zx}K_{xz}})^{-1}\mathbf{K_{zx}y} = \underbrace{\mathbf{K_{tz}}}_{\text{smoothing}} \overbrace{(\mathbf{K_z} + \mathbf{K_{zx}K_{xz}})^{-1}\mathbf{K_z}}^{\text{CNN}} \underbrace{\mathbf{c}_y}_{\text{data channel}}. \tag{D.3}$$

The sparse approximation of the posterior mean starts out with the data channel $\mathbf{c}_y$, performs some matrix multiplications $(\mathbf{K_z} + \mathbf{K_{zx}K_{xz}})^{-1}\mathbf{K_z}$, and ends with a matrix multiplication by $\mathbf{K_{tz}}$. If we replace the Gaussian kernel in step ❹ of Procedure 5.5 with $k$, then multiplication by $\mathbf{K_{tz}}$ exactly performs step ❹. Therefore, if we can implement the intermediate matrix multiplications with a convolutional neural network (CNN), the sparse approximation of the posterior mean can be implemented with a convolutional deep set discretised according to Procedure 5.5.

To begin with, since $k$ is stationary and the elements of $\mathbf{z}$ are equally spaced, $\mathbf{K_z}$ is a Toeplitz matrix, and matrix multiplication with a Toeplitz matrix can be implemented with a convolutional filter. Therefore, multiplication by $\mathbf{K_z}$ can be implemented with a convolutional filter. It remains to argue that multiplication by $(\mathbf{K_z} + \mathbf{K_{zx}K_{xz}})^{-1}$ can be approximated with a CNN.

Let $\mathbf{v} = (\mathbf{K_z} + \mathbf{K_{zx}K_{xz}})^{-1}\mathbf{u}$ and split $\mathbf{K_z} = \mathbf{D} + \mathbf{O}$ into a diagonal part $\mathbf{D}$ and off-diagonal part $\mathbf{O}$. Rearrange

$$\mathbf{v} = \mathbf{D}^{-1}(\mathbf{u} - \mathbf{O}\mathbf{v} - \mathbf{K_{zx}K_{xz}}\mathbf{v}). \tag{D.4}$$

This suggests the iterative scheme

$$\mathbf{v}_{r+1} = \overbrace{\mathbf{D}^{-1}}^{\text{diag. matrix = filter}}(\mathbf{u} - \underbrace{\mathbf{O}}_{\text{Toeplitz matrix = filter}}\mathbf{v}_r - \mathbf{K_{zx}K_{xz}}\mathbf{v}_r). \tag{D.5}$$

This iterative scheme is known as the *Jacobi method*. Note that multiplication by $\mathbf{D}^{-1}$ and multiplication by $\mathbf{O}$ can both be implemented with convolutional filters. Therefore, if $\mathbf{K_{zx}K_{xz}}\mathbf{v}_r$ can also be approximated with convolutional filters, then rolling out the iterative scheme suggests that multiplication by $(\mathbf{K_z} + \mathbf{K_{zx}K_{xz}})^{-1}$ can be implemented with a multi-layer CNN. The last piece of the puzzle is therefore to figure out whether $\mathbf{K_{zx}K_{xz}}\mathbf{v}_r$ can be approximated with convolutional filters. And it is here that the density channel comes into play.

Let $\mathbf{d}_m$ be the $m^{\text{th}}$ diagonal of the matrix $\mathbf{K_{zx}K_{xz}}$. Then

$$\mathbf{K_{zx}K_{xz}}\mathbf{v}_r = \sum_{m=-M}^{M} \mathbf{d}_m \odot \mathsf{T}_{-m}\mathbf{v}_r \tag{D.6}$$

where $\odot$ denotes the Hadamard product and $\mathsf{T}_m$ denotes shifting elementwise by $m$ positions, discarding what falls off: $\mathsf{T}_m\mathbf{x} = \mathbf{x}_{1+m:|\mathbf{x}|}$ if $m \geq 0$ and $\mathsf{T}_m\mathbf{x} = \mathbf{x}_{1:|\mathbf{x}|+m}$ otherwise. Note that the Hadamard product can be approximated with a pointwise multi-layer perceptron (MLP), and that elementwise shifts can be implemented exactly with convolutional filters. To compute $\mathbf{d}_m$, using stationarity of $k$,

$$d_{m,m'} = \sum_{n=1}^{N} k(u_{m'} - x_n)k(x_n - (u_{m'} + m\Delta)) = \sum_{n=1}^{N} w_m(u_{m'} - x_n) \tag{D.7}$$

where $w_m(\tau) = k(\tau)k(\tau - m\Delta)$. Therefore, denoting $\mathbf{W}_m = w_m(\mathbf{z} - \mathbf{z}^\mathsf{T})$, which is again a Toeplitz matrix, we find $\mathbf{d}_m = \mathbf{W}_m\mathbf{c}_1$. We conclude that

$$\mathbf{K_{zx}K_{xz}}\mathbf{v}_r = \sum_{m=-M}^{M} \mathbf{d}_m \odot \mathsf{T}_{-m}\mathbf{W}_m\mathbf{c}_1, \tag{D.8}$$

which can be approximated with convolutional filters and pointwise MLPs.

## D.3 Conclusion

By connecting the inducing point locations of a sparse Gaussian process to the discretisation of a convolutional deep set, we were able to explicitly approximate the posterior mean of a Gaussian process with a discretised convolutional deep set. Both the data channel and the density channel played an important role. The data channel communicated the values of the observations, and the density channel was necessary to appropriately calibrate the uncertainty according to the inputs of the observations. We leave it for future work to further work out this connection between ConvCNPs and Gaussian processes.

# E | Experimental Details

## E.1 Description of Models

The architectures follow the descriptions from Chapter 5. Although these descriptions are for one-dimensional inputs and outputs, the architectures are readily generalised to multidimensional inputs and outputs; we will explicitly mention wherever that generalisation requires extra care. All architectures use ReLU activation functions. All Gaussian neural processes (GNPs; Section 5.5), in addition to a covariance matrix over the target points, also output heterogeneous observation noise along the marginal means; the total covariance over the target points is thus the sum of $k_\theta(D)$ and a diagonal matrix formed from these observation noises.

**Conditional Neural Process (CNP; Garnelo et al., 2018a).** The architecture follows Model 5.1. Set the dimensionality of the encoding to $K = 256$. Parametrise $\phi_\theta$ with a three-hidden-layer multi-layer perceptron (MLP) of width $256$; and parametrise $\text{dec}_\theta$ with a six-hidden-layer MLP of width $256$. For multidimensional outputs, let $\text{dec}_\theta$ have width $512$. For multidimensional outputs where outputs can have context points at different inputs, produce a separate encoding for every output and concatenate these into one big encoding. These encoders may or may not share parameters. In our experiments, for two-dimensional outputs, parametrise separate $\text{enc}_\theta$; for higher-dimensional outputs, apply the same $\text{enc}_\theta$.

**Gaussian Neural Process (GNP; Model 5.9).** The architecture follows Model 5.1. Use the same choices for $K$, $\phi_\theta$, and $\text{dec}_\theta$ as the CNP. Set the rank of the kernel map to $R = 64$. As mentioned in the introduction, let $\text{dec}_\theta$ produce one extra dimension which forms heterogeneous observation noise; the total covariance over the target points is then the sum of $k_\theta(D)$ and a diagonal matrix formed from these observation noises. For multidimensional outputs, the same caveats as for the CNP apply.

**Neural Process (NP; Garnelo et al., 2018b).** The NP builds off the CNP. Call the existing encoder $\text{enc}_\theta$ the *deterministic encoder.* The NP adds one more encoder called the *stochastic encoder.* The stochastic encoder mimics the deterministic encoder, but outputs a $K$-dimensional vector of means and a $K$-dimensional vector of marginal variances. These are used to sample a $K$-dimensional Gaussian latent variable (the *stochastic encoding*). The $\text{dec}_\theta$ now additionally takes in the stochastic encoding. For multidimensional outputs, the same caveats as for the CNP apply.

**Attentive Conditional Neural Process (ACNP; Kim et al., 2019).** The ACNP builds off the CNP. It replaces the deterministic encoder $\text{enc}_\theta\colon \mathcal{D} \to \mathbb{R}^K$ with an eight-head attentive encoder $\text{enc}_\theta^{(\text{att})}\colon \mathcal{D} \times \mathcal{X} \to \mathbb{R}^K$ (Bahdanau et al., 2015; Vaswani et al., 2017). Unlike the original deterministic encoder $\text{enc}_\theta$, the new attentive encoder $\text{enc}_\theta^{(\text{att})}$ also takes in the target input. Let $D^{(\text{c})} = (\mathbf{x}^{(\text{c})}, \mathbf{y}^{(\text{c})}) \in \mathcal{D}^N$ be a context set and let $x^{(\text{t})} \in \mathcal{X}$ be a target input. We now descibe the computation of $\text{enc}_\theta^{(\text{att})}(D^{(\text{c})}, x^{(\text{t})})$. Parametrise $\phi_x\colon \mathcal{X} \to (\mathbb{R}^{32})^8$ and $\phi_{xy}\colon \mathcal{X} \times \mathcal{Y} \to (\mathbb{R}^{32})^8$ both with three-hidden-layer MLPs of width $256$. Compute

$$\text{the \emph{keys}:} \quad (\mathbf{k}_{h,n})_{h=1}^8 = \phi_x(x_n^{(\text{c})}) \qquad \text{for } n = 1, \ldots, N, \tag{E.1}$$

$$\text{the \emph{values}:} \quad (\mathbf{v}_{h,n})_{h=1}^8 = \phi_{xy}(x_n^{(\text{c})}, y_n^{(\text{c})}) \quad \text{for } n = 1, \ldots, N, \tag{E.2}$$

$$\text{the \emph{query}:} \quad (\mathbf{q}_h)_{h=1}^8 = \phi_x(x^{(\text{t})}). \tag{E.3}$$

Then compute

$$\mathbf{v}_h^{(\text{q})} = \sum_{n=1}^N \frac{e^{\langle \mathbf{q}_h, \mathbf{k}_{h,n} \rangle}}{\sum_{n'=1}^N e^{\langle \mathbf{q}_h, \mathbf{k}_{h,n'} \rangle}} \mathbf{v}_{h,n} \in \mathbb{R}^{256} \tag{E.4}$$

Concatenate $\mathbf{v}^{(\text{q})} = (\mathbf{v}_1^{(\text{q})}, \ldots, \mathbf{v}_8^{(\text{q})}) \in \mathbb{R}^{256}$ and $\mathbf{q} = (\mathbf{q}_1, \ldots, \mathbf{q}_8) \in \mathbb{R}^{256}$. Let $\mathbf{L}\colon \mathbb{R}^{256} \to \mathbb{R}^{256}$ be a linear layer; let $\phi^{(\text{res})}\colon \mathbb{R}^{256} \to \mathbb{R}^{256}$ be a one-hidden-layer MLP of width $256$; and let $\text{norm}_1$ and $\text{norm}_2$ be two layer normalisation layers with learned pointwise transformations (Ba et al., 2016). Then

$$\text{enc}_\theta^{(\text{att})}(D^{(\text{c})}, x^{(\text{t})}) = \text{norm}_2(\mathbf{z} + \phi^{(\text{res})}(\mathbf{z})) \quad \text{where} \quad \mathbf{z} = \text{norm}_1(\mathbf{v}^{(\text{q})} + \mathbf{L}\mathbf{q}). \tag{E.5}$$

For multidimensional outputs, the same caveats as for the CNP apply.

**Attentive Gaussian Neural Process (AGNP; Section 5.5).** The AGNP builds off the GNP It replaces the deterministic encoder with the eight-head attentive deterministic encoder of the ACNP.

**Attentive Neural Process (ANP; Kim et al., 2019).** The ANP builds off the NP. It replaces the deterministic encoder with the eight-head attentive deterministic encoder of the ACNP.

**Convolutional Conditional Neural Process (ConvCNP; Model 5.4).** The architecture follows Model 5.4 and performs the discretisation approach outlined by (5.5). Set the discretisation to an evenly spaced grid at a certain density (the *points per unit*) spanning a bit more (the *margin*) than the most extremal context and target inputs. The points per unit and margin are specified separately for every experiment. Initialise the length scales of all Gaussian kernels to twice the interpoint spacing of the discretisation. Do divide the data channel by the density channel, as described in Section 5.3. Parametrise dec$_\theta$ with a U-Net (Ronneberger et al., 2015). Before the U-turn, let the U-Net have six convolutional layers with kernel size five, stride two, and 64 output channels; and six more such layers, but using transposed convolutions, after the U-turn. The layers after the U-turn additionally take in the outputs of the layers before the U-turn in reversed order; this is the U-net structure (Figure 1; Ronneberger et al., 2015). For multidimensional outputs where outputs can have context points at different inputs, produce a separate data and density channel for every output and concatenate these into one big encoding; use separate length scales for every application of enc$_\theta$.

**Convolutional Gaussian Neural Process (ConvGNP; Model 5.12).** The architecture follows Model 5.12. Use the same choices for the discretisation, length scales, and CNN architecture as for the ConvCNP. Set the rank of the kernel map to $R = 64$. As mentioned in the introduction, let dec$_\theta$ produce one extra channel which forms heterogeneous observation noise; the total covariance over the target points is then the sum of $k_\theta(D)$ and a diagonal matrix formed from these observation noises. For multidimensional outputs, the same caveat as for the ConvCNP applies.

**Fully Convolutional Gaussian Neural Process (FullConvGNP; Model 5.13).** For the mean architecture and the kernel architecture, use the same choices for the discretisation, length scales, and CNN architecture as for the ConvCNP. Do implement the source channel with the identity matrix, as described in Section 5.5; and do apply the matrix transform $\mathbf{Z} \mapsto \mathbf{Z}\mathbf{Z}^\mathsf{T}$ to ensure positive definiteness, as also described in Section 5.5. As mentioned in the introduction, let dec$_\theta^{(m)}$ produce one extra channel which forms heterogeneous observation noise; the total covariance over the target points is then the sum of $k_\theta(D)$ and a diagonal matrix formed from these observation noises. For multidimensional outputs, in addition to the caveat for the ConvCNP, two additional caveats apply. First, for $D_\mathrm{o}$-dimensional outputs, let dec$_\theta^{(k)}$ produce $D_\mathrm{o}^2$ channels rather than just one. These channels should be interpreted as all covariance and cross-covariance matrices between all outputs. Second, when applying the matrix transform $\mathbf{Z} \mapsto \mathbf{Z}\mathbf{Z}^\mathsf{T}$, these channels should first be assembled into one total covariance matrix.

**Convolutional Neural Process (ConvNP; Foong et al., 2020).** The ConvNP builds off the ConvCNP. The ConvNP replaces the CNN architecture by two copies of this architecture placed in sequence. Inbetween the two architectures, there is a sampling step: the first architecture outputs 32 channels, comprising 16 means and 16 marginal variances, which are used to sample a 16-dimensional Gaussian latent variable; and the second architecture then takes in this sample.

**Autoregressive Conditional Neural Processes (AR CNPs; Section 5.6).** The AR CNP, AR ACNP, and AR ConvCNP use the architectures described above. Rolling out an AR CNP according to Procedure 5.14 requires an ordering of the target points. In all experiments, we choose a random ordering of the target points.

## E.2 Training, Cross-Validation, and Evaluation Protocols

A *task* consists of a context set and target set (Section 2.1). How precisely the context and target sets are generated is specific to an experiment. To train a model, we consider batches of 16 tasks at a time, compute an objective function value, and update the model parameters using ADAM (Kingma et al., 2015). The learning rate is specified separately for every experiment. We define an epoch to consist of $2^{14} \approx 16\,\text{k}$ tasks. We typically train a model for between 100 and 500 epochs.

For an experiment, we split up the meta–data set into a *training set*, a *cross-validation set*, and an *evaluation set*. The model is trained on the training set. During training, after every epoch, the model is cross-validated on the cross-validation set. Cross-validation uses $2^{12}$ fixed tasks. These $2^{12}$ are fixed, which means that cross-validation always happens with exactly the same data. The cross-validation objective is a confidence bound computed from the model objective. Suppose that model objective over all $2^{12}$ cross-validation tasks has empirical mean $\hat{\mu}$ and empirical variance $\hat{\sigma}^2$. If a higher model objective is better, then the cross-validation objective is given by $\hat{\mu} - 1.96 \cdot \hat{\sigma}/\sqrt{2^{12}}$. The model with the best cross-validation objective is selected and used for evaluation. Evaluation is performed with the evaluation set and also uses $2^{12}$ tasks.

Conditional neural processes and Gaussian neural processes are trained, cross-validated, and evaluated with the neural process objective proposed by Garnelo et al. (2018a), (2.2) in Section 2.2. We normalise the terms in the neural process objective by the target set sizes. Latent-variable neural processes (LNPs) are trained, cross-validated, and evaluated with the ELBO objective proposed by Garnelo et al. (2018b) using five samples, also normalised by the target set size. When training LNPs with the ELBO objective, but not when cross-validating and evaluating, the context set is subsumed in the target set. Additionally, LNPs are trained, cross-validated, and evaluated with the ML objective proposed by Foong et al. (2020), again normalised by the target set size. When training and cross-validating LNPs with the ML objective, we use twenty samples; and when evaluating, we use 512 samples. For completeness, LNPs trained with the ELBO objective are also evaluated with the ML objective using 512 samples.

To stabilise the numerics for GNPs, we increase the regularisation of covariance matrices for one epoch. To encourage LNPs to fit, we fix the variance of the observation noise of the decoder to $10^{-4}$ for the first three epochs.

## E.3 Synthetic Experiments

For this experiment, the learning rate is $3 \cdot 10^{-4}$, the margin is $0.1$, and the points per unit is $64$. We trained the models for 100 epochs. Due to an error in the cross-validation procedure, we did not use cross-validation, but used the model at epoch 100.

For the kernel architecture of the FullConvGNP, we reduce the points per unit and the number of channels in the U-Net by a factor two. For the ConvNP with two-dimensional inputs, we reduce the number of outputs channels in the U-Net by a factor $\sqrt{2}$; and, for training and cross-validation, we reduce the number of samples of the ELBO objective to one and the number of samples for the ML objective to five.

## E.4 Sim-to-Real Transfer with the Lotka–Volterra Equations

For this experiment, the learning rate is $1 \cdot 10^{-4}$, the margin is $1$, and the points per unit is $4$. We trained the models for 200 epochs.

The convolutional models use a U-Net architecture with seven layers instead of six where, in the first layer, the stride is one instead of two. For the kernel architecture of the FullConvGNP, we reduce the points per unit and the number of channels in the U-Net by a factor two.

## E.5 Electroencephalography Experiments

For this experiment, the learning rate is $2 \cdot 10^{-4}$, the margin is $0.1$, and the points per unit is $256$. We trained the models for 200 epochs. The training runs for the ANPs and FullConvGNP were terminated after 45 hours; these models all reached epoch 80–120.

The convolutional models use a U-Net architecture where, in the first layer, the stride is one instead of two. In addition, the number of channels are adjusted as follows: the ConvCNP and ConvGNP use 128 channels, the ConvNP uses 96 channels, and the FullConvGNP uses 64 channels. The length scales of the Gaussian kernels of the convolutional model is initialised to $0.77/256$. To scale to seven outputs, the deep set–based and attentive models reuse the same encoder for every output dimension.

## E.6 Climate Downscaling

**MLP ConvCNP and MLP ConvGNP (Section 6.5; Vaughan et al., 2022).** The architectures of the MLP ConvCNP and MLP ConvGNP follow the description in Section 6.5, with the following additional details. The decoder $\text{dec}_\theta = \text{fuse}_\theta \circ \text{dec}'_\theta$ is decomposed into a convolutional architecture $\text{dec}'_\theta$ followed by a pointwise MLP $\text{fuse}_\theta$ (Definition 6.1). Parametrise $\text{dec}'_\theta$ with a seven-layer residual convolutional neural network (He et al., 2016). Every residual layer involves one depthwise-separable convolutional filter (Chollet, 2017) with kernel size three followed by a pointwise MLP. Every layer has 128 channels, and the network also outputs 128 channels. The discretisation for $\text{dec}'_\theta$ is the grid of the ERA-Interim reanalysis variables. Parametrise $\text{fuse}_\theta$ with a three-hidden-layer MLP of width 128.

The MLP ConvCNP and MLP ConvGNP are trained with learning rate $2.5 \cdot 10^{-5}$ for 500 epochs. For the MLP ConvGNP, to encourage the covariance to fit, we fix the variance of the decoder to $10^{-4}\mathbf{I}$ for the first ten epochs.

**AR ConvCNP (Section 6.5).** The architecture of the AR ConvCNP follows the descriptions in Section 6.5 and Figure 6.12, with the following additional details. Parametrise $\text{CNN}_{\text{lr}}$ with a depthwise-separable residual convolutional neural network like in the MLP ConvCNP and MLP ConvGNP, but use six layers instead of seven. Let $\text{CNN}_{\text{lr}}$ output 64 channels. The discretisation for $\text{CNN}_{\text{lr}}$ is the grid of the ERA-Interim reanalysis variables. Parametrise $\text{CNN}_{\text{mr}}$ with a U-Net (Ronneberger et al., 2015) using an architecture similar to what we have been using. Before the U-turn, let the U-Net have five convolutional layers with kernel size five, stride one for the first layer and stride two afterwards, 64 output channels for the first three layers and 128 output channels afterwards. After the U-turn, instead of using transposed convolutions, use regular convolutions combined with an upsampling layer using bilinear interpolation. Let $\text{CNN}_{\text{mr}}$ output 64 channels. The receptive field of $\text{CNN}_{\text{mr}}$ is approximately $10°$. The discretisation for $\text{CNN}_{\text{mr}}$ is centred around the target points with margin $5°$. Parametrise $\text{CNN}_{\text{hr}}$ with a U-Net like for $\text{CNN}_{\text{hr}}$, but with four convolutional layers before the U-turn. The receptive field of $\text{CNN}_{\text{hr}}$ is approximately $0.5°$. The discretisation for $\text{CNN}_{\text{hr}}$ is centred around the target points with margin $0.25°$.

The AR ConvCNP is trained with learning rate $1 \cdot 10^{-5}$ for 500 epochs. During training and cross-validation, the target points are subsampled to lie in a $3° \times 3°$ square. For training, the number of target points is ensured to be at least ten; and for cross-validation, at least one. The size of the cross-validation set is increased ten fold.

**Fusion experiments.** The number of context points $n$ is sampled from $p(n) \propto e^{-0.01n}$. A data set is split into a context and target set by randomly selecting $n$ points as context points and letting the remainder be target points. For the AR ConvCNP, this splitting is done after subsampling the $3° \times 3°$ square.

# Index